\documentclass[blueaccent,abstractfill]{preprint_classic}
\title{Towards Agentic Robots:}
\subtitle{Model, Data, Environment, and Harness}
\hypersetup{pdftitle={Towards Agentic Robots: Model, Data, Environment, and Harness}}

\author{Chonghan Qin, Dongming Wu, Kaituo Feng, Jiaxu Wang, Xinlei Yu, Peiwen Sun\\Yijia Fan, Xilin He, Bingao Chen, Weitong Lian, Ryan Ji, 
Wenqi Ren, Xiaobin Hu\\ Dahua Lin\textdagger, Yu-Gang Jiang\textdagger, Xiangyu Yue\textdagger}

\affiliation{MMLab, CUHK}
\affiliation{FDU}
\affiliation{NovaXBot}
\affiliation{THU}
\affiliation{ZJU}
\affiliation{SYSU}

\abstract{
Foundation models are driving a paradigm shift in robotic intelligence, moving beyond low-level action prediction to goal-directed orchestration of data collection, model training, task planning, and real-world execution. We refer to robotic systems that follow this emerging paradigm as \textbf{agentic robots}. In these systems, agents reason over human instructions, interact with robotic environments, rectify execution errors autonomously, and dynamically adjust operational strategies.
Although this field develops fast, current studies are not well connected. Researchers have no consistent view of how to combine different agentic functions into a complete robot system.
\looseness=-1\par\vspace{4pt}% Keep the first paragraph from ending with a single word.
This paper provides a unified overview of existing research on agentic robots. We first trace the development from robot policies to agentic systems, covering vision-language-action models, world models, and world-action models. We then clarify the definition and scope of agentic robots and organize existing research into seventeen categories across four functional domains: \textbf{Model}, \textbf{Data}, \textbf{Environment}, and \textbf{Harness}. Across these domains, we examine how agents shape model learning and robot decisions, collect and curate robot data, construct worlds, tasks, and tests for learning and evaluation, and organize capabilities for robot use. Finally, we discuss open challenges and outline directions for future research. Looking forward, we hope this work contributes to a clearer understanding of agentic robots and supports new approaches to making them more capable, reliable, and adaptive.

}
\correspondence{\email{chonghanqin@link.cuhk.edu.hk}, \email{xyyue@ie.cuhk.edu.hk}}
\checkdata[Repository]{\href{https://github.com/chonghanqin/Awesome-Agentic-Robots}{https://github.com/chonghanqin/Awesome-Agentic-Robots}}

\coverfigure{%
\noindent\begin{minipage}{\textwidth}
  \centering
  \includegraphics[width=\linewidth]{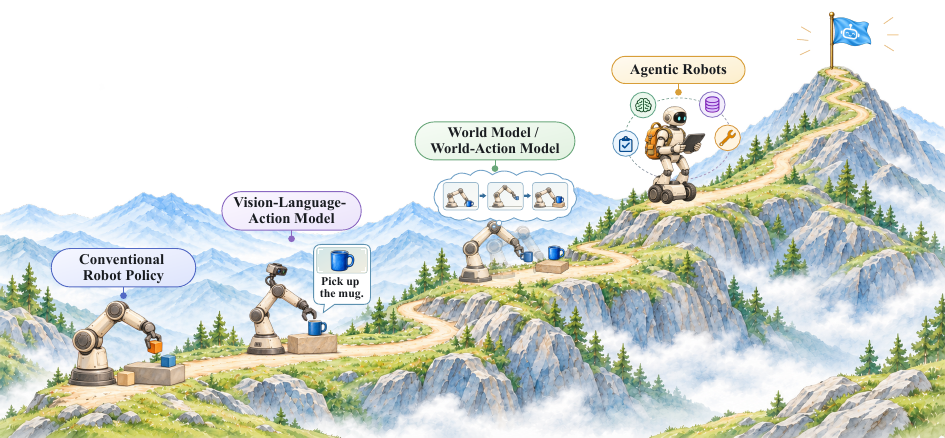}
  \captionsetup{skip=8pt, hypcap=false}
\captionof{figure}{From task-specific robot policies to agentic robots.}
\label{fig:agentic-robots-overview}
\end{minipage}
}

\usepackage{authblk}
\usepackage{array,tabularx,longtable,ragged2e,float,placeins,needspace}
\usetikzlibrary{arrows.meta,positioning,shapes.geometric,calc}
\definecolor{ruBlue}{HTML}{246A86}
\definecolor{ruGreen}{HTML}{356F58}
\definecolor{ruRed}{HTML}{A44640}
\definecolor{ruLight}{HTML}{F1F5F7}
\newcolumntype{Y}{>{\raggedright\arraybackslash}X}

\tikzset{ruBox/.style={draw=cuhkGray,rounded corners=1mm,fill=white,align=center,font=\sffamily\fontsize{9}{10.5}\selectfont,inner sep=2mm},ruArrow/.style={-{Stealth[length=1.6mm]},draw=cuhkGray,line width=.6pt},ruFeedback/.style={-{Stealth[length=1.6mm]},draw=ruBlue,line width=.9pt},ruHyp/.style={-{Stealth[length=1.6mm]},draw=ruRed,dashed,line width=.8pt}}

\NewDocumentCommand{\rewriteplaceholder}{O{!htbp} O{54mm} m m m}{%
\begin{figure}[#1]
\centering
\fbox{\begin{minipage}[c][#2][c]{\dimexpr\linewidth-2\fboxsep-2\fboxrule\relax}
\centering{\sffamily\bfseries #4}\par\medskip
{\sffamily\small PLACEHOLDER}
\end{minipage}}
\caption{#5}
\label{fig:#3}
\end{figure}}

\makeatletter
\newcommand{\manuscript@LT@start}{%
  \ifdim\dp\LT@foot>\z@
    \setbox\LT@foot\vbox{\unvbox\LT@foot\kern\z@}%
  \fi
  \manuscript@LT@originalstart}
\ExplSyntaxOn
\cs_if_exist:NF \regex_match:nVTF
  { \cs_generate_variant:Nn \regex_match:nnTF { nV } }
\tl_new:N \l__manuscript_lt_output_tl
\cs_new_protected:Npn \mechanismtableglue
  {
    \ifx\LT@start\manuscript@LT@start
    \else
      \let\manuscript@LT@originalstart\LT@start
      \let\LT@start\manuscript@LT@start
    \fi
    \tl_set:No \l__manuscript_lt_output_tl { \LT@output }
    \regex_match:nVTF { \c{vss} } \l__manuscript_lt_output_tl
      {
        \regex_replace_all:nnN { \c{vss} } { \c{vfil} }
          \l__manuscript_lt_output_tl
        \cs_set_nopar:Npx \LT@output { \exp_not:V \l__manuscript_lt_output_tl }
      }
      { }
  }
\ExplSyntaxOff
\makeatother

\usepackage{colortbl}
\definecolor{TSTModelAccent}{HTML}{55B6EE}
\definecolor{TSTModelLight}{HTML}{EAF7FF}
\definecolor{TSTModelLightest}{HTML}{F7FCFF}
\definecolor{TSTDataAccent}{HTML}{E3B331}
\definecolor{TSTDataLight}{HTML}{FFF4D6}
\definecolor{TSTDataLightest}{HTML}{FFFAEC}
\definecolor{TSTEnvironmentAccent}{HTML}{45D0AE}
\definecolor{TSTEnvironmentLight}{HTML}{E9FBF5}
\definecolor{TSTEnvironmentLightest}{HTML}{F6FEFB}
\definecolor{TSTHarnessAccent}{HTML}{A98BEA}
\definecolor{TSTHarnessLight}{HTML}{F0E8FF}
\definecolor{TSTHarnessLightest}{HTML}{FAF7FF}
\colorlet{CategoryModelTraining}{TSTModelAccent!45!black}
\colorlet{CategoryModelPerception}{TSTModelAccent!70!black}
\colorlet{CategoryModelPlanning}{TSTModelAccent}
\colorlet{CategoryModelRobotControl}{TSTModelAccent!75!white}
\colorlet{CategoryModelAdaptation}{TSTModelAccent!55!white}
\colorlet{CategoryDataCollection}{TSTDataAccent!58!black}
\colorlet{CategoryDataAnnotation}{TSTDataAccent}
\colorlet{CategoryDataRefinement}{TSTDataAccent!72!white}
\colorlet{CategoryDataFiltering}{TSTDataAccent}
\colorlet{CategoryDataCorrection}{TSTDataAccent!72!white}
\colorlet{CategoryEnvironmentReconstruction}{TSTEnvironmentAccent!55!black}
\colorlet{CategoryEnvironmentTaskGeneration}{TSTEnvironmentAccent}
\colorlet{CategoryEnvironmentBenchmarking}{TSTEnvironmentAccent!68!white}
\colorlet{CategoryHarnessOrchestration}{TSTHarnessAccent!45!black}
\colorlet{CategoryHarnessMemory}{TSTHarnessAccent!65!black}
\colorlet{CategoryHarnessMonitoringandRecovery}{TSTHarnessAccent!85!black}
\colorlet{CategoryHarnessSkills}{TSTHarnessAccent}
\colorlet{CategoryHarnessInfrastructure}{TSTHarnessAccent!80!white}
\colorlet{CategoryHarnessLayeredSystems}{TSTHarnessAccent!60!white}
\newcommand{\CategoryMarker}[1]{\raisebox{0.1ex}{\textcolor{Category#1}{\rule{1.25mm}{1.25mm}}}}
\newcommand{\CategoryLabel}[2]{\mbox{\CategoryMarker{#1}\hspace{0.7mm}\textcolor{black}{#2}}}
\newcommand{\CategoryLabelWrap}[2]{\CategoryMarker{#1}\hspace{0.7mm}\parbox[t]{\dimexpr\linewidth-2mm\relax}{\raggedright\textcolor{black}{#2}}}
\newcommand{\TableStyle}[2]{%
  \color{black}\hypersetup{linkcolor=black,urlcolor=black}%
  \captionsetup{font={small,color=black},labelfont={bf,color=black},skip=5pt}%
  \fontsize{#1}{#2}\selectfont
  \setlength{\parskip}{0pt}\hyphenpenalty=500
  \setlength{\tabcolsep}{3pt}\renewcommand{\arraystretch}{1.08}%
  \setlength{\heavyrulewidth}{0.65pt}\setlength{\lightrulewidth}{0.3pt}%
  \setlength{\aboverulesep}{2pt}\setlength{\belowrulesep}{3pt}}
\newcommand{\CompactTableStyle}{\TableStyle{8.25}{10}\setlength{\tabcolsep}{1.7pt}}
\newlength{\TableTextSep}
\AddToHook{env/longtable/begin}{%
  \setlength{\LTpre}{\TableTextSep}%
  \setlength{\LTpost}{\TableTextSep}}
\newcommand{\TableDomain}[1]{{\color{TST#1Accent}\rule{2pt}{8pt}}\hspace{4pt}\textbf{#1}}
\newcommand{\TaxonomyTableStyle}{\TableStyle{9.5}{11.4}\setlength{\aboverulesep}{1pt}\setlength{\belowrulesep}{1pt}}
\newcommand{\CategorySeparator}{\arrayrulecolor{black!22}\cmidrule[0.2pt](l{0pt}r{0pt}){2-3}\arrayrulecolor{black}}

\newlength{\LiteratureWidth}
\newcommand{\LiteratureCode}[1]{\href{#1}{\textcolor{black}{\raisebox{-0.05ex}{{\faGithub}}}}}
\newcommand{\LiteraturePaperLink}[1]{\href{#1}{\textcolor{black}{\raisebox{-0.03ex}{\faIcon[regular]{file-alt}}}}}
\newcommand{\LiteratureLinks}[3][]{\mbox{%
  \fontsize{9}{11}\selectfont%
  \LiteraturePaperLink{#2}\hspace{5pt}%
  \if\relax\detokenize{#3}\relax%
    {\setbox0=\hbox{\faGithub}\makebox[\wd0][c]{-}}%
  \else%
    \LiteratureCode{#3}#1%
  \fi}}

\newcommand{\LiteratureTopRule}[1]{\arrayrulecolor{TST#1Accent}\toprule\arrayrulecolor{black}}

\newmarks\LiteratureCategoryMarks

\makeatletter
\newcommand{\literature@continued}[2]{%
  \strut\CategoryLabel{#1}{\textbf{#2}}\enspace(continued)}
\newdimen\literature@firstdepth
\newcommand{\LiteratureContinuationSetup}{%
  \let\literature@start\LT@start
  \def\LT@start{%
    \literature@firstdepth\z@
    \ifdim\pagegoal<\maxdimen \literature@firstdepth\@maxdepth\fi
    \edef\literature@startpage{\number\c@page}%
    \literature@start
    \ifnum\c@page=\literature@startpage
      \ifdim\pagegoal<\maxdimen \pagegoal\dimexpr\pagegoal-\literature@firstdepth\relax\fi
    \fi}%
  \let\literature@output\LT@output
  \def\LT@output{%
    \edef\literature@category{\botmarks\LiteratureCategoryMarks}%
    \ifx\literature@category\@empty\else
      \setbox\LT@head\vbox{%
        \unvcopy\LT@head
        \hbox to\linewidth{\expandafter\literature@continued\literature@category\hfil}%
        \kern2.5pt}%
    \fi
    \literature@output
    \ifnum\outputpenalty>-\@Mi
      \global\maxdepth\z@
    \else
      \global\maxdepth\@maxdepth
      \ifnum\c@page=\literature@startpage
        \ifdim\pagegoal<\maxdimen
          \pagegoal\dimexpr\pagegoal+\literature@firstdepth\relax
        \fi
      \fi
    \fi}}
\makeatother

\DeclareRobustCommand{\ComparisonIconGeometry}{\tikz[x=.76pt,y=.76pt,baseline=2pt,line width=.7pt,line cap=round,line join=round,draw=black!85]{\path[use as bounding box] (0,0) rectangle (16,14);\draw[-{Stealth[length=2.2pt]}] (6,5) -- (15,5);\draw[-{Stealth[length=2.2pt]}] (6,5) -- (6,13);\draw[-{Stealth[length=2.2pt]}] (6,5) -- (1,1);}}
\DeclareRobustCommand{\ComparisonIconReal}{\raisebox{-1.5pt}{\includegraphics[height=11.5pt]{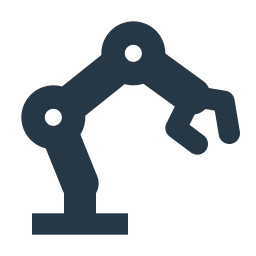}}}
\DeclareRobustCommand{\ComparisonIconSim}{\raisebox{-1.5pt}{\includegraphics[height=11.5pt]{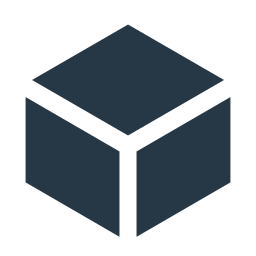}}}
\DeclareRobustCommand{\ComparisonIconFA}[1]{{\fontsize{10.5}{11}\selectfont\textcolor{black!85}{#1}}}

\DeclareRobustCommand{\ComparisonIconImage}{\tikz[x=.72pt,y=.72pt,baseline=2pt,line width=.65pt,line cap=round,line join=round,draw=black!80,fill=black!80]{\path[use as bounding box] (0,0) rectangle (16,14);\draw[rounded corners=.7pt] (1,1) rectangle (15,13);\fill (11,10) circle (1.15);\draw (2.5,3) -- (6,7.5) -- (9,4.3) -- (11,6.5) -- (13.5,3);}}
\DeclareRobustCommand{\ComparisonIconVideo}{\tikz[x=.72pt,y=.72pt,baseline=2pt,line width=.65pt,line cap=round,line join=round,draw=black!80,fill=black!80]{\path[use as bounding box] (0,0) rectangle (16,14);\draw[rounded corners=.7pt] (1,1) rectangle (15,13);\draw (3.5,1) -- (3.5,13) (12.5,1) -- (12.5,13);\draw (1,5) -- (3.5,5) (1,9) -- (3.5,9) (12.5,5) -- (15,5) (12.5,9) -- (15,9);\fill (6,4) -- (10.5,7) -- (6,10) -- cycle;}}
\DeclareRobustCommand{\ComparisonGlyph}[1]{\csname ComparisonGlyph@#1\endcsname}
\expandafter\def\csname ComparisonGlyph@geometry\endcsname{\ComparisonIconGeometry{}}
\expandafter\def\csname ComparisonGlyph@image\endcsname{\ComparisonIconImage{}}
\expandafter\def\csname ComparisonGlyph@language\endcsname{\ComparisonIconFA{\faCommentDots}}
\expandafter\def\csname ComparisonGlyph@video\endcsname{\ComparisonIconVideo{}}
\expandafter\def\csname ComparisonGlyph@real\endcsname{\ComparisonIconReal{}}
\expandafter\def\csname ComparisonGlyph@sim\endcsname{\ComparisonIconSim{}}
\expandafter\def\csname ComparisonGlyph@dataset\endcsname{\ComparisonIconFA{\faDatabase}}

\begin{document}

% Keep the author notes together in the first-page footnotes.
% Allow one extra line on the cover for the note and the complete Figure 1 caption.
\enlargethispage{\baselineskip}
\begin{NoHyper}
\renewcommand{\thefootnote}{\textdagger}
\footnotetext{Core Supervisors}
\end{NoHyper}

\maketitle

\papercontents

\noindent
\begin{minipage}{\textwidth}
  \centering
  \includegraphics[width=\linewidth]{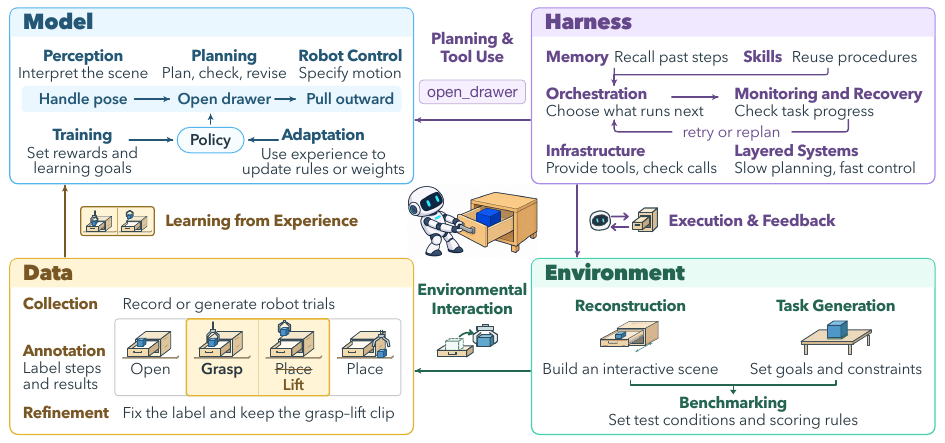}
  \captionsetup{skip=8pt,hypcap=false}
  \captionof{figure}{Overview of agentic robots across
  Model, Data, Environment, and Harness. The four domains
  connect model decisions, robot execution, environmental
  interaction, and learning from experience.}
  \label{fig:four-domain-overview}
\end{minipage}
\par

\section{Introduction}
\label{sec:introduction}

% Recent advances in robot learning and foundation models have expanded both the actions robots can perform and the ways their behavior can be organized. Reinforcement learning and imitation learning turn interaction and demonstrations into policies that map observations to actions, including visuomotor skills learned directly from images~\citep{survey_ax_1504_00702}. Scaling robot data and action models extends this approach toward multi-task and generalist policies~\citep{survey_ax_2212_06817,survey_ax_2405_12213}. Vision-language-action (VLA) models further connect robot behavior to the semantic knowledge of pretrained foundation models~\citep{survey_ax_2307_15818}. Alongside these advances in action competence, large language models (LLMs), vision-language models (VLMs), and general-purpose agents provide mechanisms for reasoning, tool use, program generation, verification, and memory~\citep{survey_ax_2210_03629,survey_ax_2209_07753,survey_ax_2305_16291,survey_ax_2603_20633}.

Over the past few years, foundation models including large language models (LLMs) and multimodal large language models (MLLMs) have greatly advanced robotic policy learning. They endow robots with novel capabilities that traditional task-specific models lack, especially in scene and task understanding and cross-scene generalization, by scaling robot datasets and model size~\citep{survey_ax_2212_06817,survey_ax_2405_12213}. As a dominant technical solution in this field, vision-language-action (VLA) models utilize either discrete action tokens or diffusion- or flow-matching-based continuous action prediction strategies to connect robotic action prediction and the semantic knowledge embedded in pre-trained LLMs~\citep{survey_ax_2307_15818,survey_ax_2410_24164,survey_ax_2504_16054}. Nevertheless, improved action prediction alone cannot guarantee reliable execution for long-horizon tasks. These tasks require robots to track task progress, adjust plans amid environmental changes, and recover from failures. VLA policies by themselves cannot consistently deliver these capabilities.

\paragraph{\textbf{Why Do Agentic Robots Matter?}}
Recently, agentic robots are attracting surging attention in the research community, which represent a new generation of embodied intelligent systems that integrate foundation model reasoning with embodied agency and environmental interaction capabilities. General-purpose models such as GPT-6 Astra~\citep{openai2026astra} and Gemini 4 Argon~\citep{google2026argon} are opening new possibilities for such systems.
For example, GPT-6 Astra demonstrates promising capabilities as a robot manipulation agent, using visual observations to generate end-effector commands and adjust its actions from execution feedback \textit{without task-specific fine-tuning}~\citep{survey_ax_2609_24170}. On RoboDojo, it achieves a 22.48\% average success rate across 42 simulation tasks, outperforming the trained robot policy DM0.5 (19.34\%)~\citep{dexmal2026dm05}, GalaxeaVLA G0.5 (14.88\%)~\citep{galaxea2026g05}, Xiaomi-Robotics-1 (13.93\%)~\citep{survey_ax_2607_15330}, OpenWAM-$\alpha$ (11.92\%)~\citep{survey_ax_2609_07398} and $\pi_{0.5}$ (6.91\%)~\citep{survey_ax_2504_16054,survey_ax_2609_24170}.
% Besides, agent-based methods have shown strong potential in improving model capabilities 
% Unlike previous approaches that focus on optimizing and scaling the model parameters, agentic robots integrate goal-oriented planning, state-based self-monitoring, and iterative refinement to support long-horizon decision-making. 
% In other words, its capability boundary also varies sharply as the external environment changes.
% These mechanisms enable robots to actively explore unstructured environments, reason about implicit task logic, and adapt to dynamic disturbances when tackling complex, open-ended, real-world tasks. 
Despite this progress, inference latency and resource costs highlight the importance of how foundation models are integrated into robotic systems. Beyond optimizing and scaling model parameters, agentic robots use foundation models to plan actions, monitor task progress, and revise decisions based on execution feedback. Their capabilities therefore depend on both the model and the tools, observations, and feedback available in the environment.
%This interaction loop allows robots to gather missing information, adjust plans when conditions change, and recover from failed actions, supporting more adaptive execution of long-horizon tasks.
For example, after pulling on a drawer, a robot can use a new observation to decide whether to reach inside or try opening it again. The agentic robot paradigm addresses the “brain in a vat” limitation of previous robotic solutions, shifting robotic intelligence from static pattern matching toward adaptive autonomous operation.

% Bringing these capabilities into robotic systems therefore broadens the design problem beyond action generation. A capable policy remains central, but the system must also determine what to attempt, when to gather more evidence, and how to respond when execution fails~\citep{survey_ax_2207_05608,survey_ax_2607_08448}. In a drawer-retrieval task, for example, a learned skill may execute the pull, while a higher-level process decides when to call it, checks whether the drawer moved, and revises the plan if it did not. Experience from execution also raises choices about future capabilities: models can direct data collection or revise policy and training code~\citep{survey_ax_2401_12963,survey_ax_2606_19980}. We use \emph{agentic robotics} to encompass these model-mediated decisions in both runtime robot use and capability building, without tying agency to a particular model architecture.

\paragraph{\textbf{Agentic Robots Need a Taxonomy.}}
% However, work on runtime decisions and capability building is spread across research traditions that emphasize different parts of the system. Robot learning and VLAs develop action competence~\citep{survey_ax_2307_15818}. Embodied reasoning and code-as-policy methods examine how models choose and compose executable behavior~\citep{survey_ax_2207_05608,survey_ax_2209_07753}. World models and world-action models connect predictions of future outcomes to action selection and generation~\citep{survey_ax_1610_00696,survey_ax_2602_15922}. Work on self-improving systems and agent harnesses examines how experience is used to revise capabilities or guide subsequent execution~\citep{survey_ax_2606_19980,survey_ax_2607_08448}. Although these lines overlap, they differ in the decisions assigned to models and the execution capabilities taken as given. Studying them separately leaves their dependencies difficult to see: how a model's decisions rely on its executor, how feedback shapes the next decision, and which changes persist beyond an interaction. A system-level account is needed to make these connections explicit
% \textbf{Agentic robots need a taxonomy. }
Despite growing interest, research on agentic robots remains fragmented across model learning, data acquisition, environment construction, and runtime system design. These directions are often studied separately, although their contributions interact within a robotic system. Related terms such as \textit{embodied agents}, \textit{autonomous robots}, and \textit{self-improving robotic systems} further blur the scope of the field by emphasizing different capabilities without consistently clarifying their boundaries. Together, these issues make it difficult to identify where agents make decisions and how those decisions contribute to the overall system. A unified taxonomy should therefore distinguish the roles of individual components while explaining how they work together. To this end, we organize agentic robotics into four functional domains: \textbf{Model}, \textbf{Data}, \textbf{Environment}, and \textbf{Harness}, tracing the agent’s decisions within each domain and their effects across the system. Figure~\ref{fig:four-domain-overview} shows how these four domains work together in agentic robot systems.

% \Needspace{8\baselineskip}
% We therefore organize our synthesis around three linked questions. \textbf{What should the agent decide?} This identifies the choices assigned to a model, from selecting a subgoal to revising how a capability is learned. \textbf{What should the robot execute?} This asks how those decisions are realized through policies, skills, solvers, and controllers. \textbf{What should experience change?} This distinguishes changes that affect only the current attempt from those retained as reusable knowledge or persistent updates to the system.

\begin{figure}[h]
\newcommand{\questionsection}[1]{\hyperref[sec:#1]{\S\,\ref*{sec:#1}}}
\begin{tcolorbox}[
  enhanced, colback=white, colframe=paperBlue!78!black,
  boxrule=1.25pt, arc=2.6pt, auto outer arc,
  boxsep=0pt, left=10pt, right=10pt, top=9pt, bottom=6pt,
  before skip=0pt, after skip=0pt,
  title={Key Questions}, fonttitle=\normalsize\bfseries,
  coltitle=white, attach boxed title to top left={xshift=6pt,yshift=-6.7pt},
  boxed title style={enhanced,frame hidden,colback=paperBlue!78!black,boxrule=0pt,
    boxsep=0pt,left=6pt,right=6pt,top=1.3pt,bottom=1.3pt,arc=1.8pt,auto outer arc}
]
\normalsize
\renewcommand{\arraystretch}{1.38}
\noindent\begin{tabular*}{\linewidth}{
  @{}>{\color{black!76}}r@{\hspace{0.55em}}
  l@{\hspace{1em}}l@{\extracolsep{\fill}}r@{}
}
1. & \textbf{Model} & What should the agent learn, understand, decide, and control? & \questionsection{model} \\
2. & \textbf{Data} & How should the agent collect, annotate, and refine robot data? & \questionsection{data} \\
3. & \textbf{Environment} & What worlds, tasks, and tests should the agent build? & \questionsection{environment} \\
4. & \textbf{Harness} & How should robot capabilities be orchestrated, recovered, and evolved? & \questionsection{harness}
\end{tabular*}
\end{tcolorbox}
% \caption{Key questions across the four functional domains of agentic robots.}
\label{fig:agentic-key-questions}
\end{figure}

The four questions above guide our discussion of how agents contribute to robotic systems. Section~\ref{sec:model} examines how agents support model learning and robot decision-making, while Section~\ref{sec:data} explains how they collect, annotate, and refine robot data. Section~\ref{sec:environment} considers the worlds, tasks, and tests agents construct for learning and evaluation. Section~\ref{sec:harness} examines how agents coordinate executable capabilities, monitor execution, recover from failures, and retain useful knowledge for subsequent tasks.

\paragraph{\textbf{Contributions.}}
This paper makes three main contributions:
\begin{itemize}

\item \textbf{A Unified Illustration for Agentic Robotics.}
We trace the evolution from robot policies to agentic
systems and clarify the role of agents in robot task
execution and capability development. We introduce a
taxonomy spanning \textbf{Model}, \textbf{Data},
\textbf{Environment}, and \textbf{Harness}, bringing
these research directions into a common framework.

\item \textbf{A Systematic Analysis of Agentic Methods.}
We synthesize the mechanisms through which agents
acquire, coordinate, and improve robotic capabilities.
By comparing representative approaches across the
taxonomy, we explain how agent decisions depend on
available tools, robot executors, and interaction
feedback, and examine the strengths and limitations
of these approaches.

\item \textbf{A Research Agenda for Practical Agentic Robots.}
Drawing on this analysis, we outline research directions
for establishing agents' contributions and translating
current advances into reliable, efficient robot operation.
The agenda connects fair evaluation with the challenges
of physical deployment and generalization across
robotic platforms.

\end{itemize}

\section{Evolution of Robot Learning}
\label{sec:policies-to-systems}

This section reviews how robot learning connects scene representations,
  multimodal knowledge, and predictions of environmental change to action,
  before turning to their coordination within agentic systems. \textbf{(1) Conventional
  Robot Policies: Learning Executable Behavior.} These policies learn actions
  from demonstrations or reward-based experience. We distinguish image-based and
  3D-aware approaches by how they represent the scene for action prediction. \textbf{(2)
  Vision-Language-Action Models: Connecting Knowledge to Action.} VLAs connect
  pretrained vision-language representations to robot action generation. We
  examine action representations and how memory, intermediate reasoning, and
  future prediction guide or support action learning and execution. \textbf{(3) World
  Models and World-Action Models: Anticipating Physical Outcomes.} We examine
  world models as simulated environments, representations of evolving 3D scenes,
  and sources of visual plans, alongside world-action models that jointly learn
  future visual prediction and action generation. \textbf{(4) Agentic Robots: Organizing
  Execution and Improvement.} Agentic systems extend model-mediated decisions to
  selecting and coordinating capabilities, acquiring experience, and using
  feedback to guide subsequent execution and capability improvement. These
  approaches describe overlapping research traditions and complementary
  capabilities that can coexist in one system.

\subsection{Conventional Robot Policies}
\label{sec:conventional-policy-learning}

Conventional robot policies learn from demonstrations or reward-based experience. We group visual policies by scene representation: image-based policies use image features, while 3D-aware policies introduce geometric structure for action prediction. This distinction concerns scene representation, not action dimensionality.

\paragraph{\textbf{Image-based Policies.}}
Early policies learn actions from camera images. Guided policy search~\citep{survey_ax_1504_00702} maps images and robot state to motor torques, while QT-Opt~\citep{survey_ax_1806_10293} learns an image-conditioned action-value function for grasping. RoboMimic~\citep {survey_ax_2108_03298} studies visual behavior cloning from offline demonstrations. Visual pretraining supplies reusable policy features: R3M~\citep{survey_ax_2203_12601} pretrains an image encoder on human videos for downstream imitation learning. Beyond visual encoding, ACT~\citep{survey_ax_2304_13705} and Diffusion Policy~\citep{survey_ax_2303_04137} model action sequences with a conditional variational autoencoder and conditional diffusion, respectively. Task conditioning extends these policies to multiple behaviors: BC-Z~\citep{survey_ax_2202_02005} and RT-1~\citep{survey_ax_2212_06817} use task descriptions, while VIMA~\citep{survey_ax_2210_03094} combines multimodal prompts with object-centered image tokens. Cross-robot datasets broaden training: RT-1-X~\citep{survey_ax_2310_08864} draws on cross-robot data from Open X-Embodiment, and Octo~\citep{survey_ax_2405_12213} combines heterogeneous image observations with a transformer and a diffusion action~head.

\paragraph{\textbf{3D-aware Policies.}}
These policies use scene geometry to predict actions. Transporter Networks~\citep{survey_ax_2010_14406} construct orthographic color-height maps from RGB-D observations to predict spatial pick-and-place actions, using a 2.5D scene representation. PerAct~\citep{survey_ax_2209_05451} represents observations and action locations with 3D voxels, while RVT~\citep{survey_ax_2306_14896} renders virtual views of a reconstructed point cloud and combines their predictions in 3D. Act3D~\citep{survey_ax_2306_17817} lifts image features into a 3D feature field and samples action locations from coarse to fine. These geometric representations also support generative action models: DP3~\citep{survey_ax_2403_03954} conditions action diffusion on compact features encoded from sparse point clouds, whereas 3D Diffuser Actor~\citep{survey_ax_2402_10885} uses spatially located scene tokens and 3D attention to denoise end-effector trajectories. Mamba Policy~\citep{survey_ax_2409_07163} retains point-cloud conditioning and combines selective state-space models with attention in its denoising network. Other methods learn geometric representations for downstream policies. GSRL~\citep{survey_ax_2404_07950} pretrains and freezes a generalizable Gaussian predictor, then supplies its predicted 3D Gaussian scene representation to offline reinforcement-learning algorithms. Structural latent point pretraining~\citep{survey_ax_2605_21258} regularizes point coordinates and features with a point-wise latent variational autoencoder and uses Gaussian rendering supervision; its latent representation is then supplied to an ACT policy. Scene representation, learning objective, and action decoder remain distinct, combinable design choices.

\subsection{Vision-Language-Action Models}
\label{sec:vision-language-action}

\paragraph{\textbf{General VLA Models.}}
VLAs adapt large-scale pretrained vision-language backbones to generate robot actions, as in RT-2~\citep{survey_ax_2307_15818}, OpenVLA~\citep{survey_ax_2406_09246}, and RoboFlamingo~\citep{survey_ax_2311_01378}. Earlier policies already used language conditioning and pretrained encoders; VLAs connect broader knowledge of objects, relations, and instructions to action prediction through training on robot trajectories. $\tau_0$-VLA~\citep{survey_ax_2608_16885} combines a high-level subtask policy with a flow-matching action expert, using execution memory and selective world-model-guided test-time search for subtask selection.

\paragraph{\textbf{Action Representation.}}
A central choice is how to decode robot actions. RT-2~\citep{survey_ax_2307_15818} and OpenVLA~\citep{survey_ax_2406_09246} discretize continuous actions into tokens for sequence generation. Diffusion- and flow-based models instead condition a continuous action decoder on visual and language features. Examples include
RDT-1B~\citep{survey_ax_2410_07864},
$\pi_0$~\citep{survey_ax_2410_24164},
$\pi_{0.5}$~\citep{survey_ax_2504_16054},
CogACT~\citep{survey_ax_2411_19650},
TinyVLA~\citep{survey_ax_2409_12514},
GR00T N1~\citep{survey_ax_2503_14734},
SmolVLA~\citep{survey_ax_2506_01844},
DexVLA~\citep{survey_ax_2502_05855},
InternVLA-M1~\citep{survey_ax_2510_13778},
LingBot-VLA~\citep{survey_ax_2601_18692}, and
Qwen-VLA~\citep{survey_ax_2605_30280}.
FAST~\citep{survey_ax_2501_09747} compresses action sequences into compact tokens, while OpenVLA-OFT~\citep{survey_ax_2502_19645} combines parallel decoding and action chunking to predict multiple future actions in one forward pass.
Vela~\citep{survey_ax_2610_05230} represents future motion with cubic B-splines, using a flow-matching expert to jointly predict control points and a motion-dependent temporal span.
StarVLA~\citep{survey_ax_2604_05014} provides a modular open-source framework for training and evaluating combinations of vision-language backbones and action decoders.

\paragraph{\textbf{VLA with Memory.}}
History and memory provide temporal context when a single observation does not reveal earlier actions or task progress. RoboFlamingo~\citep{survey_ax_2311_01378} aggregates successive vision-language features with an LSTM policy head. MemoryVLA~\citep{survey_ax_2508_19236} retrieves perceptual and cognitive features from a memory bank and fuses them with current representations to condition a diffusion action expert. MEM~\citep{survey_ax_2603_03596} combines a video encoder for short-term observation history in its low-level policy with language summaries of past events in its high-level policy. NativeMEM~\citep{survey_ax_2607_06678} reuses the pretrained VLA's vision encoder to compress each historical frame from each camera view into a single memory token, allowing the policy to attend to long-term visual history alongside current observations.
%These designs provide recurrent states, retrieved memories, complementary visual and textual context, or compressed history tokens to guide robot actions during execution.

% \paragraph{\textbf{VLA with Reasoning.}}
% Intermediate reasoning guides action prediction through motion descriptions, semantic subtasks, or language reasoning traces, as in RT-H~\citep{survey_ax_2403_01823}, ECoT~\citep{survey_ax_2407_08693}, $\pi_{0.5}$~\citep{survey_ax_2504_16054}, OneTwoVLA~\citep{survey_ax_2505_11917}, and Gemini Robotics 1.5~\citep{survey_ax_2510_03342}. Visual subgoals provide related guidance: CoT-VLA~\citep{survey_ax_2503_22020} predicts an intermediate goal image and conditions subsequent actions on it. These approaches structure action generation around intermediate descriptions or goals.
\paragraph{\textbf{VLA with Reasoning.}}
% Intermediate reasoning connects high-level task instructions to robot action prediction by specifying what the robot should do next and which scene information matters for execution. 
Intermediate reasoning and semantic prediction provide explicit guidance between task instructions and robot actions.
RT-H~\citep{survey_ax_2403_01823} and $\pi_{0.5}$~\citep{survey_ax_2504_16054} introduce language descriptions of motions or semantic subtasks that condition subsequent actions. ECoT~\citep{survey_ax_2407_08693} further combines task reasoning with spatial information, including object locations and end-effector positions, to guide action prediction. OneTwoVLA~\citep{survey_ax_2505_11917} and Gemini Robotics 1.5~\citep{survey_ax_2510_03342} integrate reasoning with action generation, allowing task interpretation to inform the robot's next behavior. Visual subgoals provide a complementary representation: CoT-VLA~\citep{survey_ax_2503_22020} predicts an intermediate goal image and conditions subsequent actions on the desired scene configuration. Together, these approaches provide explicit intermediate guidance at different levels, from task decomposition and motion descriptions to spatial reasoning and visual goals, connecting the robot's understanding of the task with the actions required to execute it.

\paragraph{\textbf{VLA with Imagination.}}
% Future prediction connects VLA action learning to environmental change. GR-1~\citep{survey_ax_2312_13139}, GR-2~\citep{survey_ax_2410_06158}, UP-VLA~\citep{survey_ax_2501_18867}, WorldVLA~\citep{survey_ax_2506_21539}, and RynnVLA-002~\citep{survey_ax_2511_17502} jointly learn action prediction and future-image generation. FLARE~\citep{survey_ax_2505_15659}, VLA-JEPA~\citep{survey_ax_2602_10098}, and World Guidance~\citep{survey_ax_2602_22010} instead predict future observation latents without reconstructing images. BagelVLA~\citep{survey_ax_2602_09849} combines language reasoning, visual prediction, and action generation in a shared framework.
%These developments bring predictions of environmental change into robot policy learning, connecting VLAs to the world models and world-action models discussed next.
Future prediction connects VLA action learning to how scenes evolve during robot interaction. GR-1~\citep{survey_ax_2312_13139}, GR-2~\citep{survey_ax_2410_06158}, UP-VLA~\citep{survey_ax_2501_18867}, WorldVLA~\citep{survey_ax_2506_21539}, and RynnVLA-002~\citep{survey_ax_2511_17502} combine action learning with future-image prediction, using visual changes as additional supervision for robot control. FLARE~\citep{survey_ax_2505_15659}, VLA-JEPA~\citep{survey_ax_2602_10098}, and World Guidance~\citep{survey_ax_2602_22010} instead model future observations in latent space, learning predictive features or state transitions without reconstructing images. These objectives incorporate future information into policy representations, although future images need not be generated during execution. BagelVLA~\citep{survey_ax_2602_09849} further interleaves language reasoning, visual prediction, and action generation, combining semantic task guidance with visual anticipation within the action execution loop.

\subsection{World Models and World-Action Models}
\label{sec:world-action-models}

World models learn how the environment changes in response to robot actions. Predicting future states or observations supports planning over candidate actions and policy learning from imagined interactions~\citep{survey_ax_1610_00696,survey_ax_1811_04551,survey_ax_2206_14176,survey_ax_2310_16828}. DINO-WM~\citep{survey_ax_2411_04983} and V-JEPA~2-AC~\citep{survey_ax_2506_09985} predict action-conditioned visual features and optimize action sequences toward visual goals through model-predictive control.

\paragraph{\textbf{World Model as Environment.}}
Action-conditioned video models simulate interactions for policy learning and evaluation.
UniSim~\citep{survey_ax_2310_06114} simulates visual outcomes of instructions and low-level controls to train embodied planners and policies.
Cosmos~\citep{survey_ax_2501_03575} provides pretrained diffusion and autoregressive video models for adaptation to robot-specific prediction. Cosmos-Predict2.5~\citep{survey_ax_2511_00062} unifies text-, image-, and video-conditioned generation and supports action-conditioned robot rollouts after adaptation. Genie Envisioner~\citep{survey_ax_2508_05635} and DreamDojo~\citep{survey_ax_2602_06949} use action-conditioned video prediction for policy evaluation; the latter also learns interaction dynamics from human videos with inferred latent actions before adapting to target robots.
Cosmos~3~\citep{survey_ax_2606_02800} extends the series toward shared multimodal understanding, world generation, and robot action modeling.

\paragraph{\textbf{4D World Model.}}
4D world modeling captures how 3D scene geometry evolves over time, supporting geometric planning and interaction simulation.
PointWorld~\citep{survey_ax_2601_03782} represents robot actions as 3D point flows and predicts the resulting scene-point displacements over a temporal horizon, supporting sampling-based trajectory optimization.
TesserAct~\citep{survey_ax_2504_20995} jointly generates RGB, depth, and surface-normal videos, reconstructs temporally consistent 3D point clouds, and uses a point-cloud inverse dynamics model to infer actions.
4DGen~\citep{survey_ax_2507_01099} uses cross-view pointmap supervision to generate geometrically consistent multi-view RGB-D futures, from which a pose tracker recovers end-effector trajectories.
MVISTA-4D~\citep{survey_ax_2602_09878} generates synchronized, view-consistent RGB-D futures from a single-view observation. It optimizes a trajectory latent at test time to match the imagined future and refines the decoded actions with residual inverse dynamics.
Kinema4D~\citep{survey_ax_2603_16669} converts robot commands into kinematically computed 4D trajectories and uses their projected pointmaps to condition joint RGB and pointmap generation, simulating the surrounding environment's response.

\paragraph{\textbf{World Model as Planner.}}
World models supply visual plans. Inverse dynamics maps predicted observations or latents to robot actions.
UniPi~\citep{survey_ax_2302_00111} and RoboDreamer~\citep{survey_ax_2404_12377} use inverse dynamics to translate language-conditioned visual plans into actions.
Seer~\citep{survey_ax_2412_15109} couples visual foresight and inverse dynamics in an end-to-end model, conditioning action prediction on latent representations of the forecasted visual state.
GE-Act~2.0~\citep{survey_ax_2609_05588} predicts future visual latents in a single generator pass, then combines them with current visual latents and proprioception in an inverse dynamics model to produce an action chunk.
This connection also extends to geometric plans: TesserAct~\citep{survey_ax_2504_20995} infers actions from reconstructed point-cloud sequences, while MVISTA-4D~\citep{survey_ax_2602_09878} combines test-time trajectory-latent optimization with residual inverse dynamics to refine actions toward the imagined future.
These approaches depend on both the feasibility of the predicted future and the inverse ability to translate it into robot-specific control.

\paragraph{\textbf{Joint Video--Action Modeling.}}
World-action models (WAMs) combine future visual prediction with robot action learning, linking behavior to its environmental consequences. UVA~\citep{survey_ax_2503_00200}, UWM~\citep{survey_ax_2504_02792}, and Motus~\citep{survey_ax_2512_13030} jointly learn video prediction and action generation within a shared framework.
$\tau_0$-WM~\citep{survey_ax_2606_01027} unifies video--action generation and action-conditioned simulation. At test time, predicted task progress scores candidate futures; when initial candidates are unreliable, a selected future guides action refinement.
DreamZero~\citep{survey_ax_2602_15922},
LingBot-VA~\citep{survey_ax_2601_21998},
Cosmos Policy~\citep{survey_ax_2601_16163},
GigaWorld-Policy~\citep{survey_ax_2603_17240}, and
Being-H0.7~\citep{survey_ax_2605_00078} connect action learning to future visual information through observation prediction or latent alignment.
Fast-WAM~\citep{survey_ax_2603_16666} learns efficient action generation through video co-training without future imagination at inference.
ImageWAM~\citep{survey_ax_2606_19531} guides action generation with pretrained image-editing features without decoding future images.
MaskWAM~\citep{survey_ax_2606_13515} combines optional initial mask prompts with joint future mask, RGB, and action prediction to focus control on task-relevant regions.
A recent direction jointly models language, video, and action: UniWAM~\citep{survey_ax_2610_02054} connects a physical reasoner, world generator, and action predictor through joint multimodal attention, learning language outputs alongside future visual predictions and actions.

% An important distinction is whether future prediction is used during training, action inference, or both. Joint video--action training does not always require generating future images during execution: decoupled decoding and video co-training can support action prediction directly from observed context~\citep{survey_ax_2503_00200,survey_ax_2603_16666}. These designs distinguish learning from future observations from explicitly generating them to select actions.
% Using these models within a larger task also requires deciding which capability to invoke and how to respond to execution results. This brings the discussion from predictive action models to agentic robot systems.

\subsection{Agentic Robots}
\label{sec:agentic-robot-systems}

Agentic robots organize robotic execution and capability improvement. Given a goal, they coordinate capabilities, interpret execution feedback, and adjust subsequent operations~\citep{survey_ax_2204_01691,survey_ax_2207_05608,survey_ax_2209_07753,survey_ax_2502_19417}. Learned policies, including VLAs and world-action models, supply executable capabilities within this process, shifting attention to how behavior is selected, coordinated, and revised during execution.

This perspective also extends to data collection, model training, and experimental deployment. Agentic decisions organize experience acquisition, guide learning experiments, and use evaluation feedback to improve capabilities~\citep{survey_ax_2310_12931,survey_ax_2606_19980}. These decisions connect current execution with improvements for future tasks. The following sections examine four functional domains, \textbf{Model}, \textbf{Data}, \textbf{Environment}, and \textbf{Harness}, that describe where these decisions act and how robotic capabilities are acquired, used, and improved.
\FloatBarrier

\section{Foundations of Agentic Robots}
\label{sec:framing-mapping}

% Section~\ref{sec:policies-to-systems} motivates the need to organize reasoning, planning, and action prediction capabilities around a task. We now examine the decisions that govern their use and development, and how those decisions connect to execution. We first define the scope of agentic robots, then examine the interfaces connecting model decisions to execution, and finally organize the literature by where those decisions act in the system.

% Section~\ref{sec:policies-to-systems} traces how robot policies incorporate vision--language knowledge, intermediate reasoning, and predictions of environmental change. Agentic robot systems extend these capabilities to the coordination of task execution and robot learning.
% In this section, we first define the scope of agentic robots, then examine the interfaces connecting model decisions to execution, and finally organize the literature by where those decisions act in the system.
%This section establishes a framework for describing these decisions: we define the scope of agentic robots, examine the interfaces through which model decisions guide robot execution, and organize the literature into four functional domains---Model, Data, Environment, and Harness.

The approaches discussed in Section~\ref{sec:policies-to-systems}
provide capabilities for perception, reasoning, action
prediction, and world modeling. Building agentic robots
requires coordinating these capabilities through task
planning, tool use, and execution feedback, while also
supporting the development of new robot capabilities.
To characterize these systems, this section first defines
agentic robots and their scope, then introduces a taxonomy
spanning Model, Data, Environment, and Harness, and finally
compares agentic robotics with related concepts.

\subsection{Definition and Scope}
\label{sec:definition-scope}

We define \emph{agentic robots} as robotic systems in which agents autonomously plan and coordinate task execution using environmental observations and execution feedback. Given a goal, these agents determine the steps needed to achieve it, select and coordinate actions, skills, or tools, and monitor progress without requiring a human to specify every step. As execution proceeds, they use new observations and outcomes to decide whether to continue, gather more information, revise the plan, or recover from failure. We consider such systems on both physical robots and in simulation.

These capabilities can be implemented within an integrated model or through a combination of models, policies, programs, and tools. For example, when retrieving an object from a drawer, the agent may select an opening skill, check whether the drawer has opened, and decide whether to retrieve the object or revise the opening attempt. The defining consideration is how the agent organizes and adjusts task execution to achieve the goal. Action prediction or low-level feedback control alone does not establish these capabilities.

Beyond task execution, agents can also contribute to developing robot capabilities. They may collect and refine robot data, construct tasks and environments, revise training procedures, or improve robot skills and execution software. Feedback from interaction and experiments can guide these changes and help improve performance on subsequent tasks. Our paper examines these contributions across Model, Data, Environment, and Harness, covering how agents support robot operation and capability development.

\begin{figure}[!t]
\centering
\includegraphics[width=\linewidth]{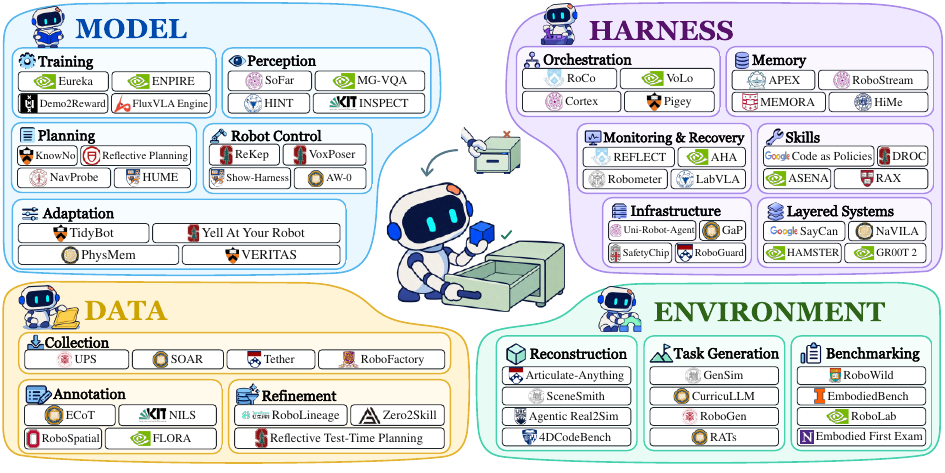}
\caption{Taxonomy and literature landscape of agentic robotic systems. Four functional domains cover seventeen non-exclusive categories: Model covers learning and robot-facing decisions. Data covers the acquisition and curation of data. Environment covers worlds, tasks, and evaluation. Harness covers how robotic capabilities are coordinated,
reused, and improved to support agentic task execution.}
\label{fig:taxonomy-landscape}
\end{figure}

\paragraph{\textbf{Comparison with Current Works.}}
PAPAV~\citep{survey_repo_showlab_papav} compares computer use and robot use through Perceive, Anticipate, Plan, Act, and Verify, emphasizing the consequences of physical embodiment. Salimpour et al.~\citep{survey_ax_2508_05294} examine model--robot integration through protocol interfaces, orchestration, and embedded policies. Jena et al.~\citep{survey_ax_2608_01851} survey robot-learning approaches spanning action-predicting weights and executable skills. Other surveys examine the software that organizes agent execution~\citep{survey_ax_2605_18747} and the sources, availability, and reliability of signals used to verify robot policies~\citep{survey_ax_2609_09250}. Our focus is how these components work together during robot use and capability building.

\subsection{Taxonomy}
\label{sec:taxonomy}
We organize existing work into four domains according to the roles agents play in developing and operating robotic systems: \emph{Model}, \emph{Data}, \emph{Environment}, and \emph{Harness}. \emph{Model} concerns how agents use and improve models for robotic perception, planning, control, and adaptation. \emph{Data} concerns how agents collect, annotate, and refine data for robot learning. \emph{Environment} concerns how agents construct scenes, generate tasks, and design evaluations for robot training and assessment. \emph{Harness} concerns how agents coordinate models, tools, and robot skills to execute tasks, using memory and execution feedback to monitor progress and recover from failures. Together, these domains describe the roles of agents across robot capability development and task execution. Figure~\ref{fig:taxonomy-landscape} presents the overall taxonomy, and Table~\ref{tab:taxonomy} defines the seventeen categories within these domains.

The taxonomy distinguishes functional contributions rather than assigning each robotic system to a single category. These contributions may span domains: Zero2Skill~\citep{survey_ax_2607_14047}, for example, links agent-directed data collection and preparation to subsequent robot policy training. They may also operate together within a domain, as in Harness VLA~\citep{survey_ax_2607_08448}, which integrates memory, orchestration, monitoring, and recovery around fixed executors. ENPIRE~\citep{survey_ax_2606_19980} illustrates how a research agent coordinates experiments to improve robot policies, connecting experiment management with capability development. We therefore discuss each work under the categories corresponding to its relevant functions, allowing a paper to appear in multiple categories when it makes contributions to more than one.

\subsection{Comparison with Related Concepts}
\label{sec:related-concepts}

Agentic robotics is closely related to embodied AI, autonomous robotics, and robot foundation models. These concepts overlap in their methods and applications, but emphasize different aspects of robotic intelligence. We clarify their relationships using the scope adopted in this paper.

\paragraph{\textbf{Agentic Robotics vs. Embodied AI.}}
Embodied AI studies intelligence through an agent's interaction with its environment, encompassing tasks such as exploration, navigation, manipulation, and embodied question answering. Both embodied AI and agentic robotics connect perception, reasoning, and action in physical or simulated environments~\citep{survey_ax_2103_04918,survey_ax_2402_02385}. The distinction concerns the emphasis placed on how these capabilities are organized. Embodied AI broadly investigates learning and acting through environmental interaction; our agentic perspective focuses on how agents select capabilities, coordinate task execution, and revise decisions using feedback. This focus is consistent with work examining foundation models as agents that invoke and coordinate robot functionality~\citep{survey_ax_2508_05294}. It also extends to capability development: AutoRT, for example, uses foundation models to propose and filter tasks for robot data collection~\citep{survey_ax_2401_12963}.

\paragraph{\textbf{Agentic Robotics vs. Autonomous Robotics.}}
Autonomous robotics concerns robots that carry out tasks without continuous human control. Its established research already integrates planning, acting, monitoring, observation, and learning to support deliberate behavior in changing environments. These functions therefore constitute shared foundations, rather than capabilities unique to agentic robotics~\citep{ingrand2017deliberation}. Within this broader field, our paper examines how foundation model agents participate in these functions by interpreting goals, coordinating skills or tools, and incorporating execution feedback into subsequent decisions. Inner Monologue illustrates this connection by returning observations and skill outcomes to a language model for continued planning~\citep{survey_ax_2207_05608}. Accordingly, ``agentic'' identifies how decisions are organized within the system; it does not imply complete independence from human guidance or a higher level of autonomy.

\paragraph{\textbf{Agentic Robotics vs. Robot Foundation Models.}}
Robot foundation models aim to acquire reusable capabilities for robotic tasks through broad training and adaptation, supporting functions ranging from perception and planning to control~\citep{survey_ax_2312_08782}. Their capabilities can include reasoning and decision-making: PaLM-E integrates multimodal observations into embodied planning, while RT-2 connects visual and linguistic knowledge to action generation and explores intermediate reasoning before action~\citep{survey_ax_2303_03378,survey_ax_2307_15818}. The distinction is therefore between the capabilities learned by a model and how those capabilities participate in an ongoing task. Agentic robotics examines how goals, available actions, and execution feedback govern subsequent decisions, whether implemented within one model or across several components. Hi Robot illustrates their compatibility by combining a VLM that updates intermediate instructions with a VLA that executes them~\citep{survey_ax_2502_19417}.

The following four sections review these domains in turn. Within each domain, we group methods by the functions they serve and explain their main approaches through representative work.

% Generated by scripts/current_manuscript.py from its approved TABLE1_DEFINITIONS.
\begin{table}[!htb]
\centering
\TaxonomyTableStyle
\caption{Operational taxonomy of agentic robots. Four functional domains organize seventeen non-exclusive categories by the part of the robotic system shaped by model-mediated decisions.}
\label{tab:taxonomy}
\begin{tabular}{@{}>{\raggedright\arraybackslash}p{\dimexpr 0.17\linewidth-1.3333333333333333\tabcolsep\relax}>{\raggedright\arraybackslash}p{\dimexpr 0.17\linewidth-1.3333333333333333\tabcolsep\relax}>{\raggedright\arraybackslash}p{\dimexpr 0.66\linewidth-1.3333333333333333\tabcolsep\relax}@{}}
\toprule
\rowcolor{black!5}
\textbf{Domain} & \textbf{Category} & \textbf{Definition} \\
\midrule
\rowcolor{TSTModelLightest}
\TableDomain{Model} & \CategoryLabelWrap{ModelTraining}{Training} & Designs or revises rewards, objectives, training recipes, or learning experiments. \\[2pt]
\CategorySeparator
\rowcolor{TSTModelLightest}
 & \CategoryLabelWrap{ModelPerception}{Perception} & Interprets observations to recover task-relevant concepts, relations, geometry, or beliefs, and decides when additional evidence is needed. \\[2pt]
\CategorySeparator
\rowcolor{TSTModelLightest}
 & \CategoryLabelWrap{ModelPlanning}{Planning} & Chooses and revises subgoals, action sequences, dependencies, or task solutions as a task progresses. \\[2pt]
\CategorySeparator
\rowcolor{TSTModelLightest}
 & \CategoryLabelWrap{ModelRobotControl}{Robot Control} & Translates selected operations into robot-facing actions, poses, programs, or motion constraints for execution. \\[2pt]
\CategorySeparator
\rowcolor{TSTModelLightest}
 & \CategoryLabelWrap{ModelAdaptation}{Adaptation} & Uses experience to update model parameters or task-specific conditioning and decision rules. \\[2pt]
\addlinespace[4pt]
\rowcolor{TSTDataLightest}
\TableDomain{Data} & \CategoryLabelWrap{DataCollection}{Collection} & Acquires or generates new robot experience for learning or later use. \\[2pt]
\CategorySeparator
\rowcolor{TSTDataLightest}
 & \CategoryLabelWrap{DataAnnotation}{Annotation} & Interprets recorded experience and adds structured supervision, such as text, spatial labels, interaction targets, or outcomes. \\[2pt]
\CategorySeparator
\rowcolor{TSTDataLightest}
 & \CategoryLabelWrap{DataRefinement}{Refinement} & Improves existing experience by selecting useful records or segments and revising unreliable supervision. \\[2pt]
\addlinespace[4pt]
\rowcolor{TSTEnvironmentLightest}
\TableDomain{Environment} & \CategoryLabelWrap{EnvironmentReconstruction}{Reconstruction} & Constructs or revises executable scenes, assets, geometry, and task-relevant physical properties. \\[2pt]
\CategorySeparator
\rowcolor{TSTEnvironmentLightest}
 & \CategoryLabelWrap{EnvironmentTaskGeneration}{Task Generation} & Defines or generates goals, constraints, curricula, or task distributions for practice, learning, or execution. \\[2pt]
\CategorySeparator
\rowcolor{TSTEnvironmentLightest}
 & \CategoryLabelWrap{EnvironmentBenchmarking}{Benchmarking} & Constructs or adapts tests, perturbations, metrics, evaluators, and evaluation procedures. \\[2pt]
\addlinespace[4pt]
\rowcolor{TSTHarnessLightest}
\TableDomain{Harness} & \CategoryLabelWrap{HarnessOrchestration}{Orchestration} & Selects models, policies, skills, or tools and governs their calls and handoffs. \\[2pt]
\CategorySeparator
\rowcolor{TSTHarnessLightest}
 & \CategoryLabelWrap{HarnessMemory}{Memory} & Stores, updates, and retrieves interaction history or knowledge for later decisions. \\[2pt]
\CategorySeparator
\rowcolor{TSTHarnessLightest}
 & \CategoryLabelWrap{HarnessMonitoringandRecovery}{Monitoring and Recovery} & Assesses progress, completion, failure, or constraints and organizes corrective responses. \\[2pt]
\CategorySeparator
\rowcolor{TSTHarnessLightest}
 & \CategoryLabelWrap{HarnessSkills}{Skills} & Creates, validates, retains, revises, or retires reusable procedural knowledge and executable behaviors. \\[2pt]
\CategorySeparator
\rowcolor{TSTHarnessLightest}
 & \CategoryLabelWrap{HarnessInfrastructure}{Infrastructure} & Exposes, serves, and gates tools and policies through interfaces, services, and runtime checks. \\[2pt]
\CategorySeparator
\rowcolor{TSTHarnessLightest}
 & \CategoryLabelWrap{HarnessLayeredSystems}{Layered Systems} & Divides information, authority, and operating rates across planners, learned policies, and analytic control. \\[2pt]
\bottomrule
\end{tabular}
\end{table}

\FloatBarrier

\clearpage
\section{Model: What Should the Agent Learn, Understand, Decide, and Control?}
\label{sec:model}

\begin{figure}[!hb]
\centering
\includegraphics[width=\linewidth]{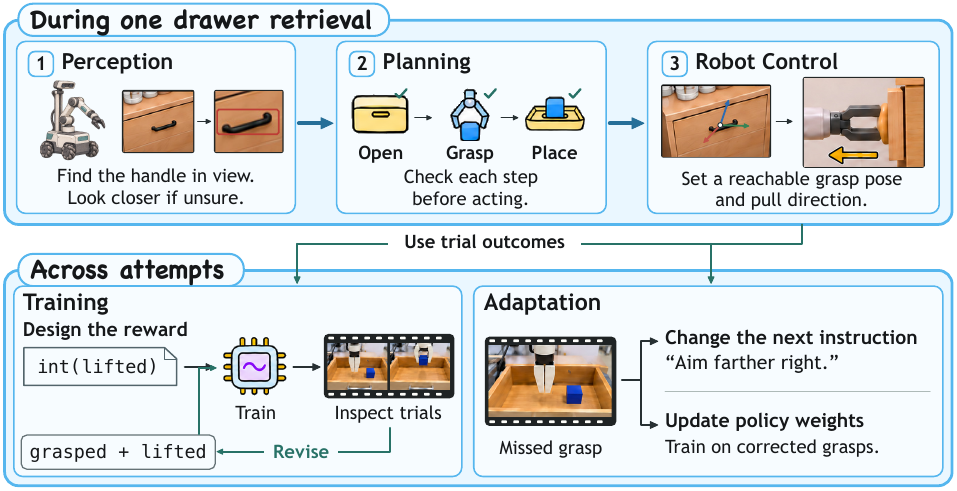}
\caption{An agent decides how to retrieve a cube from a drawer and uses trial outcomes to guide later attempts. During retrieval, it locates the handle, checks the plan, and specifies the grasp and pull. Across attempts, it can revise the training reward, change the next instruction, or update the policy weights.}
\label{fig:model-decisions}
\end{figure}

Figure~\ref{fig:model-decisions} follows a drawer retrieval from the first view of the handle to a command the robot can execute. During the attempt, Perception determines whether another view is needed, Planning checks the proposed steps, and Robot Control specifies the grasp and pull. Trial outcomes can also guide changes across attempts. Training concerns how the agent sets up learning, including what the reward measures. Adaptation concerns how experience changes the instructions or policy parameters that guide later actions.

\FloatBarrier

\subsection{Training}
\label{sec:model-training}

The agent acts on the learning process rather than in the environment. It specifies rewards, objectives, configurations, or experiments. During this process, an optimizer consumes the resulting artifact, and robot rollouts reveal whether the policy learned the desired behavior. In iterative workflows, the agent uses these results to decide what to revise next. Existing systems differ mainly in the scope of this editable artifact: 1) \textit{reward-design} methods keep the optimizer fixed and revise only the signal it consumes, whereas 2) \textit{program- and experiment-level} methods extend revision to policy code, training pipelines, and 3) the \textit{experimental workflow} itself.

\paragraph{\textbf{Reward Design.}}
Reward transfers a behavioral objective into a signal an optimizer can use. Eureka~\citep{survey_ax_2310_12931} established the basic loop: a language model writes reward code and revises it using training feedback. Such feedback, however, can confirm that a terminal predicate is satisfied without revealing whether execution actually followed the instruction. RDA~\citep{survey_ax_2606_01672} further addresses this gap with semantic analysis of the resulting behavior. It decomposes the instruction into subtasks, generates executable reward candidates, trains policies, and evaluates rollout videos alongside subtask reward logs. 

Eureka and RDA evaluate candidate rewards through full policy training, making each evaluation computationally expensive. Later methods reduce this cost in two ways: they validate rewards with cheaper evidence, or they narrow what is searched. EvoNav~\citep{survey_ax_2605_11859} screens reward programs with rules and short proxy training, assuming that the cheap tests preserve the ranking of useful candidates. Then, CRWM~\citep{survey_ax_2606_23280} conditions a one-pass reward-code generator on a causal prior derived from offline reward interventions. It pays for the prior up front, and RL outcomes never feed back into the reward. Demo2Reward~\citep{survey_ax_2606_00083} uses a VLM meta-critic to tune a VLM reward prompt on labelled demonstration prefixes before policy learning. In ReCoVLA~\citep{survey_ax_2606_09630}, a VLM maps failure observations to descriptors, a deterministic compiler binds them to predefined state-based reward components, and RL trains failure-specific residual policies while the base VLA stays frozen. Generally, each method trades the generality of full-training feedback for a cheaper but more constrained source of reliability: proxy rollouts, offline causal structure, labelled demonstrations, or predefined reward components.

A reward can also be inferred from what happened in a recorded trajectory. D\&A~\citep{survey_ax_2611_03622} uses the instruction and video to list what appears in the scene, including robot arms and objects. Given this list and the video frames, a describer reports what moved without seeing the task instruction. A separate assessor reads the instruction and description, specifies what successful completion would look like, and scores progress over time. These scores supply rewards for online RL or identify useful action segments in offline data (Section~\ref{sec:refinement}).

\paragraph{\textbf{Policy Training.}}
% Editing the policy or training process changes the learning procedures beyond reward. 
Beyond reward design, agents participate in robot policy development by implementing policies, organizing training experiments, and revising the learning process.
ENPIRE~\citep{survey_ax_2606_19980} first constructs reset, verification, and safety interfaces with human feedback, then exposes these interfaces to coding agents that inspect real rollouts and modify policy or training code. Behavior cloning, reinforcement learning, and heuristic policy synthesis are alternatives inside the research process, not simultaneous properties of every resulting controller. HARBOR~\citep{survey_ax_2606_08610} organizes simulation reinforcement learning into typed artifacts and gate-checked stages. Failed interface checks return actionable diagnostics; parallel tuning trials retain code, videos, and logs, and recurring lessons are retrieved for later experiments. ENPIRE depends on reliable physical resets; HARBOR depends on agreement among generated task, reward, and algorithm interfaces. Each exposes a different bottleneck to the experimental agent.

Real2Worlds2Real~\citep{survey_ax_2611_31098} lets the agent use failed simulation runs to decide whether to collect more demonstrations or revise a reward. In a cup-stacking experiment, the authors enabled corrective data collection in a fixed training setup. The agent's planning programs supplied expert actions at states reached by a policy trained on 2,000 demonstrations. After training on the added corrections, the policy succeeded in 73 of 100 unseen simulated scenes, up from 49. The authors also compared their system with ENPIRE~\citep{survey_ax_2606_19980} and ASPIRE~\citep{survey_ax_2607_00272} on four real-robot tasks. All three had access to the same simulation environments and 60 real demonstrations per task. Real2Worlds2Real achieved the highest success rate on each task.

GR00T~2~\citep{nvidia2026gr00t2} uses AutoRL to revise its post-training algorithm. The training logs showed that the critic predicted returns poorly. The agent replaced the critic with a baseline that compares returns within a group of rollouts. It later grouped rollouts by initial scene and seed and added past returns for each group. A group of failures could then signal a decline from earlier attempts in the same scene. Each round kept at most one candidate that beat the current best by a preset margin. Over about 35 days, the best recipe gained 10 percentage points over the starting checkpoint on the seen atomic and composite RoboCasa365 tasks.

As agents explore more training choices, researchers need to direct the search and reproduce the experiments.
\citep{survey_ax_2603_27416} uses a coding agent to improve quadruped policies. Researchers direct each round. The agent edits the code, launches experiments, and inspects the logs. Nautilus~\citep{survey_ax_2605_11665} instead standardizes policy, robot, and benchmark contracts and isolates their execution environments, allowing a coding agent to assemble and validate learning workflows.

An acting agent can also provide examples for a student policy. In EmbodyAgent~\citep{survey_ax_2611_26196}, a teacher produces successful episodes that show when to request an observation and what Python code to execute. After supervised fine-tuning (SFT), the student practices online with rewards for task completion and penalties for code errors. The authors evaluated the 9B model on 100 held-out initial states across ten LIBERO Spatial tasks. It succeeded in 14 trials after SFT and in 84 after 64 rounds of RL. RoboUniverse-35B-A3B~\citep{survey_ax_2611_31002} uses supervised training alone. About 17,000 teacher trajectories collected in simulation train the student's language backbone while its vision encoder and aligner stay fixed. With the same Claude Code harness and one attempt on each of 500 held-out tasks, success rose from 0.8\% to 21.2\%.

CaPtain-4B~\citep{survey_ax_2610_40799} learns from teacher trajectories and then from its own Python requests. During RL, a critic estimates the chance of task success from the simulator state before and after a request. It keeps the request number fixed so elapsed time alone earns no credit. Training combines the change in that estimate and the request cost with the relative outcomes of four episodes from the same initial state. The deployed model runs without the critic. Across 240 held-out initial states from 40 LIBERO tasks, goal completion rose from 55.4\% after SFT to 73.8\% after RL. When the model also had to call \texttt{finish()}, the rates were 52.1\% and 67.5\%. PEAR~\citep{survey_ax_2611_18562} instead uses a single success reward for each episode. The authors start with short LIBERO tasks, add the long tasks, and then introduce two RoboDojo tasks.

Overall, agents contribute to policy training by running experiments, connecting training tools, and producing examples for a student policy.

\paragraph{\textbf{Workflow.}}
A training workflow sets the objective and determines how progress is measured. In AgenticRobotics~\citep{survey_ax_2608_07555}, the objective stays fixed and the controller measures each result. The system can resume an interrupted run and requires a new quality check when the underlying tool changes. In LWD~\citep{survey_ax_2605_00416}, fleet successes, failures, and human interventions instead update a pretrained generalist VLA through offline-to-online post-training. The two workflows intervene at different levels: policy post-training from fleet experience and management of the experiments used to assess revisions.

That separation of decision and measurement also matters when an agent edits more than the training recipe. Reward, prompt, tool, and data edits can otherwise change both the intervention and its judge. RoboLineage~\citep{survey_ax_2606_22142} makes this transition explicit by storing rollouts, reviews, dataset decisions, training runs, policy metadata, and deployment recommendations as typed artifacts. From the perspective of foundational infrastructure, FluxVLA Engine~\citep{survey_ax_2609_17210} supplies configuration-resolved contracts that connect data, training, evaluation, and deployment. EmbodiedSkills~\citep{survey_ax_2609_01281} then provides separate interfaces through which planning, verification, and execution components can be trained or adapted. The useful design principle is to expose objectives, recipes, and experiments for revision while keeping the measurements used to judge them stable. The learning choices are analyzed here under Training; the execution contracts and measurement interfaces that support those choices are discussed as capability-building infrastructure in Section~\ref{sec:harness-infrastructure}. Related works are compared in Table~\ref{tab:model-training}.

\par
\begingroup
\mechanismtableglue
\CompactTableStyle
\setlength{\LiteratureWidth}{\dimexpr\linewidth-8\tabcolsep\relax}

\setlength{\LTpre}{4pt}
\setlength{\LTpost}{4pt}
\setlength{\LTcapwidth}{\linewidth}
\captionsetup{singlelinecheck=false,justification=raggedright}
\LiteratureContinuationSetup
\begin{longtable}{@{}>{\RaggedRight\arraybackslash}p{0.2313\LiteratureWidth}>{\centering\arraybackslash}p{0.0750\LiteratureWidth}>{\RaggedRight\arraybackslash}p{0.2313\LiteratureWidth}>{\RaggedRight\arraybackslash}p{0.2128\LiteratureWidth}>{\RaggedRight\arraybackslash}p{0.2498\LiteratureWidth}@{}}
\caption[Comparison of works on training]{Comparison of works on training (Section~\ref{sec:model-training}). Training metrics are collected during policy learning; evaluation metrics assess the resulting policies.}\label{tab:model-training}\\
\LiteratureTopRule{Model}
\rowcolor{TSTModelLight}\textbf{Work} & \textbf{Link} & \textbf{Target} & \textbf{Approach} & \textbf{Data / feedback} \\
\midrule
\endfirsthead
\multicolumn{5}{@{}l}{\tablename~\thetable\ (continued)} \\
\LiteratureTopRule{Model}
\rowcolor{TSTModelLight}\textbf{Work} & \textbf{Link} & \textbf{Target} & \textbf{Approach} & \textbf{Data / feedback} \\
\midrule
\endhead
\midrule
\multicolumn{5}{r@{}}{Continued on next page} \\
\endfoot
\bottomrule
\endlastfoot
% Source: https://arxiv.org/abs/2310.12931v2 | §§3.1–3.3, especially §3.3 Reward Reflection
Eureka \mbox{\citep{survey_ax_2310_12931}} & \LiteratureLinks{https://proceedings.iclr.cc/paper_files/paper/2024/hash/70c26937fbf3d4600b69a129031b66ec-Abstract-Conference.html}{https://github.com/eureka-research/Eureka} & Reward code & Evolutionary search & Training metrics \\[0pt]
\rowcolor{black!3}
% Source: https://arxiv.org/abs/2603.27416v1 | §IV autonomous research waves and experiment triage
ADARLR \mbox{\citep{survey_ax_2603_27416}} & \LiteratureLinks{https://arxiv.org/abs/2603.27416v1}{} & RL code; reward weights & Iterative refinement; hyperparameter search & Training metrics; failure cases \\[0pt]
% Source: https://arxiv.org/pdf/2605.00416v4 | PDF p8, Sec IV-D: full offline fine-tuning; online action-expert and value/critic updates
LWD \mbox{\citep{survey_ax_2605_00416}} & \LiteratureLinks{https://arxiv.org/abs/2605.00416v4}{} & VLA; value models & RL & Rollouts; human feedback \\[0pt]
\rowcolor{black!3}
% Source: https://arxiv.org/abs/2605.11859v1 | §§4.1–4.3 three-stage reward evolution
EvoNav \mbox{\citep{survey_ax_2605_11859}} & \LiteratureLinks{https://arxiv.org/abs/2605.11859v1}{} & Navigation rewards & Evolutionary search; iterative refinement & Offline data; evaluation metrics \\[0pt]
% Source: https://arxiv.org/abs/2605.11665v1 | §§3.1–3.2 module generation and executable validation
Nautilus \mbox{\citep{survey_ax_2605_11665}} & \LiteratureLinks{https://arxiv.org/abs/2605.11665v1}{} & Policy / benchmark modules & Pipeline composition & Execution checks \\[0pt]
\rowcolor{black!3}
% Source: https://arxiv.org/abs/2606.00083v1 | §3.3 task instruction discovery; §4.1 experimental setup
Demo2Reward \mbox{\citep{survey_ax_2606_00083}} & \LiteratureLinks{https://rlj.cs.umass.edu/2026/papers/Paper12.html}{https://github.com/cgumbsch/Demo2Reward} & Reward prompt & Iterative refinement & Demonstrations \\[0pt]
% Source: https://arxiv.org/abs/2606.01672v1 | §§4.1–4.2 evolutionary search, visual analysis and reward reflection
RDA \mbox{\citep{survey_ax_2606_01672}} & \LiteratureLinks{https://rlj.cs.umass.edu/2026/papers/Paper81.html}{} & Reward code & Evolutionary search & Rollout videos \\[0pt]
\rowcolor{black!3}
% Source: https://arxiv.org/abs/2606.08610v1 | §3 artifact-based staged RL development; §4 evaluation scope
HARBOR \mbox{\citep{survey_ax_2606_08610}} & \LiteratureLinks{https://arxiv.org/abs/2606.08610}{https://github.com/supersglzc/harbor-rl} & RL code; configurations & Iterative refinement; hyperparameter search & Execution checks; training metrics \\[0pt]
% Source: https://arxiv.org/abs/2606.09630v1 | §3 modules M1–M6 and deterministic recovery-reward compiler
ReCoVLA \mbox{\citep{survey_ax_2606_09630}} & \LiteratureLinks{https://arxiv.org/abs/2606.09630v1}{} & Recovery rewards; residual policies & Single-pass generation; RL & Failure cases; simulator state \\[0pt]
\rowcolor{black!3}
% Source: https://arxiv.org/abs/2606.19980v2 | §§2.1–2.2 embodied environment and open-ended training-code improvement
ENPIRE \mbox{\citep{survey_ax_2606_19980}} & \LiteratureLinks{https://arxiv.org/abs/2606.19980v2}{https://github.com/NVlabs/ENPIRE} & Training code & Iterative refinement & Rollouts; evaluation metrics \\[0pt]
% Source: https://arxiv.org/abs/2606.22142v1 | §3.1 lifecycle and review transitions; §4 evaluation scope
RoboLineage \mbox{\citep{survey_ax_2606_22142}} & \LiteratureLinks{https://arxiv.org/abs/2606.22142v1}{https://github.com/robolineage/robolineage} & Training data; policy records & Data curation; iterative refinement & Rollouts; evaluation metrics \\[0pt]
\rowcolor{black!3}
% Source: https://arxiv.org/abs/2606.23280v1 | §IV overview and §IV-C causal pruning for zero-shot reward generation
CRWM \mbox{\citep{survey_ax_2606_23280}} & \LiteratureLinks{https://arxiv.org/abs/2606.23280v1}{} & Reward code & Single-pass generation & Offline data \\[0pt]
% Source: https://arxiv.org/abs/2608.07555v1 | §§3.1–3.4 protocol, backend boundary and round transaction
AgenticRobotics \mbox{\citep{survey_ax_2608_07555}} & \LiteratureLinks{https://arxiv.org/abs/2608.07555v1}{} & Tool records; reusable workflows & Iterative refinement & Evaluation metrics \\[0pt]
\rowcolor{black!3}
% Source: https://arxiv.org/abs/2609.01281v1 | §4.1 separate planner and skill-policy training; §5.1 RoboTwin setup
EmbodiedSkills \mbox{\citep{survey_ax_2609_01281}} & \LiteratureLinks{https://arxiv.org/abs/2609.01281v1}{https://github.com/ZJU4EmbodiedAI/EmbodiedSkills} & Planner; VLA; scheduler & SFT & Demonstrations \\*[0pt]
% Source: https://arxiv.org/abs/2609.17210v1 | §3 overview and §3.1 registry construction and execution-layer responsibilities
FluxVLA Engine \mbox{\citep{survey_ax_2609_17210}} & \LiteratureLinks{https://arxiv.org/abs/2609.17210v1}{https://github.com/FluxVLA/FluxVLA} & Training pipelines & Pipeline composition; SFT & Demonstrations \\[0pt]
\rowcolor{black!3}
D\&A \mbox{\citep{survey_ax_2611_03622}}
& \LiteratureLinks{https://arxiv.org/abs/2611.03622v1}{https://github.com/memmelma/dna}
& Policy & Progress rewards for RL & Trajectory video; task criteria \\[1pt]

EmbodyAgent \mbox{\citep{survey_ax_2611_26196}}
& \LiteratureLinks{https://arxiv.org/abs/2611.26196v1}{}
& Code-generating VLM & SFT; online RL & Teacher trajectories; task outcomes; code errors \\[1pt]

\rowcolor{black!3}
GR00T~2~\mbox{\citep{nvidia2026gr00t2}}
& \LiteratureLinks{https://research.nvidia.com/labs/gear/gr00t-2/assets/gr00t2_whitepaper.pdf}{}
& RL training code & Test and revise RL algorithms & Training logs; candidate scores \\[1pt]

RoboUniverse-\allowbreak 35B-A3B \mbox{\citep{survey_ax_2611_31002}}
& \LiteratureLinks{https://arxiv.org/abs/2611.31002v1}{https://github.com/Einsia/RoboUniverse-10K}
& Language backbone & SFT & Teacher interactions in simulation \\[1pt]

\rowcolor{black!3}
CaPtain-4B~\mbox{\citep{survey_ax_2610_40799}}
& \LiteratureLinks{https://arxiv.org/abs/2610.40799v1}{https://github.com/waltstephen/CaPtain}
& Code-generating VLM & SFT; online RL & Teacher trajectories; simulator state; task outcomes \\[1pt]
% Source: 2611.31098v2, Section 4.2 and Appendix B.2 (Stack Cups).
Real2Worlds2Real~\mbox{\citep{survey_ax_2611_31098}}
& \LiteratureLinks{https://arxiv.org/abs/2611.31098v2}{https://github.com/real2worlds2real/real2worlds2real}
& Visuomotor policy & Imitation learning & Demonstrations; expert corrections \\[1pt]
\end{longtable}
\endgroup

\subsection{Perception}
\label{sec:model-perception}

Perception translates observations into task-relevant \textbf{concepts, relations, geometry, or beliefs}. The agentic intervention may be an interpretation, a request for another observation, or a physical action that reveals missing evidence. The representation must support the next decision: identifying a handle, for example, also requires geometry accurate enough for the intended grasp.

\paragraph{\textbf{Observations.}}
When the current view is insufficient, the agent chooses what to observe next. Situated human-robot conversation systems~\citep{survey_ax_2602_04157} combine multimodal interaction with task-level attention and tool calls, while HINT~\citep{survey_ax_2609_02653} updates intent and routes camera views at manipulation-pattern transitions, and re-localizes the same committed target when tracking confidence is low. SCOPE~\citep{survey_doi_353abec881f4d997} makes camera orientation and zoom explicit actions whose returned observations condition the next perception request, illustrating a camera agent rather than a fixed detector.

Active perception changes the evidence itself. MG-VQA (formerly Probe)~\citep{survey_ax_2608_17129} lets a VLM invoke perception, grasp-and-remove, and pushing tools to expose occluded information before answering a visual question. VAP-TAMP~\citep{survey_ax_2604_26988} uses visual predicate checks and additional viewpoints to correct the state supplied to a Planning Domain Definition Language (PDDL) planner. Manipulation can therefore gather information, while an additional observation can invalidate a symbolic planning premise. The choice to move an object or camera trades the value of new evidence against motion cost and risk.

\paragraph{\textbf{Decisions.}}
A robot may need a target position or an answer about the scene, depending on its next decision.
ABot-N1~\citep{survey_ax_2607_10383} uses a recurrent spatial reasoner to update semantic targets while a distinct action expert supplies continuous navigation behavior; Steerable VLA policies~\citep{survey_ax_2602_13193} similarly expose semantic-subtask, atomic-motion, and grounded-point or trace steering modes. A distributed conversational manipulation framework~\citep{survey_ax_2606_06061} grounds pixel predictions through depth and robot coordinates on each command, whereas RoboVista's~\citep{survey_ax_2607_04610} robot-VQA abstraction evaluates whether a model can answer perception questions across diverse application modules. Whether the output is a renewed view, a geometric target, or a symbolic answer, its quality depends on the downstream decision it must support.

Two studies test how well a model can locate a target in an image and how the camera view affects manipulation. MJev-Spatial~\citep{survey_ax_2610_22276} locates a target in an image by zooming in on one region at a time. At each step, it chooses one of nine regions and crops it from the original image. It returns the center of the final region. The paper evaluates the predicted coordinates on annotated images.
Embo-Harness~\citep{survey_ax_2611_31201} uses a frontier model to choose manipulation actions and tests how its camera views affect task success. Adding a wrist view to the external camera raised success from 30\% to 65\% on 20 RoboLab tasks with three runs per task. A separate test measured 3D position estimates directly. With one run per task, increasing image resolution from $224\times126$ to $640\times360$ reduced the mean error from 14.14 to 9.07\,cm.

A model can also decide which points in an image need a metric measurement. In SOLVE~\citep{survey_ax_2611_34750}, the agent selects pixels and receives their 3D positions. Each point comes with the source of its depth estimate. A point matched across views also carries a confidence score, so the agent can decide whether to measure~again.

On selected rigid puzzles in RoboARC~\citep{survey_ax_2611_33390}, GPT-6 Astra solved more trials with both object poses and part geometry than with either alone (Table~\ref{tab:benchmark-protocols}).

Navigation requires spatial evidence to remain accessible as the robot moves. AgenticNav~\citep{survey_ax_2606_10577} provides heading-indexed views, selective recall, a compact trajectory map, and geometric tools. The VLM can request evidence and choose a navigation target; depth-based validity checks and movement remain below that choice. SoFar~\citep{survey_ax_2502_13143} provides an earlier spatial-grounding comparison: object perception constructs a scene graph containing positions and semantic orientation vectors, and the VLM specifies desired object geometry that deterministic fitting and motion planning realize. A textual description such as ``upright'' becomes useful only when bound to the orientation representation understood by the executor.

\paragraph{\textbf{Reflections.}}
Before acting on a target, the robot may need to check its location again. Robo-Harness K1~\citep{survey_ax_2609_29389} gives the model images, object anchors, and reports on executed commands. The backend handles calibrated depth geometry. If an anchor is lost or invalid, the agent must locate the object again before using it as a motion target. CoRelNav~\citep{survey_ax_2609_27720} uses task-conditioned relations and shared observations to request complementary views before accepting a candidate. Its simulation supplies oracle instance masks and IDs. INSPECT~\citep{survey_ax_2609_20615} further evaluates learned view utility and commitment decisions through annotated video replay and a fixed robot-image lattice rather than a closed-loop physical inspection trial.

A navigation agent can use current and past observations to choose where to go next. OpenFrontier~\citep{survey_ax_2603_05377} scores visual frontiers and verifies candidate targets, returning to search when a candidate is rejected; GoalVLM~\citep{survey_ax_2603_18210} combines frontier scoring with allocation across agents and a shared map. FARE~\citep{survey_ax_2601_14681} reasons over an evolving belief graph to select global waypoints, while AnchorVLN~\citep{survey_ax_2609_12285} binds model commands to typed object handles through a ROS server. Frontier scores help choose where to explore, belief graphs preserve longer spatial dependencies, and typed handles constrain command references. Their common vulnerability is stale grounding: a remembered object identity must be re-observed when its location no longer supports the proposed action. Related works are compared in Table~\ref{tab:model-perception}.

\par\Needspace{12\baselineskip}
\begingroup
\mechanismtableglue
\CompactTableStyle
\setlength{\LiteratureWidth}{\dimexpr\linewidth-8\tabcolsep\relax}

\setlength{\LTpre}{4pt}
\setlength{\LTpost}{4pt}
\setlength{\LTcapwidth}{\linewidth}
\captionsetup{singlelinecheck=false,justification=raggedright}
\LiteratureContinuationSetup
\begin{longtable}{@{}>{\RaggedRight\arraybackslash}p{0.2313\LiteratureWidth}>{\centering\arraybackslash}p{0.0750\LiteratureWidth}>{\RaggedRight\arraybackslash}p{0.2128\LiteratureWidth}>{\RaggedRight\arraybackslash}p{0.2128\LiteratureWidth}>{\RaggedRight\arraybackslash}p{0.2682\LiteratureWidth}@{}}
\caption[Comparison of works on perception]{Comparison of works on perception (Section~\ref{sec:model-perception}). Input lists observations, excluding task instructions and queries.}\label{tab:model-perception}\\
\LiteratureTopRule{Model}
\rowcolor{TSTModelLight}\textbf{Work} & \textbf{Link} & \textbf{Input} & \textbf{Output} & \textbf{Perception process} \\
\midrule
\endfirsthead
\multicolumn{5}{@{}l}{\tablename~\thetable\ (continued)} \\
\LiteratureTopRule{Model}
\rowcolor{TSTModelLight}\textbf{Work} & \textbf{Link} & \textbf{Input} & \textbf{Output} & \textbf{Perception process} \\
\midrule
\endhead
\midrule
\multicolumn{5}{r@{}}{Continued on next page} \\
\endfoot
\bottomrule
\endlastfoot
% Source: https://arxiv.org/abs/2502.13143v2 | §3.1 position/orientation extraction and 6-DoF scene graph
SoFar \mbox{\citep{survey_ax_2502_13143}} & \LiteratureLinks{https://proceedings.neurips.cc/paper_files/paper/2025/hash/6ea83df350753236215cbf2fe9716eca-Abstract-Conference.html}{https://github.com/qizekun/SoFar} & RGB-D & Scene graph & Visual grounding; geometric estimation \\[0pt]
\rowcolor{black!3}
% Source: https://arxiv.org/abs/2601.14681v1 | §IV-A hierarchical robot belief graph; §III LiDAR observations
FARE \mbox{\citep{survey_ax_2601_14681}} & \LiteratureLinks{https://arxiv.org/abs/2601.14681v1}{} & LiDAR & Spatial map & Mapping; frontier ranking \\[0pt]
% Source: https://arxiv.org/abs/2602.04157v1 | §III real-time dialogue, attention tools and geometric view memory
MSR \mbox{\citep{survey_ax_2602_04157}} & \LiteratureLinks{https://arxiv.org/abs/2602.04157v1}{} & Audio; images & Retrieved images & Memory retrieval; view selection \\[0pt]
\rowcolor{black!3}
% Source: https://arxiv.org/abs/2602.13193v3 | §IV-A command styles; §V high-level VLM command generation
Steerable Policies \mbox{\citep{survey_ax_2602_13193}} & \LiteratureLinks{https://www.roboticsproceedings.org/rss22/p074.html}{https://github.com/steerable-policies/steerable-policies-bridge} & Images & Language commands; pixel coordinates & Visual reasoning \\[0pt]
% Source: https://arxiv.org/pdf/2603.05377v3 | PDF p4 Sec III-B; pp12-13 App A,D and Algorithm 2: depth and camera pose; measured versus estimated depth
OpenFrontier \mbox{\citep{survey_ax_2603_05377}} & \LiteratureLinks{https://www.roboticsproceedings.org/rss22/p067.html}{https://github.com/cvg/OpenFrontier} & RGB; depth; camera pose & 3D targets & Frontier ranking; visual grounding; verification \\[0pt]
\rowcolor{black!3}
% Source: https://arxiv.org/abs/2603.18210v1 | §III-B semantic mapping; §III-G2 shared map fusion
GoalVLM \mbox{\citep{survey_ax_2603_18210}} & \LiteratureLinks{https://arxiv.org/abs/2603.18210v1}{} & RGB-D; robot pose & Spatial map & Mapping; geometric estimation \\[0pt]
% Source: https://arxiv.org/pdf/2604.26988v1 | PDF Sec III; implementation link verified through first-author homepage and GitHub on 2026-10-13
VAP-TAMP \mbox{\citep{survey_ax_2604_26988}} & \LiteratureLinks{https://arxiv.org/abs/2604.26988v1}{https://github.com/aoloo-r/VAP-TAMP} & RGB-D; robot pose & Scene graph; symbolic predicates & View selection; verification \\[0pt]
\rowcolor{black!3}
% Source: https://doi.org/10.1145/3757279.3785641 | §3 system architecture and perception/PTZ tool loop
SCOPE \mbox{\citep{survey_doi_353abec881f4d997}} & \LiteratureLinks{https://doi.org/10.1145/3757279.3785641}{https://github.com/HindsboNikolaj/SCOPE} & PTZ images & Answers & VQA; view selection \\[0pt]
% Source: https://arxiv.org/abs/2606.06061v1 | §III-A vision grounding and coordinator transformation
CFHRCM \mbox{\citep{survey_ax_2606_06061}} & \LiteratureLinks{https://arxiv.org/abs/2606.06061v1}{https://github.com/cogrob-tuni/franka-llm} & RGB-D; calibration & 3D targets & Visual grounding; geometric estimation \\[0pt]
\rowcolor{black!3}
% Source: https://arxiv.org/abs/2606.10577v3 | §III-A tool interface; §III-B pixel-depth queries; Recall Tool
AgenticNav \mbox{\citep{survey_ax_2606_10577}} & \LiteratureLinks{https://arxiv.org/abs/2606.10577v3}{} & RGB; BEV; depth tool & Distances; retrieved images & Geometric estimation; memory retrieval \\[0pt]
% Source: https://arxiv.org/abs/2607.04610v1 | §III RQA framework and §III-C Robot-VQA construction
RoboVista \mbox{\citep{survey_ax_2607_04610}} & \LiteratureLinks{https://www.roboticsproceedings.org/rss22/p095.html}{https://github.com/ehehee/robovista} & Images & Answers & VQA \\[0pt]
\rowcolor{black!3}
% Source: https://arxiv.org/abs/2607.10383v3 | §4.1 slow-system inputs and target/affordance pixel interface
ABot-N1 \mbox{\citep{survey_ax_2607_10383}} & \LiteratureLinks{https://arxiv.org/abs/2607.10383v3}{https://github.com/amap-cvlab/ABot-Navigation} & Images; visual history & Pixel coordinates & Visual reasoning; visual grounding \\[0pt]
% Source: https://arxiv.org/abs/2608.17129v2 | §§3.1–3.3 manipulation-grounded questions and interaction tools
MG-VQA \mbox{\citep{survey_ax_2608_17129}} & \LiteratureLinks{https://arxiv.org/abs/2608.17129v2}{https://github.com/vineet2104/MG-VQA} & Top-down RGB & Answers; pixel coordinates & Interactive perception; visual grounding; VQA \\[0pt]
\rowcolor{black!3}
% Source: https://arxiv.org/abs/2609.02653v2 | §III-A view scheduling; §III-B semantic grounding and tracking
HINT \mbox{\citep{survey_ax_2609_02653}} & \LiteratureLinks{https://arxiv.org/abs/2609.02653v2}{https://github.com/robot-hint/HINT} & Images; state; torques & Segmentation masks & View selection; tracking; visual grounding \\[0pt]
% Source: https://arxiv.org/abs/2609.12285v1 | §§III-A–III-B object handles, geometric lift and validation gates
AnchorVLN \mbox{\citep{survey_ax_2609_12285}} & \LiteratureLinks{https://arxiv.org/abs/2609.12285v1}{https://github.com/aryanmangal769/embodied-nav-mcp} & Panorama; LiDAR & Object handles; 3D targets & Visual grounding; geometric estimation; verification \\[0pt]
\rowcolor{black!3}
% Source: https://arxiv.org/abs/2609.20615v1 | §III claim-conditioned view selection; §IV-A offline one-step replay
INSPECT \mbox{\citep{survey_ax_2609_20615}} & \LiteratureLinks{https://arxiv.org/abs/2609.20615v1}{https://github.com/Kratos-Wen/INSPECT} & RGB; camera poses & Camera viewpoints; answers & View selection; verification \\[0pt]
% Source: https://arxiv.org/abs/2609.27720v1 | §III-B system overview; §IV-A2 and §IV-A7 simulation oracle masks/IDs; §IV-B learned real-world perception
CoRelNav \mbox{\citep{survey_ax_2609_27720}} & \LiteratureLinks{https://arxiv.org/abs/2609.27720v1}{} & RGB-D; robot pose & Object handles; 3D targets & Frontier ranking; verification \\*[0pt]
\rowcolor{black!3}
% Source: https://arxiv.org/pdf/2609.29389v1 | PDF p17, App A.1 Table 3: grasp_candidates proposes poses but does not execute a grasp
Robo-Harness K1 \mbox{\citep{survey_ax_2609_29389}} & \LiteratureLinks{https://arxiv.org/abs/2609.29389v1}{https://github.com/Robo-Harness/k1} & RGB; state; geometry tools & Segmentation masks; object handles; 3D targets & Visual grounding; geometric estimation; tracking \\[0pt]
MJev-Spatial \mbox{\citep{survey_ax_2610_22276}}
& \LiteratureLinks{https://arxiv.org/abs/2610.22276v2}{https://github.com/Fr0zenCrane/MJev-Spatial}
& Images; local crops & Choices; values; image points & Candidate scoring; region refinement \\*[1pt]

SOLVE~\mbox{\citep{survey_ax_2611_34750}}
& \LiteratureLinks{https://arxiv.org/abs/2611.34750v1}{}
& Images; depth tools & 3D points & Visual grounding; geometric estimation \\[1pt]
\end{longtable}
\endgroup

\subsection{Planning}
\label{sec:model-planning}

Based on what it perceives, a robot plans a sequence of actions to complete a task. Before carrying out the plan, it needs to check whether these actions are feasible. The plan may also need to change during execution, for example when an object moves or an action fails. Planning therefore continues as the robot carries out the task and learns whether its decisions work. We discuss this process in three stages: generating a plan, checking it before execution, and updating it during execution.

\paragraph{\textbf{Generating a Plan.}}
Some planners choose one operation at a time and use its result to decide what to do next. This is useful when later steps depend on information the robot does not yet have. For example, one system provides tools for looking around, moving to a location, grasping objects, and accessing memory~\citep{survey_ax_2603_03148}. The planner reads the text returned by these tools before choosing its next operation. AgentRob~\citep{survey_ax_2602_13591} follows a similar approach, with VLM controllers observing the results of action primitives before selecting the next operation. In both systems, the planner chooses the operation, and the robot controller handles the physical movement.

The planner can also specify a subgoal and let a trained policy produce the actions needed to achieve it. Hi Robot~\citep{survey_ax_2502_19417} uses a high-level VLM to give short language commands to a VLA. These commands are updated periodically or after user intervention. Gemini Robotics-ER 1.5~\citep{survey_ax_2510_03342} similarly handles high-level reasoning and gives language instructions to the Gemini Robotics 1.5 action model. LocalNav~\citep{survey_ax_2606_27871} applies this division to navigation. Its distilled scene-graph reasoner decides whether to move toward an object, explore a room, perform frontier search, or stop. A PointGoal policy then carries out the navigation.

Beyond choosing the next operation or subgoal, a planner can also specify how several operations depend on one another. For example, \citep{survey_ax_2602_21670} decomposes multi-robot tasks into PDDL subproblems for a classical planner and uses planning failures to improve the agents' prompts. In an industrial setting, IMR-LLM~\citep{survey_ax_2603_02669} uses an LLM to help construct a disjunctive graph, which a deterministic solver uses to schedule operations. A process tree then guides the generation of executable robot programs. To connect symbolic planning with execution, EmboTeam~\citep{survey_ax_2601_11063} combines semantic processing with PDDL-based search and converts the resulting plans into behavior trees containing conditions, fallbacks, and synchronization.

A program can also guide a learned policy. AW-0~\citep{survey_ax_2610_30136} predicts a short program for the current subtask from the instruction, camera views, and earlier program segments. The program can name an object, call a reusable skill, or repeat it in a loop. Both the predicted video and the next action sequence depend on this representation. In the program annotation for a blanket-flattening demonstration, the left gripper grasps the blanket again while the right holds it. A two-arm skill then pulls the blanket taut and releases it. Section~\ref{sec:model-robot-control} describes when the generated program can itself control the robot.

\paragraph{\textbf{Checking a Plan before Execution.}}
Once actions have been proposed, the planner can check whether they are consistent with its model of the scene and the task requirements. For example, SayPlan~\citep{survey_ax_2307_06135} identifies the relevant part of a hierarchical 3D scene graph and uses feedback from a scene-graph simulator to revise its plan. Within this process, a classical path planner supplies routes between locations. GAVEL~\citep{survey_ax_2609_19315} likewise uses a graph world model to verify and repair candidate plans before execution. For constraints drawn from industrial procedures, MaCoPlanner~\citep{survey_ax_2608_28300} converts manuals into a typed intermediate representation and checks action sequences against procedural and state-transition constraints.

Before planning the actions, the system can also check whether the goal is clear and suitable for the robot. For example, STARS~\citep{survey_doi_ddc6fc88fa219f08} generates candidate goal descriptions and analyzes, repairs, and selects them before the underlying task-learning agent searches for actions to achieve the selected goal. Here, the checks concern whether the proposed goal is understandable and appropriate for the agent and its situation, rather than whether a particular action sequence can be executed.

Alongside checks based on graphs and rules, planners can also use learned models to assess candidate actions. For instance, Reflective Test-Time Planning~\citep{survey_ax_2602_21198} scores proposed high-level actions with an internal reflection model, while external and retrospective reflection later draw on actual outcomes. Another way to assess actions before execution is to imagine their visual consequences. ReflectVLM~\citep{survey_pmlr_v305_feng25b} predicts future images for a proposed action sequence and considers these predictions together with the current and goal images to select the next action primitive. A scripted controller then executes the selected primitive.

\paragraph{\textbf{Updating a Plan during Execution.}}
Even after these checks, an action may produce an unexpected result, or a change in the scene may make the remaining plan unsuitable. The planner therefore needs execution feedback to decide whether to continue or revise its next steps. For example, WCM~\citep{survey_ax_2607_22999} revises its next decision when actual outcomes differ from its predictions, the user changes the instructions, or the scene changes. When necessary, it can also ask the user for clarification.

To support these decisions, the system needs to report execution results and determine when they call for replanning. For example, \citep{survey_ax_2602_13081} provides status, event, and reflection tools for this purpose. A failed physical check can serve as a direct trigger. ActionReasoning 2.0~\citep{survey_ax_2609_05985} discards the remaining action sequence when an action fails verification and generates a new sequence from the updated scene state, retaining previously verified task progress. Similarly, ATP-VPDSG~\citep{survey_doi_07c1d81a50e5849a} checks the result of each primitive and uses execution feedback to diagnose failures and revise the remaining actions.

When selecting the next action, the planner can also draw on earlier actions and observations rather than relying only on the latest result. MaP-WAM~\citep{survey_ax_2609_11561} uses records of completed action segments as planning context and checks execution progress to determine when to update the plan and its context. For visual history, ActiveArena-Plan~\citep{survey_ax_2609_24124} predicts the current subtask and selects historical frames to retain in memory. A separately trained action policy uses the subtask, selected frames, current observation, and robot state to generate actions.

For embodied question answering, HFLEX-EQA~\citep{survey_ax_2609_26360} combines past visual observations with a floorplan prior and an incrementally constructed scene graph to choose among room exploration, object inspection, and room discovery. Dedicated low-level planners carry out these operations. Likewise, NavProbe~\citep{survey_ax_2609_27526} retrieves historical images and geometric records to revise intermediate goals and assess whether they have been satisfied before selecting the next navigation action.

However, revising the action sequence may be insufficient if the planning model does not describe the situation the robot encounters. To address this problem, HUME~\citep{survey_ax_2607_06501} proposes possible additions to the domain model, treats them as uncertain hypotheses, and specifies conditions for checking them. Perception then confirms or rejects these hypotheses, and rejected hypotheses guide subsequent replanning. In this case, execution feedback changes both the plan and the model used to generate it.

Overall, how far ahead a robot plans should depend on how reliably it can predict the results of its actions. When those results are predictable, longer plans can organize dependencies across several operations. When outcomes are uncertain or the scene changes, planning fewer steps ahead allows new observations to guide subsequent decisions. In either case, checks before execution and feedback during execution help determine whether to continue with the plan, revise its actions, or update the model used to generate it. Related works are compared in Table~\ref{tab:model-planning}.

\par\Needspace{12\baselineskip}
\begingroup
\mechanismtableglue
\CompactTableStyle
\setlength{\LiteratureWidth}{\dimexpr\linewidth-8\tabcolsep\relax}
\setlength{\LTpre}{4pt}
\setlength{\LTpost}{4pt}
\setlength{\LTcapwidth}{\linewidth}
\captionsetup{singlelinecheck=false,justification=raggedright}
\LiteratureContinuationSetup
\begin{longtable}{@{}>{\RaggedRight\arraybackslash}p{0.230\LiteratureWidth}>{\centering\arraybackslash}p{0.060\LiteratureWidth}>{\RaggedRight\leftskip=10pt\relax\hyphenpenalty=10000\exhyphenpenalty=10000\arraybackslash}p{0.235\LiteratureWidth}>{\RaggedRight\arraybackslash}p{0.155\LiteratureWidth}>{\RaggedRight\arraybackslash}p{0.320\LiteratureWidth}@{}}
\caption[Comparison of works on planning]{Comparison of works on planning (Section~\ref{sec:model-planning}). Manipulation includes mobile manipulation; Model lists planning, replanning, or verification backbones.}\label{tab:model-planning}\\
\LiteratureTopRule{Model}
\rowcolor{TSTModelLight}\textbf{Work} & \textbf{Link} & \textbf{Task} & \textbf{Plan Output} & \textbf{Model} \\
\midrule
\endfirsthead
\multicolumn{5}{@{}l}{\tablename~\thetable\ (continued)} \\
\LiteratureTopRule{Model}
\rowcolor{TSTModelLight}\textbf{Work} & \textbf{Link} & \textbf{Task} & \textbf{Plan Output} & \textbf{Model} \\
\midrule
\endhead
\midrule
\multicolumn{5}{r@{}}{Continued on next page} \\
\endfoot
\bottomrule
\endlastfoot
% Source: https://arxiv.org/abs/2307.01928v2 | §2 multiple-choice planning; §§3.2–3.3 calibrated prediction sets
KnowNo \mbox{\citep{survey_ax_2307_01928}} & \LiteratureLinks{https://proceedings.mlr.press/v229/ren23a.html}{https://github.com/google-research/google-research/tree/master/language_model_uncertainty} & Manipulation & Action & PaLM-2-L \\[1pt]
% Source: https://arxiv.org/abs/2307.06135v2 | §3.3 semantic search and iterative replanning
\rowcolor{black!3}
SayPlan \mbox{\citep{survey_ax_2307_06135}} & \LiteratureLinks{https://proceedings.mlr.press/v229/rana23a.html}{} & Manipulation & Action & GPT-4 \\[1pt]
% Source: https://arxiv.org/abs/2307.04738v1 | §§3.1–3.3 dialogue, subtask validation and multi-arm planning
RoCo \mbox{\citep{survey_ax_2307_04738}} & \LiteratureLinks{https://ieeexplore.ieee.org/document/10610855/}{https://github.com/MandiZhao/robot-collab} & Manipulation & Action; Trajectory & GPT-4 \\[1pt]
% Source: https://proceedings.mlr.press/v305/feng25b.html | §§4.2–4.3 diffusion dynamics and inference-time reflective planning
\rowcolor{black!3}
ReflectVLM \mbox{\citep{survey_pmlr_v305_feng25b}} & \LiteratureLinks{https://proceedings.mlr.press/v305/feng25b.html}{https://github.com/yunhaif/reflect-vlm} & Manipulation & Action & LLaVA-1.5-13B \\[1pt]
% Source: https://arxiv.org/abs/2502.19417v2 | §§4.1–4.2 hierarchical inference and user interaction
HiRobot \mbox{\citep{survey_ax_2502_19417}} & \LiteratureLinks{https://proceedings.mlr.press/v267/shi25d.html}{} & Manipulation & Goal & PaliGemma-3B \\[1pt]
% Source: https://arxiv.org/abs/2510.03342v3 | §2.1 agentic system architecture and embodied thinking
\rowcolor{black!3}
Gemini Robotics 1.5 \mbox{\citep{survey_ax_2510_03342}} & \LiteratureLinks{https://arxiv.org/abs/2510.03342v3}{} & Manipulation & Goal & Gemini Robotics-ER 1.5 \\[1pt]
% Source: https://arxiv.org/abs/2601.11063v2 | §3 three-stage architecture and reactive execution layer
EmboTeam \mbox{\citep{survey_ax_2601_11063}} & \LiteratureLinks{https://arxiv.org/abs/2601.11063v2}{} & Manipulation & Program & GPT-4o \\[1pt]
% Source: https://arxiv.org/abs/2601.18492v1 | §IV-E true–false and masked-entity verification
\rowcolor{black!3}
DV-VLN \mbox{\citep{survey_ax_2601_18492}} & \LiteratureLinks{https://arxiv.org/abs/2601.18492v1}{https://github.com/PlumJun/DV-VLN} & Navigation & Action & Llama-2-7B \\[1pt]
% Source: https://arxiv.org/abs/2602.13081v1 | §3 planner/executor tool interface and termination behavior
AARC \mbox{\citep{survey_ax_2602_13081}} & \LiteratureLinks{https://ojs.aaai.org/index.php/AAAI-SS/article/view/42578}{} & Manipulation; Navigation & Action & GPT-o3 \\[1pt]
% Source: https://arxiv.org/abs/2602.13591v1 | §3.4 robot-layer iterative tool-calling loop | Model: official README default doubao-seed-1-8-251228, commit 386f2d28e2f883262c8afa011c377ac8606e612c; not a paper-reported experiment checkpoint.
\rowcolor{black!3}
AgentRob \mbox{\citep{survey_ax_2602_13591}} & \LiteratureLinks{https://arxiv.org/abs/2602.13591v1}{https://github.com/PKU-LLM-DS-LAB/AgentRob} & Human--Robot Interaction & Action & Doubao-Seed-1.8 \\[1pt]
% Source: https://arxiv.org/abs/2602.21198v3 | §§3.1–3.2 reflection-in-action and reflection-on-action
Reflective Test-Time Planning \mbox{\citep{survey_ax_2602_21198}} & \LiteratureLinks{https://arxiv.org/abs/2602.21198v3}{https://github.com/Reflective-Test-Time-Planning/Reflective-Test-Time-Planning} & Manipulation & Action & LLaVA-3D-7B; Qwen2.5-VL-3B \\[1pt]
% Source: https://arxiv.org/abs/2602.21670v2 | §IV hierarchical decomposition, PDDL planning and prompt optimization
\rowcolor{black!3}
HLMF \mbox{\citep{survey_ax_2602_21670}} & \LiteratureLinks{https://arxiv.org/abs/2602.21670v2}{} & Manipulation & Action & GPT-4o \\[1pt]
% Source: https://arxiv.org/abs/2603.03148v1 | §III tool-based task plans, episodic memory and online recovery
FLA \mbox{\citep{survey_ax_2603_03148}} & \LiteratureLinks{https://arxiv.org/abs/2603.03148v1}{https://github.com/ShinasShaji/llm-robot-cognition} & Manipulation & Action & GPT-4.1; Claude Sonnet 4 \\[1pt]
% Source: https://arxiv.org/abs/2603.02669v1 | §IV-A disjunctive-graph task planning
\rowcolor{black!3}
IMR-LLM \mbox{\citep{survey_ax_2603_02669}} & \LiteratureLinks{https://arxiv.org/abs/2603.02669v1}{} & Manipulation & Schedule; Program & GPT-4o; Qwen3-32B \\[1pt]
% Source: https://arxiv.org/abs/2603.02772v2 | §III system overview; §IV ILAD and MARS
SERP \mbox{\citep{survey_ax_2603_02772}} & \LiteratureLinks{https://arxiv.org/abs/2603.02772v2}{} & Navigation & Trajectory & Qwen3-VL-Plus; Qwen3.7-Plus \\[1pt]
% Source: https://arxiv.org/abs/2604.13533v2 | §III-A–III-D environment interpreter, planner and error localization
\rowcolor{black!3}
EEAgent \mbox{\citep{survey_ax_2604_13533}} & \LiteratureLinks{https://arxiv.org/abs/2604.13533v2}{} & Manipulation & Action & GPT-4o \\[1pt]
% Source: https://arxiv.org/abs/2606.27871v1 | §3 online scene graph and four-action navigation interface
LocalNav \mbox{\citep{survey_ax_2606_27871}} & \LiteratureLinks{https://arxiv.org/abs/2606.27871v1}{} & Navigation & Goal & Qwen3.5-4B \\[1pt]
% Source: https://arxiv.org/abs/2607.06501v1 | §IV hypothesis-driven model expansion and verification
\rowcolor{black!3}
HUME \mbox{\citep{survey_ax_2607_06501}} & \LiteratureLinks{https://www.roboticsproceedings.org/rss22/p181.html}{https://github.com/threefruits/HUME} & Manipulation & Action & GPT-4.1 \\[1pt]
% Source: https://arxiv.org/abs/2607.22999v2 | §II-A1 Logic layer and outcome-validation rule
WCM \mbox{\citep{survey_ax_2607_22999}} & \LiteratureLinks{https://arxiv.org/abs/2607.22999v2}{} & Manipulation & Action & - \\[1pt]
% Source: https://www.frontiersin.org/journals/computer-science/articles/10.3389/fcomp.2026.1848830/full | §§3.2.2 and 4.1.2 task learning and Analysis/Repair
\rowcolor{black!3}
STARS \mbox{\citep{survey_doi_ddc6fc88fa219f08}} & \LiteratureLinks{https://www.frontiersin.org/journals/computer-science/articles/10.3389/fcomp.2026.1848830/full}{https://github.com/Center-for-Integrated-Cognition/STARS} & Manipulation & Goal & GPT-3.5 (text-davinci-002); GPT-4 \\[1pt]
% Source: https://arxiv.org/abs/2608.28300v1 | §3.1 symbolic rollout, LTL/Safety-FSM checking and targeted repair
MaCoPlanner \mbox{\citep{survey_ax_2608_28300}} & \LiteratureLinks{https://arxiv.org/abs/2608.28300v1}{https://github.com/XinGP/MaCoPlanner} & Manipulation & Action & GPT-4o \\[1pt]
% Source: https://arxiv.org/abs/2608.30396v1 | §3.1 agent–executor interface; §3.4 source-grounded evidence ledger
\rowcolor{black!3}
NavMCP \mbox{\citep{survey_ax_2608_30396}} & \LiteratureLinks{https://arxiv.org/abs/2608.30396v1}{} & Navigation; Embodied Question Answering & Goal & Qwen3.6-Plus \\[1pt]
% Source: https://www.mdpi.com/1424-8220/26/17/5595 | §3.1 video priors, dynamic scene graph and closed-loop replanning
ATP-VPDSG \mbox{\citep{survey_doi_07c1d81a50e5849a}} & \LiteratureLinks{https://www.mdpi.com/1424-8220/26/17/5595}{} & Manipulation & Action & Qwen3-VL \\[1pt]
% Source: https://arxiv.org/abs/2609.05985v1 | §III framework overview and execution-verification loop
\rowcolor{black!3}
ActionReasoning 2.0 \mbox{\citep{survey_ax_2609_05985}} & \LiteratureLinks{https://arxiv.org/abs/2609.05985v1}{} & Manipulation & Action & Qwen3.7-Plus / Qwen3-7B \\[1pt]
% Source: https://arxiv.org/abs/2609.08444v1 | §4.2 risk predictor, REPLAN signal and conservative factor
SafeMem \mbox{\citep{survey_ax_2609_08444}} & \LiteratureLinks{https://arxiv.org/abs/2609.08444v1}{https://github.com/lty759/SafeMem} & Manipulation & Action & GPT-4o \\[1pt]
% Source: https://arxiv.org/abs/2609.11561v1 | §§3.1–3.3 memory-grounded planning and joint progress prediction
\rowcolor{black!3}
MaP-WAM \mbox{\citep{survey_ax_2609_11561}} & \LiteratureLinks{https://arxiv.org/abs/2609.11561v1}{} & Manipulation & Goal & Qwen3.5-4B; Wan2.2-5B \\[1pt]
% Source: https://arxiv.org/abs/2609.13335v1 | §III-A–III-B MetaTool planning and episode scratchpad
BTA \mbox{\citep{survey_ax_2609_13335}} & \LiteratureLinks{https://arxiv.org/abs/2609.13335v1}{} & Navigation; Human--Robot Interaction & Program & GPT-OSS-120B \\[1pt]
% Source: https://arxiv.org/abs/2609.19315v1 | §IV-A graph verification, repair and belief-aware ordering
\rowcolor{black!3}
GAVEL \mbox{\citep{survey_ax_2609_19315}} & \LiteratureLinks{https://arxiv.org/abs/2609.19315v1}{} & Manipulation & Action; Schedule & Qwen3-8B \\[1pt]
% Source: https://arxiv.org/abs/2609.24124v2 | §4.4 ActiveArena-Plan; no separate formal plan checker described
ActiveArena (Plan) \mbox{\citep{survey_ax_2609_24124}} & \LiteratureLinks{https://arxiv.org/abs/2609.24124v2}{https://github.com/leeibo/ActiveArena} & Manipulation & Goal & Qwen3-VL-2B \\[1pt]
% Source: https://arxiv.org/abs/2609.26360v1 | §IV hierarchical exploration, scene graph and answer decision
\rowcolor{black!3}
HFLEX-EQA \mbox{\citep{survey_ax_2609_26360}} & \LiteratureLinks{https://arxiv.org/abs/2609.26360v1}{https://github.com/ntnu-arl/hflex_eqa} & Embodied Question Answering & Goal & GPT-5.5; Gemini 3.5 Pro \\[1pt]
% Source: https://arxiv.org/abs/2609.27720v1 | §III-B acquisition-aware frontier ranking and instance verification
CoRelNav \mbox{\citep{survey_ax_2609_27720}} & \LiteratureLinks{https://arxiv.org/abs/2609.27720v1}{} & Navigation & Goal & Qwen3.7-Plus \\[1pt]
% Source: https://arxiv.org/abs/2609.27526v2 | §III-A formulate–probe–verify–act cycle and task-state updates
\rowcolor{black!3}
NavProbe \mbox{\citep{survey_ax_2609_27526}} & \LiteratureLinks{https://arxiv.org/abs/2609.27526v2}{https://github.com/liujy25/NavProbe} & Navigation & Goal; Action & GPT-5.5 \\[1pt]
AW-0 \mbox{\citep{survey_ax_2610_30136}}
& \LiteratureLinks{https://arxiv.org/abs/2610.30136v1}{https://github.com/ZhuoyangLiu2005/agentic-wam}
& Manipulation & Program & AW-0 code expert (SDAR-1.7B) \\[1pt]
\end{longtable}
\endgroup

\subsection{Robot Control}
\label{sec:model-robot-control}
Robot control turns the steps chosen during planning into physical movement. How much detail the model needs to provide depends on what the execution layer can already do. This layer may handle inverse kinematics, collision checks, and low-level control. Below, we discuss three ways for the model to guide the robot's movement: action and language commands, target poses and control programs, and motion constraints and cost functions. We then discuss how action models, language models, and fixed controllers handle execution feedback.

\paragraph{\textbf{Action and Language Commands.}}
A model can control a robot by choosing from a set of predefined actions. For example, \citep{survey_ax_2607_26148} gives navigation agents basic movement commands and evaluates waypoint tools as an alternative or additional interface. A fixed execution interface carries out the selected movements. Show-Harness~\citep{survey_ax_2609_10522} provides commands for translation, rotation, and gripper control, using fixed rules to convert them into target positions, orientations, and gripper settings. ETA~\citep{survey_ax_2608_03924} translates commands shared across robot types into actions supported by each robot. It also requires a new observation after every tool call that changes the scene and blocks subsequent actions when the grasp or attachment evidence they require is missing.
Demonstrations can also show the model how these commands affect the robot. RoboDawn~\citep{survey_ax_2609_22966} uses task demonstrations to guide a VLM's choice of translation, rotation, and gripper commands. Even when no task demonstrations are provided, the model still receives examples showing the effects of the basic commands.

The execution layer can also accept language instructions. For example, \citep{survey_ax_2606_12299} searches for instruction sequences that improve a frozen VLA's performance and distills them into a language-feedback policy. During deployment, this policy supplies new instructions only when a calibrated predictor judges that they are likely to help; otherwise, the VLA keeps the original instruction. ART~\citep{survey_ax_2608_14047} instead adds reasoning and tool use to the VLA itself, while retaining a separate output path for continuous robot actions.

\paragraph{\textbf{Target Poses and Control Programs.}}
When predefined actions or language instructions do not specify enough spatial detail, a model can provide target positions and orientations directly. For instance, VIA~\citep{survey_ax_2607_11119} lets an agent adjust a virtual gripper, preview its pose, and submit it as a waypoint. Screenshots, coordinates shown when hovering, and pose readouts help the agent choose the target without processing raw arrays of 3D points.
GPT-Policy~\citep{survey_ax_2609_19138} uses a VLM to request robot movements and gripper operations in a structured format. A context compiler assembles task references and interaction history for the VLM, while demonstration actions provide examples for its reasoning. The execution software interpolates between poses, solves inverse kinematics, sets motion timing, and checks execution. These checks determine whether the requested motion is kinematically valid and whether the robot has settled near the target pose. Collision avoidance and task completion require separate checks.

Embo-Harness~\citep{survey_ax_2611_31201} tests who should choose the movements between target poses. In its synchronous setting, GPT-6 Astra reads the camera views, proposes an end-effector target, and receives a new observation once the motion finishes. With the motion fixed at eight frames, letting the controller interpolate between the current and target poses gave 58.3\% success. Asking the model to generate all eight frames gave 40.0\% on the same 20 RoboLab tasks, with three runs per task.

Pose interfaces differ in how much movement the model specifies before it sees the robot again. A model trained with PEAR~\citep{survey_ax_2611_18562} reads the cameras and arm state, then sets one pose target per arm. Inverse kinematics and a controller move the arm until it settles or is blocked. SOLVE~\citep{survey_ax_2611_34750} lets the model send up to five ordered poses at once. It selects the poses and the gripper action at each one. The executor plans the motion between them. JustRUA~\citep{survey_ax_2611_16009} limits each planned movement to its first 20 control steps. The model can use the next images and measured state to change a target before the arm reaches it.

AMP1~\citep{survey_ax_2611_09892} lets the model set the timing as well as the poses. From a fresh Python session for each episode, it can inspect images and robot state, then submit a sequence of timed waypoints and gripper commands. The harness interpolates between waypoints and solves inverse kinematics. When the call returns, the model can inspect the result and adjust the next movement.

Other systems help the model choose and check motion targets. World Action Agent (WAA)~\citep{survey_ax_2609_29964} lets an agent preview and edit proposed movements in a visual workspace, using scene geometry and motion-planner feedback before execution. Textual procedures and reference images stored as reusable skills can guide later actions. Robo-Harness K1~\citep{survey_ax_2609_29389} provides calibrated spatial measurements and persistent visual anchors to guide motion requests. Its tools support both bounded incremental movements and direct moves toward a target, returning achieved motion and residual error after execution.

AW-0~\citep{survey_ax_2610_30136} can control the robot with either predicted actions or a program. By default, the robot follows the continuous actions predicted alongside the program. The hybrid configuration checks whether the proposed program fits the current stage, uses available variables, and calls allowed functions. If it passes, the program drives the robot through those functions. If it fails the checks, the action model supplies the movement. A program that starts running and then raises an error triggers a new observation and planning~step.

Feedback can reach a fixed program or the model that writes it. In ACE-CaP-0~\citep{survey_ax_2611_16872}, the authors freeze the task program for evaluation. The program still reads live observations to choose its actions. In CaPtain-4B~\citep{survey_ax_2610_40799}, by contrast, the model uses feedback to write the next Python cell. After a movement, it can read how far the arm went and why it stopped.

Another approach is to generate motion from templates or keyframes prepared in advance. For example, ManiSkillFormer~\citep{survey_ax_2609_16331} uses human-defined skill schemas to guide agents in generating geometric requirements and motion templates before deployment. During execution, perception supplies the requested geometry, which the templates use to produce robot waypoints. For whole-body motion, KINO~\citep{survey_ax_2609_18869} uses a VLM to select task-relevant keyframes from a predefined library. A retargeter adjusts the selected keyframes to the current scene, and a keyframe-conditioned policy generates joint-level actions.

A model can also write a program that coordinates several movements and checks their results. MEMENTO~\citep{survey_ax_2607_22832} evolves executable robot policies using rollout feedback and memory-guided program revision. Its demonstrated policies use finite-state control, grasp and placement checks, and retries to handle execution failures. FAEA~\citep{survey_ax_2601_20334} generates Python programs that use APIs to take actions, read observations, reset the environment, and check success. It uses states and errors observed during simulator rollouts to revise these programs. Its evaluation provides ground-truth object and gripper states, so applying the same approach to camera images would require perception to supply the state information expected by the APIs.
In manufacturing, A-RAM~\citep{survey_ax_2609_19347} uses an LLM to interpret manufacturing goals and identify which planning variables to explore. Specialized tools then generate G-code and check inverse kinematics, trajectory timing, and joint motion before execution. During execution, a digital shadow records robot and extrusion states and compares them with the selected plan.

\paragraph{\textbf{Motion Constraints and Cost Functions.}}
Instead of specifying a target pose or a sequence of commands, a model can define requirements for the motion and let a solver find a trajectory. ReKep~\citep{survey_ax_2409_01652} generates constraints on relationships between keypoints, both at subgoals and along the path. For pouring tea, the spout must stay aligned with the cup as the teapot tilts. The model writes functions that score these relations. A solver uses the tracked 3D keypoints to choose an end-effector pose and a path to it. If the cup moves, the solver can realign the spout. If the teapot leaves the gripper, the stage checks can send the robot back to grasp it again. The same constraint functions guide both responses.

VoxPoser~\citep{survey_ax_2307_05973} writes programs that build spatial value maps for where to move, which regions to avoid, how to orient the gripper, how fast to move, and when to open or close the gripper. These maps are updated from new observations during model-predictive control. In both ReKep and VoxPoser, the motion solver can respond to changes in geometry without calling the language model at every control step.

For manipulation that involves substantial physical contact, CoRAL~\citep{survey_ax_2605_02600} uses an LLM to generate cost functions and contact strategies for motion optimization, while a VLM provides initial estimates of physical parameters. Interaction feedback supports updates to the physical model and revisions to the cost function or contact strategy. This separates two kinds of adjustment: the solver can find a different motion under the current requirements, or the model can change those requirements when they prove unsuitable.

Contact Harness~\citep{survey_ax_2611_29377} divides the response to contact between the model and the controller. The model can request images, force readings, or tactile readings where available. It chooses the direction and extent of a move and can set a soft force limit. During the move, the controller drops any movement component that would increase the load along an axis already over that limit. An impedance controller tracks the resulting target.

\paragraph{\textbf{Execution Feedback.}}

As the robot moves, the system needs to monitor what happens and decide how to respond. Some systems process this feedback within the action model, while others send it to a language model. For example, ARMS~\citep{survey_ax_2609_28429} adds a visual trigger, robot-state predictions, and a compressed history of past actions to a VLA. These components help the VLA decide when to act and how to respond, while perception and memory continue to update during movement. However, remembering past actions does not guarantee that stored object locations are still correct. If an object moves without being observed, the robot needs a new observation to update its location.

Feedback can also pass from a motion controller to a language model. COMPASS~\citep{survey_ax_2609_28247} uses this approach in simulated multi-agent flocking, separating slower language-model reasoning from faster motion control. A spatial-transformer controller generates actions and sends learned feedback tokens to the language model. The language model then uses these tokens to update its reasoning.

Other systems give the language model a more limited role before execution. In one such system, the LLM interprets the instruction and resolves ambiguity, then a deterministic finite-state controller carries out the selected operation~\citep{survey_ax_2608_29379}. The execution layer maps validated tool calls to predefined motion-planning templates. Before moving the robot, it checks kinematic feasibility and possible collisions. Here, the described checks take place before movement, while the fixed controller manages execution.

Overall, the model's output should match what the execution layer can do. For example, if the required skill is already available, the model may only need to name the action or provide a language instruction. When these commands are not enough, it can specify target poses, write control programs, or define motion constraints and cost functions. In each case, the system can check whether the robot can carry out the requested movement before execution. It can then use feedback during and after execution to check whether the action worked and decide what to do next. Related works are compared in Table~\ref{tab:model-robot-control}.

% Robot Control: Control Output / Executor / Feedback.
\par\Needspace{12\baselineskip}
\begingroup
\mechanismtableglue
\CompactTableStyle
\setlength{\LiteratureWidth}{\dimexpr\linewidth-8\tabcolsep\relax}
\setlength{\LTpre}{4pt}
\setlength{\LTpost}{4pt}
\setlength{\LTcapwidth}{\linewidth}
\captionsetup{singlelinecheck=false,justification=raggedright}
\LiteratureContinuationSetup

\begin{longtable}{@{}
>{\RaggedRight\arraybackslash}p{0.220\LiteratureWidth}
>{\centering\arraybackslash}p{0.070\LiteratureWidth}
>{\RaggedRight\leftskip=6pt\relax\hyphenpenalty=10000\exhyphenpenalty=10000\arraybackslash}p{0.220\LiteratureWidth}
>{\RaggedRight\hyphenpenalty=10000\exhyphenpenalty=10000\arraybackslash}p{0.245\LiteratureWidth}
>{\RaggedRight\hyphenpenalty=10000\exhyphenpenalty=10000\arraybackslash}p{0.245\LiteratureWidth}@{}}

\caption[Comparison of works on robot control]{Comparison of works on robot control (Section~\ref{sec:model-robot-control}). Control Output lists the commands or representations passed to the execution module. Executor identifies the module that turns these outputs into robot actions. Feedback lists the information returned during execution.}
\label{tab:model-robot-control}\\

\LiteratureTopRule{Model}
\rowcolor{TSTModelLight}
\textbf{Work} & \textbf{Link} & \textbf{Control Output} & \textbf{Executor} & \textbf{Feedback} \\
\midrule
\endfirsthead

\multicolumn{5}{@{}l}{\tablename~\thetable\ (continued)} \\
\LiteratureTopRule{Model}
\rowcolor{TSTModelLight}
\textbf{Work} & \textbf{Link} & \textbf{Control Output} & \textbf{Executor} & \textbf{Feedback} \\
\midrule
\endhead

\midrule
\multicolumn{5}{r@{}}{Continued on next page} \\
\endfoot

\bottomrule
\endlastfoot

VoxPoser \mbox{\citep{survey_ax_2307_05973}}
& \LiteratureLinks{https://proceedings.mlr.press/v229/huang23b.html}{https://github.com/huangwl18/VoxPoser}
& Cost & Optimizer & Vision; State \\[1pt]

\rowcolor{black!3}
ReKep \mbox{\citep{survey_ax_2409_01652}}
& \LiteratureLinks{https://proceedings.mlr.press/v270/huang25g.html}{https://github.com/huangwl18/ReKep}
& Constraint & Optimizer & Vision; State \\[1pt]

FAEA \mbox{\citep{survey_ax_2601_20334}}
& \LiteratureLinks{https://arxiv.org/abs/2601.20334v2}{https://github.com/robiemusketeer/faea-sim}
& Program & Interpreter & State; Execution Status \\[1pt]

\rowcolor{black!3}
CoRAL \mbox{\citep{survey_ax_2605_02600}}
& \LiteratureLinks{https://www.roboticsproceedings.org/rss22/p054.html}{}
& Cost & Optimizer & State \\[1pt]

SCOPE \mbox{\citep{survey_doi_353abec881f4d997}}
& \LiteratureLinks{https://doi.org/10.1145/3757279.3785641}{https://github.com/HindsboNikolaj/SCOPE}
& Action & Controller & Text; Execution Status \\[1pt]

\rowcolor{black!3}
LWSV \mbox{\citep{survey_ax_2606_12299}}
& \LiteratureLinks{https://arxiv.org/abs/2606.12299v1}{}
& Language & Learned Policy & Vision; State \\[1pt]

ABot-N1 \mbox{\citep{survey_ax_2607_10383}}
& \LiteratureLinks{https://arxiv.org/abs/2607.10383v3}{https://github.com/amap-cvlab/ABot-Navigation}
& Language; Target & Learned Policy & Vision \\[1pt]

\rowcolor{black!3}
VIA \mbox{\citep{survey_ax_2607_11119}}
& \LiteratureLinks{https://arxiv.org/abs/2607.11119v2}{https://github.com/hengyuan-hu/via}
& Target & Controller & Vision; State; Execution Status \\[1pt]

MEMENTO \mbox{\citep{survey_ax_2607_22832}}
& \LiteratureLinks{https://arxiv.org/abs/2607.22832v1}{https://github.com/sygkounas/MEMENTO}
& Program & Interpreter & State; Execution Status \\[1pt]

\rowcolor{black!3}
ETA \mbox{\citep{survey_ax_2608_03924}}
& \LiteratureLinks{https://arxiv.org/abs/2608.03924v1}{https://github.com/OpenMOSS/OpenETA}
& Action & Controller & Vision; State; Execution Status \\[1pt]

ART \mbox{\citep{survey_ax_2608_14047}}
& \LiteratureLinks{https://arxiv.org/abs/2608.14047v3}{}
& Action & Controller & Vision; State \\[1pt]

\rowcolor{black!3}
BSP \mbox{\citep{survey_ax_2608_29379}}
& \LiteratureLinks{https://arxiv.org/abs/2608.29379v1}{}
& Action & Controller & State; Execution Status \\[1pt]

Show-Harness \mbox{\citep{survey_ax_2609_10522}}
& \LiteratureLinks{https://arxiv.org/abs/2609.10522v1}{https://github.com/showlab/Show-Harness}
& Action & Controller & Vision; State; Execution Status \\[1pt]

\rowcolor{black!3}
ManiSkillFormer \mbox{\citep{survey_ax_2609_16331}}
& \LiteratureLinks{https://arxiv.org/abs/2609.16331v3}{}
& Target & Controller & State; Execution Status \\[1pt]

A-RAM \mbox{\citep{survey_ax_2609_19347}}
& \LiteratureLinks{https://arxiv.org/abs/2609.19347v1}{}
& Trajectory & Controller & State \\[1pt]

\rowcolor{black!3}
GPT-Policy \mbox{\citep{survey_ax_2609_19138}}
& \LiteratureLinks{https://arxiv.org/abs/2609.19138v1}{https://github.com/cheng-haha/GPT-Policy}
& Target & Controller & Vision; State; Execution Status \\[1pt]

KINO \mbox{\citep{survey_ax_2609_18869}}
& \LiteratureLinks{https://arxiv.org/abs/2609.18869v1}{}
& Target & Learned Policy & Vision; State; Execution Status \\[1pt]

\rowcolor{black!3}
RoboDawn \mbox{\citep{survey_ax_2609_22966}}
& \LiteratureLinks{https://arxiv.org/abs/2609.22966v1}{https://github.com/Hugo-AGI/RoboDawn}
& Action & Controller & Vision; State; Execution Status \\[1pt]

ARMS \mbox{\citep{survey_ax_2609_28429}}
& \LiteratureLinks{https://arxiv.org/abs/2609.28429v1}{}
& Feature & Learned Policy & Vision; State \\[1pt]

\rowcolor{black!3}
COMPASS \mbox{\citep{survey_ax_2609_28247}}
& \LiteratureLinks{https://arxiv.org/abs/2609.28247v1}{}
& Feature & Learned Policy & State; Feature \\[1pt]

Robo-Harness K1 \mbox{\citep{survey_ax_2609_29389}}
& \LiteratureLinks{https://arxiv.org/abs/2609.29389v1}{https://github.com/Robo-Harness/k1}
& Target & Controller & Vision; State; Execution Status \\[1pt]

\rowcolor{black!3}
WAA \mbox{\citep{survey_ax_2609_29964}}
& \LiteratureLinks{https://arxiv.org/abs/2609.29964v1}{}
& Target & Controller & Vision; Execution Status \\[1pt]

MIP~\mbox{\citep{survey_ax_2607_26148}}
& \LiteratureLinks{https://arxiv.org/abs/2607.26148v4}{https://github.com/jianzhou0420/MIP}
& Action & Controller & Vision; Execution Status \\[1pt]

\rowcolor{black!3}
AW-0 \mbox{\citep{survey_ax_2610_30136}}
& \LiteratureLinks{https://arxiv.org/abs/2610.30136v1}{https://github.com/ZhuoyangLiu2005/agentic-wam}
& Action; Program & Controller; Interpreter & Vision \\[1pt]

AMP1 \mbox{\citep{survey_ax_2611_09892}}
& \LiteratureLinks{https://arxiv.org/abs/2611.09892v3}{https://github.com/apple-aiml-research/ml-amp}
& Trajectory & Controller & Vision; State; Execution Status \\[1pt]

\rowcolor{black!3}
Embo-Harness \mbox{\citep{survey_ax_2611_31201}}
& \LiteratureLinks{https://arxiv.org/abs/2611.31201v1}{https://github.com/DravenALG/Embo-Harness}
& Target & Controller & Vision \\[1pt]

PEAR~\mbox{\citep{survey_ax_2611_18562}}
& \LiteratureLinks{https://arxiv.org/abs/2611.18562v2}{https://github.com/mit-han-lab/pear}
& Target & Controller & Vision; State \\[1pt]

\rowcolor{black!3}
SOLVE~\mbox{\citep{survey_ax_2611_34750}}
& \LiteratureLinks{https://arxiv.org/abs/2611.34750v1}{}
& Trajectory & Controller & Vision; State; Execution Status \\[1pt]

Contact Harness~\mbox{\citep{survey_ax_2611_29377}}
& \LiteratureLinks{https://arxiv.org/abs/2611.29377v1}{https://github.com/NTUMARS/ContactHarness}
& Target & Controller & Vision; State; Execution Status \\[1pt]

\rowcolor{black!3}
JustRUA~\mbox{\citep{survey_ax_2611_16009}}
& \LiteratureLinks{https://arxiv.org/abs/2611.16009v1}{https://github.com/ZefanW/JustRUA}
& Target & Controller & Vision; State; Execution Status \\[1pt]
\end{longtable}
\endgroup

\subsection{Adaptation}
\label{sec:model-adaptation}

Adaptation enables robotic systems to improve from experience by modifying inference-time conditioning, planner configurations, or learned model parameters. These mechanisms differ in their scope and persistence, but can be combined to support both immediate recovery and long-term improvement.

\begingroup\setlength{\emergencystretch}{1em}\paragraph{Experience-Conditioned Decision Making.}
Prior experience can guide decisions without updating model parameters. TidyBot~\citep{adaptation_ax_2305_05658} infers user preferences from examples, while DROC~\citep{survey_ax_2311_10678} extracts reusable knowledge from language corrections. CMMR-VLN~\citep{survey_ax_2603_07997} converts retrieved navigation experience into prioritized rules, whereas PhysMem~\citep{adaptation_ax_2602_20323} maintains and revises hypotheses about physical interactions. These methods connect adaptation to memory (Section~\ref{sec:harness}) by conditioning subsequent decisions on stored experience. Similarly, GPT-Policy~\citep{survey_ax_2609_19138} conditions a frozen VLM on demonstrations, action references, and interaction history to support within-task adaptation. Demonstrations can guide a fixed policy at inference or serve as training examples for updating its parameters~\citep{survey_ax_2609_36012}. AMP1+~\citep{survey_ax_2611_09892} tested 42 practiced tasks on new layouts. Success rose from 26.4\% with an empty memory to 84.1\% with notes and code from a selected practice run for each task (Table~\ref{tab:retained-experience}). However, improvements that depend on retained context do not necessarily indicate persistent capability acquisition.
\par\endgroup
\paragraph{Planner Adaptation.}
Experience can also improve planning without modifying the underlying policy. SERP~\citep{survey_ax_2603_02772} combines gradient-based optimization of planner parameters with VLM-guided adjustments when optimization stalls. Its MARS extension further retrieves prior recovery experience to inform subsequent configurations. Such methods primarily address failures in planning and search rather than limitations in the policy's learned capabilities.

\paragraph{Policy and Model Updates.}
Parameter updates enable more persistent behavioral changes. OLAF~\citep{adaptation_ax_2310_17555} uses verbal corrections to relabel actions for policy training. Yell At Your Robot~\citep{adaptation_ax_2403_12910} instead updates a high-level policy while keeping the low-level controller fixed. Uncertainty-Aware Policy Steering (UPS)~\citep{survey_ax_2602_22474} separates instruction clarification from residual policy learning using corrective demonstrations. These approaches target different sources of error, including task interpretation, action selection, and execution.

Autonomous interaction provides an additional learning signal. Self-Improving Embodied Foundation Models~\citep{survey_ax_2509_15155} uses a frozen supervised checkpoint for progress rewards and success detection while optimizing a separate policy with reinforcement learning. Its physical experiments use LanguageTable, whereas ALOHA self-improvement is evaluated in simulation. Language Model Predictive Control~\citep{adaptation_ax_2402_11450} fine-tunes a code-generating LLM through simulated teaching interactions. Reflective Test-Time Planning~\citep{survey_ax_2602_21198} updates high-level action selection and reflection from retrospective evaluations while keeping the external evaluator fixed.

SEES~\citep{survey_ax_2609_32698} uses repeated failures to choose which stage to train next. It finds the stage that most often blocks the task and restarts the simulation from states just before that stage. It then trains an adapter shared by related skills, using a generated test to check for success. A checkpoint is kept when it improves validation success on the selected skills. In a cutting-board task, scrubbing was the first target and navigation became the next. After two equal-budget rounds, this choice led to 32 successes on 50 held-out scenes, compared with 14 when stages were chosen at random.
Together, these methods show how feedback can guide updates to a policy, a planner, or a skill shared across tasks.

\paragraph{Learning across Tasks and Deployments.}
Experience collected during execution can support subsequent training. PAFF~\citep{adaptation_ax_2212_07398} relabels exploratory trajectories with language instructions, while RoboCat~\citep{survey_ax_2306_11706} incorporates data generated by task-adapted policies into successive generalist training iterations. VERITAS~\citep{survey_ax_2606_18247} selects verified successful executions, whereas HERO~\citep{survey_ax_2607_26809} combines heuristic execution and exemplar reuse before policy training. VLAW~\citep{survey_ax_2602_12063} iteratively improves a world model using real interactions and filters generated trajectories for policy updates. At deployment scale, RECAP~\citep{adaptation_ax_2511_14759} and Learning While Deploying~\citep{survey_ax_2605_00416} leverage autonomous experience and human corrections to improve policies. These approaches connect adaptation to data collection and selection (Section~\ref{sec:data}), with performance depending on supervision quality, experience diversity, and interaction costs.

Before replacing a deployed policy, an agent can test the candidate on the target skills and on skills learned earlier. Robo-COP~\citep{survey_ax_2610_09228} lets a language-model controller decide when to fine-tune a candidate VLA on all examples curated for that task so far. A reviewer checks whether the candidate itself performs the skills named by the controller. Task success may fall at most 10 percentage points below the current policy's recent rate. The same limit applies to each earlier skill with enough calls to compare completion rates. A candidate replaces the current policy only if all three checks pass. Across ten simulated RoboLab tasks, mean held-out success was 73.8\%. The same harness reached 64.8\% with a frozen VLA and 65.8\% when it trained every 30 trials and accepted every update.

RoboEvo$^2$~\citep{survey_ax_2611_34675} first repairs skills, memory, or routing rules after a failure and checks the result. Successful corrections train the policy. When a failed and a successful continuation share a starting state, the update also favors the successful one. Across 800 held-out LIBERO-PRO episodes, success for $\pi_{0.5}$ alone rose from 11.0\% to 84.1\% over eight rounds. With the harness, success rose from 30.4\% to 90.5\%.

The choice of adaptation mechanism depends on the source of failure. Perception errors may require additional observations, planning errors may require replanning, and persistent execution failures may motivate policy updates. Evaluation should distinguish within-episode recovery, cross-task generalization, and improvements following retraining, while assessing whether previously acquired capabilities are preserved. Related works are compared in Table~\ref{tab:model-adaptation}.

% Model Adaptation: Adapted Component / Adaptation Method / Adaptation Signal.
\par\Needspace{12\baselineskip}
\begingroup
\mechanismtableglue
\CompactTableStyle
\setlength{\LiteratureWidth}{\dimexpr\linewidth-8\tabcolsep\relax}
\setlength{\LTpre}{4pt}
\setlength{\LTpost}{4pt}
\setlength{\LTcapwidth}{\linewidth}
\captionsetup{singlelinecheck=false,justification=raggedright}
\LiteratureContinuationSetup

\begin{longtable}{@{}
>{\RaggedRight\arraybackslash}p{0.245\LiteratureWidth}
>{\centering\arraybackslash}p{0.065\LiteratureWidth}
>{\RaggedRight\leftskip=6pt\relax\hyphenpenalty=10000\exhyphenpenalty=10000\arraybackslash}p{0.220\LiteratureWidth}
>{\RaggedRight\hyphenpenalty=10000\exhyphenpenalty=10000\arraybackslash}p{0.270\LiteratureWidth}
>{\RaggedRight\hyphenpenalty=10000\exhyphenpenalty=10000\arraybackslash}p{0.200\LiteratureWidth}@{}}

\caption[Comparison of works on adaptation]{Comparison of works on adaptation (Section~\ref{sec:model-adaptation}). Demonstrations include task examples and action trajectories; Corrections include corrective instructions and interventions. Execution Outcomes include success, failure, and execution feedback.}
\label{tab:model-adaptation}\\

\LiteratureTopRule{Model}
\rowcolor{TSTModelLight}
\textbf{Work} & \textbf{Link} & \textbf{Adapted Component} & \textbf{Adaptation Method} & \textbf{Adaptation Signal} \\
\midrule
\endfirsthead

\multicolumn{5}{@{}l}{\tablename~\thetable\ (continued)} \\
\LiteratureTopRule{Model}
\rowcolor{TSTModelLight}
\textbf{Work} & \textbf{Link} & \textbf{Adapted Component} & \textbf{Adaptation Method} & \textbf{Adaptation Signal} \\
\midrule
\endhead

\midrule
\multicolumn{5}{r@{}}{Continued on next page} \\
\endfoot

\bottomrule
\endlastfoot

PAFF \mbox{\citep{adaptation_ax_2212_07398}}
& \LiteratureLinks{https://openaccess.thecvf.com/content/CVPR2023/html/Ge_Policy_Adaptation_From_Foundation_Model_Feedback_CVPR_2023_paper.html}{https://github.com/geyuying/PAFF_code}
& Model Weights & Supervised Learning & Demonstrations; Execution Outcomes \\[1pt]

\rowcolor{black!3}
TidyBot \mbox{\citep{adaptation_ax_2305_05658}}
& \LiteratureLinks{https://link.springer.com/article/10.1007/s10514-023-10139-z}{https://github.com/jimmyyhwu/tidybot}
& Context & Summarization & Demonstrations \\[1pt]

RoboCat \mbox{\citep{survey_ax_2306_11706}}
& \LiteratureLinks{https://arxiv.org/pdf/2306.11706v2}{}
& Model Weights & Supervised Learning & Demonstrations; Execution Outcomes \\[1pt]

\rowcolor{black!3}
OLAF \mbox{\citep{adaptation_ax_2310_17555}}
& \LiteratureLinks{https://arxiv.org/abs/2310.17555v1}{https://github.com/UT-Austin-RPL/olaf}
& Model Weights & Supervised Learning & Corrections \\[1pt]

DROC \mbox{\citep{survey_ax_2311_10678}}
& \LiteratureLinks{https://arxiv.org/abs/2311.10678v2}{https://github.com/Stanford-ILIAD/droc}
& Context; Memory & In-Context Learning; Summarization; Experience Retrieval & Corrections \\[1pt]

\rowcolor{black!3}
LMPC \mbox{\citep{adaptation_ax_2402_11450}}
& \LiteratureLinks{https://roboticsproceedings.org/rss20/p125.html}{}
& Model Weights; Context & Supervised Learning; In-Context Learning & Corrections; Execution Outcomes \\[1pt]

Yell At Your Robot \mbox{\citep{adaptation_ax_2403_12910}}
& \LiteratureLinks{https://www.roboticsproceedings.org/rss20/p025.html}{https://github.com/yay-robot/yay_robot}
& Model Weights & Supervised Learning & Corrections \\[1pt]

\rowcolor{black!3}
SIEFM \mbox{\citep{survey_ax_2509_15155}}
& \LiteratureLinks{https://proceedings.neurips.cc/paper_files/paper/2025/hash/a3ac59dc0bf607fa35535f546110fc34-Abstract-Conference.html}{https://github.com/self-improving-efms/self-improving-efms.github.io/blob/main/pointmass_notebook.ipynb}
& Model Weights & Reinforcement Learning & Rewards \\[1pt]

RECAP \mbox{\citep{adaptation_ax_2511_14759}}
& \LiteratureLinks{https://arxiv.org/abs/2511.14759v2}{}
& Model Weights & Reinforcement Learning & Corrections; Rewards \\[1pt]

\rowcolor{black!3}
VLAW \mbox{\citep{survey_ax_2602_12063}}
& \LiteratureLinks{https://proceedings.mlr.press/v306/guo26i.html}{https://github.com/Robert-gyj/Ctrl-World}
& Model Weights & Reinforcement Learning & Rewards; Execution Outcomes \\[1pt]

PhysMem \mbox{\citep{adaptation_ax_2602_20323}}
& \LiteratureLinks{https://arxiv.org/abs/2602.20323v6}{https://github.com/haoyangli16/PhysMem}
& Memory & Hypothesis Testing & Execution Outcomes \\[1pt]

\rowcolor{black!3}
Reflective Test-Time Planning \mbox{\citep{survey_ax_2602_21198}}
& \LiteratureLinks{https://arxiv.org/abs/2602.21198v3}{https://github.com/Reflective-Test-Time-Planning/Reflective-Test-Time-Planning}
& Model Weights; Memory & Supervised Learning; Reinforcement Learning; Reflection & Rewards; Execution Outcomes \\[1pt]

UPS \mbox{\citep{survey_ax_2602_22474}}
& \LiteratureLinks{https://www.roboticsproceedings.org/rss22/p142.html}{https://github.com/CMU-IntentLab/uncertainty_aware_policy_steering}
& Model Weights & Supervised Learning & Corrections \\[1pt]

\rowcolor{black!3}
SERP \mbox{\citep{survey_ax_2603_02772}}
& \LiteratureLinks{https://arxiv.org/abs/2603.02772v2}{}
& Planner Settings; Memory & Parameter Tuning; Experience Retrieval & Execution Outcomes \\[1pt]

CMMR-VLN \mbox{\citep{survey_ax_2603_07997}}
& \LiteratureLinks{https://arxiv.org/abs/2603.07997v1}{}
& Memory & Reflection; Experience Retrieval & Execution Outcomes \\[1pt]

\rowcolor{black!3}
Learning While Deploying \mbox{\citep{survey_ax_2605_00416}}
& \LiteratureLinks{https://arxiv.org/abs/2605.00416v4}{}
& Model Weights & Reinforcement Learning & Corrections; Rewards \\[1pt]

VERITAS \mbox{\citep{survey_ax_2606_18247}}
& \LiteratureLinks{https://www.roboticsproceedings.org/rss22/p079.html}{https://github.com/princeton-prism/veritas}
& Model Weights & Supervised Learning & Demonstrations; Execution Outcomes \\[1pt]

\rowcolor{black!3}
HERO \mbox{\citep{survey_ax_2607_26809}}
& \LiteratureLinks{https://arxiv.org/abs/2607.26809v1}{}
& Model Weights; Memory & Supervised Learning; Experience Retrieval & Demonstrations; Execution Outcomes \\[1pt]

EmbodiedSkills \mbox{\citep{survey_ax_2609_01281}}
& \LiteratureLinks{https://arxiv.org/abs/2609.01281v1}{https://github.com/ZJU4EmbodiedAI/EmbodiedSkills}
& Model Weights & Supervised Learning & Demonstrations \\[1pt]

\rowcolor{black!3}
GPT-Policy \mbox{\citep{survey_ax_2609_19138}}
& \LiteratureLinks{https://arxiv.org/abs/2609.19138v1}{https://github.com/cheng-haha/GPT-Policy}
& Context & In-Context Learning & Demonstrations; Execution Outcomes \\[1pt]

RoboDawn \mbox{\citep{survey_ax_2609_22966}}
& \LiteratureLinks{https://arxiv.org/abs/2609.22966v1}{https://github.com/Hugo-AGI/RoboDawn}
& Memory & In-Context Learning & Execution Outcomes \\[1pt]

\rowcolor{black!3}
SEES \mbox{\citep{survey_ax_2609_32698}}
& \LiteratureLinks{https://arxiv.org/abs/2609.32698v1}{}
& Model Weights & Reinforcement Learning & Execution Outcomes; Rewards \\[1pt]

Robo-COP~\mbox{\citep{survey_ax_2610_09228}}
& \LiteratureLinks{https://arxiv.org/abs/2610.09228v1}{}
& Model Weights & Supervised Learning & Execution Outcomes \\[1pt]

\rowcolor{black!3}
RoboEvo$^2$~\mbox{\citep{survey_ax_2611_34675}}
& \LiteratureLinks{https://arxiv.org/abs/2611.34675v1}{https://github.com/kaiknower/RoboEvo2}
& Model Weights & Supervised Learning; Preference Learning & Corrections; Execution Outcomes \\[1pt]
\end{longtable}
\endgroup

\section{Data: How Should the Agent Collect, Annotate, and Refine Robot Data?}
\label{sec:data}

% Progress toward agentic robots depends in part on how models shape the experience available for learning and future decisions. Models can direct further collection, interpret demonstrations, and use execution feedback to select or revise supervision. We organize these contributions into three interventions. \textit{Collection} determines what new interactions or demonstrations to acquire. \textit{Annotation} attaches task meaning, temporal structure, or learning targets to the resulting records. \textit{Refinement} assesses data quality, selects useful records, and corrects unreliable content or supervision. A single system may collect a failed attempt, annotate the cause of failure, retain a useful recovery segment, and correct an inaccurate success label. The following discussion traces what the model decides, what evidence informs that decision, and how the resulting data support later robot behavior. Some methods close a feedback loop around data acquisition or revision. Others provide annotations and supervision needed to develop these capabilities.

\begin{figure}[H]
\centering
\includegraphics[width=\linewidth]{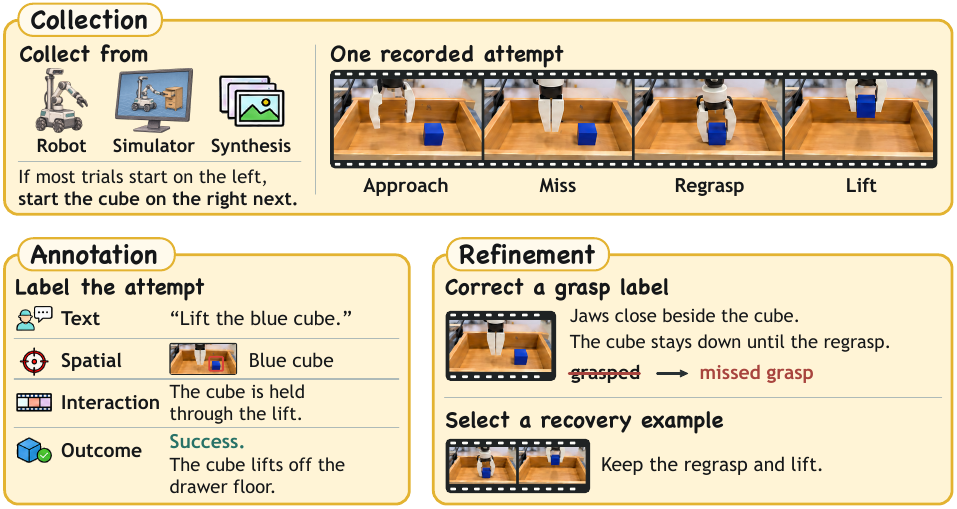}
\caption{An agent can turn an attempt to lift a cube into training data. Collection varies where the cube starts. Annotation marks the attempt as successful once the cube is lifted. Refinement relabels the missed grasp and keeps the regrasp and lift as a recovery example.}
\label{fig:experience-interventions}
\end{figure}

Agents shape the data available for robot learning by deciding which interactions to collect, how to annotate them, and which records to correct or retain. Collection can draw on physical robots, simulation, or synthetic data. Annotation and Refinement can use the same recording for different purposes. In Figure~\ref{fig:experience-interventions}, the robot misses its first grasp but ultimately lifts the cube. The successful outcome is a useful label for the whole attempt, while the missed grasp and subsequent recovery matter when correcting labels and selecting training examples.

\subsection{Collection}
\label{sec:data-collection}

% Collection gives a model influence over which tasks, states, and behaviors enter a robot's experience. This influence can take the form of selecting another attempt, requesting a demonstration, or specifying behavior to generate. We organize the methods by the source of the resulting experience, covering physical interaction, simulator execution, and generative synthesis. A collection pipeline may combine these sources by executing a generated motion in simulation and recording the resulting trajectory. Each source can also support targeted collection around failures. The comparison concerns how model decisions shape the acquired data and how execution results inform those decisions. Environment construction and policy updates are covered in Environment and Model Adaptation, respectively.

In agent-based data collection, autonomous agents generate robot trajectories or demonstrations. We group these methods into three categories.
\textit{Physical Data Collection} gathers data from real‑robot interaction and human demonstration.
\textit{Simulation Data Generation} produces trajectories by running tasks, policies, or programs within a simulator.
\textit{Generative Data Synthesis} uses generative models to create artificial observations and motions.
%For each category, we examine what experience models choose to collect, how data is obtained, and how feedback guides further collection.

\paragraph{\textbf{Physical Data Collection.}}
% Early work used foundation models to choose tasks for real-world data collection. 
An early example, AutoRT~\citep{survey_ax_2401_12963}, uses language models to propose collection tasks drawing upon existing visual scene descriptions and the capabilities of available collection policies. The selected tasks are then attempted through scripted picking, RT-2~\citep{survey_ax_2307_15818}, or human teleoperation to collect real-world trajectories.
SOAR~\citep{survey_ax_2407_20635} uses CogVLM~\citep{survey_ax_2311_03079} to identify tasks that can be performed in the current scene, then prioritizes less-attempted tasks to collect data across a wider range of tasks.

Sustained agent-based data collection involves key design choices for task execution, scene reset, and human intervention. Zero2Skill~\citep{survey_ax_2607_14047} uses Seed1.8~\citep{survey_ax_2603_20633} to process images captured after data‑collection and scene‑reset stages. Based on the resulting trial‑outcome evaluation, the data‑collection loop either proceeds onward or retries the ongoing trial, with human assistance invoked once retries reach the predefined limit.
RoboClaw~\citep{survey_ax_2603_11558} combines visual observations with task progress in a VLM controller, which invokes a paired reset policy after confirming task completion. 
%These outcomes update the task counts that guide subsequent collection.

Agents can also direct additional data collection toward underrepresented tasks and weaknesses revealed by robot execution. 
Tether~\citep{survey_ax_2603_03278} prioritizes tasks lacking sufficient successful demonstrations and generates scene‑adapted preparatory actions. In its early collection experiments, the pipeline used Gemini 2.5 Flash~\citep{survey_ax_2507_06261} to plan task sequences for scene preparation leading to the target collection task, while GPT-4.1 judged task success. Success evaluation compared camera views before and after execution, with an additional check matching the source demonstration's final keypoint to the gripper's final image position. The paper later adopts Gemini Robotics-ER 1.5 for both task selection and success evaluation, and reports a post-hoc evaluation without the additional correspondence check.
RoboLineage~\citep{survey_ax_2606_22142} reviews recorded video segments and keyframes, together with operator notes and robot logs, to identify where task execution fails. It combines these findings with data-coverage and policy-evaluation summaries to recommend additional collection for weak task phases.
FIND~\citep{survey_ax_2609_32069} also uses a VLM to judge each attempt as successful or unsuccessful from the task instruction and before-and-after scene images. 
%For each task, it maintains a rolling window of 20 outcomes, replacing the oldest result after each new attempt. The resulting success rate determines task-sampling weights, giving currently feasible tasks with lower success rates more practice.
UPS~\citep{survey_ax_2602_22474} also checks whether the policy can complete a task and requests a human correction when that check fails.
%These methods use execution feedback to decide which tasks need more data and when human demonstrations are needed.

An agent can also collect demonstrations for a smaller policy. Kim et al.~\citep{survey_ax_2610_03615} used GPT-6 Astra to collect 20 successful real-robot demonstrations for each of four tasks. They added 20 human demonstrations per task to train local policies. These policies completed 41 of 80 test trials, the same total as policies trained on 40 human demonstrations per task. The study did not test 20 human demonstrations alone, so the gain from adding the generated examples remains unknown.

\paragraph{\textbf{Simulation Data Generation.}}
RoboGen~\citep{survey_ax_2311_01455} is an early example that introduces a robotic agent that proposes tasks, constructs simulation environments, and generates training supervision.
Specifically, the agent decomposes the task into subtasks, selects and arranges the required objects in Genesis~\citep{Genesis}, and uses methods such as motion planning to generate robot behavior in simulation. 
% Executing these subtasks in Genesis~\citep{Genesis} produces robot interaction data, and repeating the process extends data collection to different tasks and environments.
To generate tasks grounded in everyday human activities, Video2Policy~\citep{survey_ax_2502_09886} uses internet videos as the source of task specifications. 
It reconstructs the manipulated objects and estimates their poses, then uses a vision-language model to generate executable task code for Isaac Gym~\citep{survey_ax_2108_10470}. 
%The model iteratively revises the rewards based on policy training and evaluation logs, and successful trajectories from the resulting policies are collected to train a general manipulation policy.
GenSim2~\citep{survey_ax_2410_03645} uses an agent to create manipulation tasks and guide motion planning, then executes the motions in SAPIEN~\citep{survey_ax_2003_08515} to collect robot trajectories. RoboFactory~\citep{survey_ax_2503_16408} focuses on collaboration between multiple robots, using a language-model planner to assign subtasks and coordinate their execution in ManiSkill~\citep{survey_ax_2410_00425}. 
% RoboTwin 2.0~\citep{survey_ax_2506_18088} and HumanoidGen~\citep{survey_ax_2507_00833} generate bimanual manipulation data in SAPIEN with different emphases. RoboTwin 2.0 uses visual feedback to repair generated collection programs and runs them across randomized scenes. HumanoidGen targets coordinated arm and dexterous hand movements, using simulation feedback to revise failed steps.

More recent work broadens the types of simulation data collected and gives agents greater responsibility for the collection process. RoboTalk~\citep{survey_ax_2609_23997} generates communication messages alongside joint robot actions, validates their sequences, and replays them in RoboCasa365~\citep{survey_ax_2603_04356} to collect the corresponding visual observations. ARSTAG~\citep{survey_ax_2609_24563} and EmbodiedSWE-Gen~\citep{survey_ax_2609_27308} automate the preparation and expansion of collection programs. 
%Starting from a scene image and an instruction, ARSTAG coordinates agents to reconstruct the scene, generate robot behavior, and collect task-specific data in Isaac Sim~\citep{IsaacSim}. EmbodiedSWE-Gen starts from a solution tested by a coding agent and expands it into diverse trajectories in Isaac Lab~\citep{survey_ax_2511_04831} by varying scene layouts, grasp choices, and action sequences.
RoboMite~\citep{survey_ax_2611_37041} starts from 17 human-written task suites. An LLM writes demonstration programs and revises them using execution feedback. Running the programs yields 1,700 demonstrations in which later instructions depend on earlier events.

Policy failures can also guide what the agent collects next. EmbodiRSI~\citep{survey_ax_2609_38905} uses failures in Isaac Sim to guide data collection for policy training. It generates recovery trajectories and collects additional data in difficult scene configurations. The policy is then updated and evaluated again, allowing the agent to adjust subsequent collection to the robot's remaining weaknesses.

Agentic-Teacher~\citep{survey_ax_2611_17394} starts from a failed policy rollout and identifies the frame where the error began. It restores a simulator state, tests a corrective program, and uses the outcome to revise it. In one example, the policy missed a bowl. The repair program realigned the gripper and lifted the bowl.

EmbodyAgent~\citep{survey_ax_2611_26196} turns 100 successful teacher episodes in LIBERO Spatial into 1,475 single-turn training examples (Section~\ref{sec:model-training}). Each pairs the history up to that turn with the teacher's explanation and next Python program, which can request an observation. RoboUniverse-10K~\citep{survey_ax_2611_31002} collects about 17,000 multi-turn GPT-6 Astra trajectories on training tasks and holds out 500 tasks for evaluation.

\paragraph{\textbf{Generative Data Synthesis.}}
Unlike previous methods like RoboTransfer~\citep{survey_ax_2505_23171}, Real2Edit2Real~\citep{survey_ax_2512_19402}, RoboVIP~\citep{survey_ax_2601_05241}, and RoboDream~\citep{survey_ax_2606_02577} that synthesize robot data through predefined generation and trajectory-processing pipelines, agent-based methods use large language models or multimodal large language models to decide what data to generate and coordinate the generation process.
RoboCurate~\citep{survey_ax_2602_18742} leverages Gemini 3 Pro to derive manipulation tasks from visual scene observations and produce prompts for video generation. Additionally, it checks whether simulator replays of the predicted actions are consistent with the motion in the corresponding generated videos.
% V-Dreamer~\citep{survey_ax_2603_18811} uses Qwen-Max~\citep{QwenMax} to specify scene objects and combines language-model task decomposition with video generation. It converts object movements in generated videos into robot trajectories and executes them in Genesis~\citep{Genesis}, allowing video generation to supply the motions used for collecting simulation data.
V-Dreamer~\citep{survey_ax_2603_18811} uses Qwen-Max~\citep{QwenMax} to turn a scene description into a list of object descriptions, which guides 3D asset generation and scene construction. 
For long instructions, a foundation model divides the task into subtasks that are processed in sequence. 
A video model generates a manipulation video from an image of the simulated scene and the current subtask instruction. The pipeline tracks points on the target object across video frames and reconstructs their three-dimensional motion. It combines this motion with a generated grasp pose to compute a robot trajectory, then executes the trajectory in Genesis~\citep{Genesis} to collect demonstrations for policy training.
Wh0~\citep{survey_ax_2606_22136} uses two agents to generate manipulation instructions: one expands object and attribute vocabulary, while the other selects less-used words based on usage counts. 
Qwen3-VL describes the required hand and object motions to guide human-hand video generation from edited workspace images. Wrist poses and finger-joint movements recovered from these videos provide action labels, which are combined with real robot demonstrations to train a dexterous policy.
These collection routes concern experience records. RATs~\citep{survey_ax_2606_19419} and Teach and Grow~\citep{survey_ax_2608_17209} instead primarily retain programs, lessons, or behavior-library entries. Their practice-to-skill workflows are therefore discussed in Section~\ref{sec:harness-skills}, which examines the construction, validation, and reuse of procedural artifacts. Related works are compared in Table~\ref{tab:data-collection}.

% Generated from survey_data/data_comparison_index.json; source-linked fixed tags.
\par\Needspace{10\baselineskip}
\begingroup
\mechanismtableglue
\CompactTableStyle\hyphenpenalty=10000
\setlength{\LiteratureWidth}{\dimexpr\linewidth-8\tabcolsep-3.3pt\relax}
\setlength{\LTpre}{3pt}
\setlength{\LTpost}{4pt}
\setlength{\LTcapwidth}{\linewidth}
\begin{longtable}{@{}>{\RaggedRight\arraybackslash}p{0.19\LiteratureWidth}>{\centering\arraybackslash}p{0.075\LiteratureWidth}!{\color{black!30}\vrule width .3pt\hspace{3pt}}>{\RaggedRight\arraybackslash}p{0.075\LiteratureWidth}>{\RaggedRight\arraybackslash}p{0.18\LiteratureWidth}>{\RaggedRight\arraybackslash}p{0.48\LiteratureWidth}@{}}
\caption[Comparison of works on data collection]{Comparison of works on data collection (Section~\ref{sec:data-collection}). Type: \raisebox{-1.5pt}{\includegraphics[height=10.5pt]{assets/table-icons/solid-robot.png}} physical interaction; \raisebox{-1.5pt}{\includegraphics[height=10.5pt]{assets/table-icons/solid-sim.png}} simulator execution; \raisebox{-1.5pt}{\includegraphics[height=10.5pt]{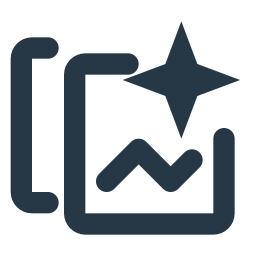}} generative-model synthesis. Counts retain each paper's units and scope.}\label{tab:data-collection}\label{tab:literature-data}\\
\LiteratureTopRule{Data}
\rowcolor{TSTDataLight}\textbf{Work} & \textbf{Link} & \textbf{Type} & \textbf{Collected data} & \textbf{Reported scale} \\
\midrule
\endfirsthead
\multicolumn{5}{@{}l}{\tablename~\thetable\ (continued)} \\
\LiteratureTopRule{Data}
\rowcolor{TSTDataLight}\textbf{Work} & \textbf{Link} & \textbf{Source} & \textbf{Collected data} & \textbf{Reported scale} \\
\midrule
\endhead
\midrule
\multicolumn{5}{r@{}}{Continued on next page} \\
\endfoot
\bottomrule
\endlastfoot
% work: ax-2311.01455 | survey_ax_2311_01455
% source: https://arxiv.org/html/2311.01455v3 | §3.1–3.4; §4.1
RoboGen~\citep{survey_ax_2311_01455} & \LiteratureLinks{https://proceedings.mlr.press/v235/wang24cc.html}{https://github.com/Genesis-Embodied-AI/RoboGen} & \raisebox{-1.5pt}{\includegraphics[height=10.5pt]{assets/table-icons/solid-sim.png}} & Robot trajectories & \textbf{106} generated tasks \\[3pt]
\rowcolor{black!3}
% work: ax-2407.20635 | survey_ax_2407_20635
% source: https://arxiv.org/html/2407.20635v2 | §3; §4; Appendix SOAR-Data
SOAR~\citep{survey_ax_2407_20635} & \LiteratureLinks{https://proceedings.mlr.press/v270/zhou25b.html}{https://github.com/rail-berkeley/soar} & \raisebox{-1.5pt}{\includegraphics[height=10.5pt]{assets/table-icons/solid-robot.png}} & Robot trajectories & \textbf{$\sim$30,500} trajectories in the released dataset. \\[3pt]
% work: ax-2502.09886 | survey_ax_2502_09886
% source: https://arxiv.org/html/2502.09886v1 | §3.2–3.3; §4 general-policy learning
Video2Policy~\citep{survey_ax_2502_09886} & \LiteratureLinks{https://arxiv.org/abs/2502.09886}{} & \raisebox{-1.5pt}{\includegraphics[height=10.5pt]{assets/table-icons/solid-sim.png}} & Robot trajectories & \textbf{100} trajectories/policy from \textbf{100} lifting videos \\[3pt]
\rowcolor{black!3}
% work: ax-2503.16408 | survey_ax_2503_16408
% source: https://arxiv.org/html/2503.16408v1 | §4.1; §4.3; §5.1
RoboFactory~\citep{survey_ax_2503_16408} & \LiteratureLinks{https://openaccess.thecvf.com/content/ICCV2025/html/Qin_RoboFactory_Exploring_Embodied_Agent_Collaboration_with_Compositional_Constraints_ICCV_2025_paper.html}{https://github.com/MARS-EAI/RoboFactory} & \raisebox{-1.5pt}{\includegraphics[height=10.5pt]{assets/table-icons/solid-sim.png}} & Robot trajectories & Up to \textbf{150} demos/task across \textbf{11} tasks \\[3pt]
% work: ax-2507.00833 | survey_ax_2507_00833
% source: https://arxiv.org/html/2507.00833v2 | §2.2–2.3; §5.1–5.2
HumanoidGen~\citep{survey_ax_2507_00833} & \LiteratureLinks{https://proceedings.neurips.cc/paper_files/paper/2025/hash/e4ef7454447baa15a424314e6284441b-Abstract-Conference.html}{https://github.com/TeleHuman/HumanoidGen} & \raisebox{-1.5pt}{\includegraphics[height=10.5pt]{assets/table-icons/solid-sim.png}} & Robot trajectories & \textbf{100} demos/task across \textbf{20} HGen-Bench tasks \\[3pt]
\rowcolor{black!3}
% work: ax-2602.18742 | survey_ax_2602_18742
% source: https://arxiv.org/html/2602.18742v1 | §2 (inverse dynamics models); §3.1–3.3; §4 (10K pre-filtering trajectories)
RoboCurate~\citep{survey_ax_2602_18742} & \LiteratureLinks{https://arxiv.org/abs/2602.18742v1}{} & \raisebox{-1.5pt}{\includegraphics[height=10.5pt]{assets/table-icons/solid-gen.png}} & Robot trajectories & \textbf{10,000} generated trajectories before filtering \\[3pt]
% work: ax-2602.22474 | survey_ax_2602_22474
% source: https://arxiv.org/html/2602.22474v2 | §IV-C; §V-B.1–2; Appendix A-F (teleoperated actions, residual targets, intervention labels)
UPS~\citep{survey_ax_2602_22474} & \LiteratureLinks{https://www.roboticsproceedings.org/rss22/p142.html}{https://github.com/CMU-IntentLab/uncertainty_aware_policy_steering} & \raisebox{-1.5pt}{\includegraphics[height=10.5pt]{assets/table-icons/solid-robot.png}}\hspace{3pt}\raisebox{-1.5pt}{\includegraphics[height=10.5pt]{assets/table-icons/solid-sim.png}} & Teleoperated demonstrations & -- \\[0pt]
\rowcolor{black!3}
% work: ax-2603.03278 | survey_ax_2603_03278
% source: https://arxiv.org/html/2603.03278v1 | §3.2; §4.3
Tether~\citep{survey_ax_2603_03278} & \LiteratureLinks{https://openreview.net/pdf?id=FqDmvMZish}{https://github.com/tether-research/tether} & \raisebox{-1.5pt}{\includegraphics[height=10.5pt]{assets/table-icons/solid-robot.png}} & Robot trajectories & \textbf{1,085} successes in \textbf{1,946} attempts \\[3pt]
% work: ax-2603.18811 | survey_ax_2603_18811
% source: https://arxiv.org/html/2603.18811v1 | §III-A–C; §IV data scaling
V-Dreamer~\citep{survey_ax_2603_18811} & \LiteratureLinks{https://arxiv.org/abs/2603.18811}{} & \raisebox{-1.5pt}{\includegraphics[height=10.5pt]{assets/table-icons/solid-sim.png}}\hspace{3pt}\raisebox{-1.5pt}{\includegraphics[height=10.5pt]{assets/table-icons/solid-gen.png}} & Robot trajectories & Up to \textbf{2,500} generated trajectories for lifting \\[3pt]
\rowcolor{black!3}
% work: ax-2606.22142 | survey_ax_2606_22142
% source: https://arxiv.org/html/2606.22142v1 | §3.5; §4.4
RoboLineage~\citep{survey_ax_2606_22142} & \LiteratureLinks{https://arxiv.org/abs/2606.22142}{https://github.com/robolineage/robolineage} & \raisebox{-1.5pt}{\includegraphics[height=10.5pt]{assets/table-icons/solid-robot.png}} & Robot trajectories & \textbf{30} demos/task/round over \textbf{3} rounds \\[3pt]
% work: ax-2606.22136 | survey_ax_2606_22136
% source: https://arxiv.org/html/2606.22136v2 | §3; Appendix A
Wh0~\citep{survey_ax_2606_22136} & \LiteratureLinks{https://arxiv.org/abs/2606.22136v2}{https://github.com/chenyt31/Wh0} & \raisebox{-1.5pt}{\includegraphics[height=10.5pt]{assets/table-icons/solid-gen.png}} & Human-hand videos;\allowbreak{} Hand poses & \textbf{50,000} generated video episodes in WM-H \\[3pt]
\rowcolor{black!3}
% work: ax-2607.14047 | survey_ax_2607_14047
% source: https://arxiv.org/html/2607.14047v3 | §3; §4.2; Table 1
Zero2Skill~\citep{survey_ax_2607_14047} & \LiteratureLinks{https://arxiv.org/abs/2607.14047}{https://github.com/open-gigaai/Zero2Skill} & \raisebox{-1.5pt}{\includegraphics[height=10.5pt]{assets/table-icons/solid-robot.png}} & Robot trajectories & \textbf{50} successful demos for desktop clearing \\[3pt]
% work: ax-2609.24563 | survey_ax_2609_24563
% source: https://arxiv.org/html/2609.24563v1 | §III-B; §III-D; §IV-A
ARSTAG~\citep{survey_ax_2609_24563} & \LiteratureLinks{https://arxiv.org/abs/2609.24563}{} & \raisebox{-1.5pt}{\includegraphics[height=10.5pt]{assets/table-icons/solid-sim.png}} & Robot trajectories & \textbf{1,000} successful demos/task for sim-to-real transfer \\[3pt]
\rowcolor{black!3}
% work: ax-2609.23997 | survey_ax_2609_23997
% source: https://arxiv.org/html/2609.23997v2 | §III (tool calling); §IV, Multi-Agent Data Generation (skill calls and inter-agent messages)
RoboTalk~\citep{survey_ax_2609_23997} & \LiteratureLinks{https://arxiv.org/abs/2609.23997v2}{} & \raisebox{-1.5pt}{\includegraphics[height=10.5pt]{assets/table-icons/solid-sim.png}} & Inter-robot messages & \textbf{7,950} validated trajectories across \textbf{53} tasks \\[3pt]
% work: ax-2609.32069 | survey_ax_2609_32069
% source: https://arxiv.org/html/2609.32069v2 | §III-B; Abstract; autonomous-interaction evaluation
FIND~\citep{survey_ax_2609_32069} & \LiteratureLinks{https://arxiv.org/abs/2609.32069v2}{} & \raisebox{-1.5pt}{\includegraphics[height=10.5pt]{assets/table-icons/solid-robot.png}} & Robot trajectories & \textbf{456} episodes in six hours \\[3pt]
\rowcolor{black!3}
% work: ax-2609.38905 | survey_ax_2609_38905
% source: https://arxiv.org/html/2609.38905v3 | §3.3.4; §4, training budgets; Appendix D.1–D.2 (operator-corrected training rollouts)
EmbodiRSI~\citep{survey_ax_2609_38905} & \LiteratureLinks{https://arxiv.org/abs/2609.38905v3}{} & \raisebox{-1.5pt}{\includegraphics[height=10.5pt]{assets/table-icons/solid-robot.png}}\hspace{3pt}\raisebox{-1.5pt}{\includegraphics[height=10.5pt]{assets/table-icons/solid-sim.png}} & Robot trajectories & \textbf{400} sim + \textbf{10} real trajectories/subtask \\[3pt]

\rowcolor{black!3}
Kim et al.~\citep{survey_ax_2610_03615}
& \LiteratureLinks{https://arxiv.org/abs/2610.03615v1}{https://github.com/pearls-lab/2bridge}
& \raisebox{-1.5pt}{\includegraphics[height=10.5pt]{assets/table-icons/solid-robot.png}} & Robot trajectories & \textbf{20} autonomous and \textbf{20} human demos/task, \textbf{4} tasks \\[1pt]

Agentic-Teacher~\citep{survey_ax_2611_17394}
& \LiteratureLinks{https://arxiv.org/abs/2611.17394v1}{}
& \raisebox{-1.5pt}{\includegraphics[height=10.5pt]{assets/table-icons/solid-sim.png}} & Corrective trajectories & \textbf{139} successes per task on average over \textbf{2} RoboCasa tasks \\[1pt]

\rowcolor{black!3}
RoboMite~\citep{survey_ax_2611_37041}
& \LiteratureLinks{https://arxiv.org/abs/2611.37041v1}{}
& \raisebox{-1.5pt}{\includegraphics[height=10.5pt]{assets/table-icons/solid-sim.png}} & Multi-turn trajectories & \textbf{1,700} demos across \textbf{17} task suites \\[1pt]
\end{longtable}
\endgroup

\subsection{Annotation}
\label{sec:data-annotation}

While data collection yields videos paired with robot‑state sequences, such records are often deficient in textual descriptions, spatial markers, interaction targets, and outcome annotations. To remedy this gap, agent‑driven annotation employs intelligent agents to interpret raw records, orchestrate annotation utilities, and produce as well as validate the missing labels. We categorize existing agent‑based annotation approaches into four groups: \textit{text annotation}, \textit{spatial annotation}, \textit{interaction annotation}, and \textit{outcome annotation}, based on their respective annotation objectives.

% \paragraph{\textbf{Action Annotation.}}
% To determine the next action, a robot agent needs to recognize the current task step and whether it has finished. RoboAnnotatorX~\citep{RoboAnnotatorX} provides supervision for this capability by labeling actions and their start and end times in long robot videos. Its Vicuna-7B-based model combines scene images, short clips, and information from the full video to interpret individual actions within the task sequence. Other methods focus on key interaction events rather than complete action segments. Keypose Exploration~\citep{survey_ax_2606_29028} uses Qwen3-VL-235B to identify these events in an example trajectory, then aligns them with end-effector movements and gripper transitions to label other trajectories of the same task. LLM Trainer~\citep{survey_ax_2509_20070} uses GPT-4o to select key timestamps and describe the robot's relation to relevant objects. The recorded poses at these timestamps guide trajectory adaptation when the objects are rearranged.

% \paragraph{\textbf{Text Annotation.}}
\paragraph{\textbf{Text Annotation.}}
The most common agent-based text annotation is to produce subtask descriptions and video captions that mark task stages and observable events, respectively.
For example, NILS~\citep{survey_ax_2410_17772} generates task-level language instructions for robot video segments. Specifically, it detects segment boundaries from object movements, state changes, and gripper interactions, then uses Gemini Pro to describe the manipulation stage performed in each segment.
IVLR~\citep{survey_ax_2605_00438} generates subtask-level video captions. It divides a robot trajectory into stages, then queries Qwen3-VL to describe the detailed gripper-object interaction or spatial relation in each stage. 
%The final frame of each stage serves as the corresponding keyframe. 
EgoTools~\citep{survey_ax_2609_39378} adopts Gemini 3 Flash to caption consecutive video segments, propagating prior captions to subsequent segments for improved temporal consistency.

FLORA~\citep{report_flora_2026} marks when each action starts and ends and describes what the person or robot is trying to do. It finds action boundaries in overlapping video windows, then joins the intervals across the episode. It uses frames from neighboring events to describe each action. DenseCaptionBench scores the boundaries, the captions for reference intervals, and the two predictions together. On 100 human videos, FLORA scored 37.4\% on joint F1 and GPT-6 Astra scored 29.7\%. On the robot videos (100 with parallel grippers and 100 with dexterous hands), GPT-6 Astra led with a mean joint F1 of 57.9\% against FLORA's 51.0\%.

One action segment can also support several instructions. In RoboCasa, Kim et al.~\citep{survey_ax_2610_03615} use rules to find segment boundaries from gripper and arm motion. Most object labels come from recorded simulator state. Given the segment type and object identity, Opus~5 generates instructions in four formats for each segment. The four formats cover arm or gripper motion, spatial destinations, short commands, and everyday phrasing. The training set includes reworded versions as well.

Another form of text annotation adopts question‑answer (QA) pairs to capture scene properties, action sequences, and fine‑grained execution details. For example, RoboAnnotatorX~\citep{RoboAnnotatorX} uses GPT-4o to generate QA about the objects present, their spatial relationships, and the order of robot actions, using scene descriptions and timestamped action annotations as the source information.
Meanwhile, AnnotateAnything~\citep{survey_ax_2606_17446} creates robot‑oriented QA pairs from simulation trajectory logs. A VLM varies question phrasing for target objects, interacting parts, approach directions and execution outcomes, with answers sourced from simulator records.
REAL~\citep{survey_ax_2607_13653} and RoboTalk~\citep{survey_ax_2609_23997} use Gemini models to generate dialogue annotations alongside robot actions, covering clarification of ambiguous instructions and communication between cooperating robots.
Methods such as SpatialVLM~\citep{survey_ax_2401_12168}, SpaceVista~\citep{survey_ax_2510_09606}, and Exemplar2VQA~\citep{survey_ax_2609_37655} combine visual models or coding agents with 3D geometry to generate question-answer pairs about object positions, distances, and spatial relationships. RoboSpatial~\citep{survey_ax_2411_16537} derives spatial answers from 3D geometry, projection, and ray casting, and constructs question-answer pairs with deterministic templates.

%Here, a VLM diversifies question phrasing, whereas answers are sourced from simulator logs and sampled interaction candidates.

Beyond question‑answer pairs, reasoning‑trace annotations augment recorded behaviors with high‑level plans and explanatory rationales. 
Specifically, ECoT~\citep{survey_ax_2407_08693} and TIC-VLA~\citep{survey_ax_2602_02459} use Gemini 1.0 and GPT-5, respectively, to generate reasoning traces that explain the robot's current subtask and next action in manipulation and navigation tasks.
Along this line, ERVLA~\citep{survey_ax_2606_03784} utilizes Qwen3.5‑397B to derive shared goals and subtask sequences per episode, alongside fine‑grained reasoning for selected timesteps.

Labels can also be executable programs. AW-0~\citep{survey_ax_2610_30136} matches program fragments to the parts of a demonstration that show each subtask. One agent writes a program for each subtask, and another identifies skill calls shared across tasks. Once the library is fixed, a new segment can reuse a program if its contact and motion match. The authors evaluated the labels by training and testing a policy. On the real card-dealing task, the average rubric score was 58.3\% with full programs and 42.0\% when skill calls were removed, over three sets of 16 rollouts per configuration.

% Finally, dialogue‑driven annotation embeds communicative clarification and coordination exchanges within raw robot trajectories. In this direction, REAL~\citep{survey_ax_2607_13653} applies Gemini 3 Pro to inject clarifying questions and simulated user responses under ambiguous instructions or uncertain target objects. Meanwhile, RoboTalk~\citep{survey_ax_2609_23997} uses Gemini 3 Flash to generate interactive dialogue messages and reasoning traces co‑occurring with robot joint actions. Invalid candidate sequences covering both actions and communications get filtered out, and rejected candidates undergo iterative revision guided by validation feedback.

\paragraph{\textbf{Spatial Annotation.}}
Robotic agents perform spatial annotation by interpreting demonstrations, invoking perception tools, and verifying spatial labels.
NILS~\citep{survey_ax_2410_17772} and FSD~\citep{survey_ax_2505_08548} use language-model-guided pipelines to identify objects for annotation. NILS uses Grounding DINO~\citep{survey_ax_2303_05499} to obtain object bounding boxes and DEVA~\citep{survey_ax_2309_03903} to track object masks across video frames. FSD uses Grounded SAM~\citep{survey_ax_2401_14159} to obtain bounding boxes and segmentation masks.
AnnotateAnything~\citep{survey_ax_2606_17446} and EmbodiedGen V2~\citep{survey_ax_2607_07459} combine 3D segmentation with VLM-based interpretation to annotate object parts, with EmbodiedGen V2 additionally checking and refining the resulting labels.
SPARC~\citep{survey_ax_2606_13497} and Keypose Exploration~\citep{survey_ax_2606_29028} interpret interaction events through language models and align them with visual tracks or robot states to annotate trajectories and keyposes.
AutoFocus-IL~\citep{survey_ax_2511_18617} and Workspace Models~\citep{survey_ax_2609_20820} identify task-relevant objects and events through VLM reasoning and localize them to generate spatial saliency annotations.

\paragraph{\textbf{Interaction Annotation.}}
Interaction annotation captures physical engagements between hands, robotic grippers, tools and target objects, covering typical behaviors such as touching, grasping, pushing, and pressing.
The annotation pipeline generally consists of reasoning over interaction intent and retrieving contact‑related evidence via external tools.
For affordance annotation, several representative works, such as AFUN~\citep{survey_ax_2606_02551}, EgoAfford~\citep{survey_ax_2608_04533}, Affordance2Action~\citep{survey_ax_2606_04172}, A4-Agent~\citep{survey_ax_2512_14442}, and Affordance Agent Harness~\citep{survey_ax_2605_00663} integrate model reasoning with tool‑aided localization. In these works, agents or LLM‑driven workflows first identify object regions suitable for the desired manipulation, then call detection and segmentation modules to pinpoint these functional areas.
For dexterous hands, DextER~\citep{survey_ax_2601_16046} and OmniDexVLG~\citep{survey_ax_2512_03874} combine model reasoning with tool-derived contact evidence to annotate finger- and link-level contacts with object parts and interpret their grasp functions.

\paragraph{\textbf{Outcome Annotation.}}
Outcome annotation labels task achievement in robot executions, including intermediate progress, final success or failure, and safety violations.
Agents or model-guided workflows interpret evaluation criteria and inspect execution evidence to produce these labels.
For task achievement, RoboReward~\citep{survey_ax_2601_00675} and Zero2Skill~\citep{survey_ax_2607_14047} use visual-language judgments against explicit task criteria to assign completion scores or success labels.
For intermediate progress, ROVER~\citep{survey_ax_2508_01943} and ProcVLM~\citep{survey_ax_2605_08774} organize visual reasoning around subtasks to annotate their completion states and estimate task progress.
ARS~\citep{survey_ax_2609_34484} further uses agents to invoke video-inspection tools, verify proposed task events, and revise frame-level progress annotations.
FLORA~\citep{report_flora_2026} describes what the robot is trying to do and scores how well it does it. Its reward model labels each half-second as normal, suboptimal, or failed. In a toy-airplane assembly, both FLORA and GPT-6 Astra describe a part falling off. FLORA scores the step as a failure, while GPT-6 Astra scores it as suboptimal. On 4,861 labeled steps from 40 held-out bimanual episodes, FLORA detected 53.3\% of failure steps and labeled about 11\% of normal steps as failures.

Embodied First Exam (EFE)~\citep{project_embodied_first_exam} examines benchmark attempts after grading to identify which stages the robot reached. One labeling agent built stage graphs for 427 of the 456 core tasks. Labeling agents used these graphs to mark when each stage condition held in a replay. For failed attempts, they identified the failure stage when possible and linked each proposed cause to a video timestamp or log entry.

For safety, ShieldVLA~\citep{survey_ax_2609_13231} uses an agentic workflow to refine safety criteria from misranked examples and generate calibrated frame-level risk labels. Related works are compared in Table~\ref{tab:data-annotation}.

% These four categories add textual descriptions, spatial labels, interaction targets, and outcome labels to robot records. The following subsection examines how agents use such annotations to select data and correct unreliable records.

% <<<<<<< HEAD

% ||||||| 2854b9c
% Annotation methods are compared in Table~\ref{tab:data_annotation}.
% =======
% Annotation methods are compared in Table~\ref{tab:data-annotation}.
% >>>>>>> c58aed7a10ee2d3f89d8abc964f8e2be3197f741
% Generated from survey_data/data_comparison_index.json; source-linked fixed tags.
\par
\begingroup
\mechanismtableglue
\DeclareRobustCommand{\DataInputImage}{\tikz[x=.72pt,y=.72pt,baseline=2pt,line width=.65pt,line cap=round,line join=round,draw=black!80,fill=black!80]{\path[use as bounding box] (0,0) rectangle (16,14);\draw[rounded corners=.7pt] (1,1) rectangle (15,13);\fill (11,10) circle (1.15);\draw (2.5,3) -- (6,7.5) -- (9,4.3) -- (11,6.5) -- (13.5,3);}}
\DeclareRobustCommand{\DataInputVideo}{\tikz[x=.72pt,y=.72pt,baseline=2pt,line width=.65pt,line cap=round,line join=round,draw=black!80,fill=black!80]{\path[use as bounding box] (0,0) rectangle (16,14);\draw[rounded corners=.7pt] (1,1) rectangle (15,13);\draw (3.5,1) -- (3.5,13) (12.5,1) -- (12.5,13);\draw (1,5) -- (3.5,5) (1,9) -- (3.5,9) (12.5,5) -- (15,5) (12.5,9) -- (15,9);\fill (6,4) -- (10.5,7) -- (6,10) -- cycle;}}
\DeclareRobustCommand{\DataInputRobot}{\tikz[x=.72pt,y=.72pt,baseline=2pt,line width=.65pt,line cap=round,line join=round,draw=black!80,fill=black!80]{\path[use as bounding box] (0,0) rectangle (16,14);\draw (1,1) -- (9,1) (3,1) -- (3,3) -- (7,3) -- (7,1);\draw (5,3) -- (3.5,8) -- (8,12) -- (12,9);\draw[fill=white] (3.5,8) circle (1.3) (8,12) circle (1.3);\draw (11,7.5) -- (12,9) -- (14,8) (11,7.5) -- (12,6.5) (14,8) -- (15,7);}}
\DeclareRobustCommand{\DataInputCloud}{\tikz[x=.72pt,y=.72pt,baseline=2pt,line width=.65pt,line cap=round,line join=round,draw=black!80,fill=black!80]{\path[use as bounding box] (0,0) rectangle (16,14);\foreach \x/\y in {2/5,3/9,5/2,5/7,6/12,8/5,9/9,10/2,12/6,12/12,15/9}{\fill (\x,\y) circle (1.05);}}}
\CompactTableStyle\hyphenpenalty=10000
\setlength{\LiteratureWidth}{\dimexpr\linewidth-10\tabcolsep-6.6pt\relax}
\setlength{\LTpre}{3pt}
\setlength{\LTpost}{4pt}
\setlength{\LTcapwidth}{\linewidth}
\begin{longtable}{@{}>{\RaggedRight\arraybackslash}p{0.19\LiteratureWidth}>{\centering\arraybackslash}p{0.07\LiteratureWidth}!{\color{black!30}\vrule width .3pt\hspace{3pt}}>{\RaggedRight\arraybackslash}p{0.1\LiteratureWidth}>{\RaggedRight\arraybackslash}p{0.15\LiteratureWidth}!{\color{black!30}\vrule width .3pt\hspace{3pt}}>{\RaggedRight\arraybackslash}p{0.2\LiteratureWidth}>{\RaggedRight\arraybackslash}p{0.29\LiteratureWidth}@{}}
\caption[Comparison of works on data annotation]{Comparison of works on data annotation (Section~\ref{sec:data-annotation}). Input: \DataInputImage{} Image; \DataInputVideo{} Video; \DataInputRobot{} Robot data; \DataInputCloud{} Point cloud. Robot data includes demonstration metadata and logs. The Models column lists main semantic or reasoning backbones; the Tools column lists supporting perception models, algorithms, and simulation interfaces. Representative annotation configurations are shown. 
}
\label{tab:data-annotation}\\
\LiteratureTopRule{Data}
\rowcolor{TSTDataLight}\textbf{Work} & \textbf{Link} & \textbf{Input} & \textbf{Output} & \textbf{Models} & \textbf{Tools} \\
\midrule
\endfirsthead
\multicolumn{6}{@{}l}{\tablename~\thetable\ (continued)} \\
\LiteratureTopRule{Data}
\rowcolor{TSTDataLight}\textbf{Work} & \textbf{Link} & \textbf{Input} & \textbf{Output} & \textbf{Models} & \textbf{Tools} \\
\midrule
\endhead
\midrule
\multicolumn{6}{r@{}}{Continued on next page} \\
\endfoot
\bottomrule
\endlastfoot
\rowcolor{TSTDataLight}\multicolumn{6}{@{}l}{\textbf{Text Annotation}} \\* [3pt]
% work: ax-2401.12168 | survey_ax_2401_12168
% source: https://arxiv.org/html/2401.12168v1 | §3.1–3.2 (expert-model extraction and geometric QA)
SpatialVLM \citep{survey_ax_2401_12168} & \LiteratureLinks{https://openaccess.thecvf.com/content/CVPR2024/html/Chen_SpatialVLM_Endowing_Vision-Language_Models_with_Spatial_Reasoning_Capabilities_CVPR_2024_paper.html}{} & \DataInputImage{} & QA pairs & FlexCap & RPN;\allowbreak{} ShapeMask;\allowbreak{} ZoeDepth;\allowbreak{} Geometric QA \\[0pt]
\rowcolor{black!3}
% work: ax-2407.08693 | survey_ax_2407_08693
% source: https://arxiv.org/pdf/2407.08693v3 | §4.2 (synthetic reasoning-chain generation)
ECoT \citep{survey_ax_2407_08693} & \LiteratureLinks{https://proceedings.mlr.press/v270/zawalski25a.html}{https://github.com/MichalZawalski/embodied-CoT} & \DataInputVideo{}\hspace{4pt}\DataInputRobot{} & Reasoning traces & Prismatic-7B;\allowbreak{} Gemini 1.0 & Grounding DINO;\allowbreak{} OWLv2;\allowbreak{} SAM \\[0pt]
% work: ax-2410.17772 | survey_ax_2410_17772
% source: https://arxiv.org/html/2410.17772v2 | §2.1–2.3; Appendix A.2
NILS \citep{survey_ax_2410_17772} & \LiteratureLinks{https://proceedings.mlr.press/v270/blank25a.html}{https://github.com/intuitive-robots/NILS} & \DataInputVideo{} & Task descriptions & Gemini 1.5;\allowbreak{} Gemini Pro & Grounding DINO;\allowbreak{} DEVA;\allowbreak{} Depth Anything;\allowbreak{} GMFlow;\allowbreak{} SigLIP \\[0pt]
\rowcolor{black!3}
% work: ax-2411.16537 | survey_ax_2411_16537
% source: https://arxiv.org/html/2411.16537v5 | §3.2.1–3.2.3 (deterministic spatial extraction/templates)
RoboSpatial \citep{survey_ax_2411_16537} & \LiteratureLinks{https://openaccess.thecvf.com/content/CVPR2025/html/Song_RoboSpatial_Teaching_Spatial_Understanding_to_2D_and_3D_Vision-Language_Models_CVPR_2025_paper.html}{https://github.com/NVlabs/RoboSpatial} & \DataInputImage{} & QA pairs & {} & Geometric rules;\allowbreak{} QA templates \\[0pt]
% work: ax-2505.08548 | survey_ax_2505_08548
% source: https://arxiv.org/html/2505.08548v3 | §3.3 weak-to-strong data construction; Appendix A
FSD \citep{survey_ax_2505_08548} & \LiteratureLinks{https://openreview.net/pdf?id=yngvAamNQi}{https://github.com/pickxiguapi/Embodied-FSD} & \DataInputImage{}\hspace{4pt}\DataInputVideo{} & QA pairs & GPT-4o & Grounded SAM;\allowbreak{} Metric3Dv2;\allowbreak{} Geometric QA \\[0pt]
% work: cvf-iccv2025-roboannotatorx | RoboAnnotatorX
% source: https://openaccess.thecvf.com/content/ICCV2025/papers/Kou_RoboAnnotatorX_A_Comprehensive_and_Universal_Annotation_Framework_for_Accurate_Understanding_ICCV_2025_paper.pdf | §3.2 automatic QA generation
% Commented out at author request, 2026-10-13.
% RoboAnnotatorX \citep{RoboAnnotatorX} & \LiteratureLinks{https://openaccess.thecvf.com/content/ICCV2025/html/Kou_RoboAnnotatorX_A_Comprehensive_and_Universal_Annotation_Framework_for_Accurate_Understanding_ICCV_2025_paper.html}{https://github.com/LongXinKou/RoboannotatorX} & \DataInputRobot{} & QA pairs & GPT-4o & \textemdash \\[0pt]
\rowcolor{black!3}
% work: ax-2605.00438 | survey_ax_2605_00438
% source: https://arxiv.org/html/2605.00438v1 | §3.4; Appendix A
IVLR \citep{survey_ax_2605_00438} & \LiteratureLinks{https://arxiv.org/abs/2605.00438v1}{} & \DataInputVideo{} & Captions & Qwen3-VL & UVD \\[0pt]
% work: ax-2606.13497 | survey_ax_2606_13497
% source: https://arxiv.org/html/2606.13497v1 | §3.1–3.3 (fixed detector/tracker/scoring stages)
SPARC \citep{survey_ax_2606_13497} & \LiteratureLinks{https://arxiv.org/abs/2606.13497}{https://github.com/intuitive-robots/sparc} & \DataInputVideo{}\hspace{4pt}\DataInputRobot{} & Task descriptions & Qwen3.6-30B & Gripper-event parsing \\[0pt]
\rowcolor{black!3}
% work: ax-2606.17446 | survey_ax_2606_17446
% source: https://arxiv.org/html/2606.17446v2 | §3.1; Appendix G.3–G.4 (QA from simulation logs and scene annotations)
% component evidence: Section 3.1 names Qwen3.5-VL for semantic annotation; G.3 does not identify the VLM used to rewrite QA questions.
AnnotateAnything \citep{survey_ax_2606_17446} & \LiteratureLinks{https://arxiv.org/abs/2606.17446v2}{} & \DataInputImage{}\hspace{4pt}\DataInputRobot{}\hspace{4pt}\DataInputCloud{} & QA pairs & NR & Simulation-log queries \\[0pt]
% work: ax-2607.13653 | survey_ax_2607_13653
% source: https://arxiv.org/html/2607.13653v2 | Appendix B.4 (social interaction annotation); B.5 (expert trajectory annotation)
% component evidence: Appendix B.5 names Gemini-3-Pro for reasoning traces; B.4 does not identify the dialogue-insertion LLM. Simulator trajectory replay supplies aligned observations.
REAL \citep{survey_ax_2607_13653} & \LiteratureLinks{https://link.springer.com/chapter/10.1007/978-3-032-37422-6_36}{https://github.com/InternRobotics/REAL} & \DataInputImage{}\hspace{4pt}\DataInputRobot{} & Dialogue;\allowbreak{} Reasoning traces & Gemini 3 Pro & Trajectory replay \\[0pt]
\rowcolor{black!3}
% work: ax-2609.37655 | survey_ax_2609_37655
% source: https://arxiv.org/html/2609.37655v1 | §3.1–3.3 (generated QA programs; geometric answers)
Exemplar2VQA \citep{survey_ax_2609_37655} & \LiteratureLinks{https://arxiv.org/abs/2609.37655}{https://github.com/yingjiayu12/Exemplar2VQA} & \DataInputRobot{} & QA pairs & Qwen3-Coder-30B-A3B-Instruct & Geometry APIs;\allowbreak{} AI2-THOR \\[0pt]
% work: ax-2609.39378 | survey_ax_2609_39378
% source: https://arxiv.org/html/2609.39378v2 | §3.2 dense-caption pipeline
EgoTools \citep{survey_ax_2609_39378} & \LiteratureLinks{https://arxiv.org/abs/2609.39378v2}{https://github.com/Ropedia/EgoTools} & \DataInputVideo{} & Captions & Gemini 3 Flash & {} \\[0pt]
% Source: 2610.30136v1, Section 2.3, Figure 3 and Appendix K.
\rowcolor{black!3}
AW-0 \citep{survey_ax_2610_30136} & \LiteratureLinks{https://arxiv.org/abs/2610.30136v1}{https://github.com/ZhuoyangLiu2005/agentic-wam} & \DataInputVideo{}\hspace{4pt}\DataInputRobot{} & Programs aligned with video segments & Gemini 3.1 Pro; Claude Opus 4.8 & Skill library and program cache \\[0pt]
% Source: 2610.31779v1, Sections 4.1 and 5.
FLORA \citep{report_flora_2026} & \LiteratureLinks{https://arxiv.org/abs/2610.31779v1}{https://github.com/NVlabs/Flora} & \DataInputVideo{} & Action intervals and captions & Qwen3.5-9B and 27B & Merge action intervals across windows \\[0pt]

\rowcolor{black!3}
Kim et al.~\citep{survey_ax_2610_03615}
& \LiteratureLinks{https://arxiv.org/abs/2610.03615v1}{https://github.com/pearls-lab/2bridge}
& \DataInputRobot{} & Action instructions & Opus 5 & Motion rules; simulator object labels \\[1pt]
\rowcolor{TSTDataLight}\multicolumn{6}{@{}l}{\textbf{Spatial Annotation}} \\* [3pt]
% work: ax-2407.08693 | survey_ax_2407_08693
% source: https://arxiv.org/pdf/2407.08693v3 | §4.2 (synthetic reasoning-chain generation)
ECoT \citep{survey_ax_2407_08693} & \LiteratureLinks{https://proceedings.mlr.press/v270/zawalski25a.html}{https://github.com/MichalZawalski/embodied-CoT} & \DataInputVideo{}\hspace{4pt}\DataInputRobot{} & Bounding boxes;\allowbreak{} Gripper positions & Prismatic-7B & Grounding DINO;\allowbreak{} OWLv2;\allowbreak{} SAM;\allowbreak{} RANSAC \\[0pt]
\rowcolor{black!3}
% work: ax-2410.17772 | survey_ax_2410_17772
% source: https://arxiv.org/html/2410.17772v2 | §2.1–2.3; Appendix A.2
NILS \citep{survey_ax_2410_17772} & \LiteratureLinks{https://proceedings.mlr.press/v270/blank25a.html}{https://github.com/intuitive-robots/NILS} & \DataInputVideo{} & Bounding boxes;\allowbreak{} Object masks & Gemini 1.5 & Grounding DINO;\allowbreak{} OWLv2;\allowbreak{} DEVA \\[0pt]
% work: ax-2505.08548 | survey_ax_2505_08548
% source: https://arxiv.org/html/2505.08548v3 | §3.3 weak-to-strong data construction; Appendix A
FSD \citep{survey_ax_2505_08548} & \LiteratureLinks{https://openreview.net/pdf?id=yngvAamNQi}{https://github.com/pickxiguapi/Embodied-FSD} & \DataInputImage{}\hspace{4pt}\DataInputVideo{} & Bounding boxes;\allowbreak{} Object masks & GPT-4o & Grounded SAM \\[0pt]
\rowcolor{black!3}
% work: ax-2606.13497 | survey_ax_2606_13497
% source: https://arxiv.org/html/2606.13497v1 | §3.1–3.3 (fixed detector/tracker/scoring stages)
SPARC \citep{survey_ax_2606_13497} & \LiteratureLinks{https://arxiv.org/abs/2606.13497}{https://github.com/intuitive-robots/sparc} & \DataInputVideo{}\hspace{4pt}\DataInputRobot{} & 3D object trajectories & Qwen3.6-30B & LLMDet;\allowbreak{} SAM2.1 Hiera-Small;\allowbreak{} RobotSeg;\allowbreak{} AllTracker;\allowbreak{} MoGe-2 \\[0pt]
% work: ax-2606.17446 | survey_ax_2606_17446
% source: https://arxiv.org/html/2606.17446v2 | §3.1; Appendix G.3–G.4 (QA from simulation logs and scene annotations)
% component evidence: Section 3.1 names Qwen3.5-VL for semantic annotation; G.3 does not identify the VLM used to rewrite QA questions.
AnnotateAnything \citep{survey_ax_2606_17446} & \LiteratureLinks{https://arxiv.org/abs/2606.17446v2}{} & \DataInputImage{}\hspace{4pt}\DataInputRobot{}\hspace{4pt}\DataInputCloud{} & 3D part masks;\allowbreak{} Semantic labels & Qwen3.5-VL & P3-SAM;\allowbreak{} X-Part \\[0pt]
\rowcolor{black!3}
% work: ax-2607.07459 | survey_ax_2607_07459
% source: https://arxiv.org/html/2607.07459v3 | §2.3 affordance autolabeling
EmbodiedGen V2 \citep{survey_ax_2607_07459} & \LiteratureLinks{https://arxiv.org/abs/2607.07459}{https://github.com/HorizonRobotics/EmbodiedGen} & \DataInputImage{} & 3D part masks;\allowbreak{} Semantic labels & GPT-5.4 & P3-SAM \\[0pt]
\rowcolor{TSTDataLight}\multicolumn{6}{@{}l}{\textbf{Interaction Annotation}} \\* [3pt]
% work: ax-2512.14442 | survey_ax_2512_14442
% source: https://arxiv.org/html/2512.14442v1 | §4.3–4.5; §5 implementation details (GPT-4o, Rex-Omni, SAM2-Large)
A4-Agent \citep{survey_ax_2512_14442} & \LiteratureLinks{https://link.springer.com/book/10.1007/978-3-032-37086-0}{https://github.com/EnVision-Research/A4-Agent} & \DataInputImage{} & Affordance masks & GPT-4o & Qwen-Image-Editing;\allowbreak{} Rex-Omni;\allowbreak{} SAM2-Large \\*[0pt]
\rowcolor{black!3}
% work: ax-2601.16046 | survey_ax_2601_16046
% source: https://arxiv.org/html/2601.16046v2 | §3.4 (physics-buffer contacts; VLM grasp descriptions)
% component evidence: Appendix A.3 names Gemma-3 for grasp-instruction annotation; Section 3.4 specifies MuJoCo contacts.
DextER \citep{survey_ax_2601_16046} & \LiteratureLinks{https://openaccess.thecvf.com/content/CVPR2026/html/Lee_DextER_Language-driven_Dexterous_Grasp_Generation_with_Embodied_Reasoning_CVPR_2026_paper.html}{https://github.com/junha-l/dexter} & \DataInputImage{}\hspace{4pt}\DataInputRobot{} & Contact points;\allowbreak{} Grasp descriptions & Gemma 3 & MuJoCo contact queries \\*[0pt]
% work: ax-2605.00663 | survey_ax_2605_00663
% source: https://arxiv.org/html/2605.00663v2 | §3.1 (binary or soft affordance masks); §3.2–3.3 (routing and verification)
Affordance Agent Harness \citep{survey_ax_2605_00663} & \LiteratureLinks{https://arxiv.org/abs/2605.00663v2}{https://github.com/tenplusgood/A-Harness} & \DataInputImage{} & Affordance masks & GPT-4o & Qwen3-VL-235B-A22B;\allowbreak{} SAM2-Large;\allowbreak{} Retrieval;\allowbreak{} Interaction imagination \\*[0pt]
\rowcolor{black!3}
% work: ax-2606.04172 | survey_ax_2606_04172
% source: https://arxiv.org/html/2606.04172v2 | §3.1 A2A-AffordGen (SAG/MAG)
% component evidence: Section 3.1: Qwen3.5-9B LoRA click policy with SAM3 and Grounding DINO. Qwen3-VL-32B generates task instructions, outside the masks-only row.
Affordance2Action \citep{survey_ax_2606_04172} & \LiteratureLinks{https://arxiv.org/abs/2606.04172v2}{https://github.com/arc-l/a2a} & \DataInputImage{} & Affordance masks & Qwen3.5-9B & SAM3;\allowbreak{} Grounding DINO \\[0pt]
\rowcolor{TSTDataLight}\multicolumn{6}{@{}l}{\textbf{Outcome Annotation}} \\* [3pt]
% work: ax-2508.01943 | survey_ax_2508_01943
% source: https://arxiv.org/html/2508.01943v2 | §3 and §4 (recursive frame/subtask reasoning)
ROVER \citep{survey_ax_2508_01943} & \LiteratureLinks{https://proceedings.nips.cc/paper_files/paper/2025/hash/cdb1f1b572b616042f01b57fd13db878-Abstract-Conference.html}{https://github.com/Philip-MIT/rover-vlm} & \DataInputVideo{} & Progress scores & Gemini 1.5 Pro & {} \\*[0pt]
\rowcolor{black!3}
% work: ax-2607.14047 | survey_ax_2607_14047
% source: https://arxiv.org/html/2607.14047v3 | Appendix B.1 (collection/reset verification)
Zero2Skill \citep{survey_ax_2607_14047} & \LiteratureLinks{https://arxiv.org/abs/2607.14047}{https://github.com/open-gigaai/Zero2Skill} & \DataInputImage{} & Success labels & Seed1.8 & {} \\*[0pt]
% work: ax-2609.34484 | survey_ax_2609_34484
% source: https://arxiv.org/html/2609.34484v1 | §3.2–3.5 (event verification and progress editing)
ARS \citep{survey_ax_2609_34484} & \LiteratureLinks{https://arxiv.org/abs/2609.34484v1}{} & \DataInputVideo{} & Progress scores & Qwen3.6-27B & Video inspection;\allowbreak{} Timeline editing;\allowbreak{} Progress editing \\[0pt]
% Source: 2610.31779v1, Sections 4.2 and 5.
\rowcolor{black!3}
FLORA \citep{report_flora_2026} & \LiteratureLinks{https://arxiv.org/abs/2610.31779v1}{https://github.com/NVlabs/Flora} & \DataInputVideo{} & Step quality, episode and rubric scores & Qwen3.5-27B & {} \\[0pt]
\end{longtable}
\endgroup

\subsection{Refinement}
\label{sec:refinement}

Data refinement improves the suitability of recorded robot demonstrations through data selection and supervision revision. Agent-based systems combine model reasoning with execution evidence to determine which trajectories or segments to retain and whether earlier annotations need correction.

RoboLineage~\citep{survey_ax_2606_22142} and Zero2Skill~\citep{survey_ax_2607_14047} use model-assisted review to determine whether recorded trajectories meet the criteria for training. Selection can also operate within a trajectory: HELP~\citep{survey_ax_2607_09776} and ARS~\citep{survey_ax_2609_34484} use estimated task progress to retain useful segments while excluding unproductive actions, allowing parts of unsuccessful attempts to support learning. HELP relies on its VLAC-CUT critic to distinguish progress and recovery from idle or failure-inducing behavior, while ARS uses agents to call video-inspection tools and verify task events before progress estimates guide segment selection.

Robo-COP~\citep{survey_ax_2610_09228} can keep a successful pickup from an episode that later fails. A text judge reads the code, console output, and outcome to label the skill. A video judge checks sampled frames for success or clear progress before the segment enters training. Agentic-Teacher~\citep{survey_ax_2611_17394} instead filters trajectories after joining a successful repair to the original prefix. Motion thresholds remove idle frames, and the frozen base VLA screens out trajectories with unusually high flow-matching loss.

D\&A~\citep{survey_ax_2611_03622} trains a bimanual policy to retrieve a book using selected parts of demonstrations. All 150 demonstrations reached the goal, but 100 included detours such as pulling at another book. D\&A estimated progress from the videos, ranked action chunks across demonstrations by their estimated advantage, and trained the policy on the top 35\%. Successful runs took 18.8 seconds on average, compared with 28.5 seconds for the policy trained on the full dataset.

Later feedback can change how an earlier action is judged. Zero2Skill~\citep{survey_ax_2607_14047} applies human-corrected success criteria to previously collected demonstrations and updates their labels. Reflective Test-Time Planning~\citep{survey_ax_2602_21198} uses later execution outcomes to revise earlier action evaluations, retaining the updated reflections and scores as supervision for subsequent learning. Related works are compared in Table~\ref{tab:data-refinement}.

% Generated from survey_data/data_comparison_index.json; source-linked fixed tags.
\par\Needspace{10\baselineskip}
\begingroup
\mechanismtableglue
\CompactTableStyle\hyphenpenalty=10000
\setlength{\LiteratureWidth}{\dimexpr\linewidth-6\tabcolsep-6.6pt\relax}
\setlength{\LTpre}{3pt}
\setlength{\LTpost}{4pt}
\setlength{\LTcapwidth}{\linewidth}
\begin{longtable}{@{}>{\RaggedRight\arraybackslash}p{0.325\LiteratureWidth}>{\centering\arraybackslash}p{0.075\LiteratureWidth}!{\color{black!30}\vrule width .3pt\hspace{3pt}}>{\RaggedRight\arraybackslash}p{0.3\LiteratureWidth}!{\color{black!30}\vrule width .3pt\hspace{3pt}}>{\RaggedRight\arraybackslash}p{0.3\LiteratureWidth}@{}}
\caption[Comparison of works on data refinement]{Comparison of works on data refinement (Section~\ref{sec:refinement}). Model lists backbones used to review data or revise supervision.}\label{tab:data-refinement}\\
\LiteratureTopRule{Data}
\rowcolor{TSTDataLight}\textbf{Work} & \textbf{Link} & \textbf{Refinement Target} & \textbf{Model} \\
\midrule
\endfirsthead
\multicolumn{4}{@{}l}{\tablename~\thetable\ (continued)} \\
\LiteratureTopRule{Data}
\rowcolor{TSTDataLight}\textbf{Work} & \textbf{Link} & \textbf{Refinement Target} & \textbf{Model} \\
\midrule
\endhead
\midrule
\multicolumn{4}{r@{}}{Continued on next page} \\
\endfoot
\bottomrule
\endlastfoot
% work: ax-2602.21198 | survey_ax_2602_21198
% source: https://arxiv.org/html/2602.21198v3 | §3.2.1–3.2.2 (hindsight-corrected scores/reflections)
% model source: Appendices P.7 and Q.2
% model evidence note: Models follow Appendix P.7 (Household: LLaVA-3D-7B) and Q.2 (Cupboard Fitting: Qwen2.5-VL-3B). Appendix K describes a 3B Household model, conflicting with P.7; the table follows the explicit architecture specification.
Reflective Test-Time Planning~\citep{survey_ax_2602_21198} & \LiteratureLinks{https://arxiv.org/abs/2602.21198}{https://github.com/Reflective-Test-Time-Planning/Reflective-Test-Time-Planning} & Action scores and reflections & \mbox{LLaVA-3D-7B} /\allowbreak{} \mbox{Qwen2.5-VL-3B} \\[0pt]
\rowcolor{black!3}
% work: ax-2606.22142 | survey_ax_2606_22142
% source: https://arxiv.org/html/2606.22142v1 | §3.4–3.5; Appendix admission routing
% model source: §4.1; Appendix I.2; Appendix K, Model backend
RoboLineage~\citep{survey_ax_2606_22142} & \LiteratureLinks{https://arxiv.org/abs/2606.22142}{https://github.com/robolineage/robolineage} & Robot trajectories & \mbox{Claude Sonnet 4.6} \\[0pt]
% work: ax-2607.14047 | survey_ax_2607_14047
% source: https://arxiv.org/html/2607.14047v3 | Appendix B.2 (retroactive re-evaluation)
% model source: §4.1; Appendix A, Offline re-judging; Appendix B.1–B.2
Zero2Skill~\citep{survey_ax_2607_14047} & \LiteratureLinks{https://arxiv.org/abs/2607.14047}{https://github.com/open-gigaai/Zero2Skill} & Robot trajectories;\allowbreak{} Outcome labels & \mbox{Seed1.8} \\[0pt]
\rowcolor{black!3}
% work: ax-2609.34484 | survey_ax_2609_34484
% source: https://arxiv.org/html/2609.34484v1 | §3.2–3.6 (agent inspection, verification, and external progress auditing); §3.7 (threshold-based chunk selection)
% model source: §4 experimental settings; Appendix F.2
ARS~\citep{survey_ax_2609_34484} & \LiteratureLinks{https://arxiv.org/abs/2609.34484v1}{} & Action segments;\allowbreak{} Progress annotations & \mbox{Qwen3.6-27B} \\[0pt]
% Source: 2607.09776v2, Sections 3.3.1 and 4.1; Appendix A.4.
HELP~\citep{survey_ax_2607_09776} & \LiteratureLinks{https://arxiv.org/abs/2607.09776v2}{https://github.com/Sage-PhysicalAI/VLAC-cut} & Progress and recovery segments & VLAC-CUT (Qwen3-VL-30B-A3B) \\[0pt]

\rowcolor{black!3}
Robo-COP~\citep{survey_ax_2610_09228}
& \LiteratureLinks{https://arxiv.org/abs/2610.09228v1}{}
& Skill segments & Text and video judges; backbones unspecified \\[1pt]
\end{longtable}
\endgroup

\section{Environment: What Worlds, Tasks, and Tests Should the Agent Build?}
\label{sec:environment}

\begin{figure}[!t]
\centering
\includegraphics[width=\linewidth]{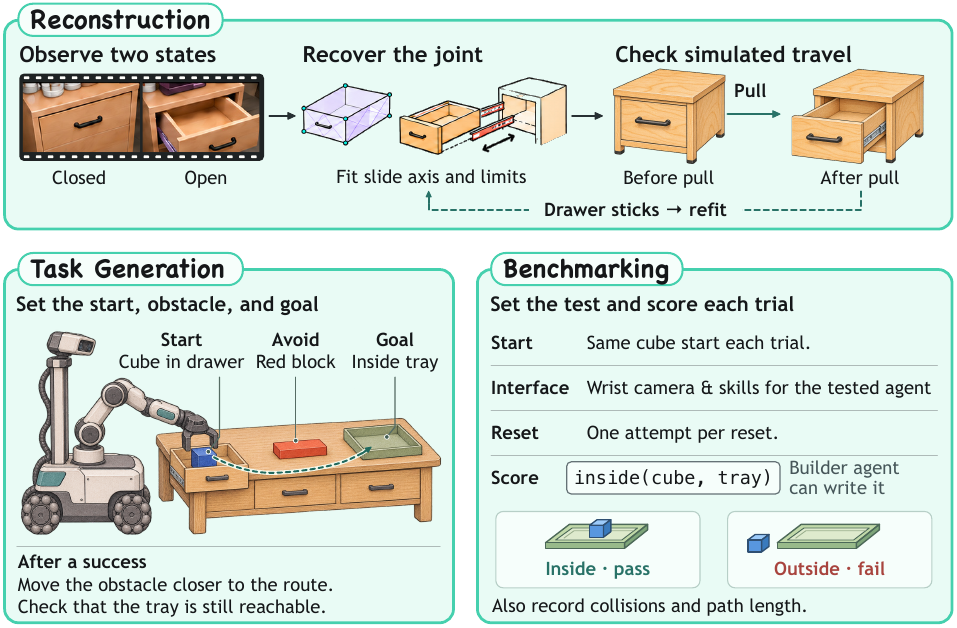}
\caption{A drawer task needs a working scene, a reachable goal, and a rule for success. Reconstruction tests the simulated drawer with a pull. Task Generation moves the red obstacle toward the path and checks that the tray is still reachable. Benchmarking allows one attempt per reset and scores whether the cube ends in the tray.}
\label{fig:environment-objects}
\end{figure}

The Environment branch covers the world, task, and evaluator that shape robot interaction. Figure~\ref{fig:environment-objects} illustrates why each needs its own check. If the simulated drawer sticks, Reconstruction must revisit its joint model. If moving the obstacle makes the tray unreachable, Task Generation has produced an infeasible task. Benchmarking fixes the conditions under which the robot is tested and the evidence used to score it. Under the illustrated protocol, a cube that finishes outside the tray counts as a failed attempt.

\subsection{Reconstruction}
\label{sec:environment-reconstruction}

Reconstruction turns observations, demonstrations, or a requested layout into an executable scene. It must recover the structure that a task uses, the changes caused by interaction, and an interface through which an agent can inspect and revise the result. Checks should expose errors that matter for execution rather than reward visual similarity alone.

\paragraph{\textbf{Static Structure.}}
Static structure includes objects, poses, supports, and the properties that make a scene stable. ARSTAG~\citep{survey_ax_2609_24563} reconstructs instruction-relevant objects and support relations, links identities to segmentation masks, and repairs poses before collecting demonstrations. SimFoundry~\citep{survey_ax_2606_28276} decomposes video into meshes and articulated parts, stabilizes the scene, and creates digital-cousin variants; Genie Sim 3.0~\citep{survey_ax_2601_02078} assembles scenes from an asset library. RoboLab~\citep{survey_ax_2604_09860} returns displacement errors from stability simulation to the generator, while RoboPlayground~\citep{survey_ax_2604_05226} settles a goal configuration and repairs task code when its success predicate fails. GIF~\citep{survey_ax_2609_05927} checks the requested relation after settling objects before generating demonstrations. These checks keep reconstruction tied to support, contact, and task feasibility.

The assembly environments in GR00T~2~\citep{nvidia2026gr00t2} illustrate why one object can need two collision models. For grasping the toy tractor body, the simulation needs only its overall shape. Insertion and threading require a more detailed model of the hole and its threads. Human experts reviewed the interactions, and the authors replayed real robot demonstrations in simulation to test grasping, insertion, and fastening.

\paragraph{\textbf{Dynamic and Articulated Structure.}}
A replayable scene must also respond correctly to new actions. Agentic Real2Sim~\citep{survey_ax_2607_19190} identifies task-relevant objects, reconstructs their meshes, and tracks 6-DoF poses, with optional VLM refinement. Video2World~\citep{survey_ax_2610_04432} evaluates coding agents that recover a scene and a control program from embodied video, while 4DCodeBench~\citep{survey_ax_2610_03715} tests code-based dynamic reconstruction with analytic motion, keyframes, or simulation. Matching observed motion does not establish correct response to a new action, so these are separate checks.

Real2Worlds2Real~\citep{survey_ax_2611_31098} builds a digital twin from RGB images or video and can vary the objects and layout while preserving the interactions a task requires. When a recording shows an object sliding, the agent tests friction values in simulation and compares the resulting motion with the recording. To test gear insertion on a real robot, the authors used Real2Worlds2Real, PolaRiS, and SimFoundry to build three twins from the same videos. They started with a policy trained on 60 real demonstrations and refined it in each twin using the same procedure and data budget. In 30 trials per method, gear insertion succeeded 28 times with Real2Worlds2Real, 18 with PolaRiS, and 12 with SimFoundry. The starting policy alone succeeded 13 times, and further training on the real robot gave 26 successes. The authors attribute the poor transfer from SimFoundry to missing gear teeth and incorrect shaft spacing in its reconstruction.

Gym-Anything for Robotics~\citep{survey_ax_2610_40565} gives a coding agent a physical activity to simulate. For wine pouring, the builder models how air enters the tilted bottle and checks its emptying rate against measurements. Other agents test the simulation, including whether legal robot motions can break it. The builder uses their reports to revise the scene and its physics.

Articulated assets require kinematic tests. Articulate-Anything~\citep{survey_ax_2410_13882} renders generated link and joint code, compares it with the input, and exports a URDF after revision. Articraft~\citep{survey_ax_2605_15187} specifies parts, joints, and tests in code, using compilation and geometric diagnostics to guide repair. Deformable assets need different representations: CoDimRecon~\citep{survey_ax_2609_36024} tests curves, surfaces, and volumes through simulation; DiagGen~\citep{survey_ax_2609_23103} routes diagnostic failures to segmentation, material, or mesh repair; and DexAgent~\citep{survey_ax_2609_35318} selects reconstruction skills and verifiers for rigid, articulated, or deformable objects before generating trajectories and training data. These procedures establish simulated behavior and task feasibility, not real-world validity.

\paragraph{\textbf{Construction Interfaces and Revision.}}
The interface determines which scene changes an agent can make. SceneSmith~\citep{survey_ax_2602_09153} edits structured scene objects through designer, critic, and orchestrator roles with rollback. SAGE~\citep{survey_ax_2602_10116} exposes placement, motion, and deletion operations and rejects or repairs unstable arrangements. EmbodiedGen V2~\citep{survey_ax_2607_07459} provides typed scene-graph operations over a persistent world state. Code interfaces allow more expressive changes: ChronoAgentic~\citep{survey_ax_2605_14398} generates and repairs PyChrono code; SimWorld Studio's SimCoder~\citep{survey_ax_2605_09423} uses compilation, collision, physics, and visual feedback; LEGO-Anything~\citep{survey_ax_2609_36380} rebuilds an image in Blender through repeated code execution; and Code4Scene~\citep{survey_ax_2609_36777} checks Unreal scene edits for support, penetration, and navigation utility. A constrained API narrows the error space, whereas generated code needs software, geometric, and physical checks.

LiteReality-Agent~\citep{survey_ax_2610_01863} edits a room reconstructed from an RGB-D scan. A coding agent compares rendered views with photographs from the same viewpoints. It then checks whether objects intersect and whether the furniture has support. Across ten rooms, mean absolute depth error over 611 views was 10.50\,cm, compared with 15.03\,cm for LiteReality. The authors separately demonstrate shaking and grasping in a MuJoCo kitchen.

RoboUniverse-10K~\citep{survey_ax_2611_31002} checks whether a reconstructed scene can support a particular task. From a real demonstration, a coding agent builds an Isaac Sim scene, a reference controller, and a verifier. A second agent compares the simulated interaction with the video, then resets the scene and runs the controller again. It also checks whether the verifier rejects failed attempts and accepts other valid solutions. Failed checks send the environment back for revision. Of the 10,807 environments, 10,307 are training tasks and 500 are held out for evaluation.

Revision should match the intended downstream use. RAPID~\citep{survey_ax_2609_30249} screens perturbed simulator variants with agent-written feasibility filters and keeps the resulting primitive and strategy code fixed at deployment. Awomo-SimDataEngine~\citep{survey_ax_2610_02274} builds worlds in stages with acceptance checks and dependency-aware repair. DexSim2Real~\citep{survey_ax_2605_05241} uses a VLM realism critic and an LLM-derived skill curriculum to tune simulation randomization for sim-to-real transfer. Appearance and stability checks cannot replace contact, friction, reachability, or downstream execution tests. Related works are compared in Table~\ref{tab:environment-reconstruction}.

\par
\begingroup
\mechanismtableglue
\CompactTableStyle
\newcommand{\ComparisonInputSlot}[1]{\makebox[12pt][c]{\raisebox{\dimexpr\depth-1.5pt\relax}{\resizebox*{!}{9pt}{\ComparisonGlyph{#1}}}}}
\setlength{\LiteratureWidth}{\dimexpr\linewidth-10\tabcolsep\relax}

\setlength{\LTpre}{4pt}
\setlength{\LTpost}{4pt}
\setlength{\LTcapwidth}{\linewidth}
\captionsetup{singlelinecheck=false,justification=raggedright}
\LiteratureContinuationSetup
\begin{longtable}{@{}>{\RaggedRight\arraybackslash}p{0.218\LiteratureWidth}>{\centering\arraybackslash}p{0.047\LiteratureWidth}>{\RaggedRight\arraybackslash}p{0.092\LiteratureWidth}>{\RaggedRight\arraybackslash}p{0.168\LiteratureWidth}>{\RaggedRight\arraybackslash}p{0.231\LiteratureWidth}>{\RaggedRight\arraybackslash}p{0.244\LiteratureWidth}@{}}
\caption[Comparison of works on reconstruction]{Comparison of works on reconstruction (Section~\ref{sec:environment-reconstruction}). Selected inputs and representations may co-occur. Scene includes graphs and layouts; Params.: parameters. Engines lists construction or verification software, excluding export-only targets. Input: \mbox{\ComparisonGlyph{language}\hspace{3pt}Text};\allowbreak\enspace \mbox{\ComparisonGlyph{image}\hspace{3pt}Image};\allowbreak\enspace \mbox{\ComparisonGlyph{video}\hspace{3pt}Video};\allowbreak\enspace \mbox{\ComparisonGlyph{geometry}\hspace{3pt}3D assets}.}\label{tab:environment-reconstruction}\\
\LiteratureTopRule{Environment}
\rowcolor{TSTEnvironmentLight}\textbf{\mbox{Work}} & \textbf{\mbox{Link}} & \textbf{\mbox{Input}} & \textbf{\mbox{Representation}} & \textbf{\mbox{Agent} \mbox{Operations}} & \textbf{\mbox{Engines}} \\
\midrule
\endfirsthead
\multicolumn{6}{@{}l}{\tablename~\thetable\ (continued)} \\
\LiteratureTopRule{Environment}
\rowcolor{TSTEnvironmentLight}\textbf{\mbox{Work}} & \textbf{\mbox{Link}} & \textbf{\mbox{Input}} & \textbf{\mbox{Representation}} & \textbf{\mbox{Agent} \mbox{Operations}} & \textbf{\mbox{Engines}} \\
\midrule
\endhead
\midrule
\multicolumn{6}{r@{}}{Continued on next page} \\
\endfoot
\bottomrule
\endlastfoot
\addlinespace[3pt]
\rowcolor{TSTEnvironmentLightest}\multicolumn{6}{@{}l}{\textbf{Assets}} \\*[3pt]
% Source: https://arxiv.org/html/2410.13882v4 | Sections 3, 4.1-4.4, 5 (Implementation Details)
Articulate-Anything \mbox{\citep{survey_ax_2410_13882}} & \mbox{\LiteraturePaperLink{https://proceedings.iclr.cc/paper_files/paper/2025/hash/2c8047bf3ed8ef6905351608d641f02f-Abstract-Conference.html}\hspace{5pt}\LiteratureCode{https://github.com/vlongle/articulate-anything}} & \mbox{\ComparisonInputSlot{video}\hspace{2pt}\ComparisonInputSlot{geometry}} & \mbox{URDF}; \mbox{Code} & \mbox{Retrieve}; \mbox{Fit}; \mbox{Edit} & \mbox{PyBullet}; \mbox{SAPIEN} \\[0pt]
% Source: https://arxiv.org/html/2605.15187v1 | Sections 3.1-3.2
\rowcolor{black!3}
Articraft \mbox{\citep{survey_ax_2605_15187}} & \mbox{\LiteraturePaperLink{https://arxiv.org/abs/2605.15187v1}\hspace{5pt}\LiteratureCode{https://github.com/mattzh72/articraft}} & \mbox{\ComparisonInputSlot{language}} & \mbox{URDF}; \mbox{Code} & \mbox{Generate}; \mbox{Edit} & \mbox{CadQuery} \\[0pt]
% Source: https://arxiv.org/html/2609.23103v1 | Sections III-A-III-D, IV
DiagGen \mbox{\citep{survey_ax_2609_23103}} & \mbox{\LiteraturePaperLink{https://arxiv.org/abs/2609.23103}\hspace{5pt}\LiteratureCode{https://github.com/diaggen/diaggen}} & \mbox{\ComparisonInputSlot{image}} & \mbox{Mesh}; \mbox{Params.} & \mbox{Generate}; \mbox{Edit} & \mbox{Genesis} \\[0pt]
\addlinespace[3pt]
\rowcolor{TSTEnvironmentLightest}\multicolumn{6}{@{}l}{\textbf{Scenes}} \\*[3pt]
% Source: https://arxiv.org/html/2502.09886v1 | Sections 3.1-3.3, 4 (Experimental Setup); Appendix A.2
Video2Policy \mbox{\citep{survey_ax_2502_09886}} & \mbox{\LiteraturePaperLink{https://arxiv.org/abs/2502.09886}\hspace{5pt}{\setbox0=\hbox{\faGithub}\makebox[\wd0][c]{-}}} & \mbox{\ComparisonInputSlot{video}} & \mbox{Mesh}; \mbox{URDF} & \mbox{Generate}; \mbox{Fit}; \mbox{Assemble} & \mbox{Isaac Gym} \\[0pt]
% Source: https://arxiv.org/html/2601.02078v4 | Section III-A (Scene Generation)
\rowcolor{black!3}
Genie Sim 3.0 \mbox{\citep{survey_ax_2601_02078}} & \mbox{\LiteraturePaperLink{https://arxiv.org/abs/2601.02078}\hspace{5pt}\LiteratureCode{https://github.com/AgibotTech/genie_sim}} & \mbox{\ComparisonInputSlot{language}} & \mbox{Scene}; \mbox{Code} & \mbox{Retrieve}; \mbox{Generate}; \mbox{Assemble} & \mbox{Isaac Sim}; \mbox{OpenUSD} \\[0pt]
% Source: https://arxiv.org/html/2602.09153v2 | Sections 3.1-3.4, 4.1; Appendix J
SceneSmith \mbox{\citep{survey_ax_2602_09153}} & \mbox{\LiteraturePaperLink{https://proceedings.mlr.press/v306/pfaff26a.html}\hspace{5pt}\LiteratureCode{https://github.com/nepfaff/scenesmith}} & \mbox{\ComparisonInputSlot{language}} & \mbox{Mesh}; \mbox{Scene} & \mbox{Generate}; \mbox{Assemble}; \mbox{Edit} & \mbox{Drake} \\[0pt]
% Source: https://arxiv.org/html/2602.10116v2 | Sections 3.1.1-3.1.2, 4.1
\rowcolor{black!3}
SAGE \mbox{\citep{survey_ax_2602_10116}} & \mbox{\LiteraturePaperLink{https://openaccess.thecvf.com/content/CVPR2026/html/Xia_SAGE_Scalable_Agentic_3D_Scene_Generation_for_Embodied_AI_CVPR_2026_paper.html}\hspace{5pt}\LiteratureCode{https://github.com/NVlabs/sage}} & \mbox{\ComparisonInputSlot{language}} & \mbox{Mesh}; \mbox{Scene} & \mbox{Generate}; \mbox{Assemble}; \mbox{Edit} & \mbox{Isaac Sim} \\[0pt]
% Source: https://arxiv.org/html/2604.05226v1 | Section III; Appendix A.1-A.6
RoboPlayground \mbox{\citep{survey_ax_2604_05226}} & \mbox{\LiteraturePaperLink{https://arxiv.org/abs/2604.05226}\hspace{5pt}{\setbox0=\hbox{\faGithub}\makebox[\wd0][c]{-}}} & \mbox{\ComparisonInputSlot{language}} & \mbox{Code} & \mbox{Generate}; \mbox{Edit} & \mbox{MuJoCo} \\[0pt]
% Source: https://arxiv.org/html/2604.09860v4 | Section III-A1; Appendix C
\rowcolor{black!3}
RoboLab \mbox{\citep{survey_ax_2604_09860}} & \mbox{\LiteraturePaperLink{https://www.roboticsproceedings.org/rss22/p096.html}\hspace{5pt}\LiteratureCode{https://github.com/NVlabs/RoboLab}} & \mbox{\ComparisonInputSlot{language}} & \mbox{Scene} & \mbox{Assemble}; \mbox{Edit} & \mbox{Isaac Sim} \\[0pt]
% Source: https://arxiv.org/html/2605.05241v1 | Sections 3.1, 4.1
DexSim2Real \mbox{\citep{survey_ax_2605_05241}} & \mbox{\LiteraturePaperLink{https://arxiv.org/abs/2605.05241}\hspace{5pt}{\setbox0=\hbox{\faGithub}\makebox[\wd0][c]{-}}} & \mbox{\ComparisonInputSlot{image}} & \mbox{Params.} & \mbox{Optimize} & \mbox{Isaac Sim} \\[0pt]
% Source: https://arxiv.org/html/2605.09423v2 | Section 2.1; Appendix E.2
\rowcolor{black!3}
SimWorld Studio \mbox{\citep{survey_ax_2605_09423}} & \mbox{\LiteraturePaperLink{https://arxiv.org/abs/2605.09423}\hspace{5pt}{\setbox0=\hbox{\faGithub}\makebox[\wd0][c]{-}}} & \mbox{\ComparisonInputSlot{language}\hspace{2pt}\ComparisonInputSlot{image}} & \mbox{Scene}; \mbox{Code} & \mbox{Generate}; \mbox{Assemble}; \mbox{Edit} & \mbox{Unreal Engine} \\[0pt]
% Source: https://arxiv.org/html/2606.28276v4 | Sections 3-4; Appendices E.2-E.4, I.2
SimFoundry \mbox{\citep{survey_ax_2606_28276}} & \mbox{\LiteraturePaperLink{https://arxiv.org/abs/2606.28276}\hspace{5pt}\LiteratureCode{https://github.com/NVlabs/SimFoundry}} & \mbox{\ComparisonInputSlot{video}} & \mbox{Mesh}; \mbox{URDF}; \mbox{Scene} & \mbox{Generate}; \mbox{Fit}; \mbox{Assemble} & \mbox{PyBullet}; \mbox{CoACD} \\[0pt]
% Source: https://arxiv.org/html/2607.07459v3 | Sections 2.2 (Geometry processing and Cross-simulator export), 2.4 (BFS spatial placement)
\rowcolor{black!3}
EmbodiedGen V2 \mbox{\citep{survey_ax_2607_07459}} & \mbox{\LiteraturePaperLink{https://arxiv.org/abs/2607.07459}\hspace{5pt}\LiteratureCode{https://github.com/HorizonRobotics/EmbodiedGen}} & \mbox{\ComparisonInputSlot{language}\hspace{2pt}\ComparisonInputSlot{image}} & \mbox{URDF}; \mbox{Scene} & \mbox{Generate}; \mbox{Assemble} & \mbox{SAPIEN}; \mbox{CoACD} \\[0pt]
% Source: https://arxiv.org/html/2609.05927v2 | Sections 3.1-3.4
GIF \mbox{\citep{survey_ax_2609_05927}} & \mbox{\LiteraturePaperLink{https://arxiv.org/abs/2609.05927}\hspace{5pt}{\setbox0=\hbox{\faGithub}\makebox[\wd0][c]{-}}} & \mbox{\ComparisonInputSlot{language}\hspace{2pt}\ComparisonInputSlot{image}} & \mbox{Mesh}; \mbox{Params.} & \mbox{Generate}; \mbox{Fit}; \mbox{Optimize} & \mbox{MuJoCo}; \mbox{Drake} \\[0pt]
% Source: https://arxiv.org/html/2609.24563v1 | Sections III-A, IV (Implementation Details)
\rowcolor{black!3}
ARSTAG \mbox{\citep{survey_ax_2609_24563}} & \mbox{\LiteraturePaperLink{https://arxiv.org/abs/2609.24563}\hspace{5pt}{\setbox0=\hbox{\faGithub}\makebox[\wd0][c]{-}}} & \mbox{\ComparisonInputSlot{language}\hspace{2pt}\ComparisonInputSlot{image}} & \mbox{Mesh}; \mbox{Scene} & \mbox{Generate}; \mbox{Fit}; \mbox{Assemble}; \mbox{Edit} & \mbox{Isaac Sim} \\[0pt]
% Source: https://arxiv.org/html/2609.30249v1 | Sections V-A-V-C, VI
RAPID \mbox{\citep{survey_ax_2609_30249}} & \mbox{\LiteraturePaperLink{https://arxiv.org/abs/2609.30249}\hspace{5pt}{\setbox0=\hbox{\faGithub}\makebox[\wd0][c]{-}}} & \mbox{\ComparisonInputSlot{language}\hspace{2pt}\ComparisonInputSlot{video}} & \mbox{Mesh}; \mbox{Scene} & \mbox{Generate}; \mbox{Fit}; \mbox{Assemble} & \mbox{MuJoCo} \\[0pt]
% Source: https://arxiv.org/html/2609.36024v3 | Sections 3.1-3.3, 4.1; Appendix B.4
\rowcolor{black!3}
CoDimRecon \mbox{\citep{survey_ax_2609_36024}} & \mbox{\LiteraturePaperLink{https://arxiv.org/abs/2609.36024}\hspace{5pt}{\setbox0=\hbox{\faGithub}\makebox[\wd0][c]{-}}} & \mbox{\ComparisonInputSlot{image}} & \mbox{Mesh}; \mbox{Scene} & \mbox{Generate}; \mbox{Fit}; \mbox{Edit} & \mbox{Blender}; \mbox{MuJoCo} \\[0pt]
% Source: https://arxiv.org/html/2609.35318v2 | Sections III-C, IV-A, V
DexAgent \mbox{\citep{survey_ax_2609_35318}} & \mbox{\LiteraturePaperLink{https://arxiv.org/abs/2609.35318}\hspace{5pt}{\setbox0=\hbox{\faGithub}\makebox[\wd0][c]{-}}} & \mbox{\ComparisonInputSlot{language}\hspace{2pt}\ComparisonInputSlot{video}} & \mbox{URDF}; \mbox{Scene} & \mbox{Generate}; \mbox{Fit}; \mbox{Edit} & \mbox{MuJoCo}; \mbox{Isaac Sim} \\[0pt]
% Source: https://arxiv.org/html/2609.36380v1 | Sections 3, 4.2, 5 (Implementation)
\rowcolor{black!3}
LEGO-Anything \mbox{\citep{survey_ax_2609_36380}} & \mbox{\LiteraturePaperLink{https://arxiv.org/abs/2609.36380}\hspace{5pt}{\setbox0=\hbox{\faGithub}\makebox[\wd0][c]{-}}} & \mbox{\ComparisonInputSlot{image}} & \mbox{Code}; \mbox{Scene} & \mbox{Generate}; \mbox{Edit} & \mbox{Blender} \\[0pt]
% Source: https://arxiv.org/html/2609.36777v1 | Sections 2.1-2.3; Appendices C.1, C.3
Code4Scene \mbox{\citep{survey_ax_2609_36777}} & \mbox{\LiteraturePaperLink{https://arxiv.org/abs/2609.36777}\hspace{5pt}{\setbox0=\hbox{\faGithub}\makebox[\wd0][c]{-}}} & \mbox{\ComparisonInputSlot{language}\hspace{2pt}\ComparisonInputSlot{image}} & \mbox{Scene} & \mbox{Generate}; \mbox{Assemble}; \mbox{Edit} & \mbox{Unreal Engine} \\[0pt]
% Source: https://arxiv.org/html/2610.02274v1 | Sections 4.1, 5-6; Section 7 (scene-graph instantiation)
\rowcolor{black!3}
Awomo-SimDataEngine \mbox{\citep{survey_ax_2610_02274}} & \mbox{\LiteraturePaperLink{https://arxiv.org/abs/2610.02274}\hspace{5pt}{\setbox0=\hbox{\faGithub}\makebox[\wd0][c]{-}}} & \mbox{\ComparisonInputSlot{language}\hspace{2pt}\ComparisonInputSlot{image}} & \mbox{URDF}; \mbox{Scene} & \mbox{Generate}; \mbox{Assemble}; \mbox{Edit} & \mbox{Isaac Sim} \\[0pt]
% Source: 2610.01863v1, Sections 3.1--3.3, 4.3--4.6.
LiteReality-Agent \mbox{\citep{survey_ax_2610_01863}} & \LiteratureLinks{https://arxiv.org/abs/2610.01863v1}{https://github.com/LiteReality/LiteReality-Agent} & \mbox{\ComparisonInputSlot{image}} & \mbox{Scene}; \mbox{Code} & \mbox{Generate}; \mbox{Edit} & \mbox{Blender} \\[0pt]
% Source: 2611.31002v1, Sections 2.2--2.3, A.2--A.3.
\rowcolor{black!3}
RoboUniverse-10K \mbox{\citep{survey_ax_2611_31002}} & \LiteratureLinks{https://arxiv.org/abs/2611.31002v1}{https://github.com/Einsia/RoboUniverse-10K} & \mbox{\ComparisonInputSlot{video}\hspace{2pt}\ComparisonInputSlot{geometry}} & \mbox{Scene}; \mbox{Code} & \mbox{Generate}; \mbox{Edit} & \mbox{Isaac Sim} \\[0pt]
\addlinespace[3pt]
\rowcolor{TSTEnvironmentLightest}\multicolumn{6}{@{}l}{\textbf{Dynamics}} \\*[3pt]
% Source: https://arxiv.org/html/2605.14398v3 | Sections 2.1-2.4, 3.1
ChronoAgentic \mbox{\citep{survey_ax_2605_14398}} & \mbox{\LiteraturePaperLink{https://arxiv.org/abs/2605.14398}\hspace{5pt}{\setbox0=\hbox{\faGithub}\makebox[\wd0][c]{-}}} & \mbox{\ComparisonInputSlot{language}\hspace{2pt}\ComparisonInputSlot{image}} & \mbox{Code} & \mbox{Generate}; \mbox{Edit} & \mbox{PyChrono} \\*[0pt]
% Source: https://arxiv.org/html/2607.19190v4 | Sections III-A-III-B (DROID conversion)
\rowcolor{black!3}
Agentic Real2Sim \mbox{\citep{survey_ax_2607_19190}} & \mbox{\LiteraturePaperLink{https://arxiv.org/abs/2607.19190v4}\hspace{5pt}\LiteratureCode{https://github.com/agentic-real2sim/agentic_real2sim}} & \mbox{\ComparisonInputSlot{video}} & \mbox{Mesh}; \mbox{Params.} & \mbox{Generate}; \mbox{Fit}; \mbox{Optimize} & \mbox{MuJoCo} \\*[0pt]
% Source: https://arxiv.org/html/2610.03715v1 | Sections 2.1, 2.4, 3.1; Appendices G.1-G.3
4DCodeBench \mbox{\citep{survey_ax_2610_03715}} & \mbox{\LiteraturePaperLink{https://arxiv.org/abs/2610.03715}\hspace{5pt}\LiteratureCode{https://github.com/4DCodeBench/4DCodeBench}} & \mbox{\ComparisonInputSlot{video}} & \mbox{Code} & \mbox{Generate}; \mbox{Edit} & \mbox{Blender} \\*[0pt]
% Source: https://arxiv.org/html/2610.04432v2 | Sections 2, 3.2-3.3; Appendices 7.2-7.3
\rowcolor{black!3}
Video2World \mbox{\citep{survey_ax_2610_04432}} & \mbox{\LiteraturePaperLink{https://arxiv.org/abs/2610.04432v2}\hspace{5pt}\LiteratureCode{https://github.com/AetherLabsAI/Video2World}} & \mbox{\ComparisonInputSlot{video}} & \mbox{Code}; \mbox{Scene} & \mbox{Generate}; \mbox{Edit} & \mbox{SAPIEN}; \mbox{Isaac Sim}; \mbox{MuJoCo} \\[0pt]
% Source: 2610.40565v1, Section 3, Figure 4, Table 1, and Section 6.1.
Gym-Anything for Robotics~\mbox{\citep{survey_ax_2610_40565}} & \LiteratureLinks{https://arxiv.org/abs/2610.40565v1}{https://github.com/gym-anything/robotics} & \mbox{\ComparisonInputSlot{language}} & \mbox{Scene}; \mbox{Code} & \mbox{Generate}; \mbox{Edit} & \mbox{Custom solvers} \\[0pt]
% Source: 2611.31098v2, Sections 3.1 and 3.2.1--3.2.4.
\rowcolor{black!3}
Real2Worlds2Real~\mbox{\citep{survey_ax_2611_31098}} & \LiteratureLinks{https://arxiv.org/abs/2611.31098v2}{https://github.com/real2worlds2real/real2worlds2real} & \mbox{\ComparisonInputSlot{language}\hspace{1pt}\ComparisonInputSlot{image}\hspace{1pt}\ComparisonInputSlot{video}} & \mbox{Scene}; \mbox{Params.} & \mbox{Generate}; \mbox{Fit}; \mbox{Edit} & \mbox{Blender}; \mbox{PhysX}; \mbox{Newton} \\[0pt]
\end{longtable}
\endgroup

\subsection{Task Generation}
\label{sec:environment-task-generation}

Task generation specifies the goal, initial state, constraints, and success conditions that a robot must execute. It covers both construction of executable tasks and adjustment of task distributions as the learner changes.

\paragraph{\textbf{Executable Task Construction.}}
GenSim~\citep{survey_ax_2310_01361} writes simulation code for task libraries. RoboGen~\citep{survey_ax_2311_01455} couples task proposals with scenes and supervision, while Video2Policy~\citep{survey_ax_2502_09886} derives task code and rewards from internet video. RoboLab~\citep{survey_ax_2604_09860} composes tasks from competency axes, predicates, and asset constraints, then feeds syntax and asset errors back to the generator. HARBOR~\citep{survey_ax_2606_08610} checks scene stages, initialization, termination, actions, and observations. Scene2Demo~\citep{survey_ax_2602_12065} links subtasks through an object--action graph and generates executable initial and goal conditions. PARTNR~\citep{survey_ax_2411_00081} filters generated instructions and initial states and builds evaluators for propositions, dependencies, and temporal constraints.

RoboUniverse-10K~\citep{survey_ax_2611_31002} uses a demonstration video to define what the robot must do. In one example, the lid starts on the table and must end on the pot. The robot can choose how to move it. The scene builder and the verifier use the same task specification.

Once the physics is fixed, a single scene can support several goals. In Gym-Anything for Robotics~\citep{survey_ax_2610_40565}, an agent chooses an initial state, writes an instruction, and builds a verifier. One wine-pouring task checks how much reaches the target glass, whether any spills, and whether the robot returns the bottle to its coaster.

Checks can also govern generation of human interaction and physical feasibility. FATE~\citep{survey_ax_2603_01505} separates static semantic, geometric, and precondition checks from dynamic execution checks and repairs the failing component. GenPHRI~\citep{survey_ax_2604_08664} uses a text2sim2real pipeline to generate a human, a scene, and an assistive task; collision and kinematic checks constrain the resulting placements and motions. These repairs can change task difficulty or meaning, so feasibility and objective preservation should be checked together.

Generating a solution does not necessarily generate a new task. EmbodiedSmith~\citep{survey_ax_2610_07969} composes skills into task graphs and revises plans and initial placements while keeping assets, layout, and success criteria fixed. Scene2Demo likewise revises action flows from rollout critiques and restarts the current subtask from a saved state. The resulting demonstrations belong to data generation (Section~\ref{sec:data-collection}), not to a new task definition.

\paragraph{\textbf{Adaptive Task Distributions.}}
CurricuLLM~\citep{survey_ax_2409_18382} generates a subtask sequence, candidate reward and goal-distribution code, and passes the selected policy to the next subtask. RATs~\citep{survey_ax_2606_19419} selects practice tasks from scene information, a code-skill library, and failure memory, balancing novelty and difficulty. SimWorld Studio~\citep{survey_ax_2605_09423} instead feeds embodied-agent performance to an environment generator that changes environments and task difficulty. RATs changes the next goal; SimWorld Studio changes the task distribution. Evaluation on separately specified tasks is needed because a curriculum optimized from the learner's own success signal can become easier without improving the target ability.

RALLY~\citep{survey_ax_2611_37676} pairs a task proposer, Alice, with a coding agent, Bob. Alice reads Bob's current program to decide what Bob should practice next. In adversarial play, Alice changes a LIBERO scene to expose a weakness in Bob's program while keeping the original goal. In cooperative play, Alice writes a new goal and success conditions for a task that shares steps with one Bob has practiced. With the authors' coding harness, adversarial practice raised Bob's average success on held-out LIBERO perturbations by 17.7 percentage points. Related works are compared in Table~\ref{tab:environment-task-generation}.

\par\Needspace{12\baselineskip}
\begingroup
\mechanismtableglue
\CompactTableStyle
\setlength{\LiteratureWidth}{\dimexpr\linewidth-10\tabcolsep\relax}

\setlength{\LTpre}{4pt}
\setlength{\LTpost}{4pt}
\setlength{\LTcapwidth}{\linewidth}
\captionsetup{singlelinecheck=false,justification=raggedright}
\LiteratureContinuationSetup
\begin{longtable}{@{}>{\RaggedRight\arraybackslash}p{0.180\LiteratureWidth}>{\centering\arraybackslash}p{0.048\LiteratureWidth}>{\RaggedRight\arraybackslash}p{0.183\LiteratureWidth}>{\RaggedRight\arraybackslash}p{0.173\LiteratureWidth}>{\RaggedRight\arraybackslash}p{0.214\LiteratureWidth}>{\RaggedRight\arraybackslash}p{0.202\LiteratureWidth}@{}}
\caption[Comparison of works on task generation]{Comparison of works on task generation (Section~\ref{sec:environment-task-generation}). Output lists generated or revised artifacts. Eval.: success/evaluation code; Config.: initialization/randomization settings; Plan: executable solution, which may preserve the task. UE5: Unreal Engine 5; MolmoSp.: MolmoSpaces. Multiple (HARBOR): Isaac Lab, ManiSkill, Genesis, MJLab.}\label{tab:environment-task-generation}\\
\LiteratureTopRule{Environment}
\rowcolor{TSTEnvironmentLight}\textbf{\mbox{Work}} & \textbf{\mbox{Link}} & \textbf{\mbox{Input}} & \textbf{\mbox{Output}} & \textbf{\mbox{Agent} \mbox{Operations}} & \textbf{\mbox{Tools}} \\
\midrule
\endfirsthead
\multicolumn{6}{@{}l}{\tablename~\thetable\ (continued)} \\
\LiteratureTopRule{Environment}
\rowcolor{TSTEnvironmentLight}\textbf{\mbox{Work}} & \textbf{\mbox{Link}} & \textbf{\mbox{Input}} & \textbf{\mbox{Output}} & \textbf{\mbox{Agent} \mbox{Operations}} & \textbf{\mbox{Tools}} \\
\midrule
\endhead
\midrule
\multicolumn{6}{r@{}}{Continued on next page} \\
\endfoot
\bottomrule
\endlastfoot
\addlinespace[3pt]
\rowcolor{TSTEnvironmentLightest}\multicolumn{6}{@{}l}{\textbf{Tasks}} \\*[3pt]
% Source: https://arxiv.org/html/2310.01361v2 | Sections 2.1-2.2, 3.1; Appendices A.1, D.5-D.6
GenSim \mbox{\citep{survey_ax_2310_01361}} & \mbox{\LiteraturePaperLink{https://proceedings.iclr.cc/paper_files/paper/2024/hash/143ea4a156ef64f32d4d905206cf32e1-Abstract-Conference.html}\hspace{5pt}\LiteratureCode{https://github.com/liruiw/GenSim}} & \mbox{Goal}; \mbox{Examples} & \mbox{Task}; \mbox{Curriculum} & \mbox{Generate}; \mbox{Select} & \mbox{Ravens} \\[0pt]
% Source: https://arxiv.org/html/2311.01455v3 | Sections 3.1-3.4, 4.1; Appendix A.3
\rowcolor{black!3}
RoboGen \mbox{\citep{survey_ax_2311_01455}} & \mbox{\LiteraturePaperLink{https://proceedings.mlr.press/v235/wang24cc.html}\hspace{5pt}\LiteratureCode{https://github.com/Genesis-Embodied-AI/RoboGen}} & \mbox{Assets}; \mbox{Examples} & \mbox{Task}; \mbox{Reward}; \mbox{Plan} & \mbox{Generate}; \mbox{Compose} & \mbox{Genesis}; \mbox{OMPL} \\[0pt]
% Source: https://arxiv.org/html/2502.09886v1 | Sections 3.1-3.2, 4 (Experimental Setup); Appendix A.1
Video2Policy \mbox{\citep{survey_ax_2502_09886}} & \mbox{\LiteraturePaperLink{https://arxiv.org/abs/2502.09886}\hspace{5pt}{\setbox0=\hbox{\faGithub}\makebox[\wd0][c]{-}}} & \mbox{Demo} & \mbox{Task}; \mbox{Reward}; \mbox{Eval.} & \mbox{Generate}; \mbox{Repair}; \mbox{Select} & \mbox{Isaac Gym} \\[0pt]
% Source: https://arxiv.org/html/2411.00081v1 | Sections 3.1-3.2, 4; Appendices A.4-A.6
\rowcolor{black!3}
PARTNR \mbox{\citep{survey_ax_2411_00081}} & \mbox{\LiteraturePaperLink{https://proceedings.iclr.cc/paper_files/paper/2025/hash/a3cf318fbeec1126da21e9185ae9908c-Abstract-Conference.html}\hspace{5pt}\LiteratureCode{https://github.com/facebookresearch/partnr-planner}} & \mbox{Scene}; \mbox{Examples} & \mbox{Task}; \mbox{Eval.} & \mbox{Generate}; \mbox{Vary} & \mbox{Habitat 3.0} \\[0pt]
% Source: https://arxiv.org/html/2604.05226v1 | Sections III-A-III-D, IV-A; Appendix A1-A2
RoboPlayground \mbox{\citep{survey_ax_2604_05226}} & \mbox{\LiteraturePaperLink{https://arxiv.org/abs/2604.05226}\hspace{5pt}{\setbox0=\hbox{\faGithub}\makebox[\wd0][c]{-}}} & \mbox{Goal} & \mbox{Task}; \mbox{Eval.} & \mbox{Generate}; \mbox{Vary}; \mbox{Repair} & \mbox{MuJoCo} \\[0pt]
% Source: https://arxiv.org/html/2604.09860v4 | Sections III-A1-III-A2, III-B, IV-A; Appendix D
\rowcolor{black!3}
RoboLab \mbox{\citep{survey_ax_2604_09860}} & \mbox{\LiteraturePaperLink{https://www.roboticsproceedings.org/rss22/p096.html}\hspace{5pt}\LiteratureCode{https://github.com/NVlabs/RoboLab}} & \mbox{Scene}; \mbox{Examples} & \mbox{Task}; \mbox{Eval.} & \mbox{Generate}; \mbox{Compose}; \mbox{Repair} & \mbox{Isaac Sim} \\[0pt]
% Source: https://arxiv.org/html/2603.01505v1 | Sections 3.1-3.3, 4.1-4.4, 5.1; Appendix D
FATE \mbox{\citep{survey_ax_2603_01505}} & \mbox{\LiteraturePaperLink{https://arxiv.org/abs/2603.01505}\hspace{5pt}{\setbox0=\hbox{\faGithub}\makebox[\wd0][c]{-}}} & \mbox{Scene}; \mbox{Goal}; \mbox{Policy} & \mbox{Task}; \mbox{Reward} & \mbox{Repair} & \mbox{Isaac Sim} \\[0pt]
% Source: https://arxiv.org/html/2606.08610v1 | Section 3, 4.1; Appendices C.3-C.5, D.3
\rowcolor{black!3}
HARBOR \mbox{\citep{survey_ax_2606_08610}} & \mbox{\LiteraturePaperLink{https://arxiv.org/abs/2606.08610}\hspace{5pt}\LiteratureCode{https://github.com/supersglzc/harbor-rl}} & \mbox{Goal}; \mbox{Code} & \mbox{Task}; \mbox{Reward}; \mbox{Config.} & \mbox{Generate}; \mbox{Repair} & \mbox{Multiple} \\[0pt]
% Source: 2610.40565v1, Section 4 and Appendix C.
Gym-Anything for Robotics~\mbox{\citep{survey_ax_2610_40565}} & \LiteratureLinks{https://arxiv.org/abs/2610.40565v1}{https://github.com/gym-anything/robotics} & \mbox{Scene} & \mbox{Task}; \mbox{Config.}; \mbox{Eval.} & \mbox{Generate}; \mbox{Repair} & \mbox{RoboWorld} \\[0pt]
\addlinespace[3pt]
\rowcolor{TSTEnvironmentLightest}\multicolumn{6}{@{}l}{\textbf{Curricula}} \\*[3pt]
% Source: https://arxiv.org/html/2409.18382v2 | Sections IV-A-IV-C, V-A-V-B; Appendix VII-A
CurricuLLM \mbox{\citep{survey_ax_2409_18382}} & \mbox{\LiteraturePaperLink{https://arxiv.org/abs/2409.18382}\hspace{5pt}\LiteratureCode{https://github.com/labicon/CurricuLLM}} & \mbox{Goal}; \mbox{Scene} & \mbox{Curriculum}; \mbox{Reward} & \mbox{Generate}; \mbox{Schedule}; \mbox{Select} & \mbox{Gymnasium}; \mbox{Isaac Lab} \\[0pt]
% Source: https://arxiv.org/html/2605.09423v2 | Sections 2.1-2.2, 3.3; Appendices C.2, G.2-G.4
\rowcolor{black!3}
SimWorld Studio \mbox{\citep{survey_ax_2605_09423}} & \mbox{\LiteraturePaperLink{https://arxiv.org/abs/2605.09423}\hspace{5pt}{\setbox0=\hbox{\faGithub}\makebox[\wd0][c]{-}}} & \mbox{Goal}; \mbox{Scene}; \mbox{History} & \mbox{Task}; \mbox{Curriculum} & \mbox{Generate}; \mbox{Vary}; \mbox{Schedule} & \mbox{UE5}; \mbox{Gymnasium} \\[0pt]
% Source: https://arxiv.org/html/2606.19419v1 | Sections 3.2, 4.1; Appendix A.1
RATs \mbox{\citep{survey_ax_2606_19419}} & \mbox{\LiteraturePaperLink{https://arxiv.org/abs/2606.19419}\hspace{5pt}\LiteratureCode{https://github.com/Playful-RATs/RATs}} & \mbox{Scene}; \mbox{Skills}; \mbox{History} & \mbox{Task} & \mbox{Generate}; \mbox{Select} & \mbox{LIBERO-PRO}; \mbox{MolmoSp.} \\[0pt]
% Source: 2611.37676v1, Sections 3.2, 4.1--4.3 and Appendix A.
RALLY~\mbox{\citep{survey_ax_2611_37676}} & \LiteratureLinks{https://arxiv.org/abs/2611.37676v1}{} & \mbox{Scene}; \mbox{History} & \mbox{Config.}; \mbox{Task}; \mbox{Eval.} & \mbox{Vary}; \mbox{Generate} & \mbox{LIBERO} \\[0pt]
\addlinespace[3pt]
\rowcolor{TSTEnvironmentLightest}\multicolumn{6}{@{}l}{\textbf{Plans}} \\*[3pt]
% Source: https://arxiv.org/html/2609.24563v1 | Sections III-B-III-C, IV-A; Appendix B
ARSTAG \mbox{\citep{survey_ax_2609_24563}} & \mbox{\LiteraturePaperLink{https://arxiv.org/abs/2609.24563}\hspace{5pt}{\setbox0=\hbox{\faGithub}\makebox[\wd0][c]{-}}} & \mbox{Image}; \mbox{Goal} & \mbox{Plan}; \mbox{Config.} & \mbox{Compose}; \mbox{Vary} & \mbox{Isaac Sim} \\[0pt]
% Source: https://arxiv.org/html/2602.12065v2 | Sections 3.1-3.6, 4.1-4.2; Appendices A.1-A.3
\rowcolor{black!3}
Scene2Demo \mbox{\citep{survey_ax_2602_12065}} & \mbox{\LiteraturePaperLink{https://arxiv.org/abs/2602.12065}\hspace{5pt}{\setbox0=\hbox{\faGithub}\makebox[\wd0][c]{-}}} & \mbox{Image}; \mbox{Goal} & \mbox{Task}; \mbox{Eval.}; \mbox{Plan} & \mbox{Generate}; \mbox{Compose}; \mbox{Repair} & \mbox{OmniGibson} \\[0pt]
% Source: https://arxiv.org/html/2604.08664v2 | Sections 3.1-3.2, 4, 5.1
GenPHRI \mbox{\citep{survey_ax_2604_08664}} & \mbox{\LiteraturePaperLink{https://arxiv.org/abs/2604.08664v2}\hspace{5pt}{\setbox0=\hbox{\faGithub}\makebox[\wd0][c]{-}}} & \mbox{Goal} & \mbox{Config.}; \mbox{Plan} & \mbox{Generate}; \mbox{Repair} & \mbox{Genesis}; \mbox{SceneSmith} \\*[0pt]
% Source: https://arxiv.org/html/2610.07969v1 | Sections 3.2-3.3, 4.2; Appendices 8.4, 8.6
\rowcolor{black!3}
EmbodiedSmith \mbox{\citep{survey_ax_2610_07969}} & \mbox{\LiteraturePaperLink{https://arxiv.org/abs/2610.07969}\hspace{5pt}{\setbox0=\hbox{\faGithub}\makebox[\wd0][c]{-}}} & \mbox{Goal}; \mbox{Robot} & \mbox{Plan}; \mbox{Config.} & \mbox{Compose}; \mbox{Repair} & \mbox{Isaac Sim}; \mbox{cuRobo} \\[0pt]
\end{longtable}
\endgroup

\subsection{Benchmarking}
\label{sec:environment-benchmarking}

Benchmarking fixes the task, interface, reset protocol, and scoring rule used to measure an ability. Agents may be the systems under test, or they may construct and revise the evaluation itself. Table~\ref{tab:benchmark-protocols} shows how reported success varies with observation access, control interfaces, execution feedback, and retries. Each study compares the same model on the same tasks.

\paragraph{\textbf{Evaluating Agents in Robot Control.}}
Planning benchmarks test whether an agent converts an instruction into robot operations. EmbodiedBench~\citep{survey_ax_2502_09560} varies action abstraction and returns environment feedback. Embodied Agent Arena~\citep{survey_ax_2610_00854} uses multi-round interaction in its Planning and Manipulation domains. MissionBench~\citep{survey_ax_2607_22014} evaluates an MLLM that selects motion primitives, magnitudes, and stopping decisions for aerial missions. KinDER~\citep{survey_ax_2604_25788} evaluates open-loop composition of parameterized skills, so its score does not measure feedback-driven replanning.

Tool and programming benchmarks evaluate executable decisions. EmbodiedToolBench~\citep{survey_ax_2605_26637} separates tool recognition, selection, execution, output use, and tool-chain composition. CaP-X~\citep{survey_ax_2603_22435} varies state access, API abstraction, examples, and visual grounding while executing generated control code. With noisy perception and the same high-level commands, CaP-Bench either requires one program or lets the agent revise it after reading the execution output. EmbodiedSWE~\citep{survey_ax_2609_27308} gives coding agents simulator state, rendering, and execution feedback, then grades the resulting program in a separate environment.

Benchmarks for coding agents differ in what the agent can observe and when it can reset or try again. In EFE~\citep{project_embodied_first_exam}, the agent chooses its own control method. In the Standard setting, the agent observes the environment through the robot's sensors and can reset only when the task allows it. The simulator normally pauses between requests, but it kept running during inference on 50 of the 391 tasks in the comparison. GPT-6 Astra solved more tasks with saved states and retries than with exact object states alone. With save and reset access, 66\% of its successes came after it resumed from a saved state. It solved the most tasks when it had both permissions.

RoboARC~\citep{survey_ax_2611_33390} separates two ways to solve a physical puzzle. Standard gives the agent camera and robot feedback in a single attempt. Privileged also gives it the simulator code, full state, and resets. An earlier set of runs identified 15 rigid puzzles that GPT-6 Astra at high reasoning effort solved only with Privileged access. On fresh runs without resets, the agent solved 11 of 45 attempts with Standard access and 27 of 45 when given both object poses and part geometry (Table~\ref{tab:benchmark-protocols}).

In RoboUniverse-Verified~\citep{survey_ax_2611_31002}, each agent gets one attempt per task and cannot query intermediate scores. During benchmark construction, a separate agent resets each scene and replays a reference solution to check the task. The authors sampled 500 tasks across embodiments and task families after automated checks, then had experts verify them.

Embodied Playground~\citep{survey_ax_2610_27039} checks both the robot's movements and the final state. A program can produce the right order of moves in Tower of Hanoi yet displace a ring during contact. EP-100 tests each task under four initial conditions that vary factors such as object pose or lighting. GPT-6 Astra in Codex could read privileged state and had up to two valid attempts per condition, each in a fresh session. The reported pass rates give equal weight to the five task suites. In 574 of the 645 scored failures, the authors identified a rule or threshold that the instructions had left unclear.

Existing robot suites can host these evaluations. LIBERO~\citep{survey_ax_2306_03310} provides a standard lifelong-learning suite, and LIBERO-PRO~\citep{survey_ax_2510_03827} changes the protocol to reduce memorization and improve robustness. CaP-X evaluates CaP-Agent0 on LIBERO-PRO with position and instruction perturbations, while ASPIRE~\citep{survey_ax_2607_00272} uses LIBERO-PRO, robosuite, and BEHAVIOR-1K with separate protocols for fixed programs and trace-based code revision. The agent score therefore depends on whether the program is fixed before evaluation or revised from the current trace.

Harness VLA~\citep{survey_ax_2607_08448} tests agent selection of analytic primitives with a frozen VLA on LIBERO, LIBERO-PRO, RoboCasa365, and RoboTwin C2R. RoboHarness~\citep{survey_ax_2607_18060} tests heterogeneous policy orchestration, perturbation robustness, and long-horizon execution on LIBERO variants and adds tasks for policy handoffs. In both cases, the task suite alone does not determine the measured ability; the execution and adaptation protocol does as well.
RoboDojo~\citep{survey_ax_2607_04434} combines 42 simulation tasks with 18 real-world tasks and evaluates generalization, memory, precision, long-horizon execution, and open-vocabulary instruction following. 
RobotWorld~\citep{survey_ax_2610_10409} extends this setting across manipulation, mobile manipulation, locomotion, driving, and aerial control. Its 84 simulation tasks specify interaction budgets and executable success checks, and its trace analysis links outcomes to the agent's perception, computation, feedback use, and completion decisions.

The scoring rule determines which actions count as success, while the control interface and step budget determine what the agent can try. RoboWild~\citep{xlanglab2026robowild} tests whether the robot followed the requested procedure. For a food dispenser, five pellets in the bowl could mean that the robot operated the dispenser or simply poured them from the box. The evaluator records events during execution and checks them alongside the final state. A coding agent implements the evaluator from the public specification, and a separate verifier tests it against private annotated cases. Across the full benchmark, 239 task setups yield 341 setup--embodiment pairs. The authors first average the outcomes across the robots tested on each setup. They then average the 239 setup scores, giving each setup equal weight.

On EBench~\citep{survey_ax_2610_35832}, the authors tested GPT-6 Astra on the same episodes with direct waypoint control alone or with policy calls also available. On Bottle and Shop, direct control had a 3,000-step limit. The authors scored each policy-assisted run at both 3,000 and 5,000 steps. Each task comes with one annotated demonstration. In 31 of the 510 episodes, it showed the test scene itself, so the agent had already seen that scene before taking control.

The SenseNova study~\citep{survey_ax_2610_22471} examines how GPT-6 Astra uses demonstrations and extra time across 12 benchmarks. Where training demonstrations are available, the model can use them to prepare notes and code. Each test session then starts with a fixed copy of these materials. When the step budget is extended, the agent continues from the scene left by its earlier actions.

Some physical tests score a single move or choice within a longer task. DexHoldem~\citep{survey_ax_2605_18727} asks whether each primitive succeeds and leaves the table usable for the next move. Each policy is scored on 80 scheduled primitive trials. If a pickup fails, the corresponding put-down also counts as a failure. A separate test uses 36 perception problems drawn from recorded poker hands. With the GPT-5.5 agent and the $\pi_0$ policy, four of 33 poker hands were completed. Only one of those four did so without a retry or a request for human help. In RAB~\citep{survey_ax_2610_34649}, touching the right object can earn credit for discovering a task or interpreting a preference. Completing the household activity is scored separately from this choice.

LLM4Robot-Bench~\citep{survey_ax_2610_34527} tests the same 16 manipulation tasks in simulation and on hardware, with three trials per task in each setting. External and wrist RGB views, robot state, and execution feedback let the agent check contact without a force reading. On hardware, GPT-6 Astra succeeded in 12 of 15 trials (80\%) on L1 placement and alignment tasks. It completed none of the 12 L3 trials, which required changing contact while regulating force. Simulation and hardware gave the same success count on 10 of the 16 tasks. On seven of those ten, all three trials failed in both settings.

RoboStressBench~\citep{survey_ax_2606_00828} tests perception with fixed images. It asks how well a model answers questions when the material, lighting, viewpoint, or geometry changes. It uses natural images and synthesized variants. Depending on the question, the model selects an answer, marks a point, or returns a box. A point must fall inside the annotated mask. A box is scored by its overlap with the reference box.

RoboARCade~\citep{survey_ax_2611_08209} compares two ways to aim the gripper on 22 manipulation tasks. With calibrated RGB-D images, the agent commands changes in gripper position and rotation. In the point cloud, it places a virtual gripper and executes waypoints. Once the scene settles, the grader checks individual subtasks. For a PCR machine, placing the plate earns credit even if the agent fails to close the lid, secure the latch, or turn the knob.

Active-perception tests evaluate information gathering. VABench~\citep{survey_ax_2609_19554} asks an MLLM to inspect a scene, issue camera and Cartesian commands, and revise decisions after new observations. ActiveArena-Plan~\citep{survey_ax_2609_24124} uses a planner to select subtasks and history for a VLA executor. The former tests observation gathering directly; the latter also tests how selected history guides task progress.

EmbodiedLab~\citep{survey_ax_2610_35385} asks an agent to measure a hidden physical quantity on a simulated instrument bench. To find the focal length of a lens, for example, the agent must operate the bench, read a scale, and submit its answer with evidence. The grader checks the claimed readings against the instrument log as well as the result and required final state.

Memory tests ask whether earlier observations remain useful. RoboGraph~\citep{survey_ax_2608_08036} controls dependencies, observation access, and external events to test state maintenance and updating. RoboMemArena~\citep{survey_ax_2605_10921} uses a VLM planner to select subtasks and retained observations for a low-level executor. EvoNav-Bench~\citep{survey_ax_2609_08292} introduces hidden scene changes that can make retained evidence stale. MEMOBench~\citep{survey_ax_2609_07047} adds executable checkpoints for memory storage, update, and compression, separating memory fidelity from final manipulation success. Although its reported subjects are VLA policies, the same process-level checks can be applied to agent-controlled rollouts; memory mechanisms are discussed in Section~\ref{sec:harness}.

RoboMite~\citep{survey_ax_2611_37041} gives instructions one at a time in 17 kitchen task suites. A later instruction can ask for an object the robot saw while doing an unrelated task. The benchmark scores both the completed steps and whether the robot finished every task in the sequence.

Mind2Act World~\citep{survey_ax_2611_26444} tests memory and reasoning, physical control, and tasks that require both. For example, the robot must remember a decoration pattern, reproduce it on a rotating cake, and check that the cake has stopped before covering it. Its 30 tasks span three difficulty levels and two scene settings, giving 180 configurations. For each failed episode of three representative systems, the authors identified the first error from which the system did not recover. They used simulator checks first, a language model for unresolved cases, and human review for uncertain or conflicting judgments.

Safety tests separate hazard avoidance from task completion. ReactHuman~\citep{survey_ax_2609_10895} scores an MLLM's motor plan for a sudden event against simulator ground truth. ManiGuard~\citep{survey_ax_2608_17386} uses specification-grounded runtime monitors and reports safety separately from task success. Its published evaluations use VLA policies, but the monitors can also score agent-controlled rollouts under the same task protocol. ReactHuman pauses simulation during inference, so it measures the hazard-response decision rather than real-time reaction.

RoboRisk~\citep{survey_ax_2611_14115} labels each run as harmful, safe and successful, or safe but incomplete. Refusing an unsafe instruction or stopping in time when an unexpected hazard appears counts as success. On tasks that require a safer plan, the robot must still complete the instruction. The study uses three seeds per task on 92 simulated tasks and one run per task and system on 13 physical tasks.

Some benchmarks keep the simulation running while agents plan. RoboTeam~\citep{survey_ax_2611_22138} tests how several arms divide a job under this time constraint. Across 36 simulated tasks, it compares one agent for all arms, a supervisor with one worker per arm, workers that coordinate directly, and a team that can reassign arms. As agents reason and exchange messages, the scene can change before their next command takes effect.

A benchmark score therefore depends on the full execution contract: agent decision, executor, feedback path, observation access, controller assistance, and interaction budget.

\paragraph{\textbf{Agent-Directed Evaluation Construction.}}
An agent that builds an evaluation must specify the ability under test, the task conditions, and the evidence used for scoring. ManipEvalAgent~\citep{survey_iclr_2026_8833c8aa} decomposes an evaluation request, generates tasks and evaluation tools, and uses rollout results to propose further tests or summarize the tested policy. Embodied-BenchForge~\citep{survey_ax_2609_13082} generates task templates, data, and executable verifiers; interactive tasks require restorable states and terminal checks, while offline tasks require grounded evidence and executable scoring. Its dependency graph supports targeted rollback and repair. A2Eval~\citep{survey_ax_2602_01640} groups embodied-reasoning questions by capability, generates inference and scoring code, and repairs that code from execution feedback.

RoboWorld~\citep{survey_ax_2610_40565} includes 25 test tasks built with Gym-Anything for Robotics. Each starts from a specified state. Agents see camera images and robot feedback during an attempt. A verifier scores success and partial progress after the attempt ends. GPT-6 Astra, GPT-6.1 Terra, and Claude Opus~5.5 each failed every test task. All three solved some of the easier tasks drawn from the training split.

Adaptive evaluation can reveal weaknesses, but its conclusions depend on test selection, access to prior results, and score validation. Fixing test conditions independently of the evaluated system supports broader claims. Improving the evaluator and improving the robot are separate outcomes.

World construction, task generation, and benchmarking diagnose different failures: unstable geometry requires scene repair, an infeasible goal requires task revision, and an invalid success signal requires evaluator repair. Keeping these checks separate makes failures easier to locate and simulated conditions easier to validate on hardware. Related works are compared in Table~\ref{tab:environment-benchmarking}.

\par\Needspace{12\baselineskip}
\begingroup
\mechanismtableglue
\CompactTableStyle
\newcommand{\ComparisonInputSlot}[1]{\makebox[12pt][c]{\raisebox{\dimexpr\depth-1.5pt\relax}{\resizebox*{!}{9pt}{\ComparisonGlyph{#1}}}}}
\setlength{\LiteratureWidth}{\dimexpr\linewidth-10\tabcolsep\relax}

\setlength{\LTpre}{4pt}
\setlength{\LTpost}{4pt}
\setlength{\LTcapwidth}{\linewidth}
\captionsetup{singlelinecheck=false,justification=raggedright}
\LiteratureContinuationSetup
\begin{longtable}{@{}>{\RaggedRight\arraybackslash}p{0.236\LiteratureWidth}>{\centering\arraybackslash}p{0.046\LiteratureWidth}>{\RaggedRight\arraybackslash}p{0.078\LiteratureWidth}>{\RaggedRight\arraybackslash}p{0.178\LiteratureWidth}>{\RaggedRight\arraybackslash}p{0.150\LiteratureWidth}>{\RaggedRight\arraybackslash}p{0.312\LiteratureWidth}@{}}
\caption[Comparison of works on benchmarking]{Comparison of works on benchmarking (Section~\ref{sec:environment-benchmarking}). Interface describes tested-system outputs: Answer (predictions), Action (motor commands), Call (tools/skills), Code (programs); evaluator-generated code is excluded. RobotWorld's Code is optional. Engines lists simulation backends; Multiple denotes mixed suites. NR: not reported; N/A: no simulator execution. Collab.: collaboration; UE5: Unreal Engine 5. Setting: \mbox{\ComparisonGlyph{real}\hspace{3pt}Real};\allowbreak\enspace \mbox{\ComparisonGlyph{sim}\hspace{3pt}Sim};\allowbreak\enspace \mbox{\ComparisonGlyph{dataset}\hspace{3pt}Offline data}.}\label{tab:environment-benchmarking}\\
\LiteratureTopRule{Environment}
\rowcolor{TSTEnvironmentLight}\textbf{\mbox{Work}} & \textbf{\mbox{Link}} & \textbf{\mbox{Setting}} & \textbf{\mbox{Capability}} & \textbf{\mbox{Interface}} & \textbf{\mbox{Engines}} \\
\midrule
\endfirsthead
\multicolumn{6}{@{}l}{\tablename~\thetable\ (continued)} \\
\LiteratureTopRule{Environment}
\rowcolor{TSTEnvironmentLight}\textbf{\mbox{Work}} & \textbf{\mbox{Link}} & \textbf{\mbox{Setting}} & \textbf{\mbox{Capability}} & \textbf{\mbox{Interface}} & \textbf{\mbox{Engines}} \\
\midrule
\endhead
\midrule
\multicolumn{6}{r@{}}{Continued on next page} \\
\endfoot
\bottomrule
\endlastfoot
\addlinespace[3pt]
\rowcolor{TSTEnvironmentLightest}\multicolumn{6}{@{}l}{\textbf{Benchmarks}} \\*[3pt]
% Source: https://arxiv.org/html/2306.03310v2 | Sections 2, 3, 4.1, 4.4, 5; Appendices C-D
LIBERO \mbox{\citep{survey_ax_2306_03310}} & \mbox{\LiteraturePaperLink{https://proceedings.neurips.cc/paper_files/paper/2023/hash/8c3c666820ea055a77726d66fc7d447f-Abstract-Datasets_and_Benchmarks.html}\hspace{5pt}\LiteratureCode{https://github.com/Lifelong-Robot-Learning/LIBERO}} & \mbox{\ComparisonInputSlot{sim}} & \mbox{Transfer} & \mbox{Action} & \mbox{robosuite} \\[0pt]
% Source: https://arxiv.org/html/2411.00081v1 | Sections 3.1, 4.1; Appendices A.4.1, A.7-A.8
\rowcolor{black!3}
PARTNR \mbox{\citep{survey_ax_2411_00081}} & \mbox{\LiteraturePaperLink{https://proceedings.iclr.cc/paper_files/paper/2025/hash/a3cf318fbeec1126da21e9185ae9908c-Abstract-Conference.html}\hspace{5pt}\LiteratureCode{https://github.com/facebookresearch/partnr-planner}} & \mbox{\ComparisonInputSlot{sim}} & \mbox{Planning}; \mbox{Collab.} & \mbox{Call} & \mbox{Habitat 3.0} \\[0pt]
% Source: https://arxiv.org/html/2502.09560v3 | Sections 4.1-4.3, 5.1; Appendix C.1-C.4, I.2-I.4
EmbodiedBench \mbox{\citep{survey_ax_2502_09560}} & \mbox{\LiteraturePaperLink{https://proceedings.mlr.press/v267/yang25f.html}\hspace{5pt}\LiteratureCode{https://github.com/EmbodiedBench/EmbodiedBench}} & \mbox{\ComparisonInputSlot{sim}} & \mbox{Planning}; \mbox{Reasoning} & \mbox{Action}; \mbox{Call} & \mbox{AI2-THOR}; \mbox{Habitat}; \mbox{CoppeliaSim} \\[0pt]
% Source: https://arxiv.org/html/2510.03827v2 | Sections 3.1-3.2, 4.1-4.3
\rowcolor{black!3}
LIBERO-PRO \mbox{\citep{survey_ax_2510_03827}} & \mbox{\LiteraturePaperLink{https://arxiv.org/abs/2510.03827}\hspace{5pt}{\setbox0=\hbox{\faGithub}\makebox[\wd0][c]{-}}} & \mbox{\ComparisonInputSlot{sim}} & \mbox{Robustness} & \mbox{Action} & \mbox{LIBERO} \\[0pt]
% Source: https://arxiv.org/html/2604.09860v4 | Sections III-A to III-C, IV; benchmark implementation details
RoboLab \mbox{\citep{survey_ax_2604_09860}} & \mbox{\LiteraturePaperLink{https://www.roboticsproceedings.org/rss22/p096.html}\hspace{5pt}\LiteratureCode{https://github.com/NVlabs/RoboLab}} & \mbox{\ComparisonInputSlot{sim}} & \mbox{Robustness}; \mbox{Control} & \mbox{Action} & \mbox{Isaac Sim} \\[0pt]
% Source: https://arxiv.org/html/2603.22435v2 | Sections 3.1-3.3, 4; Appendices I-K
\rowcolor{black!3}
CaP-X \mbox{\citep{survey_ax_2603_22435}} & \mbox{\LiteraturePaperLink{https://proceedings.mlr.press/v306/fu26u.html}\hspace{5pt}{\setbox0=\hbox{\faGithub}\makebox[\wd0][c]{-}}} & \mbox{\ComparisonInputSlot{sim}\hspace{2pt}\ComparisonInputSlot{real}} & \mbox{Coding} & \mbox{Code} & \mbox{robosuite}; \mbox{LIBERO-PRO}; \mbox{BEHAVIOR} \\[0pt]
% Source: https://arxiv.org/html/2604.25788v2 | Sections IV-VI; Appendix -C baseline implementation
KinDER \mbox{\citep{survey_ax_2604_25788}} & \mbox{\LiteraturePaperLink{https://www.roboticsproceedings.org/rss22/p184.html}\hspace{5pt}\LiteratureCode{https://github.com/Princeton-Robot-Planning-and-Learning/kindergarden}} & \mbox{\ComparisonInputSlot{sim}} & \mbox{Reasoning} & \mbox{Action}; \mbox{Call} & \mbox{Pymunk}; \mbox{PyBullet}; \mbox{MuJoCo} \\[0pt]
% Source: https://arxiv.org/html/2605.10921v1 | Sections 3.1, 3.4, 4, 5.1-5.2; Appendices 8-9
\rowcolor{black!3}
RoboMemArena \mbox{\citep{survey_ax_2605_10921}} & \mbox{\LiteraturePaperLink{https://arxiv.org/abs/2605.10921}\hspace{5pt}\LiteratureCode{https://github.com/OpenHelix-Team/RoboMemArena}} & \mbox{\ComparisonInputSlot{sim}\hspace{2pt}\ComparisonInputSlot{real}} & \mbox{Memory} & \mbox{Action}; \mbox{Call} & \mbox{NR} \\[0pt]
% Source: https://arxiv.org/html/2605.26637v1 | Section 3.2; Appendices F-G
EmbodiedTool \mbox{\citep{survey_ax_2605_26637}} & \mbox{\LiteraturePaperLink{https://arxiv.org/abs/2605.26637}\hspace{5pt}{\setbox0=\hbox{\faGithub}\makebox[\wd0][c]{-}}} & \mbox{\ComparisonInputSlot{dataset}} & \mbox{Tools} & \mbox{Answer}; \mbox{Call} & \mbox{N/A} \\[0pt]
% Source: https://arxiv.org/html/2607.22014v1 | Sections 3.1-3.4; Appendix simulation setup
\rowcolor{black!3}
MissionBench \mbox{\citep{survey_ax_2607_22014}} & \mbox{\LiteraturePaperLink{https://arxiv.org/abs/2607.22014}\hspace{5pt}\LiteratureCode{https://github.com/FraunhoferIVI/MissionBench}} & \mbox{\ComparisonInputSlot{sim}} & \mbox{Planning}; \mbox{Perception} & \mbox{Answer}; \mbox{Action} & \mbox{Cosys-AirSim}; \mbox{UE5} \\[0pt]
% Source: https://arxiv.org/html/2608.08036v1 | Task-State Horizon Compiler; Experiments: Setup and Closed-loop evaluation; backend-validation appendix
RoboGraph \mbox{\citep{survey_ax_2608_08036}} & \mbox{\LiteraturePaperLink{https://arxiv.org/abs/2608.08036}\hspace{5pt}{\setbox0=\hbox{\faGithub}\makebox[\wd0][c]{-}}} & \mbox{\ComparisonInputSlot{sim}} & \mbox{Memory} & \mbox{Call} & \mbox{RoboTwin}; \mbox{RoboCasa} \\[0pt]
% Source: https://arxiv.org/html/2609.08292v2 | Sections 3.1-3.2, 4.1; Implementation Details
\rowcolor{black!3}
EvoNav-Bench \mbox{\citep{survey_ax_2609_08292}} & \mbox{\LiteraturePaperLink{https://arxiv.org/abs/2609.08292}\hspace{5pt}{\setbox0=\hbox{\faGithub}\makebox[\wd0][c]{-}}} & \mbox{\ComparisonInputSlot{sim}} & \mbox{Memory}; \mbox{Robustness} & \mbox{Action} & \mbox{AI2-THOR} \\[0pt]
% Source: https://arxiv.org/html/2609.10895v1 | Sections 3.1, 4-5
ReactHuman \mbox{\citep{survey_ax_2609_10895}} & \mbox{\LiteraturePaperLink{https://arxiv.org/abs/2609.10895}\hspace{5pt}{\setbox0=\hbox{\faGithub}\makebox[\wd0][c]{-}}} & \mbox{\ComparisonInputSlot{sim}} & \mbox{Reasoning}; \mbox{Safety} & \mbox{Action} & \mbox{Genesis} \\[0pt]
% Source: https://arxiv.org/html/2609.19554v1 | Sections 3.1-3.5, 4; Appendix C protocol and controller details
\rowcolor{black!3}
VABench \mbox{\citep{survey_ax_2609_19554}} & \mbox{\LiteraturePaperLink{https://arxiv.org/abs/2609.19554}\hspace{5pt}{\setbox0=\hbox{\faGithub}\makebox[\wd0][c]{-}}} & \mbox{\ComparisonInputSlot{sim}} & \mbox{Perception}; \mbox{Control} & \mbox{Action} & \mbox{NR} \\[0pt]
% Source: https://arxiv.org/html/2609.24124v2 | Section 2: ActiveArena-Sim; Section 3; Appendix D.4.4
ActiveArena \mbox{\citep{survey_ax_2609_24124}} & \mbox{\LiteraturePaperLink{https://arxiv.org/abs/2609.24124}\hspace{5pt}\LiteratureCode{https://github.com/leeibo/ActiveArena}} & \mbox{\ComparisonInputSlot{sim}} & \mbox{Perception}; \mbox{Memory} & \mbox{Action}; \mbox{Call} & \mbox{RoboTwin 2.0} \\[0pt]
% Source: https://arxiv.org/html/2609.27308v2 | Sections 2.1-2.2, 3; Appendix B implementation
\rowcolor{black!3}
EmbodiedSWE \mbox{\citep{survey_ax_2609_27308}} & \mbox{\LiteraturePaperLink{https://arxiv.org/abs/2609.27308v2}\hspace{5pt}\LiteratureCode{https://github.com/EmbodiedSWE/EmbodiedSWE}} & \mbox{\ComparisonInputSlot{sim}} & \mbox{Coding}; \mbox{Control} & \mbox{Code} & \mbox{Isaac Lab} \\[0pt]
% Source: https://arxiv.org/html/2610.00854v2 | Sections 3.1-3.3, 4; Appendices A, B.1-B.3
Embodied Agent Arena \mbox{\citep{survey_ax_2610_00854}} & \mbox{\LiteraturePaperLink{https://arxiv.org/abs/2610.00854}\hspace{5pt}\LiteratureCode{https://github.com/embodied-agent-arena/embodied-agent-arena}} & \mbox{\ComparisonInputSlot{dataset}\hspace{2pt}\ComparisonInputSlot{sim}} & \mbox{Reasoning}; \mbox{Planning} & \mbox{Action}; \mbox{Call}; \mbox{Code} & \mbox{Multiple} \\[0pt]
% Source: https://arxiv.org/html/2607.04434v3 | Sections 3-4; Appendix F.1
\rowcolor{black!3}
RoboDojo \mbox{\citep{survey_ax_2607_04434}} & \mbox{\LiteraturePaperLink{https://arxiv.org/abs/2607.04434}\hspace{5pt}\LiteratureCode{https://github.com/RoboDojo-Benchmark/RoboDojo}} & \mbox{\ComparisonInputSlot{sim}\hspace{2pt}\ComparisonInputSlot{real}} & \mbox{Control}; \mbox{Robustness} & \mbox{Action} & \mbox{Isaac Lab}; \mbox{Isaac Sim} \\[0pt]
% Source: https://arxiv.org/html/2608.17386v1 | Sections 3, 4.1-4.3, 5; Appendix C
ManiGuard \mbox{\citep{survey_ax_2608_17386}} & \mbox{\LiteraturePaperLink{https://arxiv.org/abs/2608.17386}\hspace{5pt}{\setbox0=\hbox{\faGithub}\makebox[\wd0][c]{-}}} & \mbox{\ComparisonInputSlot{sim}} & \mbox{Safety} & \mbox{Action} & \mbox{Isaac Sim}; \mbox{OmniGibson} \\[0pt]
% Source: https://arxiv.org/html/2609.07047v1 | Sections 3.2-3.3, 4, 5.1; Appendices A, E
\rowcolor{black!3}
MEMOBench \mbox{\citep{survey_ax_2609_07047}} & \mbox{\LiteraturePaperLink{https://arxiv.org/abs/2609.07047}\hspace{5pt}\LiteratureCode{https://github.com/Collab-Gen/MEMOBench}} & \mbox{\ComparisonInputSlot{sim}} & \mbox{Memory} & \mbox{Action} & \mbox{LIBERO}; \mbox{robosuite} \\[0pt]
% Source: https://arxiv.org/html/2610.10409v1 | Sections 3.1-3.3, 4.1-4.2; Appendices A-D
RobotWorld \mbox{\citep{survey_ax_2610_10409}} & \mbox{\LiteraturePaperLink{https://arxiv.org/abs/2610.10409v1}\hspace{5pt}{\setbox0=\hbox{\faGithub}\makebox[\wd0][c]{-}}} & \mbox{\ComparisonInputSlot{sim}} & \mbox{Planning}; \mbox{Control} & \mbox{Action}; \mbox{Code} & \mbox{Multiple} \\[0pt]
% Source: 2605.18727v3, Sections 3.1--3.3 and 4.2--4.5.
DexHoldem \mbox{\citep{survey_ax_2605_18727}} & \LiteratureLinks{https://arxiv.org/abs/2605.18727v3}{https://github.com/DexHoldem/Dexholdem-Policy} & \mbox{\ComparisonInputSlot{real}\hspace{2pt}\ComparisonInputSlot{dataset}} & \mbox{Control}; \mbox{Perception} & \mbox{Action}; \mbox{Call}; \mbox{Answer} & \mbox{N/A} \\[0pt]
% Source: 2606.00828v1, Sections 4 and 6.1, Appendix B.
\rowcolor{black!3}
RoboStressBench \mbox{\citep{survey_ax_2606_00828}} & \LiteratureLinks{https://arxiv.org/abs/2606.00828v1}{https://github.com/YUEVII/RoboStressBench} & \mbox{\ComparisonInputSlot{dataset}} & \mbox{Perception}; \mbox{Robustness} & \mbox{Answer} & \mbox{N/A} \\[0pt]
% Source: 2610.27039v2, Sections 3.1--3.2 and 5--6.
Embodied Playground \mbox{\citep{survey_ax_2610_27039}} & \LiteratureLinks{https://arxiv.org/abs/2610.27039v2}{https://github.com/Winniechen2002/Embodied-Playground} & \mbox{\ComparisonInputSlot{sim}} & \mbox{Planning}; \mbox{Control} & \mbox{Code} & \mbox{MuJoCo}; \mbox{Isaac Lab}; \mbox{ManiSkill} \\[0pt]
% Source: 2610.34649v1, Sections 2.1--2.3 and 3.4.
\rowcolor{black!3}
RAB~\mbox{\citep{survey_ax_2610_34649}} & \LiteratureLinks{https://arxiv.org/abs/2610.34649v1}{https://github.com/AetherLabsAI/RobotAgentBench} & \mbox{\ComparisonInputSlot{sim}\hspace{2pt}\ComparisonInputSlot{real}} & \mbox{Planning}; \mbox{Control} & \mbox{Action}; \mbox{Code} & \mbox{Isaac Sim} \\[0pt]
% Source: 2610.35385v2, Sections 3.2--3.4.
EmbodiedLab \mbox{\citep{survey_ax_2610_35385}} & \LiteratureLinks{https://arxiv.org/abs/2610.35385v2}{https://github.com/HexaFuture/EmbodiedLab} & \mbox{\ComparisonInputSlot{sim}} & \mbox{Tools}; \mbox{Reasoning} & \mbox{Code}; \mbox{Answer} & \mbox{Isaac Sim} \\[0pt]
% Source: 2610.34527v3, Sections 4--6 and Appendix C.
\rowcolor{black!3}
LLM4Robot-Bench \mbox{\citep{survey_ax_2610_34527}} & \LiteratureLinks{https://arxiv.org/abs/2610.34527v3}{https://github.com/LLM4Robot/LLM4Robot-Bench} & \mbox{\ComparisonInputSlot{sim}\hspace{2pt}\ComparisonInputSlot{real}} & \mbox{Control} & \mbox{Action}; \mbox{Call} & \mbox{Isaac Sim} \\[0pt]
% Source: 2611.12177v1, Sections 2.1--2.3 and 3.2.
EFE~\mbox{\citep{project_embodied_first_exam}} & \LiteratureLinks{https://arxiv.org/abs/2611.12177v1}{https://github.com/mll-lab-nu/embodied-first-exam} & \mbox{\ComparisonInputSlot{sim}} & \mbox{Control}; \mbox{Coding} & \mbox{Action}; \mbox{Code} & \mbox{Multiple} \\[0pt]
% Source: RoboWild report 40a24c8, Sections 2.2--3.1 and Appendix E.
\rowcolor{black!3}
RoboWild \mbox{\citep{xlanglab2026robowild}} & \LiteratureLinks{https://huggingface.co/datasets/xlangai/RoboWild/blob/40a24c8f22fe567da47de49a8cddba7c51ae5560/RoboWild.pdf}{https://github.com/xlang-ai/RoboWild} & \mbox{\ComparisonInputSlot{sim}} & \mbox{Planning}; \mbox{Control} & \mbox{Call}; \mbox{Code} & \mbox{Isaac Sim}; \mbox{Isaac Lab} \\[0pt]
% Source: 2611.31002v1, Sections 2.2 and 3.1, Appendix B.1.
RoboUniverse-Verified \mbox{\citep{survey_ax_2611_31002}} & \LiteratureLinks{https://arxiv.org/abs/2611.31002v1}{https://github.com/Einsia/RoboUniverse-10K} & \mbox{\ComparisonInputSlot{sim}} & \mbox{Control} & \mbox{Code}; \mbox{Action} & \mbox{Isaac Sim} \\[0pt]
% Source: 2610.40565v1, Sections 4, 6.1--6.2 and Appendix E.
\rowcolor{black!3}
RoboWorld~\mbox{\citep{survey_ax_2610_40565}} & \LiteratureLinks{https://arxiv.org/abs/2610.40565v1}{https://github.com/gym-anything/robotics} & \mbox{\ComparisonInputSlot{sim}} & \mbox{Control} & \mbox{Action}; \mbox{Call} & \mbox{Multiple} \\[0pt]
% Source: 2611.14115v1, Sections 3.2--3.3, 5.1, and Supplement B.
RoboRisk~\mbox{\citep{survey_ax_2611_14115}} & \LiteratureLinks{https://arxiv.org/abs/2611.14115v1}{https://github.com/zhuqinfeng1999/LEASH} & \mbox{\ComparisonInputSlot{sim}\hspace{2pt}\ComparisonInputSlot{real}} & \mbox{Safety} & \mbox{Call}; \mbox{Action} & \mbox{MuJoCo} \\[0pt]
% Source: 2611.37041v1, Sections 3.1--3.4 and 4.3.
\rowcolor{black!3}
RoboMite~\mbox{\citep{survey_ax_2611_37041}} & \LiteratureLinks{https://arxiv.org/abs/2611.37041v1}{} & \mbox{\ComparisonInputSlot{sim}} & \mbox{Memory}; \mbox{Reasoning} & \mbox{Call} & \mbox{MuJoCo} \\[0pt]
% Source: 2611.08209v2, Sections 3.1--3.2 and 4.1.
RoboARCade~\mbox{\citep{survey_ax_2611_08209}} & \LiteratureLinks{https://arxiv.org/abs/2611.08209v2}{https://github.com/Satvik1701/RoboARCade} & \mbox{\ComparisonInputSlot{sim}} & \mbox{Perception}; \mbox{Control} & \mbox{Action}; \mbox{Call} & \mbox{MuJoCo}; \mbox{Superdex} \\[0pt]
% Source: 2611.33390v1, Sections 3.2--3.3, 4.1 and 4.3.
\rowcolor{black!3}
RoboARC~\mbox{\citep{survey_ax_2611_33390}} & \LiteratureLinks{https://arxiv.org/abs/2611.33390v1}{https://github.com/mll-lab-nu/RoboARC} & \mbox{\ComparisonInputSlot{sim}\hspace{2pt}\ComparisonInputSlot{real}} & \mbox{Reasoning}; \mbox{Control} & \mbox{Action} & \mbox{MuJoCo} \\[0pt]
% Source: 2611.22138v2, Sections 3.1--3.4.
RoboTeam~\mbox{\citep{survey_ax_2611_22138}} & \LiteratureLinks{https://arxiv.org/abs/2611.22138v2}{https://github.com/changxuanfan/RoboTeam} & \mbox{\ComparisonInputSlot{sim}} & \mbox{Collab.}; \mbox{Control} & \mbox{Call} & \mbox{MuJoCo} \\[0pt]
% Source: 2611.26444v1, Sections 3.2--3.3 and 4.1--4.3.
\rowcolor{black!3}
Mind2Act World~\mbox{\citep{survey_ax_2611_26444}} & \LiteratureLinks{https://arxiv.org/abs/2611.26444v1}{https://github.com/Mind2Act/Mind2Act_World} & \mbox{\ComparisonInputSlot{sim}} & \mbox{Memory}; \mbox{Reasoning}; \mbox{Control} & \mbox{Action}; \mbox{Call} & \mbox{Isaac Sim} \\[0pt]
\addlinespace[3pt]
\rowcolor{TSTEnvironmentLightest}\multicolumn{6}{@{}l}{\textbf{Systems}} \\*[3pt]
% Source: https://arxiv.org/html/2607.00272v1 | Sections 2, 3.1-3.3, 3.6; Appendices B-D
ASPIRE \mbox{\citep{survey_ax_2607_00272}} & \mbox{\LiteraturePaperLink{https://arxiv.org/abs/2607.00272}\hspace{5pt}\LiteratureCode{https://github.com/NVlabs/ASPIRE}} & \mbox{\ComparisonInputSlot{sim}\hspace{2pt}\ComparisonInputSlot{real}} & \mbox{Coding}; \mbox{Transfer} & \mbox{Code} & \mbox{CaP-X} \\*[0pt]
% Source: https://arxiv.org/html/2607.08448v5 | Sections 2.1-2.3, 3; Appendices A-D
\rowcolor{black!3}
Harness VLA \mbox{\citep{survey_ax_2607_08448}} & \mbox{\LiteraturePaperLink{https://arxiv.org/abs/2607.08448}\hspace{5pt}\LiteratureCode{https://github.com/RLinf/RPent}} & \mbox{\ComparisonInputSlot{sim}} & \mbox{Memory}; \mbox{Robustness} & \mbox{Call} & \mbox{robosuite}; \mbox{RoboCasa}; \mbox{RoboTwin} \\*[0pt]
% Source: https://arxiv.org/html/2607.18060v3 | Sections 4.3-4.4, 5; Appendices C.1-C.3, D.1-D.4
RoboHarness \mbox{\citep{survey_ax_2607_18060}} & \mbox{\LiteraturePaperLink{https://arxiv.org/abs/2607.18060v3}\hspace{5pt}{\setbox0=\hbox{\faGithub}\makebox[\wd0][c]{-}}} & \mbox{\ComparisonInputSlot{sim}\hspace{2pt}\ComparisonInputSlot{real}} & \mbox{Planning}; \mbox{Robustness} & \mbox{Call} & \mbox{LIBERO}; \mbox{RoboTwin} \\*[0pt]
% Source: https://arxiv.org/html/2609.08402v1 | AGOS-Bench: Simulator Platform and Action Space; AGOS-Agent; Appendices A.1, B prompts
\rowcolor{black!3}
AGOS-Agent \mbox{\citep{survey_ax_2609_08402}} & \mbox{\LiteraturePaperLink{https://arxiv.org/abs/2609.08402}\hspace{5pt}{\setbox0=\hbox{\faGithub}\makebox[\wd0][c]{-}}} & \mbox{\ComparisonInputSlot{sim}} & \mbox{Planning}; \mbox{Collab.} & \mbox{Call} & \mbox{CARLA-Air} \\[0pt]
% Source: 2610.22471v2, Sections 2--4 and Appendix D.
SenseNova study~\mbox{\citep{survey_ax_2610_22471}} & \LiteratureLinks{https://arxiv.org/abs/2610.22471v2}{https://github.com/OpenSenseNova/SenseNova-RoboRSI} & \mbox{\ComparisonInputSlot{sim}\hspace{2pt}\ComparisonInputSlot{real}} & \mbox{Control}; \mbox{Planning} & \mbox{Action}; \mbox{Call}; \mbox{Code} & \mbox{Multiple} \\[0pt]
% Source: 2610.35832v1, Sections 4.1--4.3 and Appendix B.2.
\rowcolor{black!3}
EBench~\mbox{\citep{survey_ax_2610_35832}} & \LiteratureLinks{https://arxiv.org/abs/2610.35832v1}{} & \mbox{\ComparisonInputSlot{sim}} & \mbox{Control}; \mbox{Planning} & \mbox{Action}; \mbox{Call} & \mbox{Isaac Sim} \\[0pt]
\addlinespace[3pt]
\rowcolor{TSTEnvironmentLightest}\multicolumn{6}{@{}l}{\textbf{Evaluators}} \\*[3pt]
% Source: https://arxiv.org/html/2604.05226v1 | Sections III-IV; Appendices Code Generation, Environment Configuration and Model Training Details
RoboPlayground \mbox{\citep{survey_ax_2604_05226}} & \mbox{\LiteraturePaperLink{https://arxiv.org/abs/2604.05226}\hspace{5pt}{\setbox0=\hbox{\faGithub}\makebox[\wd0][c]{-}}} & \mbox{\ComparisonInputSlot{sim}} & \mbox{Robustness} & \mbox{Action} & \mbox{MuJoCo} \\*[0pt]
% Source: https://arxiv.org/html/2602.01640v1 | Sections 3.1-3.2, 4.1; evaluation prompt appendix
\rowcolor{black!3}
A2Eval \mbox{\citep{survey_ax_2602_01640}} & \mbox{\LiteraturePaperLink{https://arxiv.org/abs/2602.01640}\hspace{5pt}{\setbox0=\hbox{\faGithub}\makebox[\wd0][c]{-}}} & \mbox{\ComparisonInputSlot{dataset}} & \mbox{Perception}; \mbox{Reasoning} & \mbox{Answer} & \mbox{N/A} \\*[0pt]
% Source: https://proceedings.iclr.cc/paper_files/paper/2026/file/8833c8aa10542d24d693bbaf6a4598f5-Paper-Conference.pdf | Sections 3.1-3.4, 4.1.1; Appendix A.1.1
ManipEvalAgent \mbox{\citep{survey_iclr_2026_8833c8aa}} & \mbox{\LiteraturePaperLink{https://proceedings.iclr.cc/paper_files/paper/2026/hash/8833c8aa10542d24d693bbaf6a4598f5-Abstract-Conference.html}\hspace{5pt}{\setbox0=\hbox{\faGithub}\makebox[\wd0][c]{-}}} & \mbox{\ComparisonInputSlot{sim}} & \mbox{Robustness} & \mbox{Action} & \mbox{LIBERO}; \mbox{RoboTwin} \\*[0pt]
% Source: https://arxiv.org/html/2609.13082v1 | Sections 3-5; interactive task and action-interface appendix
\rowcolor{black!3}
Embodied-BenchForge \mbox{\citep{survey_ax_2609_13082}} & \mbox{\LiteraturePaperLink{https://arxiv.org/abs/2609.13082}\hspace{5pt}{\setbox0=\hbox{\faGithub}\makebox[\wd0][c]{-}}} & \mbox{\ComparisonInputSlot{dataset}\hspace{2pt}\ComparisonInputSlot{sim}} & \mbox{Reasoning}; \mbox{Planning} & \mbox{Answer}; \mbox{Call} & \mbox{AI2-THOR} \\[0pt]
\end{longtable}
\endgroup

\par\Needspace{12\baselineskip}
\begingroup
\mechanismtableglue
\CompactTableStyle
\setlength{\tabcolsep}{3pt}
\setlength{\aboverulesep}{1.2pt}
\setlength{\belowrulesep}{1.5pt}
\setlength{\LiteratureWidth}{\dimexpr\linewidth-10\tabcolsep\relax}
\setlength{\LTcapwidth}{\linewidth}
\captionsetup{singlelinecheck=false,justification=raggedright}
\LiteratureContinuationSetup
\begin{longtable}{@{}>{\RaggedRight\arraybackslash}p{.225\LiteratureWidth}>{\RaggedRight\arraybackslash}p{.184\LiteratureWidth}>{\RaggedRight\arraybackslash}p{.175\LiteratureWidth}>{\RaggedRight\arraybackslash}p{.231\LiteratureWidth}>{\centering\arraybackslash}p{.0925\LiteratureWidth}>{\centering\arraybackslash}p{.0925\LiteratureWidth}@{}}
\caption[How evaluation protocols change reported task success]{Task success under different evaluation protocols for the same model and tasks within each study.}\label{tab:benchmark-protocols}\\
\LiteratureTopRule{Environment}
\rowcolor{TSTEnvironmentLight}\textbf{Study} & \textbf{Model} & \textbf{Tasks} & \textbf{Protocol} & \multicolumn{2}{c@{}}{\textbf{Success (\%)}} \\
\midrule
\endfirsthead
\multicolumn{6}{@{}l}{\tablename~\thetable\ (continued)} \\
\LiteratureTopRule{Environment}
\rowcolor{TSTEnvironmentLight}\textbf{Study} & \textbf{Model} & \textbf{Tasks} & \textbf{Protocol} & \multicolumn{2}{c@{}}{\textbf{Success (\%)}} \\
\midrule
\endhead
\midrule
\multicolumn{6}{r@{}}{Continued on next page} \\
\endfoot
\bottomrule
\endlastfoot
% RoboWild, Figure 4, Table 3 and Appendix E.5.
\textbf{RoboWild}~\citep{xlanglab2026robowild} & Claude Opus 5.5 (max) & 239 Dexmate setups & ReAct (one motion call) & \multicolumn{2}{c@{}}{6.28} \\*
 & & RGB observations & Claude Code (two calls) & \multicolumn{2}{c@{}}{12.13} \\
\midrule[.25pt]
% EBench, Section 6.2 and Appendix B.2. Both cutoffs score the same policy runs.
\textbf{EBench}~\citep{survey_ax_2610_35832} & GPT-6 Astra (high) & 26 tasks & & \multicolumn{2}{c@{}}{\emph{Bottle and Shop cutoff}} \\*
 & & 510 episodes & \emph{Control interface} & \textbf{3,000 steps} & \textbf{5,000 steps} \\*
 & & Equal task weights & Direct waypoints & 46.7 & \emph{Not run} \\*
 & & & Waypoints + $\pi_{0.5}$ & 52.4 & 53.0 \\*
 & & & Waypoints + InternW0-$\Delta$ & 55.8 & 56.0 \\
\midrule[.25pt]
% EFE, Section 3.2 and Appendix G, Table 5.
\textbf{EFE}~\citep{project_embodied_first_exam} & GPT-6 Astra in Codex & 391 core tasks & Standard interface & \multicolumn{2}{c@{}}{28} \\*
 & & & Add exact object states & \multicolumn{2}{c@{}}{45} \\*
 & & & Save/load (1,000-reset limit) & \multicolumn{2}{c@{}}{53} \\*
 & & & Add both & \multicolumn{2}{c@{}}{68} \\
\midrule[.25pt]
% RoboARC, Section 5.1 and Appendix E.4, Table 13; three runs per puzzle and condition.
\textbf{RoboARC}~\citep{survey_ax_2611_33390} & GPT-6 Astra in Codex (high) & 15 selected rigid puzzles & Standard, no resets & \multicolumn{2}{c@{}}{24.4} \\*
 & & 3 runs per puzzle & Add object poses & \multicolumn{2}{c@{}}{28.9} \\*
 & & & Add part geometry & \multicolumn{2}{c@{}}{37.8} \\*
 & & & Add both & \multicolumn{2}{c@{}}{60.0} \\
\midrule[.25pt]
% CaP-X, Table 1 and Figure 17, tiers S2 and M1; each cell uses 100 trials.
\textbf{CaP-Bench}~\citep{survey_ax_2603_22435} & GPT-5.2 & Cube Stack & Submit one program & \multicolumn{2}{c@{}}{45} \\*
 & & 100 trials per protocol & Revise after feedback & \multicolumn{2}{c@{}}{69} \\*
 & & Cube Lift & Submit one program & \multicolumn{2}{c@{}}{87} \\*
 & & 100 trials per protocol & Revise after feedback & \multicolumn{2}{c@{}}{87} \\
\midrule[.25pt]
% Embodied Playground, Sections 6.1--6.2 and Table 5.
\textbf{Embodied Playground}~\citep{survey_ax_2610_27039} & GPT-6 Astra in Codex & 100 tasks & First valid attempt & \multicolumn{2}{c@{}}{13.80} \\*
 & & 4 conditions per task & Within two valid attempts & \multicolumn{2}{c@{}}{19.94} \\
\end{longtable}
\endgroup

\section{Harness: How Should Robot Capabilities Be Orchestrated, Recovered, and Evolved?}
\label{sec:harness}

\begin{figure}[!t]
\centering
\includegraphics[width=\linewidth]{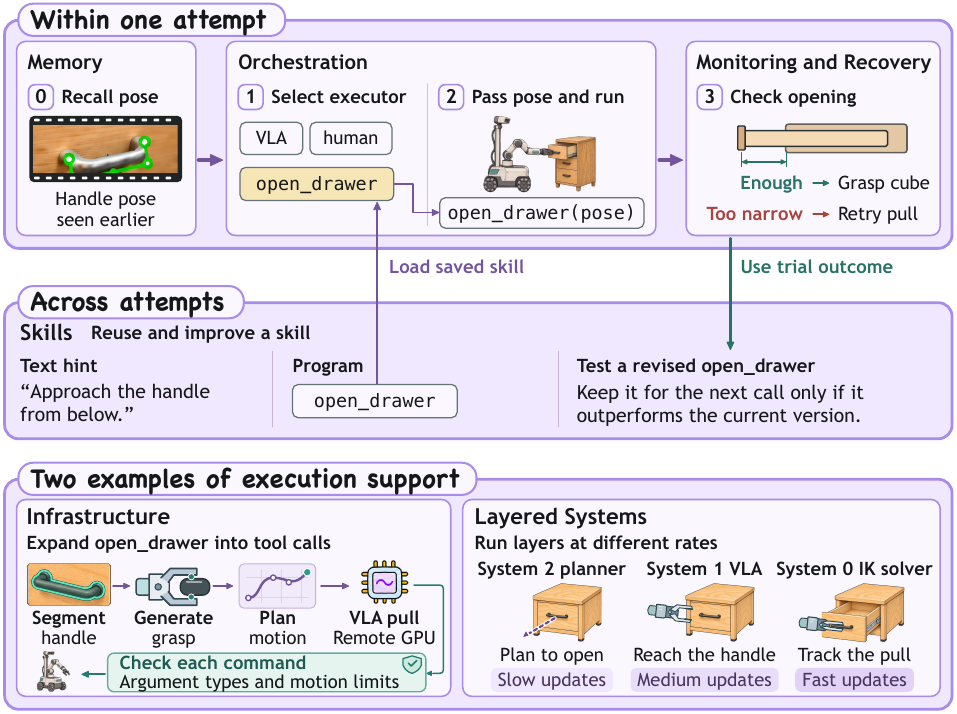}
% \caption{Harness functions around robot execution. The central row follows a task from observations and retrieved context through decision and dispatch to an executor library and robot execution with feedback. Above, Memory supplies relevant history, Orchestration selects executors and governs handoffs, Monitoring assesses progress and completion, and Recovery responds through retries, executor switches, replanning, or resets. Below, Skill Synthesis turns behavior specifications into callable skills, while Evolution revises the connecting software, including orchestration logic, monitors, tools, and context builders.}
\caption{The harness manages how a drawer-opening skill is called, checked, and reused. It retries the pull if the opening is too narrow. It keeps a revised skill for later attempts only if it outperforms the current version. The bottom row shows command checks and control layers running at different rates.}
\label{fig:harness-operation}
\end{figure}

A robot may have capable models and policies yet still fail if they are poorly organized around execution. We use the \emph{harness} to denote the software that manages which components are called, what information they receive, how execution is checked, and what can be reused. Figure~\ref{fig:harness-operation} distinguishes retrying a pull within one attempt from retaining a revised drawer-opening skill for later attempts. During the attempt, \textit{Orchestration} governs calls and handoffs, \textit{Memory} supplies retained information, and \textit{Monitoring and Recovery} assess progress and respond to failure. \textit{Skills} package behavior for reuse across attempts. To support execution, \textit{Infrastructure} makes tools callable and checks commands, and \textit{Layered Systems} divide responsibility among planners, learned policies, and analytic controllers operating at different rates.

\subsection{Orchestration}
\label{sec:orchestration}

Orchestration decides which model, policy, skill, or tool runs, how long it keeps control, and how control passes to the next component.
We illustrate orchestration by following a single handoff order through four steps, where the harness chooses the next executor, defines what the function calling takes and returns, makes sure the next executor can start from the resulting state, and decides when control comes back.

\paragraph{\textbf{Routing and Dispatch.}}
In an agentic harness, the planner rarely acts on the robot directly. Instead, it decides which executor should handle the next step, such as a learned policy, an analytic tool, or a human, and passes the task to it. A key design choice is when this decision is made. VoLo~\citep{survey_ax_2606_07723} makes it during execution. It monitors a running VLA and, when the action goes wrong, halts the VLA and switches to a hand-designed skill guided by perception. MistyPilot~\citep{survey_ax_2608_15549} makes it in advance. It generates persistent bindings between sensor events and skills for social-robot behavior, so that later events trigger skills without a new planning step. AeroWeaver~\citep{survey_ax_2609_18520} makes it at the level of the team, assigning body-specific aerial skills to distributed roles in a simulated swarm. M2HRI~\citep{survey_ax_2604_11975} further includes humans among the parties to route to, adding a coordinator for human participation and an interaction planner conditioned on memory. Together, these systems place the routing decision at different points, from a running action to a standing rule to a team-level role.

Meanwhile, routing is reliable only when the dispatcher knows what each executor can and cannot do. This knowledge can come from a description of the robot's body or from the results of earlier calls. EMOS~\citep{survey_ax_2410_22662} takes the first approach for teams of different robots. It builds a Robot Resume for each robot from its URDF model and forward-kinematics tools, and each robot's agent uses this resume to assess an assigned subtask and report whether it can perform it before calling its skill library. Harness VLA~\citep{survey_ax_2607_08448} and Physical Agency~\citep{survey_ax_2607_21725} take the second approach on a single robot, placing analytic tools and a VLA behind one dispatch interface and choosing the next executor according to the evidence each call returns, such as grasp checks and typed failure messages. Guava~\citep{survey_ax_2606_18363} applies the same idea across levels of control, choosing between semantic operations, such as grasping and alignment, and numerical motion tools based on visual observations, the robot's state, and earlier tool outcomes.

Some decisions concern who controls a particular arm. In RoboTeam~\citep{survey_ax_2611_22138}, the adaptive system lets the lead agent give an arm to a delegate. The delegate inherits the lead's history and returns the arm when its job ends. ACAP~\citep{survey_ax_2611_31205} also lets a lead agent distribute work. Both ACAP and its fixed-delegation baseline can change assignments and timing. Only the ACAP lead can take back the controls. In towel folding, each member secures a grasp, then the lead moves both arms through the fold.

\paragraph{\textbf{Function Protocol.}}
Robot capabilities are usually exposed to the agent as functions, which the agent calls with a name and some arguments without seeing how they work inside. A function protocol fills this gap by stating how a function should be called, what it reports back, and when a call counts as successful. The first of these defines what counts as a valid call, and can be written for the agent to follow or enforced by the harness. EmbodiedTool~\citep{survey_ax_2605_26637} writes it for the agent, giving each tool a typed description of its inputs, outputs, and usage conditions that the agent's call must match. ROSClaw~\citep{survey_ax_2603_26997} enforces it in the harness, checking each call to a physical TurtleBot3 or Go2 robot against velocity limits and a list of allowed interfaces, so that an invalid call is stopped even if the agent produces it.

Once a call has run, the protocol defines what comes back to the agent. A simple success flag says little about the robot's situation, so some protocols return more. Physical Agentic AI~\citep{survey_ax_2608_22657} returns each skill outcome with the robot state at that moment, which later dispatch can use without calling the foundation model again. Physical Agency~\citep{survey_ax_2607_21725} returns typed failures, so that the agent can respond to each kind differently, for example by setting up an operation again or switching to another executor.

Even a detailed result is still the executor's own report. Success conditions add an independent check by stating what should be true after a call succeeds. Thea~\citep{survey_ax_2608_11246} attaches such conditions to each tool, and a task evaluator separate from the planner checks them against current observations.

\paragraph{\textbf{State Connection.}}
Each executor works reliably only from certain starting states, so a correctly chosen executor can still fail if the previous step leaves the robot elsewhere. 
State connection makes sure that the state left by one executor connects to the state the next one needs.
Existing systems make this connection in four ways. The first two ways adjust either the state or the executor. RoboHarness~\citep{survey_ax_2607_18060} adjusts the state to fit the executor. It retrieves past trajectories from memory and uses a motion-planned memory bridge to move the robot to the starting state expected by the next policy. RoboRouter~\citep{survey_ax_2603_07892} adjusts the executor to fit the state. It selects a policy that suits the current state from several candidates, and uses summaries of past runs to improve later selections.

The other two ways set up the connection before the handoff, which is common when several robots share a task. In this case, the next robot must also know what the others have done and which resources it may use. Some systems check before each handoff whether the next robot is ready. RoCo~\citep{survey_ax_2307_04738} lets robots agree on a plan through dialogue, then checks whether the plan is feasible before they move.
RoboTalk~\citep{survey_ax_2609_23997} checks each message and joint action against symbolic preconditions and resource ownership, so that a robot acts only when what it was told and what it may use are consistent. Other systems fix the state connection structure in advance. AGOS-Agent~\citep{survey_ax_2609_08402} divides a task into search for candidates, route planning on the ground, and verification from ground-level views, so that the input each stage receives is set at design time. Together, these approaches either repair a mismatch when it appears or prevent it by checking or fixing the handoff ahead of time.

% \paragraph{\textbf{Successor state.}}
% A well-chosen executor can still fail if the previous action leaves the robot in a state it cannot handle. RoboHarness~\citep{survey_ax_2607_18060} fixes this by changing the state, using a motion-planned Memory Bridge built from past trajectories to move the robot to the starting state the next policy expects. RoboRouter~\citep{survey_ax_2603_07892} instead changes the executor, choosing among several policies based on summaries of past executions. Either way, the system must check whether the robot's current state suits the next policy.

% With several robots, this state also includes what the other robots are doing, so a robot must understand incoming messages and know which resources it may use. RoCo~\citep{survey_ax_2307_04738} reaches this agreement through dialogue and feasibility checks, AGOS-Agent~\citep{survey_ax_2609_08402} fixes the division of work in advance, and RoboTalk~\citep{survey_ax_2609_23997} checks each message and joint action against symbolic preconditions and resource ownership. However, these systems are tested in simplified settings, such as a robot paired with a human (RoCo) or known maps with perfect communication (AGOS-Agent). An open problem is to keep these coordination rules when the shared state comes from real, noisy execution results.

\paragraph{\textbf{Handoff Timing.}}
Handoff timing decides when control returns from an executor to the decision process. Returning control too late lets a failing action continue, while returning it too early interrupts an action that would have succeeded. Existing systems differ mainly in the signal that triggers this return, which can be task progress, an execution budget, or a periodic or event-driven schedule.
Progress-based triggers return control when the observed state of the task changes. HarnessWAM~\citep{survey_ax_2608_09516} uses a fast progress estimator to detect such changes, and a VLM task manager then decides whether to move to the next subtask, gather new observations, replan, or reset the robot. Cortex~\citep{survey_ax_2607_05377} checks visual progress continuously while the robot moves, and keeps a running semantic memory to decide whether the current stage should continue.

A monitor can also return control to the agent that chose the executor. Robo-COP~\citep{survey_ax_2610_09228} writes code that can call scripted motions or a VLA. A visual monitor can interrupt either kind of call. A judge then checks the outcome and can ask the agent to revise the code.

A handoff can also occur while a skill is still running. In DynaHarness~\citep{survey_ax_2609_40306}, the VLM names a skill and its target. The scheduler locates the target and can reject a request it cannot resolve. It checks at 2\,Hz while the controller runs at 20\,Hz and can interrupt the skill to switch executors or request a new plan. A comparison kept the same skill library and allowed replanning after each successful skill using the updated observation. It disabled replanning after failure, skill substitution or reordering, recovery steps, and branching on a verifier's result. On the same 800 LIBERO-PRO development episodes, success was 74.0\% with the full scheduler and 63.9\% with these responses disabled.

Other systems limit each call in time or coordinate components that run at different rates. VLAs-as-Tools~\citep{survey_ax_2605_13119} runs the VLA as a tool with a limited execution budget, during which the VLA reports progress to an outside monitor that can move on or replan. Retriever~\citep{survey_ax_2607_17213} runs perception, belief updates, planning, the VLA, and control asynchronously, with separate periodic or event-driven clocks and explicit input-synchronization and commit rules that preserve causality. Generally, these triggers of handoff timing differ in what the harness waits for before taking control back, whether an observed change in progress, the end of a budget, or the next tick of a clock. Related works are compared in Table~\ref{tab:harness-orchestration}.

% \paragraph{\textbf{Timing the handoff.}}
% Once the right executor is running from a suitable state, the harness must decide when to take control back. Systems differ mainly in the signal that triggers this return. HarnessWAM~\citep{survey_ax_2608_09516} uses a fast progress estimator, with a VLM task manager deciding whether to move on, replan, or reset. Cortex~\citep{survey_ax_2607_05377} checks visual progress continuously, and VLAs-as-Tools~\citep{survey_ax_2605_13119} gives the VLA a limited budget and reports its progress to an outside monitor. Retriever~\citep{survey_ax_2607_17213} instead keeps perception, planning, the VLA, and control on a shared clock. Checking more often helps only if the observations are recent enough to act on. Otherwise, the harness may interrupt motions that have not finished.

% Revised executor-type matrix.
% Orchestration comparison, styled after the Data annotation table.
\par\Needspace{.6\textheight}
\begingroup
\mechanismtableglue
% Match the reference survey: Dingbats check/cross, green #009900 and red #B20000.
\DeclareRobustCommand{\Y}{\textcolor[HTML]{009900}{{\fontencoding{U}\fontfamily{pzd}\fontseries{m}\fontshape{n}\selectfont\char51}}}
\DeclareRobustCommand{\N}{\textcolor[HTML]{B20000}{{\fontencoding{U}\fontfamily{pzd}\fontseries{m}\fontshape{n}\selectfont\char55}}}
\newcommand{\OrchGroup}[1]{\rowcolor{TSTHarnessLight}\multicolumn{7}{l}{\textbf{#1}} \\*}
\CompactTableStyle\hyphenpenalty=10000
\renewcommand{\arraystretch}{1.18}
\setlength{\LiteratureWidth}{\dimexpr\linewidth-14\tabcolsep-.6pt\relax}
\setlength{\LTleft}{0pt}
\setlength{\LTright}{0pt}
\setlength{\LTpre}{3pt}
\setlength{\LTpost}{4pt}
\setlength{\LTcapwidth}{\linewidth}
\begin{longtable}{>{\RaggedRight\arraybackslash}p{0.26\LiteratureWidth}>{\centering\arraybackslash}p{0.08\LiteratureWidth}!{\color{black!30}\vrule width .3pt}>{\centering\arraybackslash}p{0.10\LiteratureWidth}>{\centering\arraybackslash}p{0.10\LiteratureWidth}>{\centering\arraybackslash}p{0.10\LiteratureWidth}>{\centering\arraybackslash}p{0.155\LiteratureWidth}!{\color{black!30}\vrule width .3pt}>{\RaggedRight\arraybackslash}p{0.205\LiteratureWidth}}
\caption[Comparison of works on orchestration]{Comparison of works on orchestration (Section~\ref{sec:orchestration}), grouped by the four categories in the text. The executor columns mark which types of component receive control, namely learned policies such as VLAs and WAMs (Policy), callable utilities such as motion planners and controllers (Tool), reusable behaviors from a skill library (Skill), and delegated agents, other robots, or humans (Other Agent). \emph{Selection} shows what decides which executor runs, namely an LLM or VLM agent (Model-based), a model whose calls are checked or triggered by a rule layer (Hybrid), or a preset schedule (Rule-based). Icons link to the paper and, when available, the code.}\label{tab:harness-orchestration}\\
\LiteratureTopRule{Harness}
\rowcolor{TSTHarnessLight}\textbf{Work} & \textbf{Link} & \textbf{Policy} & \textbf{Tool} & \textbf{Skill} & \mbox{\textbf{Other Agent}} & \textbf{Selection} \\
\midrule
\endfirsthead
\multicolumn{7}{l}{\tablename~\thetable\ (continued)} \\
\LiteratureTopRule{Harness}
\rowcolor{TSTHarnessLight}\textbf{Work} & \textbf{Link} & \textbf{Policy} & \textbf{Tool} & \textbf{Skill} & \mbox{\textbf{Other Agent}} & \textbf{Selection} \\
\midrule
\endhead
\midrule
\multicolumn{7}{r}{Continued on next page} \\
\endfoot
\bottomrule
\endlastfoot
\OrchGroup{Routing and Dispatch}
VoLo \citep{survey_ax_2606_07723} & \LiteratureLinks{https://arxiv.org/abs/2606.07723}{https://github.com/NVlabs/VoLoAgent} & \Y & \Y & \N & \N & Model-based \\
\rowcolor{black!3}
MistyPilot \citep{survey_ax_2608_15549} & \LiteratureLinks{https://arxiv.org/abs/2608.15549}{https://github.com/WangXiaoShawn/MistyPilot} & \N & \N & \Y & \N & Hybrid \\
ABot-AgentOS \citep{survey_ax_2607_10350} & \LiteratureLinks{https://arxiv.org/abs/2607.10350}{} & \N & \N & \Y & \N & Model-based \\
\rowcolor{black!3}
AeroWeaver \citep{survey_ax_2609_18520} & \LiteratureLinks{https://arxiv.org/abs/2609.18520}{https://github.com/Admire-ljb/AeroWeaver} & \N & \N & \Y & \Y & Model-based \\
M2HRI \citep{survey_ax_2604_11975} & \LiteratureLinks{https://arxiv.org/abs/2604.11975}{https://github.com/project-m2hri/m2hri} & \N & \N & \Y & \Y & Model-based \\
\rowcolor{black!3}
EMOS \citep{survey_ax_2410_22662} & \LiteratureLinks{https://arxiv.org/abs/2410.22662}{} & \N & \N & \Y & \Y & Model-based \\
Physical Agency \citep{survey_ax_2607_21725} & \LiteratureLinks{https://arxiv.org/abs/2607.21725}{https://github.com/lianegalanti/Pigey} & \Y & \Y & \Y & \N & Model-based \\
\rowcolor{black!3}
Guava \citep{survey_ax_2606_18363} & \LiteratureLinks{https://arxiv.org/abs/2606.18363}{} & \N & \Y & \N & \N & Model-based \\
RoboTeam~\citep{survey_ax_2611_22138} & \LiteratureLinks{https://arxiv.org/abs/2611.22138v2}{https://github.com/changxuanfan/RoboTeam} & \N & \Y & \N & \Y & Model-based \\
\rowcolor{black!3}
ACAP~\citep{survey_ax_2611_31205} & \LiteratureLinks{https://arxiv.org/abs/2611.31205v1}{} & \N & \Y & \N & \Y & Model-based \\
\OrchGroup{Function Protocol}
EmbodiedTool \citep{survey_ax_2605_26637} & \LiteratureLinks{https://arxiv.org/abs/2605.26637}{} & \N & \Y & \N & \N & Model-based \\
\rowcolor{black!3}
ROSClaw \citep{survey_ax_2603_26997} & \LiteratureLinks{https://arxiv.org/abs/2603.26997}{} & \N & \Y & \N & \N & Hybrid \\
Physical Agentic AI \citep{survey_ax_2608_22657} & \LiteratureLinks{https://arxiv.org/abs/2608.22657}{https://github.com/Liuuuxy/physical-agentic-ai} & \N & \N & \Y & \Y & Hybrid \\
\rowcolor{black!3}
EmbodiedSkills \citep{survey_ax_2609_01281} & \LiteratureLinks{https://arxiv.org/abs/2609.01281}{https://github.com/ZJU4EmbodiedAI/EmbodiedSkills} & \Y & \N & \N & \N & Hybrid \\
Thea \citep{survey_ax_2608_11246} & \LiteratureLinks{https://arxiv.org/abs/2608.11246}{https://github.com/EIT-HAI/Thea} & \N & \Y & \N & \N & Model-based \\
\OrchGroup{State Connection}
RoboHarness \citep{survey_ax_2607_18060} & \LiteratureLinks{https://arxiv.org/abs/2607.18060}{} & \Y & \Y & \N & \N & Model-based \\
\rowcolor{black!3}
RoboRouter \citep{survey_ax_2603_07892} & \LiteratureLinks{https://arxiv.org/abs/2603.07892}{} & \Y & \N & \N & \N & Model-based \\
RoCo \citep{survey_ax_2307_04738} & \LiteratureLinks{https://arxiv.org/abs/2307.04738}{https://github.com/MandiZhao/robot-collab} & \N & \Y & \N & \Y & Model-based \\
\rowcolor{black!3}
RoboTalk \citep{survey_ax_2609_23997} & \LiteratureLinks{https://arxiv.org/abs/2609.23997}{} & \N & \Y & \Y & \Y & Model-based \\
AGOS-Agent \citep{survey_ax_2609_08402} & \LiteratureLinks{https://arxiv.org/abs/2609.08402}{} & \N & \Y & \N & \Y & Hybrid \\
\OrchGroup{Handoff Timing}
HarnessWAM \citep{survey_ax_2608_09516} & \LiteratureLinks{https://arxiv.org/abs/2608.09516}{} & \Y & \N & \N & \N & Model-based \\
\rowcolor{black!3}
Cortex \citep{survey_ax_2607_05377} & \LiteratureLinks{https://arxiv.org/abs/2607.05377}{https://github.com/InternRobotics/Cortex} & \Y & \N & \N & \N & Model-based \\
VLAs-as-Tools \citep{survey_ax_2605_13119} & \LiteratureLinks{https://arxiv.org/abs/2605.13119}{} & \Y & \N & \N & \N & Model-based \\
\rowcolor{black!3}
Retriever \citep{survey_ax_2607_17213} & \LiteratureLinks{https://arxiv.org/abs/2607.17213}{https://github.com/openretriever/retriever} & \Y & \Y & \N & \N & Rule-based \\
DynaHarness \citep{survey_ax_2609_40306} & \LiteratureLinks{https://arxiv.org/abs/2609.40306v1}{} & \Y & \N & \Y & \N & Hybrid \\
Robo-COP~\citep{survey_ax_2610_09228} & \LiteratureLinks{https://arxiv.org/abs/2610.09228v1}{} & \Y & \Y & \N & \N & Model-based \\
\end{longtable}
\endgroup

\subsection{Memory}
\label{sec:memory}

Memory supplies information from earlier steps to later decisions without changing model weights. Unlike reusable behavior such as rules or programs, memory records what has happened, including what the robot has tried, what it has observed, and how far the task has progressed. We group existing methods into three categories according to the stage of memory they address. \textit{Memory Writing} decides what is recorded and when. \textit{Memory Updating} keeps records consistent with a changing scene. \textit{Memory Retrieval} selects records for the next decision and reuses earlier solutions.

\paragraph{\textbf{Memory Writing.}}
Memory writing decides which parts of an episode are worth recording. Common records include failed choices, task progress, and observed experience. Some systems record failures so that the robot can return to them later. HAM-VLN~\citep{survey_ax_2607_29600} stores each failure together with the place where it happened, which supports backtracking and verified stopping during navigation. Robo-Cortex~\citep{survey_ax_2605_18729} checks imagined candidate actions against fresh real observations and keeps the results in a memory with two levels of detail. Other systems record task progress. HiMe~\citep{survey_ax_2607_03449} keeps a subtask list with a status for each item to guide planning, separate from an editable record of the episode, and uses a separate checker to judge recent visual progress. A third group records broader experience. MEMORA~\citep{survey_ax_2607_14252} stores experience as records of places, objects, activities, and inferred knowledge, which are edited during the task and merged afterward. ABot-AgentOS~\citep{survey_ax_2607_10350} stores such records as graphs linked to their sources rather than as raw history.

Beyond what to record, a harness must also decide when to write and in what form. 
Writing after every step fills memory with redundant records, while writing too rarely misses events that later decisions depend on.
MessyMem~\citep{survey_ax_2609_15976} writes after each physical interaction. A VLM reviews the action, notes outcomes such as a missed grasp or a locked drawer, and updates the matching object entry with these notes and key images. Skills in Weights, Memory in Code~\citep{survey_ax_2608_09410} instead treats the memory strategy itself as code, developing it from rollout traces and keeping it as a runnable program. These systems thus differ in whether writing is triggered by events or governed by a learned program.

RoboMite~\citep{survey_ax_2611_37041} compares two ways to record earlier tasks for later instructions about the same scene. One keeps only keyframes. The other records task progress, links images to relations in a scene graph, and summarizes what the agent has learned. Across four models and three manipulation tools, average full-sequence success rose from 35.5\% with keyframes alone to 40.4\% with the structured record. The two records performed similarly when a later instruction asked about something the robot had seen in passing.

A record can preserve both advice and a procedure the next agent can run. In AMP1+~\citep{survey_ax_2611_09892}, each agent can read notes or import code left in a shared folder after earlier episodes. On new layouts of 42 RoboDojo tasks, category-averaged success was 84.1\% when the agent could keep both code and notes. Allowing only code throughout practice gave 81.0\%. The agent could still write reminders in code comments. Removing the notes only at test time gave 64.0\%. The records also transferred between models. Claude Sonnet~5.5 achieved 47.2\% success with its own records and 69.5\% with records written by Claude Opus~5.5.

\paragraph{\textbf{Memory Updating.}}
Memory updating keeps stored records consistent with a changing scene. As objects move and earlier observations become outdated, a memory entry that was correct when written can later be wrong. Existing systems handle this in three ways, by keeping different kinds of records separate, by refreshing records regularly, or by checking a record before relying on it. Mimir~\citep{survey_ax_2608_04933} keeps task memory separate from world memory, and updates execution progress, failed hypotheses, object relations, and evidence with different operators. RoboStream~\citep{survey_ax_2603_12939} refreshes its evidence about the robot and the scene at every planning step. OVAL~\citep{survey_ax_2604_12872} checks each remembered candidate against the goal, rejecting a wrong target and resuming exploration.

Updating also decides how long records are kept. Keeping every record makes memory large and slow to search, while dropping records too early loses information that the robot may still need. MetaTool~\citep{survey_ax_2609_13335} keeps a plan and tool outcomes in a scratchpad used only within the current episode. Other records compete for a fixed amount of space. EEAgent~\citep{survey_ax_2604_13533} condenses reflections on successful and failed interactions into a text memory of fixed size, and ActiveArena~\citep{survey_ax_2609_24124} compares rules for choosing what stays in such a memory. A third kind of record persists beyond what the robot can currently see. SafeMem~\citep{survey_ax_2609_08444} keeps objects that are out of view in its scene graph, revises their relations when changes are observed, and requests replanning when the predicted risk is high. CoRelNav~\citep{survey_ax_2609_27720} retains shared evidence about object instances. World Action Agent~\citep{survey_ax_2609_29964} retains reusable relational skills, procedures, and reference images, which are checked against the current scene; scene-specific advice expires when the associated observation or control state changes.

Trial outcomes can also guide what an agent keeps for its next attempt. EmbodiedRSI~\citep{survey_ax_2610_10498} trains a memory manager to keep records, combine them, or turn recurring findings into a general lesson. On RoboCasa365 Composite-Unseen, success was 61.7\% with retrieval-based memory selection and 65.2\% when the manager learned from task reward alone. Success reached 71.3\% when the reward also accounted for the cost of the next trial and how much it reduced uncertainty about the agent's hypotheses. In Embodied-Harness~\citep{survey_ax_2611_27689}, the Actor uses current observations alongside a short-term record of task progress and recent actions. After the attempt, the Reflector consults its long-term memory to revise the guidance for the next one. On RoboSPA-Long L5, success was 59.5\% with both memories, 32.1\% without short-term memory, and 48.5\% without long-term memory. Removing the Actor's recent history caused the larger drop.

A record can also change between practice runs. In simulation, SOLVE~\citep{survey_ax_2611_34750} turns earlier rollouts into short notes for the same task. With GPT-6.1 Sol on RoboDojo-Sim, average success was 61.5\% without the notes and 68.0\% with them. The real-robot trials start without task guides or notes.

\paragraph{\textbf{Memory Retrieval.}}
Memory retrieval decides which stored records reach the next decision. As memory grows, most of what it holds does not bear on the decision at hand, and passing the full history can bury the few details that matter. Existing systems retrieve three kinds of content, evidence for the current decision, the task state needed to continue work, and earlier results that can replace new computation. The first kind is evidence for the current decision. Such evidence is usually selected by its relevance to the current goal, location, or question. NavMCP~\citep{survey_ax_2608_30396} answers questions from a frozen navigation model with keyframe-based summaries of the route so far, and keeps a list of unresolved goals across calls. HFLEX-EQA~\citep{survey_ax_2609_26360} and NavProbe~\citep{survey_ax_2609_27526} select the graph or past-view evidence most relevant to each navigation decision. EmbodiedLGR~\citep{survey_ax_2604_18271} searches a robot-built scene graph for relevant entries, and CMMR-VLN~\citep{survey_ax_2603_07997} retrieves stored image and text experience for repeated navigation. HoloAgent-0~\citep{survey_ax_2606_23565} retrieves objects, rooms, views, and task history from a 3D spatial memory, and confirms them through visual verification and active exploration. Across these systems, retrieval narrows a long history down to the few records that bear on one decision.

The second kind is the task state needed to continue work. This kind matters when the executor cannot keep track of the task itself, for example a skill that starts fresh at each call or an action model without memory. BATON~\citep{survey_ax_2608_16889} retrieves this state at skill transitions. Instead of the whole episode, it retrieves the setup and starting conditions recorded for the next skill, and checks them against a live wrist-camera view. 2AM~\citep{survey_ax_2609_11308} retrieves it during a long task. It condenses its history into the target, the destination, how many times a step has been repeated, and the current stage, and passes this summary as language and optional 2D direction hints to an action model that has no memory of its own. In both systems, the harness keeps the full history on its own side and passes the executor only the state that links the current step to the rest of the task.

The third kind is an earlier result that can replace new computation. When a similar case has been handled before, reusing the earlier result can be faster and cheaper than computing a new one, provided that the result still applies. MeCo~\citep{survey_ax_2601_20577} retrieves the solution of a similar successful multi-robot task and reuses its waypoint structure, falling back to the LLM from the failed step if validation rejects the transfer. APEX~\citep{survey_ax_2602_00551} reuses the spatial evidence produced by earlier VLM updates, so that a drone can keep exploring without waiting for each new update to finish. The two systems handle the question of whether an earlier result still applies in different ways, MeCo by validating the reused solution and APEX by acting on evidence that may be slightly older than the current scene.

Past experience can also supply a recovery goal or help an agent diagnose a problem. In UniIntervene~\citep{survey_ax_2606_12372}, each memory entry pairs a past intervention state with a higher-value state reached later in the same episode. When intervention is needed, the system retrieves a similar intervention state and gives its paired observation and robot state to the correction policy as a goal. Spotter~\citep{survey_ax_2609_36808} keeps the current episode's action and observation history for diagnosis. Its RoboCasa supervisor also retrieves lessons from a frozen library built in separate episodes. In the one-shot setting, it receives frames from one episode in a separate example library, preferably from the same policy and task. Related works are compared in Table~\ref{tab:harness-memory}.

\par\Needspace{12\baselineskip}
\begingroup
\mechanismtableglue
\CompactTableStyle
\DeclareRobustCommand{\Y}{\textcolor[HTML]{009900}{{\fontencoding{U}\fontfamily{pzd}\fontseries{m}\fontshape{n}\selectfont\char51}}}
\DeclareRobustCommand{\N}{\textcolor[HTML]{B20000}{{\fontencoding{U}\fontfamily{pzd}\fontseries{m}\fontshape{n}\selectfont\char55}}}

\setlength{\LiteratureWidth}{\dimexpr\linewidth-12\tabcolsep\relax}
\setlength{\LTleft}{0pt}
\setlength{\LTright}{0pt}
\setlength{\LTpre}{4pt}
\setlength{\LTpost}{4pt}
\setlength{\LTcapwidth}{\linewidth}
\captionsetup{singlelinecheck=false,justification=raggedright}
\newcommand{\MemGroup}[1]{\rowcolor{TSTHarnessLight!45}\multicolumn{6}{l}{\textbf{#1}} \\}
\begin{longtable}{>{\RaggedRight\arraybackslash}p{.27\LiteratureWidth}>{\centering\arraybackslash}p{.10\LiteratureWidth}>{\RaggedRight\arraybackslash}p{.17\LiteratureWidth}>{\RaggedRight\arraybackslash}p{.17\LiteratureWidth}>{\RaggedRight\arraybackslash}p{.17\LiteratureWidth}>{\centering\arraybackslash}p{.12\LiteratureWidth}}
\caption[Comparison of works on memory]{Comparison of works on memory (Section~\ref{sec:memory}). Structured includes records, indices, plans, and code; Persists means memory is retained across episodes.}\label{tab:harness-memory}\\
\LiteratureTopRule{Harness}
\rowcolor{TSTHarnessLight}\textbf{Work} & \textbf{Link} & \textbf{Stored as} & \textbf{Written at} & \textbf{Retrieved by} & \textbf{Persists} \\
\midrule
\endfirsthead
\multicolumn{6}{l}{\tablename~\thetable\ (continued)} \\
\LiteratureTopRule{Harness}
\rowcolor{TSTHarnessLight}\textbf{Work} & \textbf{Link} & \textbf{Stored as} & \textbf{Written at} & \textbf{Retrieved by} & \textbf{Persists} \\
\midrule
\endhead
\midrule
\multicolumn{6}{r}{Continued on next page} \\
\endfoot
\bottomrule
\endlastfoot
\MemGroup{Memory Writing}
HAM-VLN \citep{survey_ax_2607_29600} & \LiteratureLinks{https://arxiv.org/abs/2607.29600}{} & Graph & Step & Query & \N \\
\rowcolor{black!3}
Robo-Cortex \citep{survey_ax_2605_18729} & \LiteratureLinks{https://arxiv.org/abs/2605.18729}{} & Text & Episode end & Similarity & \Y \\
HiMe \citep{survey_ax_2607_03449} & \LiteratureLinks{https://proceedings.mlr.press/v306/ji26k.html}{https://github.com/HappyWaterXP/HiMe} & Text & Event & Query & \N \\
\rowcolor{black!3}
MEMORA \citep{survey_ax_2607_14252} & \LiteratureLinks{https://arxiv.org/abs/2607.14252v2}{https://github.com/yuzihaowashu/MEMORA} & Structured & Step & Query & \Y \\
ARMS \citep{survey_ax_2609_28429} & \LiteratureLinks{https://arxiv.org/abs/2609.28429}{} & Structured & Event & Similarity & \Y \\
\rowcolor{black!3}
MessyMem \citep{survey_ax_2609_15976} & \LiteratureLinks{https://arxiv.org/abs/2609.15976}{} & Graph & Event & Query & \N \\
HyMeS \citep{survey_ax_2608_09410} & \LiteratureLinks{https://arxiv.org/abs/2608.09410}{} & Structured & Event & Check & \N \\
\rowcolor{black!3}
AMP1+ \citep{survey_ax_2611_09892} & \LiteratureLinks{https://arxiv.org/abs/2611.09892v3}{https://github.com/apple-aiml-research/ml-amp} & Text; Structured & Episode or round end & Direct & \Y \\
\midrule
RoboMite~\citep{survey_ax_2611_37041} & \LiteratureLinks{https://arxiv.org/abs/2611.37041v1}{} & Graph; Visual; Text & Event & Direct & \N \\
\MemGroup{Memory Updating}
Mimir \citep{survey_ax_2608_04933} & \LiteratureLinks{https://arxiv.org/abs/2608.04933}{} & Structured & Step & Query & \N \\
\rowcolor{black!3}
RoboStream \citep{survey_ax_2603_12939} & \LiteratureLinks{https://link.springer.com/chapter/10.1007/978-3-032-37252-9_3}{https://github.com/yu2hi13/RoboStream} & Graph & Step & Direct & \N \\
OVAL \citep{survey_ax_2604_12872} & \LiteratureLinks{https://arxiv.org/abs/2604.12872}{} & Structured & Step & Query & \N \\
\rowcolor{black!3}
MetaTool \citep{survey_ax_2609_13335} & \LiteratureLinks{https://arxiv.org/abs/2609.13335}{} & Structured & Event & Direct & \N \\
EEAgent \citep{survey_ax_2604_13533} & \LiteratureLinks{https://arxiv.org/abs/2604.13533v2}{} & Text & Event & Direct & \Y \\
\rowcolor{black!3}
ActiveArena \citep{survey_ax_2609_24124} & \LiteratureLinks{https://arxiv.org/abs/2609.24124}{https://github.com/leeibo/ActiveArena} & Visual & Event & Direct & \N \\
SafeMem \citep{survey_ax_2609_08444} & \LiteratureLinks{https://arxiv.org/abs/2609.08444}{} & Graph & Step & Direct & \N \\
\rowcolor{black!3}
CoRelNav \citep{survey_ax_2609_27720} & \LiteratureLinks{https://arxiv.org/abs/2609.27720}{} & Structured & Step & Query & \N \\
WAA \citep{survey_ax_2609_29964} & \LiteratureLinks{https://arxiv.org/abs/2609.29964}{} & Structured & Episode end & Query & \Y \\
\rowcolor{black!3}
Robo-Harness K1 \citep{survey_ax_2609_29389} & \LiteratureLinks{https://arxiv.org/abs/2609.29389}{https://github.com/Robo-Harness/k1} & Structured & Event & Check & \N \\
EmbodiedRSI \citep{survey_ax_2610_10498} & \LiteratureLinks{https://arxiv.org/abs/2610.10498v2}{} & Structured; Graph & Episode end & Query & \Y \\
\rowcolor{black!3}
Embodied-Harness \citep{survey_ax_2611_27689} & \LiteratureLinks{https://arxiv.org/abs/2611.27689v1}{} & Structured; Text & Step; Episode end & Direct & \Y\ (guidance) \\
\midrule
\MemGroup{Memory Retrieval}
NavMCP \citep{survey_ax_2608_30396} & \LiteratureLinks{https://arxiv.org/abs/2608.30396}{} & Structured & Event & Query & \N \\
\rowcolor{black!3}
HFLEX-EQA \citep{survey_ax_2609_26360} & \LiteratureLinks{https://arxiv.org/abs/2609.26360}{https://github.com/ntnu-arl/hflex_eqa} & Graph & Step & Query & \N \\
NavProbe \citep{survey_ax_2609_27526} & \LiteratureLinks{https://arxiv.org/abs/2609.27526v2}{https://github.com/liujy25/NavProbe} & Structured & Step & Query & \N \\
\rowcolor{black!3}
EmbodiedLGR \citep{survey_ax_2604_18271} & \LiteratureLinks{https://arxiv.org/abs/2604.18271}{} & Graph & Step & Query & \N \\
CMMR-VLN \citep{survey_ax_2603_07997} & \LiteratureLinks{https://arxiv.org/abs/2603.07997}{} & Visual & Episode end & Similarity & \Y \\
\rowcolor{black!3}
HoloAgent-0 \citep{survey_ax_2606_23565} & \LiteratureLinks{https://arxiv.org/abs/2606.23565}{} & Graph & Step & Query & \Y \\
BATON \citep{survey_ax_2608_16889} & \LiteratureLinks{https://arxiv.org/abs/2608.16889}{} & Text & Episode end & Check & \Y \\
\rowcolor{black!3}
2AM \citep{survey_ax_2609_11308} & \LiteratureLinks{https://arxiv.org/abs/2609.11308}{} & Structured & Event & Direct & \N \\
MeCo \citep{survey_ax_2601_20577} & \LiteratureLinks{https://doi.org/10.65109/qpuc1624}{https://github.com/TomWang-NPU/MeCo} & Structured & Episode end & Similarity & \Y \\
\rowcolor{black!3}
APEX \citep{survey_ax_2602_00551} & \LiteratureLinks{https://openaccess.thecvf.com/content/CVPR2026/html/Zhang_APEX_A_Decoupled_Memory-based_Explorer_for_Asynchronous_Aerial_Object_Goal_CVPR_2026_paper.html}{https://github.com/4amGodvzx/apex} & Visual & Step & Direct & \N \\
UniIntervene \citep{survey_ax_2606_12372} & \LiteratureLinks{https://arxiv.org/abs/2606.12372v1}{https://github.com/Denghaoyuan123/UniIntervene} & Structured & Offline & Similarity & \Y \\
\rowcolor{black!3}
Spotter \citep{survey_ax_2609_36808} & \LiteratureLinks{https://arxiv.org/abs/2609.36808v2}{https://github.com/zqc3117/Spotter} & Text; Structured & Offline & Query & \Y \\
\end{longtable}
\endgroup

\subsection{Monitoring and Recovery}
\label{sec:monitoring-recovery}

Monitoring judges whether execution is progressing, has finished, has failed, or has violated a constraint, and recovery changes execution after a problem is found. We group existing methods into three categories that follow this process from detection to repair. \textit{Execution Monitoring} checks execution before, during, or after an action. \textit{Response Selection} turns a verdict into a response and checks that response before it runs. \textit{Recovery Actions} carry out the response, including where recovery starts and when to stop trying.

\paragraph{\textbf{Execution Monitoring.}}
Execution monitoring can check an action before it runs, while it runs, or after it ends. The timing of a check determines which problems it can catch and what kind of report it produces. 
Checks before execution act on proposed actions or routes that have not yet had any physical effect, and usually accept, reject, or rank them. SafeHarness~\citep{survey_ax_2609_20822} checks a proposed route for obstacles. If it rejects the route, it explains why so the planner can revise it. DV-VLN~\citep{survey_ax_2601_18492} answers true or false questions about candidate navigation actions and reorders them accordingly. CoVer~\citep{survey_ax_2602_12281} prepares rephrasings of the instruction in advance and uses a verifier to score candidate actions and instructions before each action runs. Since these checks only need to choose among options, their outputs are simple decisions or scores.

Choosing a skill does not settle whether each of its operations is safe. LEASH~\citep{survey_ax_2611_14115} first checks the task against camera images, then checks each proposed action or skill. Inside a skill, it checks selected operations again just before they run, such as opening the gripper. If it rejects an action or operation, the planner sees the current scene and chooses what to do next. Across 276 simulated runs per setting, adding LEASH to GPT-6 Astra lowered the harm rate from 11.23\% to 0\% and raised safe success from 78.99\% to 80.80\%.

Checks during execution give signals while the action is still under way. They mainly track constraint violations and task progress, and some systems check at several levels at once. For constraints, DoReMi~\citep{survey_ax_2307_00329} uses an LLM to write constraints for the next skill and a VLM to check them at regular intervals, stopping the skill as soon as one is broken. Code-as-Monitor~\citep{survey_ax_2412_04455} writes visual constraints as code that can detect a failure as it happens or before it happens. Meanwhile, Mission-Level Runtime Assurance~\citep{survey_ax_2607_23532} checks constraints over mission events and reports ``unknown'' when evidence is missing. For progress, OnFly~\citep{survey_ax_2603_10682} checks the task stage in parallel with flight and reacts when it detects stop or lost conditions, while Robometer~\citep{survey_ax_2603_02115} tracks progress with a score learned from expert labels and from comparisons between successful and failed runs. ABot-AgentOS~\citep{survey_ax_2607_10350} combines several levels, checking at runtime, at the skill level, and once more before allowing a task to finish. DynaHarness~\citep{survey_ax_2609_40306} saves a completion signal as soon as the execution interface detects it. At its next check, the scheduler can end the command even if the signal has disappeared. A separate signal can trigger a skill switch when progress stalls. These checks mostly output signals to stop, adjust, or report progress, since they must act before the motion ends.

Checks after execution can use the full record of what happened, which allows them to judge the outcome and explain its cause. RoboLab~\citep{survey_ax_2604_09860}, after first checking task assets and stability test results, judges the outcome with generated task code that defines when a task has succeeded or should end.
REFLECT~\citep{survey_ax_2306_15724} summarizes an execution at the sensor, event, and subgoal levels to explain a failure to a correction planner. AHA~\citep{survey_ax_2410_00371} gives failure-specific feedback in language based on sequences of key images. DQAF~\citep{survey_ax_2605_26349} writes feedback after each episode for the next human operator. 

Overall, the timing of a check thus shapes its output. Checks before execution choose among candidates, checks during execution signal whether to stop or adjust, and checks after execution explain what happened.

\paragraph{\textbf{Response Selection.}}
Response selection decides what the system does after a problem is detected. It covers three questions, how the detected problem reaches the agent, how the agent chooses a response, and whether that response is checked before it runs. A detected problem can change behavior only if the agent that chooses the next action can see it. Existing work either builds this link into the agent's tools or tests whether the link affects later decisions. Agentic AI for Robot Control~\citep{survey_ax_2602_13081} takes the first approach, giving the agent tools to query robot status and events. Diagnosing Semantic Handoff Failures~\citep{survey_ax_2607_06256} takes the second, checking whether a visual reachability check changes the agent's next skill or route.

Once the agent sees the problem, it must choose a response. Existing systems either pick from a fixed set of responses or let a planner revise the plan. Fixed responses are simple and predictable, while a planner can produce responses that were not listed in advance. Goal2Skill~\citep{survey_ax_2604_13942} picks among retrying, adjusting, and replanning, and A Closed-Loop Multi-Agent Framework for Robust Multi-Robot Manipulation~\citep{survey_ax_2607_06990} separates local pose fixes from full replanning. Among planner-based systems, the planner is triggered by an explanation of the failure in REFLECT~\citep{survey_ax_2306_15724}, by a gap between the predicted and actual outcome in WCM~\citep{survey_ax_2607_22999}, and by a lack of progress in Stop Wandering~\citep{survey_ax_2604_02318}.

A chosen response can also be wrong, so systems validate proposed actions and use execution evidence to decide whether to retry. These checks rely on perception, sub-goal evidence, or the validity of tool inputs. ETA~\citep{survey_ax_2608_03924} refuses to act when it cannot locate the target. ACE~\citep{survey_ax_2607_04162} uses tracked pick-and-place masks, execution reports, and new observations to assess sub-goal completion and guide retries. HarnessVLN~\citep{survey_ax_2609_15195} checks tool inputs and evidence before the robot moves or stops.

\paragraph{\textbf{Recovery Actions.}}
Once a failure is detected, the robot needs a recovery action that fixes it. Generally, they differ in what they change, namely the input to the current policy, the executor, the task flow, or something outside the policy. VLA-Corrector~\citep{survey_ax_2609_06508} changes the input, feeding stage-specific recovery prompts to a frozen VLA. FLARE~\citep{survey_ax_2608_26645}, VoLo~\citep{survey_ax_2606_07723}, and ReCoVLA~\citep{survey_ax_2606_09630} change the executor, switching to a trained reset adapter, a hand-designed skill guided by perception, or a simulation-trained correction policy for each known failure type. AgentChord~\citep{survey_ax_2605_11951} changes the task flow, moving to recovery branches compiled into its task graph without calling the LLM again. SERP~\citep{survey_ax_2603_02772} and Dream2Fix~\citep{survey_ax_2603_13528} change something outside the policy, adjusting motion-cost parameters and applying small learned physical corrections. In all cases, recovery is limited to the actions the system provides.

Detecting a failure and choosing a repair can be tested separately. UniIntervene~\citep{survey_ax_2606_12372} monitors how predicted value changes over a window of actions. In a two-task ablation, removing the retrieved recovery goal left the detector's offline F1 score at 0.882. Online task success fell from 95\% to 85\%, suggesting that the goal helps the policy recover once intervention begins. Spotter~\citep{survey_ax_2609_36808} screens completed action chunks while the policy continues. If its VLM confirms a problem, it calls motion primitives, predicts a visible sign of repair, and checks a subsequent observation. A retreat can restore the arm pose while leaving displaced objects where they are. In simulation, A$^2$-MEAS~\citep{survey_ax_2611_09124} uses a correction instruction, spatial evidence, execution history, and the policy's proposed action chunk to generate a corrected chunk.

The next observation can change a repair already in progress. In USB insertion, Contact Harness~\citep{survey_ax_2611_29377} uses fingertip measurements of connector bending to decide how to adjust alignment. The controller limits contact force during each commanded motion. X-Harness~\citep{survey_ax_2611_34778} compares two ways to call a VLM when a program needs help. A single call can replace the rest of the program. With several calls, the VLM can see what happened after each action and change its next one. This worked better on long-horizon RoboCasa365 tasks. In a separate test across three benchmarks, one-shot takeovers lost more successes than they gained on tasks where training had retained a code revision.

Recovery also needs to choose which state to resume from. A failed attempt often leaves the scene partly changed, with objects moved or a step left half done. The robot can continue from this state and keep the progress already made, or first restore the scene to a known state and start again. PhyAgentOS~\citep{survey_ax_2607_16636} continues from the current state, diagnosing the failed attempt and resuming with new instructions. HALTER~\citep{survey_ax_2609_19413} restores the scene first, using scene-graph histories to find what was completed, planning reset skills, and checking that the reset worked. LIBERO-RECOVER~\citep{survey_ax_2609_05178} evaluates these two cases separately, comparing recovery from a disturbed scene with an attempt that starts in a restored scene.

In real assembly tasks, Agentic GR00T~2~\citep{nvidia2026gr00t2} lets the agent take over when the WAM stalls. The agent can recover a dropped part with a code-based skill, then hand control back. With 100 trials per task and configuration, adding the full agent to the same SFT + DAgger policy raised completed toy planes from 84 to 98, tractors from 71 to 97, and cars from 78 to 97. In three of the eight remaining failures, a screw was only partly driven, yet both the WAM and the agent judged the step complete.

For each simulated task, the authors compared GR00T~2~\citep{nvidia2026gr00t2} with an ASPIRE~\citep{survey_ax_2607_00272} program on development seeds. Test episodes started with the executor that succeeded more often, with ties going to the program. The agent monitored execution and could take over. In one RoboCasa365 episode, a program grasped the oven handle instead of the drawer handle. The agent then grasped the correct handle and opened the drawer. In another episode, the agent failed to regrasp a slice of toast, and the WAM ran out of time after the handoff. Across RoboCasa365 episodes with SFT + Agent, 1,492 of 1,512 succeeded when the agent left execution alone, compared with 553 of 988 when it intervened. On RoboDojo, the corresponding counts were 1,291 of 1,320 and 188 of 773. The agent chose when to intervene based on signs of failure. The trials therefore leave open how often those same episodes would have succeeded without help.

Finally, repeated attempts can fail again, so systems need a rule for when to stop. Existing systems stop after a fixed budget, hand the task to a human, retry only under specific conditions, or aim for an end point other than task completion. PEACE~\citep{survey_ax_2606_00104} lets the planner retry after a drone tool failure only within a fixed budget. LabVLA~\citep{survey_ax_2606_13578} allows one validated retry during offline workflow-YAML generation before returning the workflow for human review. A physical agentic grasping loop~\citep{survey_ax_2604_07395} retries only when the gripper is empty, and otherwise stops or asks for clarification. SPINE~\citep{survey_ax_2607_13049} repairs robot software and settings until the robot is ready for teleoperation, rather than until a task is completed. Related works are compared in Table~\ref{tab:harness-monitoring-and-recovery}.

\par\Needspace{12\baselineskip}
\begingroup
\mechanismtableglue
\CompactTableStyle

\setlength{\LiteratureWidth}{\dimexpr\linewidth-12\tabcolsep\relax}
\setlength{\LTleft}{0pt}
\setlength{\LTright}{0pt}
\setlength{\LTpre}{4pt}
\setlength{\LTpost}{4pt}
\setlength{\LTcapwidth}{\linewidth}
\captionsetup{singlelinecheck=false,justification=raggedright}
\newcommand{\MRGroup}[1]{\rowcolor{TSTHarnessLight!45}\multicolumn{6}{l}{\textbf{#1}} \\}
\newcommand{\PhIc}[1]{{\scriptsize\textcolor{black!55}{#1}}}
\newcommand{\IcB}{\PhIc{\faHourglassStart}}
\newcommand{\IcD}{\PhIc{\faPlay}}
\newcommand{\IcA}{\PhIc{\faFlagCheckered}}
\newcommand{\PhB}{\IcB~Before}
\newcommand{\PhD}{\IcD~During}
\newcommand{\PhA}{\IcA~After}
\begin{longtable}{>{\RaggedRight\arraybackslash}p{.22\LiteratureWidth}
                  >{\centering\arraybackslash}p{.10\LiteratureWidth}
                  >{\RaggedRight\arraybackslash}p{.16\LiteratureWidth}
                  >{\RaggedRight\arraybackslash}p{.15\LiteratureWidth}
                  >{\RaggedRight\arraybackslash}p{.18\LiteratureWidth}
                  >{\RaggedRight\arraybackslash}p{.19\LiteratureWidth}}
\caption[Comparison of works on monitoring and recovery]{Comparison of works on monitoring and recovery (Section~\ref{sec:monitoring-recovery}). Check Time is relative to the checked step: before, during, or after execution.}
\label{tab:harness-monitoring-and-recovery}\\
\LiteratureTopRule{Harness}
\rowcolor{TSTHarnessLight}
\mbox{\textbf{Work}} & \mbox{\textbf{Link}} & \mbox{\textbf{Check Time}} & \mbox{\textbf{Checker}} & \mbox{\textbf{Signal}} & \mbox{\textbf{Response}} \\
\midrule
\endfirsthead
\multicolumn{6}{l}{\tablename~\thetable\ (continued)} \\
\LiteratureTopRule{Harness}
\rowcolor{TSTHarnessLight}
\mbox{\textbf{Work}} & \mbox{\textbf{Link}} & \mbox{\textbf{Check Time}} & \mbox{\textbf{Checker}} & \mbox{\textbf{Signal}} & \mbox{\textbf{Response}} \\
\midrule
\endhead
\midrule
\multicolumn{6}{r}{Continued on next page} \\
\endfoot
\bottomrule
\endlastfoot
\MRGroup{Execution Monitoring}
SafeHarness~\citep{survey_ax_2609_20822} & \LiteratureLinks{https://arxiv.org/abs/2609.20822v2}{} & \IcB\,\IcD~Before, during & Rules & Constraint & Replan \\
\rowcolor{black!3}
DV-VLN~\citep{survey_ax_2601_18492} & \LiteratureLinks{https://arxiv.org/abs/2601.18492}{https://github.com/PlumJun/DV-VLN} & \PhB & VLM & Validity & Select \\
CoVer~\citep{survey_ax_2602_12281} & \LiteratureLinks{https://arxiv.org/abs/2602.12281}{https://github.com/cover-vla/cover-vla} & \PhB & Learned & Validity & Select \\
\rowcolor{black!3}
DoReMi~\citep{survey_ax_2307_00329} & \LiteratureLinks{https://ieeexplore.ieee.org/document/10802284}{} & \PhD & VLM & Constraint & Stop, replan \\
Code-as-Monitor~\citep{survey_ax_2412_04455} & \LiteratureLinks{https://openaccess.thecvf.com/content/CVPR2025/html/Zhou_Code-as-Monitor_Constraint-aware_Visual_Programming_for_Reactive_and_Proactive_Robotic_Failure_CVPR_2025_paper.html}{} & \PhD & Code & Constraint & Report \\
\rowcolor{black!3}
MLRA~\citep{survey_ax_2607_23532} & \LiteratureLinks{https://arxiv.org/abs/2607.23532}{} & \PhD & Rules & Constraint & Report \\
OnFly~\citep{survey_ax_2603_10682} & \LiteratureLinks{https://arxiv.org/abs/2603.10682}{} & \PhD & VLM & Progress & Replan \\
\rowcolor{black!3}
Robometer~\citep{survey_ax_2603_02115} & \LiteratureLinks{https://www.roboticsproceedings.org/rss22/p140.html}{https://github.com/robometer/robometer} & \PhD & Learned & Progress & Report \\
RoboLab~\citep{survey_ax_2604_09860} & \LiteratureLinks{https://www.roboticsproceedings.org/rss22/p096.html}{https://github.com/NVlabs/RoboLab} & \PhA & Code & Progress & Report \\
\rowcolor{black!3}
REFLECT~\citep{survey_ax_2306_15724} & \LiteratureLinks{https://proceedings.mlr.press/v229/liu23g.html}{https://github.com/real-stanford/reflect} & \PhA & LLM & Failure & Report, replan \\
AHA~\citep{survey_ax_2410_00371} & \LiteratureLinks{https://proceedings.iclr.cc/paper_files/paper/2025/hash/70a06501001e1820fd1eb9ee821302d2-Abstract-Conference.html}{https://github.com/NVlabs/AHA} & \PhA & VLM & Failure & Report \\
\rowcolor{black!3}
DQAF~\citep{survey_ax_2605_26349} & \LiteratureLinks{https://arxiv.org/abs/2605.26349}{} & \PhA & VLM + human & Progress & Report \\
DynaHarness \citep{survey_ax_2609_40306} & \LiteratureLinks{https://arxiv.org/abs/2609.40306v1}{} & \PhD & Rules & Stall, risk & Switch, replan \\
LEASH~\citep{survey_ax_2611_14115} & \LiteratureLinks{https://arxiv.org/abs/2611.14115v1}{https://github.com/zhuqinfeng1999/LEASH} & \PhB & Learned & Safety & Reject, replan \\
\MRGroup{Response Selection}
AARC~\citep{survey_ax_2602_13081} & \LiteratureLinks{https://ojs.aaai.org/index.php/AAAI-SS/article/view/42578}{} & \PhD & Rules & Progress & Replan \\
\rowcolor{black!3}
DSHF~\citep{survey_ax_2607_06256} & \LiteratureLinks{https://arxiv.org/abs/2607.06256}{} & \PhA & VLM & Success & Retry, replan \\
Goal2Skill~\citep{survey_ax_2604_13942} & \LiteratureLinks{https://arxiv.org/abs/2604.13942}{} & \PhA & VLM & Failure & Retry, replan \\
\rowcolor{black!3}
CLMAF~\citep{survey_ax_2607_06990} & \LiteratureLinks{https://www.roboticsproceedings.org/rss22/p036.html}{} & \IcB\,\IcA~Before, after & VLM & Validity & Replan \\
WCM~\citep{survey_ax_2607_22999} & \LiteratureLinks{https://arxiv.org/abs/2607.22999}{} & \PhA & VLM & Failure & Replan \\
\rowcolor{black!3}
MetaNav~\citep{survey_ax_2604_02318} & \LiteratureLinks{https://arxiv.org/abs/2604.02318}{} & \PhD & LLM & Progress & Replan \\
ACE~\citep{survey_ax_2607_04162} & \LiteratureLinks{https://arxiv.org/abs/2607.04162v2}{} & \PhA & VLM + human & Success & Retry \\
\rowcolor{black!3}
HarnessVLN~\citep{survey_ax_2609_15195} & \LiteratureLinks{https://arxiv.org/abs/2609.15195}{} & \PhB & Rules & Validity & Retry, stop \\
\MRGroup{Recovery Actions}
VLA-Corrector~\citep{survey_ax_2609_06508} & \LiteratureLinks{https://arxiv.org/abs/2609.06508}{} & \PhD & Learned & Failure & Adjust \\
\rowcolor{black!3}
FLARE~\citep{survey_ax_2608_26645} & \LiteratureLinks{https://openaccess.thecvf.com/content/CVPR2026/html/Zhao_FLARE_A_Failure-Aware_Framework_for_Autonomous_Correction_and_Recovery_in_CVPR_2026_paper.html}{} & \PhD & VLM & Failure & Retry, reset \\
ReCoVLA~\citep{survey_ax_2606_09630} & \LiteratureLinks{https://arxiv.org/abs/2606.09630}{} & \PhD & VLM & Failure & Switch \\
\rowcolor{black!3}
AgentChord~\citep{survey_ax_2605_11951} & \LiteratureLinks{https://www.roboticsproceedings.org/rss22/p180.html}{https://github.com/EDEM-AI/AgentChord} & \PhD & Code & Failure & Switch \\
SERP~\citep{survey_ax_2603_02772} & \LiteratureLinks{https://arxiv.org/abs/2603.02772v2}{} & \PhA & Rules & Failure & Adjust, retry \\
\rowcolor{black!3}
Dream2Fix~\citep{survey_ax_2603_13528} & \LiteratureLinks{https://arxiv.org/abs/2603.13528}{} & \PhA & Learned & Failure & Adjust \\
PhyAgentOS~\citep{survey_ax_2607_16636} & \LiteratureLinks{https://arxiv.org/abs/2607.16636}{https://github.com/PhyAgentOS/PhyAgentOS-core} & \PhA & Rules & Success & Replan \\
\rowcolor{black!3}
HALTER~\citep{survey_ax_2609_19413} & \LiteratureLinks{https://arxiv.org/abs/2609.19413}{} & \PhA & LLM & Success & Reset \\
LIBERO-RECOVER~\citep{survey_ax_2609_05178} & \LiteratureLinks{https://arxiv.org/abs/2609.05178}{https://github.com/liulin815/LIBERO-Recovery} & -- & Benchmark & -- & -- \\
\rowcolor{black!3}
PEACE~\citep{survey_ax_2606_00104} & \LiteratureLinks{https://arxiv.org/abs/2606.00104}{https://github.com/erdemuysalx/PEACE} & \PhA & Rules & Failure & Replan \\
LabVLA~\citep{survey_ax_2606_13578} & \LiteratureLinks{https://arxiv.org/abs/2606.13578}{https://github.com/zjunlp/LabVLA} & \PhA & Code + human & Success, constraint & Reject \\
\rowcolor{black!3}
PAL~\citep{survey_ax_2604_07395} & \LiteratureLinks{https://arxiv.org/abs/2604.07395}{} & \PhA & Rules + human & Failure & Retry, stop \\
SPINE~\citep{survey_ax_2607_13049} & \LiteratureLinks{https://arxiv.org/abs/2607.13049}{} & \PhD & Rules + human & Failure & Adjust, retry \\
\rowcolor{black!3}
UniIntervene \citep{survey_ax_2606_12372} & \LiteratureLinks{https://arxiv.org/abs/2606.12372v1}{https://github.com/Denghaoyuan123/UniIntervene} & \PhD & Learned & Value stagnation & Track retrieved goal \\
Spotter \citep{survey_ax_2609_36808} & \LiteratureLinks{https://arxiv.org/abs/2609.36808v2}{https://github.com/zqc3117/Spotter} & \PhD & VLM & Suspect action & Repair, recheck \\
\rowcolor{black!3}
A$^2$-MEAS \citep{survey_ax_2611_09124} & \LiteratureLinks{https://arxiv.org/abs/2611.09124v1}{https://github.com/AutoLab-SAI-SJTU/A2MEAS} & \PhD & VLM & Anomaly & Correct action chunk \\
Agentic GR00T~2 \citep{nvidia2026gr00t2} & \LiteratureLinks{https://research.nvidia.com/labs/gear/gr00t-2/assets/gr00t2_whitepaper.pdf}{} & \PhD & VLM & Stall, drop & Pause, repair, resume \\
Contact Harness~\citep{survey_ax_2611_29377} & \LiteratureLinks{https://arxiv.org/abs/2611.29377v1}{https://github.com/NTUMARS/ContactHarness} & \PhA & VLM & Contact load & Adjust next motion \\
\rowcolor{black!3}
X-Harness~\citep{survey_ax_2611_34778} & \LiteratureLinks{https://arxiv.org/abs/2611.34778v1}{} & \PhD & Rules + VLM & Failure, progress & Replace code or revise each step \\
\end{longtable}
\endgroup

\subsection{Skills}
\label{sec:harness-skills}

Orchestration, memory, and monitoring decide how a harness handles the current attempt; skills decide what it brings to the next one. A skill is a pluggable package of procedural knowledge that the harness can load or call. Loaded into the context, a textual skill shapes how the language model reasons about the task and uses the harness's tools; called as a function, an executable skill gives the model a reusable operation on the robot. In the Skills panel of Figure~\ref{fig:harness-operation}, the same experience can become a note to approach the handle from below or a callable \texttt{open\_drawer} program, and either can be swapped in without retraining the model. Skills thus differ from memory, which records what happened, and from model weights, which a skill leaves unchanged. A recent survey ranks code-as-policy systems by whether they use execution feedback, persistent skill memory, and program search~\citep{survey_ax_2608_01851}; we focus instead on how skills enter, change within, and leave a harness.

\paragraph{\textbf{From Skills to Code as Policies.}}
Early language-model harnesses treated skills as a fixed set of pretrained primitives, and the model's role was only to choose among them, as in SayCan~\citep{survey_ax_2204_01691}. Code generation changed that role. In Code as Policies~\citep{survey_ax_2209_07753}, the model composes perception and control APIs into a program and defines missing functions on the fly, so the skill set becomes a vocabulary rather than a menu. ProgPrompt~\citep{survey_ax_2209_11302} and VoxPoser~\citep{survey_ax_2307_05973} followed the same idea, expressing plans as programs with assertions or as code that builds value maps for a motion planner. These programs were written for one task and then discarded. Later work let agents revise a program during the task. Act-Observe-Rewrite~\citep{survey_ax_2603_04466} rewrites a controller between trials using keyframes and logs. MEMENTO~\citep{survey_ax_2607_22832} searches over a single code policy. In Agent as Policy~\citep{survey_ax_2609_12541}, a coding agent writes and revises programs directly on a physical robot. Generated code thereby became a capable policy within a task, which raised the question of what to keep across tasks.

\paragraph{\textbf{Skill Libraries and Reuse.}}
Reuse is the natural answer, and it requires organizing skills so that later tasks can find them. Voyager~\citep{survey_ax_2305_16291} set the pattern in a Minecraft agent: a program joins a library once it passes self-verification, and new tasks retrieve entries by embedding similarity. Skill libraries are now common across other agent domains, where skills are distilled from agent experience and evolved alongside reinforcement learning~\citep{survey_ax_2602_08234}, or built from scientific resources~\citep{survey_ax_2604_03964} and human-created multimodal material such as tutorials, repositories, and articles~\citep{survey_ax_2606_29538}. Robot systems adopted this library structure and differ mainly in what they store and how an entry is admitted. RoboCoder~\citep{survey_ax_2406_03757} stores joint-level action code accepted by a vision-language evaluator, while DROC~\citep{survey_ax_2311_10678} and ARCHITECT~\citep{survey_ax_2607_23784} store knowledge distilled from human corrections, so that the same correction need not be given twice. Uni-Skill~\citep{survey_ax_2603_02623} fills gaps in its library by implementing missing skills from a video-derived repository, and Teach and Grow~\citep{survey_ax_2608_17209} admits a skill only after it passes a scoped outcome test on held-out states. A library can even be built once and frozen, as CaP-X~\citep{survey_ax_2603_22435} does with nine helper functions distilled from successful rollouts, or compressed into macros, as in SkillComposer~\citep{survey_ax_2608_14944}.

An existing library also changes what code an agent needs to write. Chen et al.~\citep{survey_ax_2611_09030} gave coding agents both general skills from \texttt{open-robot-skills} and scripts from human-written policies for each task. The general skills covered perception, grasping, and motion planning. The authors counted lines in each initial program that did not match the supplied code. The share was 24\% for the Claude Opus~5 configuration labeled xhigh, 18\% for Opus~5.5, and 12\% for Fable~5.1. Each of these models gained more from two rounds of feedback than the four that wrote only 4--7\% new code.

RAX~\citep{survey_ax_2610_22496} tests each skill on the robot before adding it to the library. A coding agent practices the skill and adds checks for when the program can start and whether it has finished. The skill must also complete a recorded run without human help. On an RB-Y1 mobile manipulator, the agent reused four skills in a six-stage tidy-up task. It moved a chair, stacked cups, and carried a bucket and a box. It supplied new object positions and wrote code to connect the stages. If a program encountered an unfamiliar state, the agent could inspect the failure and revise the program. In each of three trials, the agent judged from head-camera images that it had completed all six stages. It updated its library after each trial. An operator reset the scene before each run, gave advice, and re-armed the robot after controller faults in the third trial.

An agent can keep a program or a written guide for its next attempt at the same task. In ACE-CaP-0~\citep{survey_ax_2611_16872}, a coding agent tests programs on the robot and can request demonstrations or use simulation during development. The final program still reads sensors but runs without the coding agent. The frozen programs succeeded in 221 of 240 trials across eight real-robot tasks. RoboARCade~\citep{survey_ax_2611_08209} saves a text guide instead. GPT-6 Astra wrote a guide from a successful example of each task and used it in a fresh session. Across five tasks, its mean rubric score was 53.5\% without an example or a guide and 62.5\% with the first guide. Revising the guide using scored attempts raised the score to 72.0\%. With the full example in context, it scored 73.0\%.

Table~\ref{tab:retained-experience} compares task success with and without prior experience. Two studies tested whether records from earlier tasks helped on tasks the agent had yet to practice. AMP1+~\citep{survey_ax_2611_09892} arranged 42 tasks in random order and saved its memory after the first ten. On the last 12 tasks, all still unpracticed, success was 24\% with empty memory and 47\% with the saved memory. EmbodiedEvo~\citep{survey_ax_2610_18867} tested four conditions on the same 12 target tasks. Mean success was 1.67\% with no experience and 23.06\% with experience from the target task itself. Code and notes from related tasks gave 16.39\%. Reassigning the same material to mismatched targets gave 18.33\%, so matching tasks by similarity brought no average advantage here.

Real2Worlds2Real~\citep{survey_ax_2611_31098} tested skill reuse in robot control and scene building. The agent wrote candidate control programs for 20 LIBERO-PRO Long tasks. Mean success across all candidates rose from 44.1\% with no saved skills to 69.1\% with 136 skills from other tasks. For a held-out lemon-cutting scene, the authors added skills for handling fruit, modeling cut geometry, and simulating the cutting interaction. The inputs and scene checks stayed fixed. With the base library, the agent used 28.1 million tokens to build a scene that passed those checks. With the three added skills, it used 13.9 million. The two reconstructions matched the reference images about equally well (SSIM of 0.925 and 0.929).

\paragraph{\textbf{Skill Evolution.}}
When a library checks skills only on admission, a flawed skill persists until something revises it. Skill evolution supplies that revision by feeding execution results back into the library. Early systems revised skills in simple loops. LRLL~\citep{survey_ax_2406_18746} periodically turns accumulated experience into new functions, and LYRA~\citep{survey_ax_2509_18597} uses human corrections and retests earlier behavior. Because an agent can also make its own skills worse, recent systems use different tests to decide which revisions to keep. ASPIRE~\citep{survey_ax_2607_00272} debugs programs from per-primitive traces and promotes reusable repairs that pass validation, while RATs~\citep{survey_ax_2606_19419} promotes or deprecates a skill after repeated successes or failures. RoboRSI~\citep{noematrix2026roborsi} traces a failure to the first failing node in a task--skill hierarchy and checks the patch through review, functional tests, and replay of recent runs. SkillOpt~\citep{survey_ax_2605_23904} keeps an edit to a frozen agent's skill document only if its held-out score improves. Reconstruct, Practice, Go Real~\citep{survey_ax_2610_02204} checks for gains and regressions across practice tasks, while VASO~\citep{survey_ax_2606_05395} uses counterexamples from a model checker to revise the skill that produced a plan. Agents can also revise the code around a skill. RHO~\citep{survey_ax_2606_16458} edits whole policy repositories, while Zetta~\citep{survey_ax_2608_16590} diagnoses failed rollouts and tests changes to the code that monitors execution and triggers recovery. Zetta keeps its base VLA frozen.

The tests can also shape which part of a program changes. In RALLY~\citep{survey_ax_2611_37676}, one agent proposes scenes or tasks that the current program fails. The coding agent changes the relevant branch and tests the new behavior. A failed lift, for example, can lead to another grasp at a different depth. RoboEvo$^2$~\citep{survey_ax_2611_34675} tests candidate solutions to failed policy runs, then checks that the selected changes work together before adding new or revised functions to its shared skill library.

Agents can use a failed run to revise a program for later attempts. ASENA~\citep{survey_ax_2609_39207} logs code and motion commands together with the corresponding sensor readings and robot states. In the navigation experiments, agents repeat the same tasks from the same starting poses. Between passes, a separate agent can replay a failed turn and use the traces and ground-truth feedback to revise the shared programs and notes. During frozen evaluation, agents can read these files but cannot change them. Physical RSI on RoboCasa~\citep{project_physical_rsi_robocasa} uses a separate agent to revise instructions and Python functions for the agents running the simulated trials. The project log reports that all 365 tasks were solved at least once over 493 episodes. With the final library, 309 of the 365 tasks succeeded on the first attempt.

EmbodiedRSI~\citep{survey_ax_2610_10498} tests code edits, skill edits, and their combination from the same starting states. On RoboCasa365 Composite-Unseen, success was lowest with code edits alone. It reached 55.6\% with skill edits alone and 71.3\% when the agent could edit both. In Embodied-Harness~\citep{survey_ax_2611_27689}, the Reflector writes conditional instructions for the Actor's next attempt. After a collision during transfer, for example, it tells the Actor to lift the object higher before moving it. On RoboSPA-Long L5, success fell from 59.5\% to 43.0\% when the entire Reflector was disabled. Successive revisions do not always improve task performance. Explore, Execute, Evolve~\citep{survey_ax_2609_37810} reports that performance rose and fell across successive updates. In a 123-round study~\citep{survey_ax_2609_31760}, accepted changes passed replay, paired, and held-out checks, yet the agent never completed the target task. A useful skill library therefore depends as much on how revisions are tested and retired as on how they are generated.

\paragraph{\textbf{From Skill Evolution to Training.}}
All of these mechanisms leave the model that writes skills unchanged, which raises the question of whether skill experience could instead flow into weights. In practice, most systems that do so train the executor rather than the coder. Agentic Skill Discovery~\citep{survey_ax_2405_15019} trains a reinforcement-learning policy for each proposed skill, RoboClaw~\citep{survey_ax_2603_11558} and ENPIRE~\citep{survey_ax_2606_19980} fine-tune robot policies from collected or agent-designed training runs, and HarnessPAI~\citep{survey_ax_2609_29166} and RACaP~\citep{survey_ax_2609_29394} distill harness trajectories into a VLA or a smaller decision model. Training the coder itself is rarer. RoboCodeX~\citep{survey_ax_2402_16117} and Uni-Skill~\citep{survey_ax_2603_02623} fine-tune code-generating models on curated or video-derived programs, and CaP-X~\citep{survey_ax_2603_22435} applies reinforcement learning with environment rewards to a coding model on three simulated tasks before transferring it to a physical arm. Training a multimodal coder continually on its own rollouts, or folding an evolved skill library into its weights, would close the loop between skill evolution and model learning. Related works are compared in Table~\ref{tab:harness-skills}.

% Styled after Table 13 (tables/harness_orchestration.tex); hand-maintained.
% Previous generated version: git history of this file.
\par\Needspace{.4\textheight}
\begingroup
\mechanismtableglue
\newcommand{\SkillGroup}[1]{\rowcolor{TSTHarnessLight}\multicolumn{6}{l}{\textbf{#1}} \\*}
\CompactTableStyle\hyphenpenalty=10000
\renewcommand{\arraystretch}{1.18}
\setlength{\LiteratureWidth}{\dimexpr\linewidth-12\tabcolsep-.6pt\relax}
\setlength{\LTleft}{0pt}
\setlength{\LTright}{0pt}
\setlength{\LTpre}{3pt}
\setlength{\LTpost}{4pt}
\setlength{\LTcapwidth}{\linewidth}
\begin{longtable}{>{\RaggedRight\arraybackslash}p{0.200\LiteratureWidth}>{\centering\arraybackslash}p{0.080\LiteratureWidth}!{\color{black!30}\vrule width .3pt}>{\RaggedRight\arraybackslash}p{0.165\LiteratureWidth}>{\RaggedRight\arraybackslash}p{0.185\LiteratureWidth}>{\RaggedRight\arraybackslash}p{0.175\LiteratureWidth}>{\RaggedRight\arraybackslash}p{0.195\LiteratureWidth}}
\caption[Comparison of works on skills]{Comparison of works on skills (Section~\ref{sec:harness-skills}), grouped by the four parts of the text. Update distinguishes code/library changes from policy training. NR: not reported.}\label{tab:harness-skills}\\
\LiteratureTopRule{Harness}
\rowcolor{TSTHarnessLight}\textbf{Work} & \textbf{Link} & \textbf{Skill Form} & \textbf{Built From} & \textbf{Reuse} & \textbf{Update} \\
\midrule
\endfirsthead
\multicolumn{6}{l}{\tablename~\thetable\ (continued)} \\
\LiteratureTopRule{Harness}
\rowcolor{TSTHarnessLight}\textbf{Work} & \textbf{Link} & \textbf{Skill Form} & \textbf{Built From} & \textbf{Reuse} & \textbf{Update} \\
\midrule
\endhead
\midrule
\multicolumn{6}{r}{Continued on next page} \\
\endfoot
\bottomrule
\endlastfoot
\SkillGroup{From Skills to Code as Policies}
Code as Policies \citep{survey_ax_2209_07753} & \LiteratureLinks{https://arxiv.org/abs/2209.07753}{https://github.com/google-research/google-research/tree/master/code_as_policies} & Python functions & Instructions; API examples & Call earlier functions & Generate missing functions \\
\rowcolor{black!3}
ProgPrompt \citep{survey_ax_2209_11302} & \LiteratureLinks{https://arxiv.org/abs/2209.11302}{https://github.com/NVlabs/progprompt-vh} & Plans with assertions & Instructions; action examples & Compose provided actions & Within-plan recovery; skill revision NR \\
VoxPoser \citep{survey_ax_2307_05973} & \LiteratureLinks{https://arxiv.org/abs/2307.05973}{https://github.com/huangwl18/VoxPoser} & Value-map code & Instructions; visual grounding & Rerun code on new observations & Code \textbf{fixed} within each subtask \\[3pt]
\rowcolor{black!3}
Act-Observe-Rewrite \citep{survey_ax_2603_04466} & \LiteratureLinks{https://arxiv.org/abs/2603.04466}{} & Python controllers & Episode logs; keyframe images & Run controller in the next trial & Rewrite \textbf{between episodes} \\[3pt]
IMR-LLM \citep{survey_ax_2603_02669} & \LiteratureLinks{https://arxiv.org/abs/2603.02669}{} & Process tree; robot programs & Program examples; human checks & Select branches; reuse code snippets & Tree built once for the reported tests \\
\rowcolor{black!3}
MEMENTO \citep{survey_ax_2607_22832} & \LiteratureLinks{https://arxiv.org/abs/2607.22832}{https://github.com/sygkounas/MEMENTO} & Stateful policy programs & Rollout metrics; search history & Run the selected program & Mutate and combine code; retain the elite \\
AGP \citep{survey_ax_2609_12541} & \LiteratureLinks{https://arxiv.org/abs/2609.12541}{} & Task definitions; scripts & Task materials; robot observations & Reload saved task definitions & Revise scripts during execution \\
\rowcolor{black!3}
EmbodiedSWE \citep{survey_ax_2609_27308} & \LiteratureLinks{https://arxiv.org/abs/2609.27308}{https://github.com/EmbodiedSWE/EmbodiedSWE} & Executable solutions & Task materials; simulator trials & Use verified solutions as references & Test and revise task programs \\
RAPID \citep{survey_ax_2609_30249} & \LiteratureLinks{https://arxiv.org/abs/2609.30249}{} & Relational programs & One visual demonstration & Bind a fixed strategy to a new scene & Revise in simulation before deployment \\
\SkillGroup{Skill Libraries and Reuse}
Voyager \citep{survey_ax_2305_16291} & \LiteratureLinks{https://arxiv.org/abs/2305.16291}{https://github.com/MineDojo/Voyager} & Action programs & Exploration; execution feedback & Retrieve relevant programs & Add success-verified programs \\
\rowcolor{black!3}
DROC \citep{survey_ax_2311_10678} & \LiteratureLinks{https://arxiv.org/abs/2311.10678}{https://github.com/Stanford-ILIAD/droc} & Code; correction knowledge & Human corrections & Retrieve task/object knowledge & Revise code; save reusable lessons \\
RoboCoder \citep{survey_ax_2406_03757} & \LiteratureLinks{https://arxiv.org/abs/2406.03757}{} & Action programs & Basic actions; evaluator feedback & Retrieve similar actions & Repair code; add accepted actions \\
\rowcolor{black!3}
Uni-Skill \citep{survey_ax_2603_02623} & \LiteratureLinks{https://arxiv.org/abs/2603.02623}{} & On-demand skill code & Video-derived skill examples & Retrieve examples from SkillFolder & Generate missing skill implementations \\
CaP-X \citep{survey_ax_2603_22435} & \LiteratureLinks{https://arxiv.org/abs/2603.22435}{} & \textbf{9} helper functions & Successful benchmark rollouts & Call the synthesized library & Single library-synthesis pass \\[3pt]
\rowcolor{black!3}
ARCHITECT \citep{survey_ax_2607_23784} & \LiteratureLinks{https://arxiv.org/abs/2607.23784}{https://github.com/robo-architect/architect-franka} & Text rules; code patterns & Human language corrections & Load the full skill library & Append corrections as reusable skills \\
SkillComposer \citep{survey_ax_2608_14944} & \LiteratureLinks{https://arxiv.org/abs/2608.14944}{} & Callable macros & Accepted robot programs & Expand macros into primitive calls & Relearn macros after accepted prompts \\
\rowcolor{black!3}
Teach and Grow \citep{survey_ax_2608_17209} & \LiteratureLinks{https://arxiv.org/abs/2608.17209}{https://github.com/IRMVLab/TGL} & Validated behavior blocks & Successful demonstrations & Ground blocks in the current scene & Refine scope or executor; revalidate \\
LTCLRS \citep{survey_ax_2609_19906} & \LiteratureLinks{https://arxiv.org/abs/2609.19906}{} & Closed-loop software & Demonstrations; simulated trials & Use archived code as examples & Revise code; select on validation \\
\rowcolor{black!3}
RAX \citep{survey_ax_2610_22496} & \LiteratureLinks{https://arxiv.org/abs/2610.22496v1}{https://github.com/robot-agent-exploration/rax} & Python skills & Robot practice & Call with new arguments & Revise and test skill code on the robot \\
EmbodiedEvo \citep{survey_ax_2610_18867} & \LiteratureLinks{https://arxiv.org/abs/2610.18867v2}{https://github.com/Einsia/EmbodiedEvo} & Programs and lessons & Playground trials & Load frozen library in a fresh scene & Revise in practice \\
Chen et al.~\citep{survey_ax_2611_09030} & \LiteratureLinks{https://arxiv.org/abs/2611.09030v1}{} & Task program & General skills; expert scripts & Copy or adapt scripts & Edit using trial feedback \\
\rowcolor{black!3}
ACE-CaP-0~\citep{survey_ax_2611_16872} & \LiteratureLinks{https://arxiv.org/abs/2611.16872v1}{} & Task program & Robot trials; optional demos and simulation & Run on the same task & Freeze after development \\
RoboARCade~\citep{survey_ax_2611_08209} & \LiteratureLinks{https://arxiv.org/abs/2611.08209v2}{https://github.com/Satvik1701/RoboARCade} & Text guide & Successful example & Load in a fresh session & Revise from scored trials \\
% Source: 2611.31098v2, Sections 3.1, 3.3, and 4.4; Figure 5.
\rowcolor{black!3}
Real2Worlds2Real~\citep{survey_ax_2611_31098} & \LiteratureLinks{https://arxiv.org/abs/2611.31098v2}{https://github.com/real2worlds2real/real2worlds2real} & Skill guides & Execution records & Adapt to new tasks & Save verified procedures \\
\SkillGroup{Skill Evolution}
LRLL \citep{survey_ax_2406_18746} & \LiteratureLinks{https://arxiv.org/abs/2406.18746}{} & Python skill functions & Demonstrations; simulated trials & Call library functions & Group similar code; abstract and replay \\
\rowcolor{black!3}
LYRA \citep{survey_ax_2509_18597} & \LiteratureLinks{https://arxiv.org/abs/2509.18597}{https://github.com/Ghiara/LYRA} & Parameterized functions & Human-guided successful trials & Retrieve skills and examples & Extend functions with human feedback \\
VASO \citep{survey_ax_2606_05395} & \LiteratureLinks{https://arxiv.org/abs/2606.05395}{} & Labeling code; plan templates & Task; APIs; formal rules & Guide plan generation & Revise using model-checker feedback \\
\rowcolor{black!3}
RHO \citep{survey_ax_2606_16458} & \LiteratureLinks{https://arxiv.org/abs/2606.16458}{https://github.com/KE7/HELIX} & Policy repositories & Rollouts; failure feedback & Deploy the selected repository & Mutate files; select by rollout scores \\
RATs \citep{survey_ax_2606_19419} & \LiteratureLinks{https://arxiv.org/abs/2606.19419}{https://github.com/Playful-RATs/RATs} & Manipulation functions & Self-directed play & Retrieve the learned library & Curate during play; \textbf{freeze} at test time \\[3pt]
\rowcolor{black!3}
ASPIRE \citep{survey_ax_2607_00272} & \LiteratureLinks{https://arxiv.org/abs/2607.00272}{https://github.com/NVlabs/ASPIRE} & Repair notes; code sketches & Primitive-level execution traces & Load relevant repair guidance & Admit validated, reusable repairs \\
SHAPER \citep{survey_ax_2608_11350} & \LiteratureLinks{https://arxiv.org/abs/2608.11350}{} & Text; context-building code & Target-environment rollouts & Reuse guidance and context builder & Test edits; retain top candidates \\
\rowcolor{black!3}
Zetta $\zeta$ \citep{survey_ax_2608_16590} & \LiteratureLinks{https://arxiv.org/abs/2608.16590}{https://github.com/air-embodied-brain/Zetta-Embodiment} & Critic, recovery and tool packages & Failure traces; validation rollouts & Deploy versioned harness packages & Patch, validate and consolidate repairs \\
WSRSI \citep{survey_ax_2609_31760} & \LiteratureLinks{https://arxiv.org/abs/2609.31760}{} & Skill code; model adapters & Recorded failures & Call skills through typed interfaces & Replay failures; test on fresh tasks \\
\rowcolor{black!3}
RoboSkill \citep{survey_ax_2609_37810} & \LiteratureLinks{https://arxiv.org/abs/2609.37810}{https://github.com/SII-dannyXSC/RoboSkill} & Text; code; supporting records & Exploration and execution records & Select relevant skill packages & Consolidate skills after agent exit \\*[0pt]
RPG \citep{survey_ax_2610_02204} & \LiteratureLinks{https://arxiv.org/abs/2610.02204}{} & Skill code; system prompts & Simulated practice; failure diagnosis & Deploy the retained system & Test across tasks; merge and retest \\
\rowcolor{black!3}
Physical RSI on RoboCasa \citep{project_physical_rsi_robocasa} & \LiteratureLinks{https://stacx-lab.github.io/physical-rsi/}{} & Guides and functions & Project trial logs & Workers load current library & Principal revises library \\
ASENA \citep{survey_ax_2609_39207} & \LiteratureLinks{https://arxiv.org/abs/2609.39207v2}{} & Programs and notes & Sensor and command traces & Load workspace on next pass & Replay failure and edit program \\
\rowcolor{black!3}
EmbodiedRSI \citep{survey_ax_2610_10498} & \LiteratureLinks{https://arxiv.org/abs/2610.10498v2}{} & Code and skills & Matched trials & Retrieve supported changes & Revise code and skills from trial outcomes \\
Embodied-Harness \citep{survey_ax_2611_27689} & \LiteratureLinks{https://arxiv.org/abs/2611.27689v1}{} & Conditional instructions & Attempt traces & Pass guidance to Actor & Reflector revises after attempt \\
RALLY~\citep{survey_ax_2611_37676} & \LiteratureLinks{https://arxiv.org/abs/2611.37676v1}{} & Code branches & Failed scenes and tasks & Run revised policy & Test branch changes \\
\rowcolor{black!3}
RoboEvo$^2$~\citep{survey_ax_2611_34675} & \LiteratureLinks{https://arxiv.org/abs/2611.34675v1}{https://github.com/kaiknower/RoboEvo2} & Skill functions & Verified solutions & Call by name & Test combined revisions \\
\SkillGroup{From Skill Evolution to Training}
RoboCodeX \citep{survey_ax_2402_16117} & \LiteratureLinks{https://arxiv.org/abs/2402.16117}{} & Object-centered code units & Multimodal code training & Compose motion primitives & Skill-library revision NR \\
\rowcolor{black!3}
Agentic Skill Discovery \citep{survey_ax_2405_15019} & \LiteratureLinks{https://arxiv.org/abs/2405.15019}{https://github.com/xf-zhao/Agentic-Skill-Discovery} & Learned policies & Generated rewards; RL & Compose learned skills & Train missing skills on demand \\
RoboClaw \citep{survey_ax_2603_11558} & \LiteratureLinks{https://arxiv.org/abs/2603.11558}{https://github.com/RoboClaw-Robotics/RoboClaw} & Tool-use procedures & Modular skills; trained policies & Compose tool and policy calls & Procedure revision NR \\
\rowcolor{black!3}
ENPIRE \citep{survey_ax_2606_19980} & \LiteratureLinks{https://arxiv.org/abs/2606.19980}{https://github.com/NVlabs/ENPIRE} & Policy-learning code & Physical trials; automated rewards & Reuse successful training recipes & Edit code; train and retest policies \\
AgenticRobotics \citep{survey_ax_2608_07555} & \LiteratureLinks{https://arxiv.org/abs/2608.07555}{} & Skill procedures; tool records & Round logs; measured outcomes & Select skill--tool pairs & Registry promotion: live use \textbf{untested} \\[3pt]
\rowcolor{black!3}
HarnessPAI \citep{survey_ax_2609_29166} & \LiteratureLinks{https://arxiv.org/abs/2609.29166}{} & Repair notes; linked code & Between-rollout failure analysis & Retrieve repairs by workflow node & Evolve code; save failure--repair pairs \\
RACaP \citep{survey_ax_2609_29394} & \LiteratureLinks{https://arxiv.org/abs/2609.29394}{https://github.com/Robo-Harness/racap} & Policy API library & Development-task rollouts & Call APIs with runtime arguments & Paired testing; code \textbf{frozen} at deployment \\[3pt]
\end{longtable}
\endgroup

\par\Needspace{12\baselineskip}
\begingroup
\mechanismtableglue
\CompactTableStyle
\setlength{\tabcolsep}{3pt}
\setlength{\LiteratureWidth}{\dimexpr\linewidth-8\tabcolsep\relax}
\setlength{\LTcapwidth}{\linewidth}
\captionsetup{singlelinecheck=false,justification=raggedright}
\LiteratureContinuationSetup
\begin{longtable}{@{}>{\RaggedRight\arraybackslash}p{.195\LiteratureWidth}>{\RaggedRight\arraybackslash}p{.145\LiteratureWidth}>{\RaggedRight\arraybackslash}p{.375\LiteratureWidth}>{\raggedleft\arraybackslash}p{.155\LiteratureWidth}>{\raggedleft\arraybackslash}p{.13\LiteratureWidth}@{}}
\caption[Task success with and without prior experience]{Task success with and without prior experience. For Real2Worlds2Real, both columns show mean success across all candidate programs.}\label{tab:retained-experience}\\
\LiteratureTopRule{Harness}
\rowcolor{TSTHarnessLight}\textbf{Study} & \textbf{Reused material} & \textbf{Evaluation tasks} & \textbf{Without reuse (\%)} & \textbf{With reuse (\%)} \\
\midrule
\endfirsthead
\multicolumn{5}{@{}l}{\tablename~\thetable\ (continued)} \\
\LiteratureTopRule{Harness}
\rowcolor{TSTHarnessLight}\textbf{Study} & \textbf{Reused material} & \textbf{Evaluation tasks} & \textbf{Without reuse (\%)} & \textbf{With reuse (\%)} \\
\midrule
\endhead
\midrule
\multicolumn{5}{r@{}}{Continued on next page} \\
\endfoot
\bottomrule
\endlastfoot
% CaP-X, Figure 8. M4 and M4 + skill library, before parallel reasoning.
CaP-Agent0~\citep{survey_ax_2603_22435} & Code & Same seven tasks used to build the skill library & 55 & \textbf{59} \\
% AMP1+, Figure 4F and Section 4E. Same-task practice, held-out layouts.
\rowcolor{black!3}
AMP1+~\citep{survey_ax_2611_09892} & Code and notes & New layouts of 42 practiced tasks & 26.4 & \textbf{84.1} \\
% EmbodiedEvo, Appendix C, Table 11. No experience vs. related-task sources.
EmbodiedEvo~\citep{survey_ax_2610_18867} & Code and notes & 12 new tasks after practice on related ones & 1.67 & \textbf{16.39} \\
% Real2Worlds2Real, Section 4.4 and Figure 5(b), mean over all candidate programs.
\rowcolor{black!3}
Real2Worlds2Real~\citep{survey_ax_2611_31098} & 136 skills & 20 new LIBERO-PRO Long tasks & 44.1 & \textbf{69.1} \\
\end{longtable}
\endgroup

\subsection{Infrastructure}
\label{sec:harness-infrastructure}

Beneath skills, every harness function rests on software infrastructure: the interface through which it reaches robot capabilities, the perception and manipulation tools behind that interface, the services that run learned policies, and the checks that stand between a proposed action and the hardware. In the Infrastructure panel of Figure~\ref{fig:harness-operation}, \texttt{open\_drawer} may expand into a segmentation call that finds the handle, a grasp generator that proposes a pose, a motion planner that reaches it, and a VLA served on a remote GPU that performs the pull. Each of these components can fail, add latency, or hide information from the agent. We discuss the interface first, then the toolchains and policy services that dominate practical systems, and finally runtime gating.

\paragraph{\textbf{Capability Interfaces.}}
The first layer of this infrastructure is the interface, which fixes what an agent can express and what it learns back. EmbodiedTool~\citep{survey_ax_2605_26637} registers more than a hundred tools with capability cards that state inputs, outputs, and trigger conditions; agents form valid calls almost always but choose the right tool far less often. The return channel matters as much as the call: NavMCP~\citep{survey_ax_2608_30396} combines structured intent, keyframe-grounded journey summaries, and persistent memory; replacing this complete three-channel protocol with an episodic interface lowers question-answering accuracy from 74.0\% to 59.1\%. Granularity is a further choice. Show-Harness~\citep{survey_ax_2609_10522} exposes semantic Cartesian steps that an interpreter projects into safe setpoints, and halving the step size raises zero-shot success from 60\% to 82\%. The Model Context Protocol is becoming a common transport: RoboNeuron~\citep{survey_ax_2512_10394} derives typed MCP tools from ROS~2 message schemas, and the ROS MCP Server~\citep{robotmcp_rosmcpserver} exposes topics, services, and actions through rosbridge. On the robot side, Cardenas et al.'s ROSClaw~\citep{survey_ax_2603_26997} checks each call against velocity bounds and an interface allowlist before dispatch, and EMOS~\citep{survey_ax_2410_22662} derives a capability profile from the robot's URDF and forward kinematics rather than from hand-written text.

As the robot moves, the agent needs to know when it observed the scene. In uni-robot-agent~\citep{survey_ax_2611_01560}, each observation groups the image, camera calibration, and robot state from the same moment. A blocking move returns execution feedback. An asynchronous call returns a handle immediately, so the agent can continue reasoning while the robot moves. Commands using the same arm still run in order. If new evidence changes the plan, the agent can cancel commands still waiting in the queue. On five selected tasks with GPT-6 Astra, the authors timed one run per setting at the same seed and layout, excluding simulator initialization. Pipelining shortened episode time by 4.8--73.7\% relative to serial execution.

An interface also determines which calculations the agent can perform itself. JustRUA~\citep{survey_ax_2611_16009} gives the model camera images and robot readings. It can write code to compute a target in robot coordinates from image measurements, then submit a motion command as JSON. RoboDojo uses its motion planner and inverse kinematics to execute the command. The interface also determines what the agent learns from the result. CaPtain-Harness~\citep{survey_ax_2610_40799} exposes the same Python calls in simulation and on the robot. Each call returns a text record that can include the measured pose, contact force, or reason a motion stopped.

RoboWild~\citep{xlanglab2026robowild} offers two interfaces that give the agent different control over a planned motion. Both interfaces allow 100 robot calls on short tasks and 200 on long tasks. ReAct plans and executes a motion in one call. Claude Code uses separate calls, so the agent can inspect a plan and change the target before execution. Table~\ref{tab:benchmark-protocols} reports success for both interfaces on the same Dexmate setups.

\paragraph{\textbf{Perception and Manipulation Toolchains.}}
Behind the interface, code-as-policy harnesses act through a stack of pretrained tools: open-vocabulary detection and segmentation such as Grounding DINO~\citep{survey_ax_2303_05499}, SAM~2~\citep{survey_ax_2408_00714}, and SAM~3~\citep{survey_ax_2511_16719}; pointing models such as Molmo~\citep{survey_ax_2409_17146}; pose and depth estimation such as FoundationPose~\citep{survey_ax_2312_08344} and FoundationStereo~\citep{survey_ax_2501_09898}; grasp generators such as AnyGrasp~\citep{survey_ax_2212_08333}, Contact-GraspNet~\citep{survey_ax_2103_14127}, M2T2~\citep{survey_ax_2311_00926}, and GraspGen~\citep{survey_ax_2507_13097}; and GPU motion planners such as cuRobo~\citep{survey_ax_2310_17274}. CaP-X~\citep{survey_ax_2603_22435} exposes such a stack as stateless services behind a Python API, and its benchmark separates tiers that receive privileged simulator masks and poses from tiers that must perceive from RGB-D. GaP~\citep{survey_ax_2607_05369} organizes 51 perception, grasping, motion, and verification skills behind single gRPC methods with shared types for poses, masks, point clouds, and trajectories, and translates them to ROS for physical execution. Tools are also beginning to ship agent-facing endpoints: GraspGen provides a GPU server and an MCP wrapper so that an LLM agent can request grasps without a local model. Physical Agency~\citep{survey_ax_2607_21725} illustrates a complete physical deployment: detection, segmentation, stereo depth, grasp generation, and task-and-motion planning run as servers on one workstation, while a real-time control computer drives~the~arm.

These toolchains determine much of what a harness can achieve. Perception is the dominant source of physical failure in several systems: most real-robot errors of VoxPoser~\citep{survey_ax_2307_05973} come from perception, point tracking is the largest error source in ReKep~\citep{survey_ax_2409_01652}, and OK-Robot~\citep{survey_ax_2401_12202} shows how errors compound when off-the-shelf detection, segmentation, mapping, and grasping are chained without recovery. Privileged state hides this cost. Kuroki and Tang~\citep{survey_ax_2609_19906} report 65.0\% success with full simulator observations but 22.5\% with RGB alone, and CaP-X trains its coding model on the privileged tier to avoid perception noise before transferring to perceived inputs. An agent can repair a toolchain when the toolchain reports its failures. ASPIRE~\citep{survey_ax_2607_00272} provides an execution engine with primitive-level traces, debugging, and repair validation; adding the engine raises macro-average success on the evaluated LIBERO-PRO perturbations from 14\% to 62\%. Physical Agency returns typed failures and overrides a reported grasp when the gripper-width sensor shows it is empty. Across systems, coding agents repeatedly add the same hardening around tools, including fallback chains from text-prompted to point-prompted segmentation, filters on grasp orientation, inverse-kinematics checks before grasping, and sensor-checked grasps; RHO~\citep{survey_ax_2606_16458} and ASPIRE move such logic into the tool wrappers themselves. Tool interfaces that report uncertainty and failure, together with evaluations that separate privileged from perceived inputs, therefore make toolchain results comparable across harnesses.

\paragraph{\textbf{Policy Inference and Serving.}}
Learned policies form the other half of this stack: when a harness calls a VLA, the policy becomes a service with its own latency and capacity. The openpi remote-inference interface~\citep{openpi2025remote} serves a checkpoint over a WebSocket that returns action chunks, keeping robot code free of model dependencies, and LeRobot~\citep{survey_ax_2602_22818} separates a policy server from a robot client that requests a new chunk when its action queue runs low. Asynchronous execution hides inference delay: SmolVLA~\citep{survey_ax_2506_01844} uses this design to cut task time by about 30\% at a small cost in success, real-time chunking~\citep{survey_ax_2506_07339,survey_ax_2512_05964} freezes actions committed during inference, and VLASH~\citep{survey_ax_2512_01031} conditions on the state those actions will reach. Placement is a further design dimension. VLA-Perf~\citep{survey_ax_2602_18397} models latency across devices and networks and finds that, for $\pi_0$, an edge server over WiFi can outperform on-device inference; Gemini Robotics~\citep{survey_ax_2503_20020} combines a cloud backbone with an on-robot decoder, and RoboECC~\citep{survey_ax_2603_20711} partitions a VLA between edge and cloud according to predicted bandwidth. For fleets, ROSA~\citep{survey_ax_2607_01088} schedules VLA and VLM components on a shared GPU pool under latency targets, Robion~\citep{survey_ax_2609_12075} partitions one GPU between the vision-language and action stages to serve about twenty robots, and Armory~\citep{survey_ax_2608_00337} schedules batched action-chunk inference across robots running at different rates. Runtime engineering narrows the remaining gap: kernel-level optimization brings $\pi_0$ to 27\,ms on a consumer GPU~\citep{survey_ax_2510_26742}, OxyGen~\citep{survey_ax_2603_14371} shares a VLA's cache between action and language heads, vla.cpp~\citep{survey_ax_2606_08094} runs quantized VLAs on embedded boards, and BitVLA~\citep{survey_ax_2506_07530} matches a 4-bit baseline with ternary weights. These systems are mostly evaluated on policy throughput; coupling them with agent harnesses will make serving latency part of the planning budget.

\paragraph{\textbf{Runtime Gating and Versioning.}}
Once tools and policies are reachable, the harness must also decide which proposed commands may reach the hardware, and this check is best placed outside the model. RoboPAIR~\citep{survey_ax_2410_13691} motivates this placement by jailbreaking a self-driving language model, a GPT-4o-planned ground robot, and a commercial quadruped in every attempted scenario. Safety Chip~\citep{survey_ax_2309_09919} compiles expert-verified temporal-logic constraints into an automaton that prunes unsafe actions before execution, and RoboGuard~\citep{survey_ax_2503_07885} model-checks plans against specifications derived by an isolated model, reducing jailbreak success from 92.3\% to 2.3\%. SafeGate~\citep{survey_ax_2604_05427} fails closed by deferring to the user whenever a hazard is not covered by a verified template, and RoboSafe~\citep{survey_ax_2512_21220} checks each action against generated executable predicates. Other systems gate at the interface: ETA~\citep{survey_ax_2608_03924} separates what the agent wants to do from what the interface permits, SafeHarness~\citep{survey_ax_2609_20822} rejects routes that violate obstacle clearance, and a constrained-LLM design~\citep{survey_ax_2608_29379} lets the model state only symbolic intent. CORE~\citep{survey_ax_2602_19983} instead enforces inferred constraints in the controller through control barrier functions. The same gates also apply when the harness itself changes. ETA requires a revised harness to pass contract checks, isolated review, replay, and held-out evaluation, and in its reported runs no candidate passed; Harness Robotic OS~\citep{survey_ax_2609_11225} proposes versioned, rollback-capable updates without demonstrating a revised executable. Related works are compared in Table~\ref{tab:harness-infrastructure}.

\par\Needspace{12\baselineskip}
\begingroup
\mechanismtableglue
\CompactTableStyle
\setlength{\LiteratureWidth}{\dimexpr\linewidth-8\tabcolsep\relax}

\setlength{\LTpre}{4pt}
\setlength{\LTpost}{4pt}
\setlength{\LTcapwidth}{\linewidth}
\captionsetup{singlelinecheck=false,justification=raggedright}
\LiteratureContinuationSetup
\begin{longtable}{@{}>{\RaggedRight\arraybackslash}p{0.2128\LiteratureWidth}>{\centering\arraybackslash}p{0.0750\LiteratureWidth}>{\RaggedRight\arraybackslash}p{0.1665\LiteratureWidth}>{\RaggedRight\arraybackslash}p{0.1943\LiteratureWidth}>{\RaggedRight\arraybackslash}p{0.3515\LiteratureWidth}@{}}
\caption[Comparison of works on infrastructure]{Comparison of works on infrastructure (Section~\ref{sec:harness-infrastructure}). Includes supporting tools and deployment systems, which need not be autonomous agents.}\label{tab:harness-infrastructure}\\
\LiteratureTopRule{Harness}
\rowcolor{TSTHarnessLight}\textbf{Work} & \textbf{Link} & \textbf{Purpose} & \textbf{Works on} & \textbf{Main operation} \\
\midrule
\endfirsthead
\multicolumn{5}{@{}l}{\tablename~\thetable\ (continued)} \\
\LiteratureTopRule{Harness}
\rowcolor{TSTHarnessLight}\textbf{Work} & \textbf{Link} & \textbf{Purpose} & \textbf{Works on} & \textbf{Main operation} \\
\midrule
\endhead
\midrule
\multicolumn{5}{r@{}}{Continued on next page} \\
\endfoot
\bottomrule
\endlastfoot
% Source: https://arxiv.org/abs/2309.09919 | §III-C–E: temporal-logic monitoring and corrective feedback
Safety Chip \mbox{\citep{survey_ax_2309_09919}} & \LiteratureLinks{https://arxiv.org/abs/2309.09919}{https://github.com/YzyLmc/ltl_safety} & Action safety & Proposed actions & Check temporal rules; block violations and reprompt \\[0pt]
% Source: https://arxiv.org/abs/2401.12202 | §III: open-vocabulary navigation, grasping and placing modules
\rowcolor{black!3}
OK-Robot \mbox{\citep{survey_ax_2401_12202}} & \LiteratureLinks{https://www.roboticsproceedings.org/rss20/p091.html}{https://github.com/ok-robot/ok-robot} & Mobile manipulation & Navigation; grasping; placement & Connect pretrained tools through a semantic map \\[0pt]
% Source: https://arxiv.org/abs/2410.13691v2 | §4.3–4.4; Algorithm 1: attacker, target, judge and syntax checker
RoboPAIR \mbox{\citep{survey_ax_2410_13691}} & \LiteratureLinks{https://ieeexplore.ieee.org/document/11128119/}{https://github.com/arobey1/robopair} & Adversarial testing & Robot-language interfaces & Refine attack prompts using judges and API checks \\[0pt]
% Source: https://arxiv.org/abs/2503.20020 | §5: semantic action safety, ASIMOV, constitutions and post-training
\rowcolor{black!3}
Gemini Robotics \mbox{\citep{survey_ax_2503_20020}} & \LiteratureLinks{https://arxiv.org/abs/2503.20020}{} & Safety evaluation & Embodied model responses & Test with ASIMOV; add constitutions and safety training \\[0pt]
% Source: https://arxiv.org/abs/2506.01844 | §3.3; Algorithm 1: client/server inference and overlapping action queues
SmolVLA \mbox{\citep{survey_ax_2506_01844}} & \LiteratureLinks{https://arxiv.org/abs/2506.01844}{https://github.com/huggingface/lerobot} & Continuous execution & Action chunks & Overlap server inference with queued robot actions \\[0pt]
% Source: https://arxiv.org/abs/2506.07339v2 | §3.1–3.3: inference-time inpainting and soft-masked overlap
\rowcolor{black!3}
RTC \mbox{\citep{survey_ax_2506_07339}} & \LiteratureLinks{https://proceedings.neurips.cc/paper_files/paper/2025/hash/300ccb2187dedd4edcc07f7e76d8e553-Abstract-Conference.html}{https://github.com/Physical-Intelligence/real-time-chunking-kinetix} & Continuous execution & Action chunks & Freeze committed actions; fill in the next chunk \\[0pt]
% Source: https://arxiv.org/abs/2510.26742 | §3.2–3.3; §4: weight/operator fusion and kernel optimization
Running VLAs at Real-time Speed \mbox{\citep{survey_ax_2510_26742}} & \LiteratureLinks{https://arxiv.org/abs/2510.26742}{https://github.com/Dexmal/realtime-vla} & Inference acceleration & VLA GPU computation & Fuse operations and optimize GPU kernels \\[0pt]
% Source: https://arxiv.org/abs/2512.01031 | §IV-A: state roll-forward using the already queued actions
\rowcolor{black!3}
VLASH \mbox{\citep{survey_ax_2512_01031}} & \LiteratureLinks{https://arxiv.org/abs/2512.01031}{https://github.com/mit-han-lab/vlash} & Delay compensation & VLA state inputs & Condition on the predicted execution-time robot state \\[0pt]
% Source: https://arxiv.org/abs/2512.05964 | §IV; Algorithm 1: training-time action-prefix conditioning
Training-time RTC \mbox{\citep{survey_ax_2512_05964}} & \LiteratureLinks{https://arxiv.org/abs/2512.05964}{} & Delay compensation & Action-chunk policies & Train on committed action prefixes and sampled delays \\[0pt]
% Source: https://arxiv.org/abs/2512.10394v2 | §III-A–B: control/data planes, direct and persistent execution paths
\rowcolor{black!3}
RoboNeuron \mbox{\citep{survey_ax_2512_10394}} & \LiteratureLinks{https://arxiv.org/abs/2512.10394}{https://github.com/guanweifan/RoboNeuron} & Robot middleware & Tool calls; robot streams & Separate tool orchestration from streaming execution \\[0pt]
% Source: https://arxiv.org/abs/2503.07885 | §III-A–B; Algorithm 1: safety reasoning and specification-constrained synthesis
RoboGuard \mbox{\citep{survey_ax_2503_07885}} & \LiteratureLinks{https://arxiv.org/abs/2503.07885}{https://github.com/KumarRobotics/RoboGuard} & Plan safety & LLM-generated plans & Check plans against context-specific temporal rules \\[0pt]
% Source: https://arxiv.org/abs/2506.07530 | §III-A–B: quantized linear layers, custom kernels and distillation
\rowcolor{black!3}
BitVLA \mbox{\citep{survey_ax_2506_07530}} & \LiteratureLinks{https://arxiv.org/abs/2506.07530}{https://github.com/ustcwhy/BitVLA} & Model compression & VLA weights; activations & Quantize weights to ternary values and activations to INT8 \\[0pt]
% Source: https://arxiv.org/abs/2512.21220 | §4.1–4.3: backward reflection, forward prediction and executable safety logic
RoboSafe \mbox{\citep{survey_ax_2512_21220}} & \LiteratureLinks{https://openreview.net/pdf?id=wyKCkQ2GyO}{} & Action safety & Proposed actions; recent history & Run generated safety checks; block or request replanning \\[0pt]
% Source: https://arxiv.org/abs/2602.18397 | §3: roofline-based analytical model; estimates are not measured runtime
\rowcolor{black!3}
VLA-Perf \mbox{\citep{survey_ax_2602_18397}} & \LiteratureLinks{https://arxiv.org/abs/2602.18397}{https://github.com/NVlabs/vla-perf} & Deployment analysis & Model--hardware choices & Estimate inference latency and throughput analytically \\[0pt]
% Source: https://arxiv.org/abs/2602.19983 | §IV-A–C: visual predicates, spatial grounding and control-barrier-function filter
CORE \mbox{\citep{survey_ax_2602_19983}} & \LiteratureLinks{https://arxiv.org/abs/2602.19983}{} & Motion safety & Controller outputs & Ground visual safety regions; filter control commands \\[0pt]
% Source: https://arxiv.org/abs/2602.22818 | §3.1–3.5: hardware middleware, datasets, learning and inference
\rowcolor{black!3}
LeRobot \mbox{\citep{survey_ax_2602_22818}} & \LiteratureLinks{https://openreview.net/pdf?id=CiZMMAFQR3}{https://github.com/huggingface/lerobot} & Robot-learning workflow & Hardware; data; policies & Unify data collection, training and deployment interfaces \\[0pt]
% Source: https://arxiv.org/abs/2603.14371 | §3.1–3.3: shared KV cache and asymmetric expert deadlines
OxyGen \mbox{\citep{survey_ax_2603_14371}} & \LiteratureLinks{https://arxiv.org/abs/2603.14371}{https://github.com/air-embodied-brain/OxyGen} & Shared inference & Action and language experts & Reuse observation caches; prioritize action deadlines \\[0pt]
% Source: https://arxiv.org/abs/2603.20711 | §IV-A–B: structure/hardware-aware segmentation and bandwidth-aware adjustment
\rowcolor{black!3}
RoboECC \mbox{\citep{survey_ax_2603_20711}} & \LiteratureLinks{https://arxiv.org/abs/2603.20711}{https://github.com/zhengzihaoPKU/RoboECC} & Edge--cloud deployment & VLA model partitions & Choose hardware-aware splits; adjust to network changes \\[0pt]
% Source: https://arxiv.org/abs/2604.05427 | §IV-A–F: pre-execution checks and runtime safety-contract stages
SafeGate \mbox{\citep{survey_ax_2604_05427}} & \LiteratureLinks{https://arxiv.org/abs/2604.05427}{} & Task safety & Task requests; execution & Screen commands first; monitor task contracts at runtime \\[0pt]
% Source: https://arxiv.org/abs/2606.08094 | §III-A–C: common action interface, cached conditioning and robot integration
\rowcolor{black!3}
vla.cpp \mbox{\citep{survey_ax_2606_08094}} & \LiteratureLinks{https://arxiv.org/abs/2606.08094}{https://github.com/VinRobotics/vla.cpp} & Portable inference & VLA models; robot clients & Run model adapters through a shared C++ runtime \\[0pt]
% Source: https://arxiv.org/abs/2607.01088 | §3.2–3.4: centralized model serving and calibrated scheduling
ROSA \mbox{\citep{survey_ax_2607_01088}} & \LiteratureLinks{https://arxiv.org/abs/2607.01088}{} & Fleet inference & Requests from multiple robots & Pool model serving and batch requests across robots \\[0pt]
% Source: https://arxiv.org/abs/2607.05369 | §3–4.2: typed graph nodes, MORSL skills and simulation rehearsal
\rowcolor{black!3}
GaP \mbox{\citep{survey_ax_2607_05369}} & \LiteratureLinks{https://arxiv.org/abs/2607.05369}{https://github.com/graph-robots/graph-as-policy} & Reusable automation & Robot skill graphs & Rehearse and refine graphs; deploy an edge interpreter \\[0pt]
% Source: https://arxiv.org/abs/2608.00337 | §3.1–3.3: software mirrors, lookahead scheduler and serving engine
Armory \mbox{\citep{survey_ax_2608_00337}} & \LiteratureLinks{https://arxiv.org/abs/2608.00337}{https://github.com/GaTech-RL2/armory} & Fleet scheduling & Queued action chunks & Plan inference batches using robot execution mirrors \\[0pt]
% Source: https://arxiv.org/abs/2608.03924 | §3.1–3.2; §4.1: interface checks, single-action invariant and trusted receipts
\rowcolor{black!3}
ETA \mbox{\citep{survey_ax_2608_03924}} & \LiteratureLinks{https://arxiv.org/abs/2608.03924}{https://github.com/OpenMOSS/OpenETA} & Checked execution & Agent tool calls & Gate commands; return execution records and fresh observations \\[0pt]
% Source: https://arxiv.org/abs/2608.29379 | §3.1–3.4: schema-constrained protocol and kinematic/collision checks
Bridging Semantics and Physics \mbox{\citep{survey_ax_2608_29379}} & \LiteratureLinks{https://arxiv.org/abs/2608.29379}{} & Checked execution & Pouring tool calls & Validate arguments, then check motion-template feasibility \\[0pt]
% Source: https://arxiv.org/abs/2609.10522 | §3.1–3.2: discrete movement/gripper tokens and embodiment-specific interpreters
\rowcolor{black!3}
Show-Harness \mbox{\citep{survey_ax_2609_10522}} & \LiteratureLinks{https://arxiv.org/abs/2609.10522}{https://github.com/showlab/Show-Harness} & Robot action interface & VLM action tokens & Translate semantic actions into robot control deterministically \\[0pt]
% Source: https://arxiv.org/abs/2609.11225 | §3.1–3.3: hardware abstraction, autonomy services and cognitive runtime
Harness Robotic OS \mbox{\citep{survey_ax_2609_11225}} & \LiteratureLinks{https://arxiv.org/abs/2609.11225}{} & Inspection runtime & Robot services; agent context & Connect autonomy skills through shared context and interfaces \\*[0pt]
% Source: https://arxiv.org/abs/2609.12075 | §4.1.1; §4.2.1–4.2.2: stage overlap, model placement and traffic assignment
\rowcolor{black!3}
Robion \mbox{\citep{survey_ax_2609_12075}} & \LiteratureLinks{https://arxiv.org/abs/2609.12075}{} & Fleet inference & Multiple VLA models & Overlap model stages; assign robot traffic across GPUs \\[0pt]
uni-robot-agent \citep{survey_ax_2611_01560} & \LiteratureLinks{https://arxiv.org/abs/2611.01560v2}{https://github.com/Open-RUA/uni-robot-agent} & Robot-use runtime & Observations and tool calls & Record state with images and queue calls that need the same exclusive resource \\
\rowcolor{black!3}
RoboWild-Harness \citep{xlanglab2026robowild} & \LiteratureLinks{https://huggingface.co/datasets/xlangai/RoboWild/blob/40a24c8f22fe567da47de49a8cddba7c51ae5560/RoboWild.pdf}{https://github.com/xlang-ai/RoboWild-Harness} & Robot-use interfaces & Robot motion commands & ReAct combines planning and execution; Claude Code separates them on Dexmate \\
JustRUA~\citep{survey_ax_2611_16009} & \LiteratureLinks{https://arxiv.org/abs/2611.16009v1}{https://github.com/ZefanW/JustRUA} & Robot-use interface & Images; robot readings & Compute targets; call motion planner \\
\rowcolor{black!3}
CaPtain-Harness~\citep{survey_ax_2610_40799} & \LiteratureLinks{https://arxiv.org/abs/2610.40799v1}{https://github.com/waltstephen/CaPtain} & Robot-use interface & Python motion calls & Return measured state and stop reason \\
\end{longtable}
\endgroup

\subsection{Layered Systems}
\label{sec:harness-layered-systems}

Layered systems organize the tools and policies above into layers that run at different rates. In the Layered Systems panel of Figure~\ref{fig:harness-operation}, a planner decides to open the drawer, a VLA reaches for the handle, and an inverse-kinematics solver tracks the pull, each at its own rate. We describe this stack as three layers. At \emph{System~2}, VLMs, LLMs, or generated programs plan over tasks and subgoals. At \emph{System~1}, learned visuomotor policies, typically VLAs, react at control rate. \emph{System~0} covers analytic control, including inverse kinematics, motion planning, model predictive control, and whole-body control. Systems~1 and~2 follow the fast--slow distinction of dual-process theories~\citep{stanovich2000individual,kahneman2011thinking}. Taniguchi et al.~\citep{survey_ax_2503_06138} use System~0 for morphological computation. We use it here for explicit analytic control. We avoid System~3 because it already refers both to collective intelligence and to a runtime critic~\citep{survey_ax_2603_05185}. Existing VLA surveys contrast monolithic with hierarchical models~\citep{survey_ax_2508_13073} but leave analytic control outside the comparison. The harness is what connects the three layers, and this subsection examines what it passes between them, how often, and which layer holds authority.

\paragraph{\textbf{Layer Interfaces and Rates.}}
The first design choice is what each layer passes downward and how often. Helix~\citep{figure2025helix} writes a single latent vector from a vision-language model at 7--9\,Hz to a 200\,Hz policy. In GR00T~N1~\citep{survey_ax_2503_14734}, a 10\,Hz reasoning module passes embeddings to an action module that generates motor actions at 120\,Hz. The simulated GR-1 tabletop setup uses a whole-body inverse-kinematics controller. Ground Slow, Move Fast~\citep{survey_ax_2512_08186} makes all three layers explicit: a 2\,Hz planner outputs a pixel goal, a 30\,Hz policy produces waypoints, and model predictive control tracks them. Other systems pass interpretable signals. Hi Robot~\citep{survey_ax_2502_19417} sends language subtasks about once per second or when a user interjects, $\pi_{0.5}$~\citep{survey_ax_2504_16054} decodes subtask text before actions within one model, NaVILA~\citep{survey_ax_2412_04453} sends parameterized language commands to a locomotion policy, and HAMSTER~\citep{survey_ax_2502_05485} sends a few-point image path once per episode. In a controlled comparison, VLA-OS~\citep{survey_ax_2506_17561} finds that low-level policies follow visual plans more faithfully than language plans. Because the slow layer's output is stale when used, RoboDual~\citep{survey_ax_2410_08001} and HiRT~\citep{survey_ax_2410_05273} train the fast layer on delayed conditioning, and HarnessWAM~\citep{survey_ax_2608_09516} triggers deliberation on progress events, which outperforms invoking the planner at fixed intervals. Latent interfaces carry more information but are opaque to monitoring; text, paths, and waypoints can be checked at the cost of what they can express.

Separate layers also make it possible to trace a failure. DynaHarness~\citep{survey_ax_2609_40306} uses the execution record to distinguish a request whose target could not be located from a skill that ran and failed. The system tests changes to a skill or its execution logic in paired trials and runs broader regression checks before accepting~them.

\paragraph{\textbf{Agents Calling VLAs as Tools.}}
A layered system can also use separate models and expose the VLA as one tool among others. Gemini Robotics~1.5~\citep{survey_ax_2510_03342} makes its action model available to an orchestrating reasoning model through language instructions, and replacing a general model with the embodied-reasoning orchestrator halves the total failure rate. Harness VLA~\citep{survey_ax_2607_08448} places a frozen VLA behind a \texttt{vla\_act} primitive next to analytic motion primitives and verifies each call; the VLA accounts for a minority of calls on its manipulation benchmarks. Physical Agency~\citep{survey_ax_2607_21725} lets an agent choose between task-and-motion planning and a VLA, and overrides reported successes that contradict grasp sensing. VLAs-as-Tools~\citep{survey_ax_2605_13119} returns progress signals that let a monitor advance or replan with a few planner calls per episode. A systematic study of hierarchical VLA agents~\citep{survey_ax_2606_10267} finds that termination rules and VLA steerability matter as much as planner size. Compared with an integrated dual-system model, this arrangement keeps each layer replaceable and each call inspectable, but every handoff must re-establish state and intent, as discussed under Orchestration. At each handoff, the harness determines who approves an action, what the policy receives, and when to call the policy.

Two studies compare direct delegation with letting the agent review the policy's proposed actions. In the SenseNova study~\citep{survey_ax_2610_22471}, GPT-6 Astra acting alone in the authors' harness achieved 15.56\% category-averaged success on RoboDojo. The $\pi_{0.5}$ policy alone achieved 6.91\%. When GPT-6 Astra specified a subgoal and let the policy execute it, success was 5.83\%. When it could revise each proposed end-effector action, success was 23.61\%. Reviewing four actions per call gave 25.83\% and reduced token use from 7.664 million to 2.796 million.

Embo-Harness~\citep{survey_ax_2611_31201} found a similar pattern on 20 RoboLab tasks, with three trials per task. GPT-6 Astra alone achieved 90.0\% success with a reported task execution time of 10.5 minutes. Letting the model choose at each step whether to act or delegate was faster (4.4 minutes) but less successful (58.3\%). Reviewing the VLA's proposed actions took the longest (14.8 minutes) and achieved the highest success (96.7\%).

What the policy receives also varies. On EBench~\citep{survey_ax_2610_35832}, the agent chooses how many steps to delegate, but the frozen policy always receives the original task instruction. Across 510 paired episodes, 96 succeeded only with access to InternW0-$\Delta$ and 49 succeeded only under direct control. The two configurations ran at different times, once per episode.

Agentic GR00T~2~\citep{nvidia2026gr00t2} lets the upper layer specify where to act. In its visually guided configuration, the agent chooses an arm, gives a subinstruction, and marks grasp and release points on a current camera image. The WAM is trained to follow these markers and chooses the grasp pose, approach, and intermediate motion. After a release, a positive completion check, or 45 seconds without either event, the harness asks the agent whether to advance, retry, or change the plan.

A policy can also receive one instruction at a time from a plan written in advance. In one such system~\citep{survey_ax_2610_03615}, a frontier model writes the subgoals from the initial camera views. A local vision-language model compares later images with the start of the current subgoal and decides when to advance. The order of subgoals stays fixed.

In ASENA~\citep{survey_ax_2609_39207}, the agent decides when to call a navigation policy, check progress, or resume geometric planning. The policy takes a short navigation instruction and recent monocular images and returns waypoints relative to the robot. On the R2R and RxR splits of 100 episodes each, giving Claude Sonnet~5 this tool raised success by 11 percentage points on each split. The same tool lowered GPT-6 Astra's success on RxR from 93\% to 86\%, even as calls per episode fell from 40.80 to 19.98.

\paragraph{\textbf{Decision APIs.}}
Between such orchestrators and their executors, a recent development adds a fast decision layer. Decision models answer bounded questions over a supplied set of options instead of generating text. TypeSafe describes Jev~\citep{typesafe2026jev} as a System One model. It takes a textual state and a batch of questions. Each asks it to choose among options, place a case on an ordered scale, or judge whether a condition holds. It returns a probability for every answer without generating output tokens. Liquid AI's d1~\citep{liquid2026d1} offers the same three question types over text and images in a single forward pass, with text decisions in 200--300\,ms, and OpenAI has announced a comparable decision interface~\citep{openai2026decisions}. In a robot harness, such a model sits beneath a slower planner: code proposes the legal candidates, the decision model selects one with a probability, and the harness acts only above a confidence threshold. EmbodiedJev~\citep{embodiedjev2026}, an open-source simulation workbench, applies this protocol to manipulation, letting Jev choose among skills, phases, or Cartesian steps while inverse kinematics, safety checks, and success judgment remain in code. The idea of restricting a language model to scored candidates predates these products. SayCan~\citep{survey_ax_2204_01691} combines language-model likelihood with learned affordances over a fixed skill library, MOKA~\citep{survey_ax_2403_03174} asks a VLM to choose among marked keypoints, and KnowNo~\citep{survey_ax_2307_01928} and IntroPlan~\citep{survey_ax_2402_06529} turn option likelihoods into conformal prediction sets that trigger requests for help. Dedicated decision models make this pattern cheap enough to run at every step. Their thresholds are meaningful once calibration is measured on the robot's own decisions, since a candidate's probability expresses preference among the offered options rather than the chance of physical success.

\paragraph{\textbf{Code as the Upper Layer and Outlook.}}
At the top of the stack, code can itself serve as System~2. The code-as-policies line discussed under Skills composes System~0 primitives in generated programs. In ReKep~\citep{survey_ax_2409_01652}, a VLM writes keypoint-constraint functions once per task. The solver runs at about 10\,Hz, while keypoint tracking and position control run at 20\,Hz. Instruct2Act~\citep{survey_ax_2305_11176} composes perception models and motion primitives in generated code. AgentChord~\citep{survey_ax_2605_11951} compiles monitored task graphs with precompiled recovery branches and constrained solvers, so that detected violations are handled without a new language-model call. Programs that call VLAs alongside analytic primitives, as in Harness VLA, combine both: code remains inspectable and verifiable, while the VLA covers contact-rich segments that are hard to specify.

A program can keep responding to the robot between calls to a VLM. X-Harness~\citep{survey_ax_2611_34778} compares running the program alone, calling the VLM when the program needs help, and consulting it at every subgoal. On 50 RoboCasa365 tasks with Awomo-0.5I as the underlying VLA, the revised programs alone succeeded in 656 of 1,000 episodes. With VLM calls after failures or to check whether a move had stalled, 773 episodes succeeded. An earlier version that called the VLM at every subgoal succeeded in 768 of 1,000 episodes. Calling the VLM only when needed used 42\% fewer VLM tokens per episode than calling it at every subgoal. In that setting, the program finished 51.4\% of test episodes without a VLM call.

A natural direction is a stack in which VLAs serve as callable lower layers, programs provide the upper layer, and an agent revises both offline through skill evolution. Each choice across the chapter carries a cost in this stack. A bridge between executors broadens successor states but consumes motion time; retrieved history preserves a commitment but can propagate stale evidence; editable code can repair a recurring failure but can also change the monitor that judges it. The strongest comparisons isolate the changed layer or interface, hold the others fixed, and test later behavior beyond the feedback used for optimization. Related works are compared in Table~\ref{tab:harness-layered-systems}.

\par\Needspace{12\baselineskip}
\begingroup
\mechanismtableglue
\CompactTableStyle
\setlength{\LiteratureWidth}{\dimexpr\linewidth-10\tabcolsep\relax}

\setlength{\LTpre}{4pt}
\setlength{\LTpost}{4pt}
\setlength{\LTcapwidth}{\linewidth}
\captionsetup{singlelinecheck=false,justification=raggedright}
\LiteratureContinuationSetup
\begin{longtable}{@{}>{\RaggedRight\arraybackslash}p{0.1758\LiteratureWidth}>{\centering\arraybackslash}p{0.0750\LiteratureWidth}>{\RaggedRight\arraybackslash}p{0.1388\LiteratureWidth}>{\RaggedRight\arraybackslash}p{0.1480\LiteratureWidth}>{\RaggedRight\arraybackslash}p{0.1850\LiteratureWidth}>{\RaggedRight\arraybackslash}p{0.2775\LiteratureWidth}@{}}
\caption[Comparison of works on layered systems]{Comparison of works on layered systems (Section~\ref{sec:harness-layered-systems}). Update timing identifies the refreshed module; rates refer to the reported configurations. Architecture studies list alternatives. NR: not reported.}\label{tab:harness-layered-systems}\\
\LiteratureTopRule{Harness}
\rowcolor{TSTHarnessLight}\textbf{Work} & \textbf{Link} & \textbf{Upper layer} & \textbf{Lower layer} & \textbf{Passed down} & \textbf{Update timing} \\
\midrule
\endfirsthead
\multicolumn{6}{@{}l}{\tablename~\thetable\ (continued)} \\
\LiteratureTopRule{Harness}
\rowcolor{TSTHarnessLight}\textbf{Work} & \textbf{Link} & \textbf{Upper layer} & \textbf{Lower layer} & \textbf{Passed down} & \textbf{Update timing} \\
\midrule
\endhead
\midrule
\multicolumn{6}{r@{}}{Continued on next page} \\
\endfoot
\bottomrule
\endlastfoot
\rowcolor{TSTHarnessLight}\multicolumn{6}{@{}l}{\textbf{Systems and planning methods}} \\*[3pt]
% Source: https://arxiv.org/abs/2204.01691v2 | §3–4; Algorithm 1: value-grounded selection followed by skill execution
SayCan \mbox{\citep{survey_ax_2204_01691}} & \LiteratureLinks{https://proceedings.mlr.press/v205/ichter23a.html}{https://github.com/google-research/google-research/tree/master/saycan} & LLM selector & Learned skills & Skill description & Select again after each skill \\[0pt]
% Source: https://arxiv.org/abs/2305.11176 | §3.1–3.3: generated perception-to-action program; runtime LLM requery schedule not specified
\rowcolor{black!3}
Instruct2Act \mbox{\citep{survey_ax_2305_11176}} & \LiteratureLinks{https://arxiv.org/abs/2305.11176}{https://github.com/OpenGVLab/Instruct2Act} & LLM coder & Perception APIs; controller & Python program & Runtime requery schedule NR \\[0pt]
% Source: https://arxiv.org/abs/2307.01928 | §2: candidate generation, calibrated prediction set, human help and low-level execution
KnowNo \mbox{\citep{survey_ax_2307_01928}} & \LiteratureLinks{https://proceedings.mlr.press/v229/ren23a.html}{https://github.com/google-research/google-research/tree/master/language_model_uncertainty} & LLM planner & Skill executor & Selected action option & Each decision; ask for help if ambiguous \\[0pt]
% Source: https://arxiv.org/abs/2402.06529 | §2–3; Algorithm 2; §4.2 and §7: planning-dataset evaluation, not a specified low-level runtime
\rowcolor{black!3}
IntroPlan \mbox{\citep{survey_ax_2402_06529}} & \LiteratureLinks{https://proceedings.neurips.cc/paper_files/paper/2024/hash/8451a20c5a7e0ee5671dda28f7daf7f3-Abstract-Conference.html}{https://github.com/kevinliang888/IntroPlan} & LLM planner & Executor NR & Calibrated action options & Runtime schedule NR \\[0pt]
% Source: https://arxiv.org/abs/2403.03174 | §IV-A–B: point-based motion representation and per-subtask affordance inference
MOKA \mbox{\citep{survey_ax_2403_03174}} & \LiteratureLinks{https://www.roboticsproceedings.org/rss20/p062.html}{https://github.com/moka-manipulation/moka} & VLM & Motion planner & Keypoints; waypoints & New points at each subtask start \\[0pt]
% Source: https://arxiv.org/abs/2409.01652v2 | §3.3–3.4: real-time pose/path solving, keypoint tracking and VLM-generated constraints
\rowcolor{black!3}
ReKep \mbox{\citep{survey_ax_2409_01652}} & \LiteratureLinks{https://proceedings.mlr.press/v270/huang25g.html}{https://github.com/huangwl18/ReKep} & VLM & Constraint solver & Keypoint constraints & Solver replans at about \textbf{10 Hz} \\[3pt]
% Source: https://arxiv.org/abs/2410.05273 | §3.1–3.2: asynchronous latent updates and per-step visual policy
HiRT \mbox{\citep{survey_ax_2410_05273}} & \LiteratureLinks{https://proceedings.mlr.press/v270/zhang25b.html}{} & VLM & Visual policy & Cached vision-language features & Async; policy uses fresh images each step \\[0pt]
% Source: https://arxiv.org/abs/2410.08001 | §3.1–3.3: generalist latents/actions, specialist conditioning and shifted-window inference
\rowcolor{black!3}
RoboDual \mbox{\citep{survey_ax_2410_08001}} & \LiteratureLinks{https://arxiv.org/abs/2410.08001}{https://github.com/OpenDriveLab/RoboDual} & Generalist VLA & Diffusion policy & Latent features; action chunk & Async; several policy steps per VLA call \\[0pt]
% Source: https://arxiv.org/abs/2412.04453 | §II-A–B: parsed movement commands and command-dependent execution durations
NaVILA \mbox{\citep{survey_ax_2412_04453}} & \LiteratureLinks{https://roboticsproceedings.org/rss21/p018.html}{https://github.com/AnjieCheng/NaVILA} & Navigation VLM & Locomotion policy & Movement; distance or angle & Motion duration set by the command \\[0pt]
% Source: https://arxiv.org/abs/2502.05485 | §4; Fig. 2; §4.1–4.2: one high-level call, path/gripper representation and closed-loop low-level policy
\rowcolor{black!3}
HAMSTER \mbox{\citep{survey_ax_2502_05485}} & \LiteratureLinks{https://proceedings.iclr.cc/paper_files/paper/2025/hash/3bfee3bc6639c36e6e7b058db909f760-Abstract-Conference.html}{https://github.com/liyi14/HAMSTER_beta} & Path VLM & 3D robot policy & 2D path; gripper states & Path generated \textbf{once} per episode \\[3pt]
% Source: https://arxiv.org/abs/2502.19417 | §4.1–4.2: one-second periodic inference and immediate user-interaction trigger
Hi Robot \mbox{\citep{survey_ax_2502_19417}} & \LiteratureLinks{https://proceedings.mlr.press/v267/shi25d.html}{} & VLM & VLA & Language subtask & Every \textbf{1 s} or on new user feedback \\[3pt]
% Source: https://arxiv.org/abs/2503.14734v1 | §1; §2.1: reported L40 rates, VLM features and action-transformer cross-attention
\rowcolor{black!3}
GR00T N1 \mbox{\citep{survey_ax_2503_14734}} & \LiteratureLinks{https://arxiv.org/abs/2503.14734}{https://github.com/NVIDIA/Isaac-GR00T} & VLM & Action transformer & Vision-language features & VLM \textbf{10 Hz}; motor actions \textbf{120 Hz} (L40) \\[3pt]
% Source: https://arxiv.org/abs/2504.16054 | §IV-A, §IV-E: shared-model subtask/action factorization and 50 Hz command stream
$\pi_{0.5}$ \mbox{\citep{survey_ax_2504_16054}} & \LiteratureLinks{https://proceedings.mlr.press/v305/black25a.html}{https://github.com/Physical-Intelligence/openpi} & Subtask predictor & Action expert & Language subtask & Actions \textbf{50 Hz}; upper-layer rate NR \\[3pt]
% Source: https://arxiv.org/abs/2510.03342 | §2.1: orchestrator, action-model tool and success-detection-based switching
\rowcolor{black!3}
Gemini Robotics 1.5 \mbox{\citep{survey_ax_2510_03342}} & \LiteratureLinks{https://arxiv.org/abs/2510.03342}{} & VLM orchestrator & VLA tool & Language instruction & Switch subtasks on detected completion \\[0pt]
% Source: https://arxiv.org/abs/2512.08186 | §3.1–3.3: pixel-goal grounding followed by latent queries; the lower policy consumes latent goals
DualVLN \mbox{\citep{survey_ax_2512_08186}} & \LiteratureLinks{https://openreview.net/pdf?id=GK4rznYwhn}{https://github.com/InternRobotics/InternNav} & Navigation VLM & Trajectory policy & Latent goal from pixel-waypoint reasoning & Async; fast policy reuses the latest latent \\[0pt]
% Source: https://arxiv.org/abs/2603.05185 | §3.1–3.4: language subtask, independent critic and three event-driven interrupts
\rowcolor{black!3}
Critic in the Loop \mbox{\citep{survey_ax_2603_05185}} & \LiteratureLinks{https://arxiv.org/abs/2603.05185}{} & VLM planner; critic & Action expert & Language subtask & Replan on completion, anomaly or stagnation \\[0pt]
% Source: https://arxiv.org/abs/2607.08448 | §2.1–2.3: turn-based planner and one unified primitive library containing VLA and analytic tools
Harness VLA \mbox{\citep{survey_ax_2607_08448}} & \LiteratureLinks{https://arxiv.org/abs/2607.08448}{https://github.com/RLinf/RPent} & VLM planner & Analytic tools; frozen VLA & JSON primitive call & New observation after each primitive ends \\[0pt]
\rowcolor{black!3}
DynaHarness \citep{survey_ax_2609_40306} & \LiteratureLinks{https://arxiv.org/abs/2609.40306v1}{} & VLM and scheduler & Analytic skills, VLA & Skill and target & 2 Hz scheduler can interrupt 20 Hz execution \\
ASENA~\citep{survey_ax_2609_39207} & \LiteratureLinks{https://arxiv.org/abs/2609.39207v2}{} & Coding agent & Navigation policy & Navigation instruction & Delegate on demand and inspect progress \\
\rowcolor{black!3}
Agentic GR00T~2 \citep{nvidia2026gr00t2} & \LiteratureLinks{https://research.nvidia.com/labs/gear/gr00t-2/assets/gr00t2_whitepaper.pdf}{} & Reasoning agent & WAM & Subtask, arm, grasp and release points & Ask the agent on release, confirmed completion, or timeout \\
\rowcolor{TSTHarnessLight}\multicolumn{6}{@{}l}{\textbf{Architecture studies}} \\*[3pt]
% Source: https://arxiv.org/abs/2506.17561 | §3.2–3.4: action-only, implicit, explicit and hierarchical designs; representations are alternatives
VLA-OS \mbox{\citep{survey_ax_2506_17561}} & \LiteratureLinks{https://proceedings.neurips.cc/paper_files/paper/2025/hash/c7af751b5e0a407c62ac023e3cb381f9-Abstract-Conference.html}{https://github.com/HeegerGao/VLA-OS} & VLM; planning heads & Alternative action heads & Text / spatial / image plans & Varies across the compared architectures \\*[0pt]
% Source: https://arxiv.org/abs/2606.10267 | §3; Algorithm 1; §4.4: unified control loop and three tested termination rules
\rowcolor{black!3}
WMORP \mbox{\citep{survey_ax_2606_10267}} & \LiteratureLinks{https://arxiv.org/abs/2606.10267}{} & VLM planner & VLA & Language subtask & Compare fixed, success-based and VLM-timed switching \\[0pt]
SenseNova study~\citep{survey_ax_2610_22471} & \LiteratureLinks{https://arxiv.org/abs/2610.22471v2}{https://github.com/OpenSenseNova/SenseNova-RoboRSI} & GPT-6 Astra & $\pi_{0.5}$ & Subgoal or revised action & Delegate or review 1 or 4 actions \\
\rowcolor{black!3}
EBench~\citep{survey_ax_2610_35832} & \LiteratureLinks{https://arxiv.org/abs/2610.35832v1}{} & GPT-6 Astra & Frozen policy & Original task instruction & Agent chooses how long the policy runs \\
Embo-Harness \citep{survey_ax_2611_31201} & \LiteratureLinks{https://arxiv.org/abs/2611.31201v1}{https://github.com/DravenALG/Embo-Harness} & GPT-6 Astra & VLA or controller & VLA call or refined action & Select or refine before each action \\
\rowcolor{black!3}
X-Harness~\citep{survey_ax_2611_34778} & \LiteratureLinks{https://arxiv.org/abs/2611.34778v1}{} & VLM or policy code & VLA; robot tools & Subgoal or tool call & Code alone, on a trigger, or each subgoal \\
\end{longtable}
\endgroup

\section{Outlook: What Comes Next for Agentic Robots?}
\label{sec:frontiers}

Agentic robots connect general-purpose reasoning models with robot policies, tools, and physical execution. However, integrating these components does not necessarily make a robotic system more capable. It remains difficult to determine what the agent contributes, whether its decisions can be executed reliably on physical robots, and how well the system performs under limited resources or on unfamiliar platforms. These challenges motivate four research directions.

% Agentic robots extend model-mediated decisions across task execution, data collection, model training, and experimental deployment. A key next step is to connect these activities: robots should use available capabilities to complete tasks, identify limitations through interaction, and organize learning to address those limitations. This raises four research directions.

\subsection{Fair Benchmarking}

What does a task success rate actually tell us about an agentic robot? Consider two systems evaluated on the same manipulation benchmark. One may rely on a stronger VLA executor, while the other uses an agent to coordinate multiple weaker skills, gather additional observations, and recover from failures. Similar scores can reflect very different capabilities. CaP-X~\citep{survey_ax_2603_22435} evaluates agents that generate robot-control programs, while Harness VLA~\citep{survey_ax_2607_08448} and RoboHarness~\citep{survey_ax_2607_18060} study how agents coordinate robotic policies and skills. Their results depend not only on the agent but also on the underlying executors, available tools, observation access, and allowed retries. Task success alone therefore cannot show which part of the system made the difference.

To interpret a success rate, we need to know which tools the agent could use and how many steps it had. RoboWild~\citep{xlanglab2026robowild} compared two harnesses on the same 239 Dexmate setups. Both used RGB observations and Claude Opus 5.5 at maximum reasoning effort. ReAct combined motion planning and execution in one call and kept images from only the two most recent messages containing camera views. It achieved 6.28\% success at an estimated mean API cost of \$43.0 per task. Claude Code exposed separate calls for planning and execution and let the agent reopen saved images. It achieved 12.13\% success at an estimated mean API cost of \$57.1 per task.

EBench~\citep{survey_ax_2610_35832} shows why the execution budget matters as well. On the Shop task, neither direct control nor policy-assisted control completed an episode within 3,000 steps. By step 5,000, the agent had completed three of the 20 Shop episodes with $\pi_{0.5}$ and one with InternW0-$\Delta$. The authors ran only the policy-assisted configurations past 3,000 steps on Shop. Table~\ref{tab:benchmark-protocols} shows how changing the cutoff on Bottle and Shop affects the average across all 26 tasks.

JustRUA~\citep{survey_ax_2611_16009} tested how soon GPT-6 Astra should see the result of a motion. Both configurations kept the full episode history. On the same 111 RoboDojo test episodes, limiting each motion to 20 control steps raised the overall score from 34.7 to 38.0. Feedback during development serves a different purpose. X-Harness~\citep{survey_ax_2611_34778} uses training trials to revise policy code. In another study~\citep{survey_ax_2611_09030}, training episodes supply the feedback, validation episodes select the program, and the test uses held-out layouts of the same task.

Even when two methods share a simulator, they may face different tests. Mind2Act World~\citep{survey_ax_2611_26444} evaluates each end-to-end policy on 50 instances per configuration. Each agent gets the same five of those instances and four times the simulation-time budget per episode. The paper reports the two groups on separate leaderboards.

To compare agents fairly, we need to examine their decisions under comparable conditions. Using the same executors, observation access, and interaction budgets makes it easier to assess planning, tool selection, and recovery. Comparing a system with and without memory or replanning can further show what each component contributes. However, controlled comparisons alone are not enough, since choosing and combining different models and tools is itself an important capability. End-to-end evaluations should therefore report the resources available to each system, including additional observations, computational costs, execution time, and human assistance. Such comparisons can reveal whether the benefits of additional reasoning and tool use justify their costs.

Even with careful comparisons, a benchmark may tell us little about the agent if its tasks can be completed by a single fixed policy. Some robot benchmarks focus on task completion without requiring the agent to handle missing information, unsuitable executors, or failed actions. More informative benchmarks could introduce unexpected environmental changes, incomplete observations, and tasks that require coordination across different skills. For agents that collect data, update models, or develop new skills, evaluation should also examine whether these changes improve performance on unseen tasks. This would help establish when an agent is useful and whether its improvements generalize beyond the original setting.

\subsection{Real-robot Agentic Robotics}

A plan that makes sense to an agent may not always be executable by a robot. A reasoning model might request a grasp beyond the robot's reach or select a skill based on an outdated observation. Agentic robots may combine reasoning models, VLAs, world-action models, and specialized robot skills, each with different capabilities and execution requirements. ReKep~\citep{survey_ax_2409_01652} uses model-generated constraints to guide robot motion, while Harness VLA~\citep{survey_ax_2607_08448} allows an agent to call a VLA alongside other robot operations. Yet choosing a suitable executor is only part of the problem. A valid command may still fail during physical execution, and the scene may change before the agent receives feedback.

Robot interfaces therefore need to describe more than the names and inputs of available skills. An agent should know when a skill can be used, what conditions it requires, and how to tell whether execution has finished. For example, a failed grasp may require another observation, a revised grasp command, or a different skill. The right response depends on what actually happened: the object may have moved, or the previous command may still be running. Sending a stop request does not guarantee that the robot has stopped moving. Before replanning, the agent needs feedback that confirms the current state and the effects of actions already taken.

These difficulties become harder to address when moving from simulation to physical robots. Simulators can provide accurate object positions and clearly defined success conditions, while real robots must often estimate these from noisy camera images and sensor readings. Small perception errors, calibration differences, and unexpected contacts can cause an otherwise reasonable plan to fail. Future systems should recognize when their observations are insufficient, check important assumptions before execution, and use physical feedback to correct their decisions. Real-robot evaluations should also make clear what information the agent receives, what the lower-level controllers handle, and whether the agent can recover when its assumptions turn out to be wrong.

% Agentic robots may combine reasoning models, VLAs, world-action models, and specialized robot skills. Coordinating these capabilities requires deciding which component to use, what information it needs, and when execution feedback calls for a different response. Future systems should adapt these choices to the task and the robot's current capabilities. For example, a failed grasp may require another observation, a revised grasp command, or a different skill. Understanding when to continue, replan, or request assistance remains a central challenge.

\subsection{Efficient Robot Use}

A robot may complete a motion in a few seconds, yet spend much longer waiting for the agent to decide what to do next. Repeated model calls, visual observations, and tool interactions can introduce significant delays, especially when a task requires many small actions. Recent work has begun addressing this problem by reducing unnecessary high-level reasoning. HarnessWAM~\citep{survey_ax_2608_09516} uses a lightweight progress estimator to monitor execution and invokes its VLM task manager when important events occur. Similarly, VLAs-as-Tools~\citep{survey_ax_2605_13119} provides execution feedback that allows an agent to replan when needed, rather than continuously checking the robot's progress. Efficiency therefore depends not only on how fast a model runs, but also on when it is called.

Whether the world waits changes the cost of a model call. Mind2Act World~\citep{survey_ax_2611_26444} pauses physics during inference. RoboTeam~\citep{survey_ax_2611_22138} keeps physics running and reports how many tasks finish within 3, 6, 12, and 24 hours. On real robots, reasoning and motion can also overlap. ACAP~\citep{survey_ax_2611_31205} measures time spent reasoning, moving, and waiting for other agents. Compared with direct control, more of the time gain came from overlapping reasoning with other work than from moving several robot parts at once.

The main challenge is deciding when another model call is worth the delay. A robot following a familiar manipulation sequence may continue with lightweight monitoring, while an unexpected obstacle or failed action may require a new decision from the reasoning model. Calling the model too often wastes time, but waiting too long can leave the robot following a plan that no longer works. The timing should depend on task progress, uncertainty, and the possible consequences of a delayed response.

Efficiency also depends on where computation takes place. Large models running in the cloud may offer stronger reasoning but introduce communication delays, while smaller local models can respond more quickly with limited computational resources. Future systems could combine these options, using local components for time-sensitive decisions and stronger models when additional reasoning is needed. Evaluations should therefore consider the full execution process, including reasoning time, tool calls, communication overhead, and physical task completion time. Fewer model calls or faster inference do not necessarily improve efficiency if they lead to more failures, retries, or longer task execution.

\subsection{Cross-Embodiment Generalization}

An agent that works well with one robot may struggle when given a different platform. Even a simple task, such as reaching a target location, requires different actions from a wheeled robot, a humanoid, or a drone. The high-level goal may remain similar, but the available sensors, control interfaces, and physical constraints are different. RobotWorld~\citep{survey_ax_2610_10409} evaluates general-purpose agents on 84 simulated tasks spanning manipulation, mobile manipulation, locomotion, driving, and aerial control. What remains unclear is whether the same agent can adapt to a new robot without major changes to its planning and tool-use setup.

One promising direction is to enable agents to understand the capabilities of unfamiliar robots before using them. An agent could identify the robot's available sensors, movement capabilities, and supported skills, then plan around those capabilities. For example, a wheeled robot may need to navigate around obstacles to inspect an area, while a drone could approach it from above. The goal is similar, but the actions and physical constraints are different. The agent must therefore adjust its plan rather than simply translate the same commands into another robot's control interface.

Another important question is how much of an agent's previous experience can be reused. High-level strategies, such as gathering additional observations when information is missing or checking why an action failed, may remain useful across different robots. In contrast, specific motion commands and control parameters often depend on the robot's embodiment. Future systems should retain general task knowledge without assuming that the same execution details apply to every platform. Evaluations could test agents on previously unseen robots, measuring task success, the amount of additional configuration or training required, and whether experience from other robots improves performance.

\section{Conclusion}
\label{sec:conclusion}

This paper examines how agents support robot execution and learning. We trace the evolution from robot policies to agentic systems and clarify the scope of agentic robots. We organize existing work into seventeen overlapping categories across four functional domains: Model, Data, Environment, and Harness. These categories cover model learning and decisions, data collection and preparation, the construction of worlds, tasks, and tests, and the software that coordinates robot capabilities.

Across these approaches, an agent's decisions depend on what it can observe, what the robot can do, and what feedback it receives. During execution, agents coordinate tools and skills, check progress, and use feedback to decide whether to continue, gather more information, or recover from failure. Beyond the current task, stored experience can guide later decisions and changes to data, training procedures, skills, models, and software. These changes need to be tested on new tasks and conditions, while checking that previously learned capabilities are preserved.

Building on this analysis, we discuss four directions for future research: fair benchmarking, real-robot deployment, efficient robot use, and cross-embodiment generalization. Future work needs to establish when agents improve performance under comparable conditions and how they handle uncertainty and failures on physical robots. It should also examine whether these improvements justify the time and resources used, and how well agents adapt to unfamiliar robots. We hope this work helps researchers connect these questions and develop more capable agentic robots. Progress in agentic robotics could bring us closer to intelligent systems that turn knowledge into action, learn from experience, and adapt to the changing physical world.
\clearpage

\bibliographystyle{plainnat-extended}
% \bibliography{ref}
\bibliography{main}

\end{document}